\documentclass[nohyperref]{article}

\usepackage{microtype}
\usepackage{graphicx}
\usepackage{subfigure}
\usepackage{booktabs} 
\usepackage{xcolor, soul}
\usepackage{hyperref}

\usepackage[accepted]{shmicml2023}

\usepackage{amsmath}
\usepackage{amssymb}
\usepackage{mathtools}
\usepackage{amsthm}

\usepackage[capitalize,noabbrev]{cleveref}

\usepackage{xcolor}         
\usepackage[most]{tcolorbox}
\usepackage{xspace}
\usepackage{graphicx}
\usepackage{color}
\usepackage{multirow}
\usepackage{makecell}
\usepackage{bm}
\usepackage{wrapfig}
\usepackage{babel}

\usepackage{pifont}
\newcommand{\xmark}{\ding{55}}

\newcommand{\nbatch}{N_{Bt}}

\newcommand{\srcparam}{\theta^s}
\newcommand{\sourcemodel}{f(\cdot ; \srcparam)}
\newcommand{\preparam}{\theta^{\text{pre}}}
\newcommand{\premodel}{f(\cdot ; \preparam)}

\newcommand{\dm}{\boldsymbol{\delta_m}}

\newcommand{\ypl}{\widetilde{y}_i}
\newcommand{\bypls}{\widetilde{\mathbf{y}}}
\newcommand{\bx}{\mathbf{x}}

\newcommand{\xtgt}{\mathbf{x}_{tgt}^{t}}
\newcommand{\ytgt}{\dot y_{tgt}}

\newcommand{\batch}{\mathbf{X}_{B}^{t}}
\newcommand{\wasser}{D_\text{W}}
\newcommand{\R}{\mathbb{R}}

\newcommand{\sourcedata}{\mathcal{D}^{s}}
\newcommand{\targetdata}{\mathbf{X}^t}
\newcommand{\batchdata}{\mathbf{X}_{B}^{t}}

\newcommand{\batchsample}{\mathbf{x}_{i}^{t}}

\newcommand{\mal}{\mathbf{X}_{mal}^{t}}
\newcommand{\hatmal}{\widehat{\mathbf{X}}_{mal}^{t}}

\newcommand{\bntheta}{\theta_{\mathcal{B}}}
\newcommand{\adatheta}{\theta_{\mathcal{A}}}
\newcommand{\fixtheta}{\theta_{\mathcal{F}}}

\newcommand{\norm}[1]{\left\lVert#1\right\rVert}

\newcommand{\eps}{\varepsilon}

\newcommand{\pert}{\boldsymbol{\delta}}

\newcommand{\A}{\mathcal{A}}
\newcommand{\B}{\mathcal{B}}
\newcommand{\mL}{\mathcal{L}}

\newcommand{\benign}{\mathbf{X}_{B \setminus ({tgt}\cup{mal})}^{t}}

\newcommand{\benignlabel}{\mathbf{y}_{B \setminus ({tgt}\cup{mal})}^{t}}

\newcommand{\bL}{\mathbb{L}}
\newcommand{\epsproj}{\Pi_{\epsilon}}
\newcommand{\hatx}{\hat{\mathbf{x}}^{t}}
\newcommand{\rr}{\textcolor{red}}

\newcommand{\benigns}{\mathbf{X}^t_{B \setminus mal}}
\newcommand{\benignslabel}{\mathbf{y}^t_{B \setminus mal}
}

\newcommand{\gobject}{\bL(f(\cdot \ ;\theta^*(\batchdata)))}

\newcommand{\TeBN}{{\textbf{TeBN}}}

\DeclareMathOperator*{\argmax}{arg\,max}
\DeclareMathOperator*{\argmin}{arg\,min}

\newcommand{\alglinelabel}{%
  \addtocounter{ALC@line}{-1}
  \refstepcounter{ALC@line}
  \label
}
\newif\iffinal

\iffinal
    \newcommand{\tianhao}[1]{}
    \newcommand{\xiangyu}[1]{}
    \newcommand{\saeed}[1]{}
    \newcommand{\tw}[1]{}
    \newcommand{\fj}[1]{}
    
    \newcommand{\vikash}[1]{}
\else
\newcommand{\tianhao}[1]{{\color{purple}\textbf{[Tianhao: {#1}]}}}
\newcommand{\xiangyu}[1]{\textcolor{cyan}{ [Xiangyu: #1]}}
\newcommand{\saeed}[1]{\textcolor{red}{ [Saeed: #1]}}
\newcommand{\tw}[1]{{\color{teal}[Tong: {#1}]}}
\newcommand{\fj}[1]{{\color{pink}[Feiran: {#1}]}}

\newcommand{\vikash}[1]{{\color{pink}[vikash: {#1}]}}
\fi

\theoremstyle{plain}

\theoremstyle{definition}

\theoremstyle{remark}

\usepackage[textsize=tiny]{todonotes}

\icmltitlerunning{\hfill Uncovering Adversarial Risks of Test-Time Adaptation \hfill \thepage}

\begin{document}

\twocolumn[

\icmltitle{Uncovering Adversarial Risks of Test-Time Adaptation} 





\icmlsetsymbol{pton}{$\dag$}
\icmlsetsymbol{psu}{$\ddag$}

\begin{icmlauthorlist}
\icmlauthor{Tong Wu}{pton}
\icmlauthor{Feiran Jia}{psu}
\icmlauthor{Xiangyu Qi}{pton}
\icmlauthor{Jiachen T. Wang}{pton} \\
\vspace{1mm}
\icmlauthor{Vikash Sehwag}{pton}
\icmlauthor{Saeed Mahloujifar}{pton}
\icmlauthor{Prateek Mittal}{pton}
\icmlcorrespondingauthor{Tong Wu}{tongwu@princeton.edu}

\end{icmlauthorlist}

\begin{center}
\vspace{1mm}
\textbf{$^\dag$Princeton University},
\textbf{$^\ddag$Penn State University}
\end{center}

\vskip 0.3in
]




\printAffiliationsAndNotice{} 

\vspace{-3mm}
\begin{abstract}
\vspace{-1mm}
    Recently, test-time adaptation (TTA) has been proposed as a promising solution for addressing distribution shifts. 
    It allows a base model to adapt to an unforeseen distribution during inference by leveraging the information from the batch of (unlabeled) test data.
    However, we uncover a novel security vulnerability of TTA based on the insight that predictions on benign samples can be impacted by malicious samples in the same batch.
    To exploit this vulnerability, we propose \textit{\textbf{D}istribution \textbf{I}nvading \textbf{A}ttack} (DIA), which injects a small fraction of malicious data into the test batch. 
    DIA causes models using TTA to misclassify benign and unperturbed test data, providing an entirely new capability for adversaries that is infeasible in canonical machine learning pipelines.
    Through comprehensive evaluations, we demonstrate the high effectiveness of our attack on multiple benchmarks across six TTA methods.
    In response, we investigate two countermeasures to robustify the existing insecure TTA implementations, following the principle of ``security by design''. 
    Together, we hope our findings can make the community aware of the \textit{utility-security tradeoffs} in deploying TTA and provide valuable insights for developing robust TTA approaches.
\end{abstract}




\vspace{-5mm}
\section{Introduction}

%
\vspace{-1mm}
Test-time adaptation (TTA)~\cite{Wang2021TentFT, Schneider2020ImprovingRA, Goyal2022TestTimeAV} is a cutting-edge machine learning (ML) approach that addresses the problem of distribution shifts in test data~\cite{Hendrycks2019BenchmarkingNN}. Unlike conventional ML methods that rely on a fixed base model, TTA generates batch-specific models to handle different test data distributions. 
Specifically, when test data are processed batch-wise \cite{batch-prediction}, TTA first leverages them to update the base model and then makes the final predictions using the updated model.
Methodologically, TTA differs from the conventional ML (\emph{inductive learning}) and falls within the \textit{transductive learning paradigm} \cite{vapnik1998statistical}. 
TTA usually outperforms conventional ML under distribution shifts since 1) it gains the distribution knowledge from the test batch and 2) it can adjust the model adaptively.
Empirically, TTA has been shown to be effective in a range of tasks, including image classification~\cite{Wang2021TentFT}, object detection~\cite{Sinha_2023_WACV}, and visual document understanding~\cite{TTAVDU}.

\begin{figure}[t]
\setlength\abovecaptionskip{0pt}
\setlength\belowcaptionskip{-14pt}
  \centering
  \vspace{-2mm}
    \includegraphics[width=0.50\textwidth]{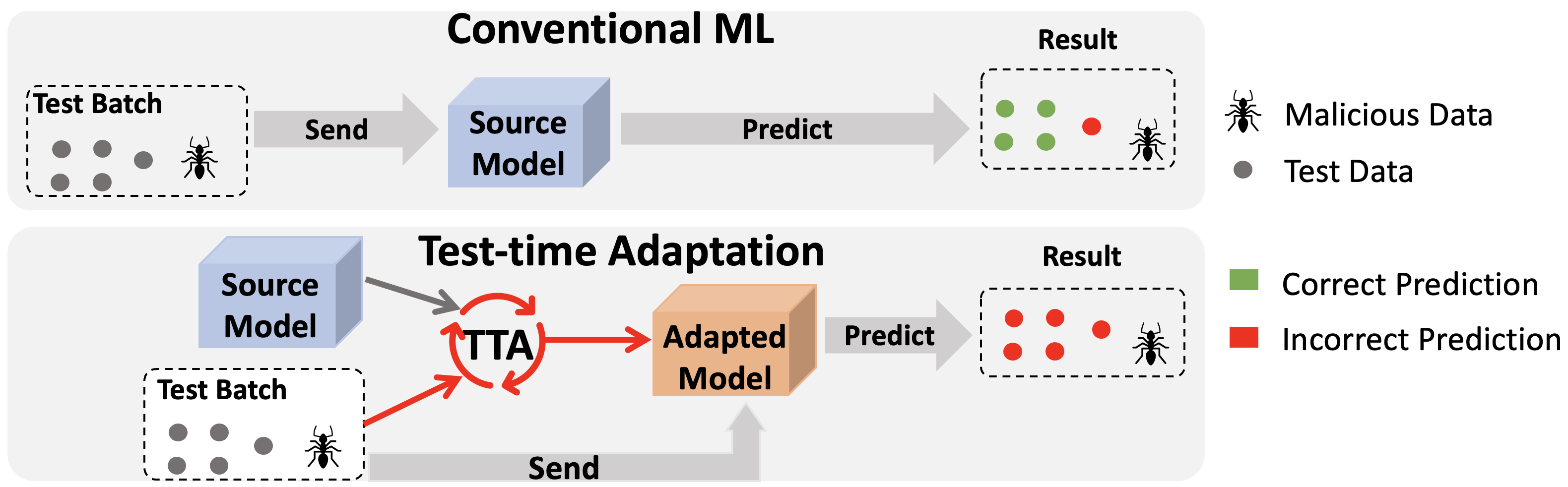}
    \vspace{-5mm}
    \caption{ Overview of conventional machine learning and test-time adaptation. Test-time adaptation adapts to the test batch, but is vulnerable to malicious data at test time (in contrast to conventional machine learning).} 
    \vspace{-7mm}
    \label{Fig:pip}
  \end{figure}


However, in this work, we highlight \textbf{a potential security vulnerability in the test-time adaptation process --- an adversary can introduce malicious behaviors into the model by crafting samples in the test batch}. 
Our key insight is that TTA generates the final predictive model based on the entire test batch rather than making independent predictions for each data as in a conventional ML pipeline.
Therefore, the prediction for one entry in a batch will be influenced by other entries in the same batch. 
As a result, an adversary may submit malicious data at test time to interfere with the generation of the final predictive model, consequently disrupting predictions on other unperturbed data submitted by benign users. 
This emphasizes the necessity of considering the \textit{utility-security tradeoffs} associated with deploying TTA.

\vspace{-1mm}\noindent\textbf{Our Contributions.} To exploit  this vulnerability, we present a novel attack called \emph{\textbf{Distribution Invading Attack (DIA)}}, which exploits TTA by introducing malicious data (Section \ref{sec:diamethod}).
Specifically, DIA crafts (or uploads) a small number of malicious samples (e.g., 5\% of the batch) to the test batch, aiming to induce the mispredictions of benign samples. 
We formulate DIA as a bilevel optimization problem with outer optimization on crafting malicious data and inner optimization on TTA updates.
Next, we transform it into a single-level optimization via approximating model parameters, which can be solved by a projected gradient descent mechanism. 
DIA is a generic framework that can achieve multiple adversarial goals, including 1) flipping the prediction of a crucial sample to a selected label (targeted attack) and 2) degrading performance on all benign data (indiscriminate attack).

We empirically illustrate that DIA achieves a high attack success rate (ASR) on various benchmarks, including CIFAR-10-C, CIFAR-100-C, and ImageNet-C~\cite{Hendrycks2019BenchmarkingNN} against a range of TTA methods, such as  \textbf{TeBN}~\cite{Nado2020EvaluatingPB}, \textbf{TENT}~\cite{Wang2021TentFT}, and \textbf{Hard PL}~\cite{lee2013pseudo} in Section \ref{sec:dia}.
Notably, we demonstrate that targeted attacks using 5\% of malicious samples in the test batch can achieve over 92\% ASR on ImageNet-C.  
Our evaluation also indicates that DIA performs well across multiple model architectures and data augmentations. 
Furthermore, the attack is still effective even when there is a requirement for the malicious inputs to be camouflaged in order to bypass the manual inspection.

In response, we explore  countermeasures to strengthen the current TTA methods by incorporating the principle of ``security by design'' (Section \ref{sec:df}). 
Given that adversarially trained models~\cite{Madry2018TowardsDL} are more resistant to perturbations, we investigate the possibility of defending against DIA using them as the base model.
In addition, since the vulnerabilities of TTA primarily stem from the insecure computation of Batch Norm (BN), we explore two methods to robustly estimate it. 
First, we leverage the BN computed during training time, which is robust to DIA, as a prior for the final BN statistics computation. 
Second, we observe that DIA impacts later BN statistics more than other layers and develop a method to adjust BN statistics accordingly. Our evaluation shows the effectiveness of combining adversarial models and robust BN estimation against DIA, providing guidance to future works for improving TTA robustness. 


Overall, our work shows that in effort to enhance utility, using test-time adaptation (transductive learning paradigm) inadvertently amplifies the security risks. Such utility-secruity tradeoff has been deeply explored in inductive learning \cite{Chakraborty2018AdversarialAA}, and we urge the community to build upon the prior works in inductive learning, by taking these security risks into account when enhancing the utility with transductive learning.




\vspace{-2mm}
\section{Background and Related Work}
\label{sec:bkg}
\vspace{-2mm}

In this section,  we introduce test-time adaptation and then review the related works in adversarial machine learning. More details can be found in \ref{sec:ttabg}.

\vspace{-1mm}
\subsection{Notation and Test-Time Adaptation}
\vspace{-1mm}

Let $\sourcedata := \left\{\left(\mathbf{x}_{i}^{s}, y_{i}^{s}\right)\right\}_{i=1}^{N_{s}}$ be the training data set from the source domain, and  $\targetdata := \left\{\mathbf{x}_{i}^{t}\right\}_{i=1}^{N_{t}}$ be the \emph{unlabeled} test data set from the target domain, where $N_{s}$ and $N_t$ denote the number of points in $\sourcedata$ and $\targetdata$, respectively. 
The goal is to learn a prediction rule $f(\cdot; \theta)$ parameterized by $ \theta$ that can correctly predict the label $y_{i}^{t}$ for each $\mathbf{x}_{i}^{t} \in \targetdata$. 
In the conventional ML (\emph{inductive learning}) setting, we find the best model parameters $\srcparam$ by training on the source dataset $\sourcedata$. 
However, in practice, training and test data distribution may shift, and a fixed model $f(\cdot; \srcparam)$ will result in poor performance \cite{Hendrycks2019BenchmarkingNN}. 
To address this issue, test-time adaptation (TTA), under \emph{transductive learning} setting, has been proposed \cite{Wang2021TentFT}.  
Specifically, TTA first obtains a $\sourcemodel$ learned from the source training set $\sourcedata$ or downloaded from an online source and then adapts to test data $\targetdata$ during inference.
In this case, TTA can  characterize the distribution of $\targetdata$, thereby boosting the performance. 

\vspace{-1mm}
\subsection{Leveraging the Test Batch in TTA}
\vspace{-2mm}

TTA techniques are typically used when the test data is processed batch by batch in an online manner.\footnote{We exclude single-sample TTA methods (e.g., \citet{Zhang2021MEMOTT}) which usually perform worse than batched version. In addition, they make independent predictions for each test data, avoiding the risks we discuss in this paper. } 
Let $\batchdata = \{\mathbf{x}_{i}^{t}\}_{i=1}^{\nbatch} \subseteq \targetdata$ be a subset (i.e., \emph{batch}) of test data with size $\nbatch$, and $\premodel$ be the pre-adapted model which is going to perform adaption immediately. 
Initially, we set $\preparam = \srcparam$.
At each iteration, a new batch of test data $\batchdata$ is available; parts of the model parameter are updated accordingly, and the final predictions are based on the updated model.
Here we introduce two mainstream techniques: 

\vspace{-1mm} \noindent \textbf{Test-time Batch Normalization (TeBN).} Batch Norm (BN)~\citep{ioffe2015batch} normalizes the input features by batch mean and variance, reducing the internal covariate shifts in DNN. 
\citet{Nado2020EvaluatingPB} replace the normalization statistics $\{\mu, \sigma^2\}$ on the training set by the statistics of the test batch $\batchdata$, denoted as $\bntheta(\mathbf{X}_{B}^{t}) := \{\mu(\mathbf{X}_{B}^{t}), \sigma^2(\mathbf{X}_{B}^{t})\}$, where $\bntheta$ is BN statistics.
This can help the model generalize better to the unseen and shifted data distribution. 


\vspace{-1mm} \noindent \textbf{Self-Learning (SL).} SL updates part of model parameters $\adatheta$ by the gradient of loss function $\mathcal{L}_{\text{TTA}}$. Some methods like \textbf{TENT}~\cite{Wang2021TentFT}  minimize the entropy of prediction distribution on a batch of test samples, where loss is $\mathcal{L}_{\text{TTA}}(\mathbf{X}_{B}^{t})$.
We then denote the remaining parameters as $\fixtheta := \srcparam \setminus (\adatheta\cup \bntheta)$, which stay fixed.
In other methods like pseudo-labeling (PL), the loss can be formulated as $\mathcal{L}_{\text{TTA}}(\mathbf{X}_{B}^{t},\bypls)$, where $\bypls = \{\ypl\}_{i=1}^{N_{Bt}}$ denote the pseudo-labels.
In the standard PL methods, the pseudo-labels $\bypls$ can be directly predicted by the teacher, known as \textbf{Hard PL}~\cite{lee2013pseudo}, or by predicting the class probabilities, known as \textbf{Soft PL}~\cite{lee2013pseudo}. 
Later, \textbf{Robust PL} proposed by \citet{rusak2022if} and \textbf{Conjugate PL} developed by \citet{Goyal2022TestTimeAV} further improve the pseudo-labels with more advanced TTA loss.  
Notably, all \textbf{SL} methods usually achieve the best performance when $\adatheta$ equals the affine transformations of BN layers~\cite{rusak2022if}. 

\vspace{-1mm}
\subsection{Related Work}
\vspace{-2mm}

Previous work on conventional ML risks has focused on adversarial attacks, such as evasion attacks~\cite{Biggio2013EvasionAA, Goodfellow2015ExplainingAH, Carlini2017TowardsET}, which perturb test data to cause mispredictions by the model during inference. However, our attack on TTA differs in that the malicious samples we construct can target benign and unperturbed data.
Another form of ML risk is data poisoning~\cite{biggio2012poisoning, Koh2017UnderstandingBP, gu2017badnets}, which involves injecting malicious data into training samples, causing models trained on them to make incorrect predictions. Our attack differs in that we only assume access to unlabeled test data, making it easier to deploy in real-world environments. 

To the best of our knowledge, this is the first work to uncover the adversarial risks of TTA methods. We discuss more related works in Appendix \ref{append:related}.

\vspace{-3mm}
\section{Threat Model}
\vspace{-1mm}
As no previous literature has studied the vulnerabilities of TTA, we start by discussing the  adversary's objective, capabilities, and knowledge for our attack. 
We consider a scenario in which a victim gets a source model, either trained on source data or obtained online, and seeks to improve its performance using TTA on a batch of test data. 

\vspace{-1mm} \noindent\textbf{Adversary's Objective.} The objective of \textbf{\textit{Distribution Invading Attack (DIA)}} is to interfere with the performance of the post-adapted model in one of the following ways: (1) \textbf{targeted attack}: misclassifying a crucial targeted sample as a specific label, (2) \textbf{indiscriminate attack}: increasing the overall error rate
on benign data in the same batch, or (3) \textbf{stealthy targeted attack}: achieving targeted attack while maintaining accuracy on other benign samples.

\vspace{-1mm} \noindent\textbf{Adversary's Capabilities and Additional Constraints.} 
The attacker can craft and upload a limited number of malicious samples to the test batch during inference.
Since the DIA does not make any perturbations on the targeted samples, in our main evaluation, we do not require a constraint for malicious data as long as it is a valid image. 
However, bypassing the manual inspection is sometimes worthwhile; the adversary may also construct camouflaged malicious samples.
Concretely, we consider two constraints on attack samples: (1) $\ell_\infty$ attacks \cite{Goodfellow2015ExplainingAH}, the most common practice in the literature; (2) adversarially generated corruptions (e.g., snow) \cite{Kang2019TestingRA}, which simulates the target distribution. 

\vspace{-1mm} \noindent\textbf{Adversary's  Knowledge.} We consider a white-box setting where the DIA adversary knows the pre-adapted model parameters $\preparam$ and has read-only access to benign samples in the test batch (e.g., a malicious insider is involved).  
However, the adversary has no access to the training data or the training process.
Furthermore, our main attacking methods (used in most experiments) do not require the knowledge of which TTA methods the victim will deploy.

\vspace{-1mm}

\section{Distribution Invading Attack}
\label{sec:diamethod}
\vspace{-0.5mm}

In this section, we first identify the detailed vulnerabilities (Section \ref{subsec:indentify}), then formulate Distribution Invading Attack as a general bilevel optimization problem (Section \ref{sec:bilevel}), and finally discuss constructing malicious samples (Section \ref{subsec:Constructing}).

\subsection{Indentifing the Vulnerabilities of TTA}
\label{subsec:indentify}
\vspace{-2mm}
 The risk of TTA stems from its transductive learning paradigm, where the predictions on test samples are no longer independent of each other. In this section, we detail how specific approaches used in TTA expose security vulnerabilities. 

\vspace{-1mm} \noindent\textbf{Re-estimating Batch Normalization Statistics.}
Most existing TTA methods \cite{Nado2020EvaluatingPB, Wang2021TentFT, rusak2022if} adopt test-time Batch Normalization (\textbf{TeBN}) as a fundamental approach for mitigating distribution shifts when the source model is CNN with BN layers.
We denote the input of $l$th BN layer by $\mathbf{z}_{l}$, and test-time BN statistics for each BN layer can be computed by 
$
\{\mu \leftarrow \mathbb{E} [\mathbf{z}_{l}], \sigma^2 \leftarrow \mathbb{E} [(\mu-\mathbf{z}_{l})^2]\}
$
through the forward path in turn, where the expectation (i.e., average) $\mathbb{E}$ is applied channel-wise on the input. 
When calculating BN statistics $\bntheta$ over a batch of test data, any perturbations to some samples in the batch can affect the statistics ${\mu, \sigma^2}$ layer by layer and potentially alter the outputs
(i.e., $\mathbf{z}_{l}^\text{out} := (\mathbf{z}_{l} - \mu)/ \sigma$) 
of the other samples in the batch. 
Hence, adversaries can leverage this property to design Distribution Invading Attacks. 

\vspace{-1mm} \noindent\textbf{Parameters Update.}
To further adapt the model to a target distribution, existing methods often update part of model parameters $\theta_{\A}$ by minimizing unsupervised objectives
defined on test samples~\cite{Wang2021TentFT, rusak2022if, Goyal2022TestTimeAV}. 
The updated parameter can be computed by: 
$\theta^*_{\A} =  \argmin_{\theta_{\A}} \mL_{\text{TTA}}(\batch; \theta_{\A})$
Hence, malicious samples inside $\batch$ may perturb the model parameters $\theta^*_{\A}$, leading to incorrect predictions later.

\vspace{-1mm}
\subsection{Formulating DIA as a Bilevel Optimization Problem}
\label{sec:bilevel}
\vspace{-2mm}

We formulate the DIA as an optimization problem to exploit both vulnerabilities mentioned above. 
Intuitively, we want to craft some malicious samples $\mal := \{\mathbf{x}_{mal, i}^{t}\}_{i=1}^{N_{m}}$ to achieve an attack objective (e.g., misclassifying  a targeted sample). Then, the test batch $\batch$ comprises $\mal$ and other benign samples $\benigns$.
The pre-adapted model parameter $\preparam$ is composed of parameters $\adatheta$ that will update, BN statistics $\bntheta$, and other fixed parameters $\fixtheta$ (i.e., $\preparam := \adatheta \cup \bntheta  \cup \fixtheta$).
Here, we use $\gobject$ to denote the general adversarial loss, and the problem for the attacker can be formulated as the following bilevel optimization: 

\vspace{-5mm}
\begin{equation}
    \centering
    \label{eq:general_attack}
    \min_{\mal} \bL(f(\cdot \ ;\theta^*(\batchdata)))
\end{equation}

\vspace{-5mm}
\begin{small}
\begin{equation} 
    \centering
    \label{eq:condition}
    \begin{aligned}
        s.t. \ &\batch = \mal \cup \benigns; \quad
         \theta'_{\B} = \{\mu(\batch), \sigma^2(\batch)\};  \\
         &\theta^*_{\A} =  \argmin_{\theta_{\A}} \mL_{\text{TTA}}(\batch; \theta_{\A}, \theta'_{\B}, \fixtheta);\\
        &\theta^*(\batch) = \adatheta^* \cup \bntheta'  \cup \fixtheta;
    \end{aligned}
\end{equation}
\end{small}

\vspace{-3mm}
\noindent where $\theta'_{\B}$ is the updated BN statistics given the test batch data, $\theta^*_{\A}$ is the parameter that is optimized over TTA loss, and $\theta^*(\batchdata)$ is the optimized model parameters containing $\theta^*_{\A}$, $\theta'_{\B}$, and other fixed parameters $\fixtheta$. 
In most cases, test-time adaptation methods perform a single-step gradient descent update \cite{Wang2021TentFT}, and the inner optimization for TTA simplifies to $\theta^*_{\A} = \theta'_{\A} =  \theta_{\A}-\partial \mL_{\text{TTA}}(\batch)/ \partial \theta_{\A}$.
Now, we discuss how we design the specific adversarial loss $\gobject$ for achieving various objectives.




\vspace{-1mm} \noindent\textbf{Targeted Attack.} We aim to cause the model to predict a specific incorrect targeted label $\ytgt$ for a targeted sample $\xtgt \in \benigns$. Thus, the objective can be formulated as: 

\vspace{-2mm}
\begin{equation} 
    \centering
    \label{eq:mainproblem} 
    \hatmal := \argmin_{\mal} \  \mL (f ( \xtgt ; \theta^*(\batchdata)), \ytgt) 
\end{equation}
\vspace{-5mm}

\noindent where $\mL$ is the cross-entropy loss.

\vspace{-1mm} \noindent\textbf{Indiscriminate Attack.} The objective turns to degrade the performance of all benign samples as much as possible. Given the correct labels of benign samples $\benignslabel$, we define the goal of indiscriminate attack as follows: 
\vspace{-2mm}
\begin{small}
\begin{equation} 
\label{eq:indis}
\hatmal := \argmin_{\mal} \  -\mL (f( \benigns ;\theta^*(\batchdata)), \benignslabel).
\end{equation}
\end{small}
\vspace{-3mm}


\vspace{-1mm} \noindent\textbf{Stealthy Targeted Attack.} In some cases, when performing \textbf{targeted attack}, the performance of the other benign samples $\benign$, which is the whole test batch excluding malicious and targeted data, may drop.  
A solution is conducting targeted attacks and maintaining the accuracy of other benign data simultaneously, which is: 

\vspace{-4mm}
\begin{small}
\begin{equation} 
\label{eq:per}
\begin{aligned}
 \hatmal & := \argmin_{\mal}  \  \mL (f( \xtgt ;\theta^*(\batchdata)), \ytgt)  + \\
 &\omega * \mL (f(  \benign ;\theta^*(\batchdata)), \benignlabel).
\end{aligned}
\end{equation}
\end{small}

\vspace{-3mm}
We introduce a new weight term $\omega$ to capture the trade-off
between these two objectives.


\subsection{Constructing Malicious Inputs via Projected Gradient Descent}
\label{subsec:Constructing}

\sethlcolor{lightgray} 
\begin{algorithm}[t]
   \caption{for constructing Distribution Invading Attack}\label{alg:attackBilevel}
    \begin{algorithmic}[1]
    \STATE \textbf{Input:} 
    Pre-adapted model parameters $\preparam = \adatheta \cup \bntheta  \cup \fixtheta$, test batch $(\mathbf{X}^t_{B};\mathbf{y}^t_B)$ which contains malicious samples $\mal$ and benign samples $\mathbf{X}^t_{B\setminus mal}$, targeted samples $\xtgt$ and incorrect targeted label $\ytgt$, 
     attack learning rates $\alpha$, 
     constraint $\epsilon$, 
     number of steps $N$, 
     TTA update rate: $\eta$,
     perturbation $\dm$=$\boldsymbol{0}$ \alglinelabel{algin-input} 
   \STATE \textbf{Output:}  Perturbed malicious input $\mal + \dm $ \alglinelabel{algin-output} 
   \STATE  \textbf{for}  step $=1, 2, \dots,  N$ \textbf{do}:
   \STATE \quad  $\batch \leftarrow (\mal + \dm) \cup \mathbf{X}^t_{B\setminus mal} $\alglinelabel{algin-dataup} 
   \STATE \quad $\theta'_{\B} \leftarrow \{\mu(\batch), \sigma^2(\batch)\}$  \alglinelabel{algin-bn} 
   \STATE \quad (Optional) $\theta'_{\A} \leftarrow  \theta_{\A} - \eta \cdot \partial \mL_{\text{TTA}}(\batch)/ \partial \theta_{\A}$ \\   \quad
   \hl{\textit{\# $\theta'_{\A}  \approx \theta_{\A}$ in the single-level version.}}
   \alglinelabel{algin-tta} 
   \STATE \quad  $\theta^* \leftarrow \adatheta' \cup \bntheta'  \cup \fixtheta$    \alglinelabel{algin-paraup} 
   \STATE \quad  $\dm \leftarrow  \Pi_{\epsilon}$ ($\dm - \alpha \cdot \mathrm{sign}(\nabla_{\dm} \gobject))$ \\  \quad
   \hl{\textit{\# $\bL$ is chosen from Eq. ({\ref{eq:mainproblem}}), Eq. ({\ref{eq:indis}}), or Eq. ({\ref{eq:per}})}} \alglinelabel{algin-malup}
   \STATE   \textbf{end for}
   \STATE \textbf{return} $\hatmal = \mal + \dm $
\end{algorithmic}
\end{algorithm}


We solve the bilevel optimization problems defined in the last section via iterative \emph{projected gradient descent}, summarized in Algorithm \ref{alg:attackBilevel}. 
Our solution generalizes across the three adversary's objectives, where $\gobject$ can be replaced by Eq. (\ref{eq:mainproblem}), Eq. (\ref{eq:indis}), or Eq. (\ref{eq:per}).   
We follow the general projected gradient descent method but involve TTA methods. Concretely, Line \ref{algin-dataup} updates the test batch, and Line \ref{algin-bn} computes the test-time BN. 
Then, Line \ref{algin-tta} and Line \ref{algin-paraup} perform parameter updates.
In Line \ref{algin-malup}, we compute the gradient toward the objective (three attacks we discussed previously) and update the perturbation $\dm$ with $\alpha$ learning rate.
Since, in most cases, we do not consider a constraint for the malicious images, the projection $\epsproj$ is to ensure the images are valid in the [0,1] range. 
For $\ell_\infty$ constrained DIA attacks, we will also leverage the $\epsproj$ to clip $\dm$ with constraint $\epsilon$ (e.g., $\epsilon=8/255$).
Finally, we get the optimal malicious output samples $\hatmal = \mal + \dm $.

\noindent \textbf{Implementation.} While Algorithm \ref{alg:attackBilevel} is intuitive, the gradient computation can be inconsistent and incur a high computational cost due to the bilevel optimization.
Furthermore, we notice that TTA methods only perform a short and one-step update on $\theta_{\A}$ parameters at each iteration (Line \ref{algin-tta}).
Therefore, we approximate  $\theta'_{\A} \approx \theta_{\A}$ in most experiments, which makes the inner TTA gradient update (Line \ref{algin-tta}) optional.\footnote{Line \ref{algin-tta} is necessary for models without BN layers.} 
Since the re-estimated $\bntheta'$ does not involve any inner gradient computation, we then decipher the problem as a single-level optimization. 
Our empirical evaluation indicates the effectiveness of our method, reaching comparable or even higher results (presented in Appendix \ref{sec:bilevelres}). 
An additional benefit of our method is we no longer need to know which TTA methods the victim will choose. We also provide some theoretical analysis in Appendix \ref{append:UBN}


\section{Evaluation of Distribution Invading Attacks }
\label{sec:dia}

In this section, we report the results of Distribution Invading Attacks. We present the experimental setup in Section \ref{sec:expset}, discuss the main results for targeted attacks in Section \ref{sec:evalres}, present the results for indiscriminate and stealthy targeted attacks in Section \ref{sec:twoatt}, and consider extra constraints in Section \ref{sec:addcon}. Further results are presented in Appendix \ref{sec:adddia}.

\vspace{-1mm}
\subsection{Experimental Setup}
\label{sec:expset}
\vspace{-1mm}
\textbf{Dataset \& Architectures.} 
We evaluate our attacks on the common distribution shift benchmarks: CIFAR-10 to CIFAR-10-C, CIFAR-100 to CIFAR-100-C, and ImageNet to ImageNet-C~\citep{Hendrycks2019BenchmarkingNN} with medium severity (level 3). 
We primarily use ResNet-26 \citep{He2016DeepRL}  for CIFAR-C\footnote{CIFAR-C denotes both CIFAR-10-C and CIFAR-100-C.}, and ResNet-50 for ImageNet-C. 

\vspace{-1mm} \noindent\textbf{Test-time Adaptation Methods.} We select six TTA methods, which are \textbf{TeBN}~\cite{Nado2020EvaluatingPB}, \textbf{TENT}~\cite{Wang2021TentFT},  \textbf{Hard PL}~\cite{lee2013pseudo},  \textbf{Soft PL}~\cite{lee2013pseudo},
\textbf{Robust PL}~\cite{rusak2022if},
 and \textbf{Conjugate PL}~\cite{Goyal2022TestTimeAV}.
We follow the settings of their experiments, where the batch size is \textbf{200}, and all other hyperparameters are also default values. 
As a result, all TTA methods significantly boost the \textbf{corruption accuracy} (i.e., the accuracy on benign corrupted test data like ImageNet-C), which is shown in Appendix \ref{sec:ttap}. 

\vspace{-1mm} \noindent\textbf{Attack Settings (Targeted Attack).} 
We consider each test batch as an \textit{individual} attacking trial where we randomly pick one targeted sample with a targeted label.\footnote{Unless otherwise specified, most experiments in this paper are conducted using targeted attack. }
The attacker can inject a small set of malicious data inside this batch without restrictions as long as their pixel values are valid. 
In total, there are 750 trials for CIFAR-C and 375 trials for ImageNet-C. 
We estimate our attack effectiveness by \textbf{attack success rate (ASR)} averaged across all trials. 
Note that except \textbf{TeBN}, for each trial, TTA methods use the current batch to update $\adatheta$, and send it to the next trial. 
Hence, $\preparam$ is different for different trials. Further experimental setup details are in Appendix \ref{sec:setup}. 

\subsection{Main Results of Targeted Attack}
\label{sec:evalres}

\begin{table*}[t]
\centering
\vspace{-4mm}
\caption{Attack success rate of Distribution Invading Attack (targeted attacks) across benchmarks and TTA methods.  $N_m$ refers to the number of malicious data, where the batch size is \textbf{200}.  
}
\resizebox{0.85\textwidth}{!}{%
\begin{tabular}{clcccccc}
\toprule
\textbf{Dataset} & $N_m$ & \textbf{TeBN(\%)} & \textbf{TENT(\%)} & \textbf{Hard PL(\%)} & \textbf{Soft PL(\%)} & \textbf{Robust PL(\%)} & \textbf{Conjugate PL(\%)} \\ \midrule
\multirow{3}{*}{\textbf{\begin{tabular}[c]{@{}c@{}}CIFAR-10-C\\ (ResNet26)\end{tabular}}}  
& 10 (5\%) & 25.87 & 23.20 &   25.33 &   23.20 & 24.80 &      23.60  \\
& 20 (10\%) & 55.47 & 45.73 &   48.13 &   47.47 & 49.47 &      45.73 \\
& 40 (20\%)& 92.80 & 83.87 &   84.27 &   82.93 & 86.93 &      85.47 \\ \hline
\multirow{3}{*}{\textbf{\begin{tabular}[c]{@{}c@{}}CIFAR-100-C\\ (ResNet26)\end{tabular}}} 
& 10 (5\%) & 46.80 & 26.40 &   31.20 &   27.60 & 32.13 &      26.13 \\             
& 20 (10\%) & 93.73 & 72.80 &   87.33 &   78.53 & 82.93 &      71.60  \\             
& 40 (20\%) & 100.00 & 100.00 &   99.87 &  100.00 & 99.87 &     100.00\\ \hline
\multirow{3}{*}{\textbf{\begin{tabular}[c]{@{}c@{}}ImageNet-C\\ (ResNet50)\end{tabular}}}  
& 5 (2.5\%) &  80.80 & 75.73  &   69.87 &   62.67 & 66.40 &      57.87   \\ 
& 10 (5\%) & 99.47   & 98.67 &   96.53 &   94.13 & 96.00 &      92.80 \\
& 20 (10\%) & 100.00 & 100.00 &  100.00 &  100.00 & 100.00 &     100.00 \\ \bottomrule                                              
\vspace{-5mm}
\end{tabular}%
}
\label{tab:maintable}
\end{table*}

\vspace{-1mm}
\noindent\textbf{Distribution Invading Attack achieves a high attack success rate (ASR) across benchmarks and TTA methods (Table \ref{tab:maintable}). }
We select $N_m$ = $\{10, 20, 40\}$ of malicious samples out of 200 data in a batch for CIFAR-C and $N_m$ = $\{5, 10, 20\}$ for ImageNet-C. 
By constructing 40 malicious samples on CIFAR-10-C and CIFAR-100-C, our proposed attack can shift the predictions of the victim sample to a random targeted label in more than 82.93\% trials. 
For ImageNet-C, just 10 malicious samples are sufficient to attack all TTA methods with an attack success rate of more than 92.80\%. 
The ASR of \textbf{TeBN} are higher than other TTA methods since the attacks here only exploited the vulnerabilities from re-estimating batch normalization statistics. 
We demonstrate that further parameter updates of other TTA methods cannot greatly alleviate the attack effectiveness. 
For example, the drop of ASR is less than 6.7\% for ImageNet-C with 10 malicious samples.

\begin{figure}[t]
\setlength\abovecaptionskip{0pt}
\setlength\belowcaptionskip{-14pt}
  \centering
  \vspace{-2mm}
    \includegraphics[width=0.30\textwidth]{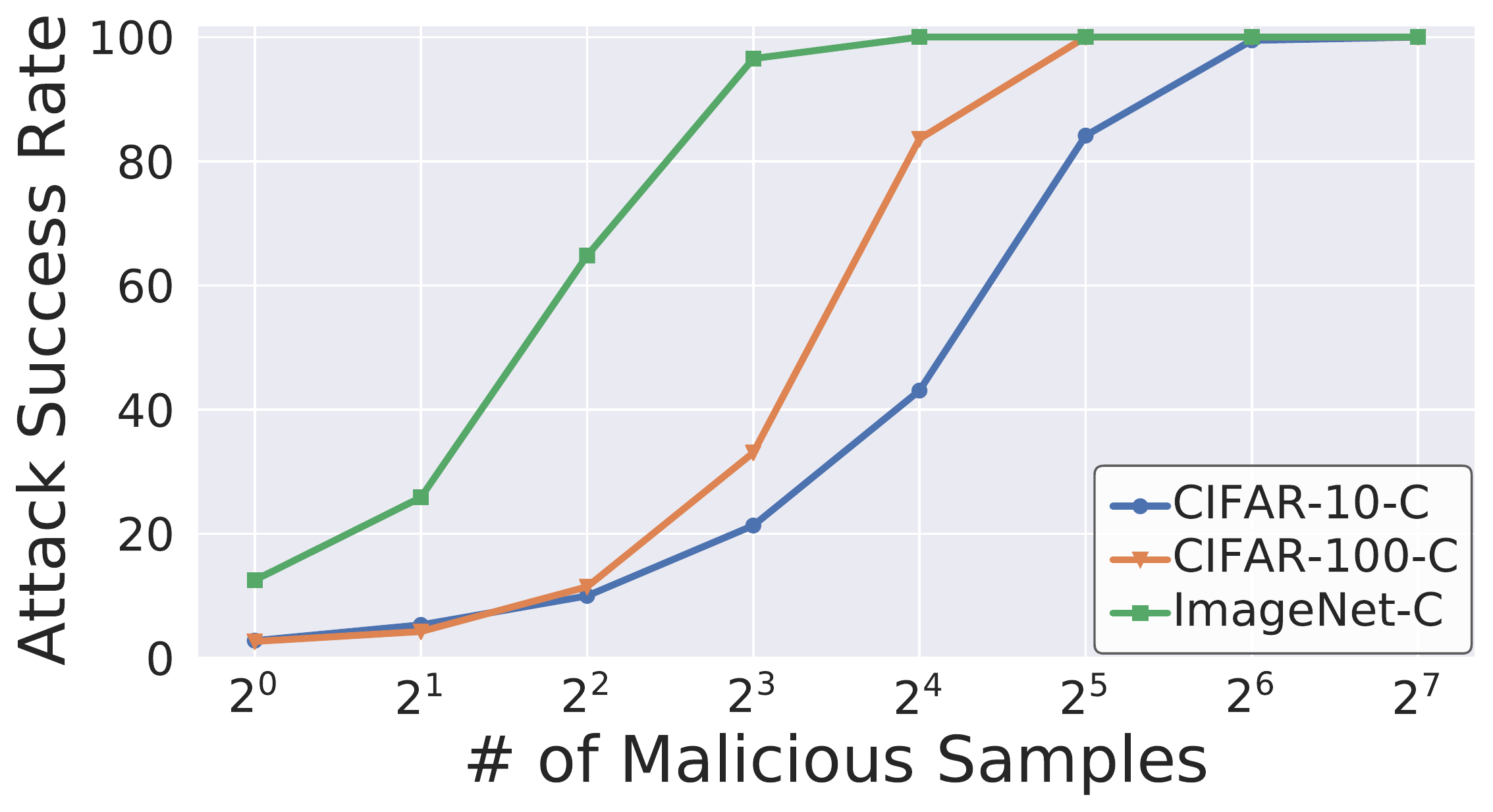}
    \vspace{-5mm}
    \caption{ Success rate of our proposed attack across numbers of malicious samples from 1 to 128 (0.5\% to 64\%). [TTA: \TeBN] }
    \label{Fig:mal_smaple}
    \vspace{-7mm}
  \end{figure}

\vspace{-1mm} \noindent\textbf{DIA reaches near-100\% ASR, with 64, 32, and 16 malicious samples for  CIFAR-10-C, CIFAR-100-C, and ImageNet-C, respectively (Figure \ref{Fig:mal_smaple}).}
We then evaluate our attack with more comprehensive experiments across the number of malicious samples. Specifically, 1 to 128 malicious samples are considered, which is 0.5\% to 64\%  of batch size. 
\textbf{TeBN} is selected as the illustrated TTA method (Appendix \ref{sec:diasample} presents more methods). Generally, the attack success rate increases when the attacker can control more samples. 
For CIFAR-10-C, we observe that the attacker has to manipulate $\sim$64 (32\% of the batch size) samples to ensure victim data will be predicted as a targeted label with a near-100\% chance. 
Although 32\% seems to be a relatively strong assumption compared to conventional poisoning attacks, it is worth noting that DIA only requires perturbing test data which might not be actively monitored.

\begin{table}[t]
\vspace{-6mm}
    \caption{ Effectiveness of the Distribution Invading Attack across various model architectures and data augmentations. [CIAFR-C: $N_m$=40; ImageNet-C: $N_m$=10]. Table \ref{tab:arch_all} presents full results.}
    \centering
    \resizebox{0.48\textwidth}{!}{%
    \begin{tabular}{@{}ccccc@{}}
    \toprule
    \textbf{Dataset} & \textbf{Architectures} & \textbf{TeBN(\%)} & \textbf{TENT(\%)} & \textbf{Hard PL(\%)} \\ \midrule
    \multirow{3}{*}{\textbf{CIFAR-10-C}}  
    & ResNet-26      & 92.80 & 83.87 &   84.27  \\
    & VGG-19      & 79.07 & 65.33 &   67.47  \\
    & WRN-28   & 93.73 & 89.60 &   90.80  \\ \midrule
    \multirow{3}{*}{\textbf{CIFAR-100-C}}  
    & ResNet-26       & 100.00 & 100.00 &   99.87  \\
    & VGG-19        &  88.00 &  82.93 &   86.53 \\
    & WRN-28     &  97.47 &  83.60 &   87.07  \\
                                          \toprule
    \textbf{Dataset} & \textbf{Augmentations} & \textbf{TeBN(\%)} & \textbf{TENT(\%)} & \textbf{Hard PL(\%)} \\ \midrule
    \multirow{3}{*}{\textbf{ImageNet-C}} 
     & Standard          & 99.47 & 98.67 &   96.53  \\
     & AugMix             & 98.40 & 96.00 &   93.60   \\
     & DeepAugment        & 96.00 & 94.67 &   91.20   \\                                  
    \bottomrule
    \end{tabular}%
    }
    \label{tab:arch}
    \end{table}

\vspace{-1mm} \noindent\textbf{DIA also works across model architectures and data augmentations (Table \ref{tab:arch}).} 
Instead of ResNet-26, we also consider two more common architectures: VGG \cite{Simonyan2015VeryDC} with 19 layers (VGG-19), and  Wide ResNet \cite{Zagoruyko2016WideRN} with a depth of 28 and a width of 10 (WRN-28) on CIFAR-C dataset. 
Our proposed attack is also effective across different architectures, despite the attack success rates suffering some degradations ($<$20\%) for VGG-19. 
We hypothesize that the model with more batch normalization layers is more likely to be exploited by attackers, where ResNet-26, VGG-19, and WRN-28 contain 28, 16, and 25 BN layers, respectively. We further refine the statement through more experiments in Appendix \ref{append:arch}. 
In addition, we also select two more data augmentation methods, AugMix \cite{Hendrycks2020AugMixAS} and DeepAugment \cite{Hendrycks2021TheMF}, on top of ResNet-50, which is known to improve the model generalization on the corrupted dataset. Strong data augmentation techniques provide mild mitigations against our attacks (with 10 malicious samples) but still mistakenly predict the targeted data in more than 91\% of trials.

\vspace{-2mm}
\subsection{ Alternating Attacking Objective}
\label{sec:twoatt}
Our previous DIA evaluations focus on targeted attacks. However, as we discussed in Section \ref{sec:bilevel}, malicious actors can also leverage our attacking framework to achieve alternative goals by slightly modifying the loss function. 

\vspace{-2mm}
\subsubsection{ Indiscriminate Attack } 
The first alternative objective is to degrade the performance on all benign samples, which is done by leveraging Eq. (\ref{eq:indis}). Here, we adopt the \textbf{corruption error rate} on the benign corrupted dataset as the attack evaluation metric. 


\begin{table}[t]
\centering
\vspace{-3mm}
\caption{Average benign corruption error rate of TTA when deploying DIA indiscriminate attack. Table \ref{tab:all_all} presents full results.}
\resizebox{0.48\textwidth}{!}{%
\begin{tabular}{clccc}
\toprule
\textbf{Dataset} & \textbf{ $N_m$} & \textbf{TeBN(\%)} & \textbf{TENT(\%)} & \textbf{Hard PL(\%)}   \\ \midrule
\multirow{2}{*}{\textbf{\begin{tabular}[c]{@{}c@{}}CIFAR-10-C\\ (ResNet-26)\end{tabular}}}  
& 0 (0\%)   & 10.73 & 10.47 &   11.05 \\
& 40 (20\%)  & 28.02 & 27.01 &   27.91 \\
 \hline
\multirow{2}{*}{\textbf{\begin{tabular}[c]{@{}c@{}}CIFAR-100-C\\ (ResNet-26)\end{tabular}}} 
& 0 (0\%)   &  34.52 & 33.31 &   34.66 \\
& 40 (20\%) &  58.41 & 54.44 &   56.59\\
\hline
\multirow{2}{*}{\textbf{\begin{tabular}[c]{@{}c@{}}ImageNet-C\\ (ResNet-50)\end{tabular}}}  
& 0 (0\%) & 47.79 & 45.45 &   43.42 \\
& 20 (10\%) & 79.03 & 75.01 &   70.26 \\
\bottomrule                   
\end{tabular}%
}
\vspace{-5mm} 
\label{tab:all}
\end{table}

\vspace{-1mm} \noindent\textbf{By injecting a small set of malicious samples, the error rate grows (Table \ref{tab:all}).}
Besides attack effectiveness, we report the error rate for 0 malicious samples (which stands for no attacks) as the standard baseline. 
Our attack causes the error rate on benign samples to rise from $\sim$11\% to  $\sim$28\% and from $\sim$34\% to  $\sim$56\% for the CIFAR-10-C and CIFAR-100-C benchmarks, respectively. 
Furthermore, only 20 malicious samples (10\%) for ImageNet-C boost the error rate to more than 70\%. Since all benign samples remain unperturbed, the increasing error rate demonstrates the  extra risks of TTA methods compared to the conventional ML. 


\subsubsection{Stealthy Targeted Attack}

In Table \ref{tab:maintable}, $\sim$40 samples are needed for the CIFAR-C dataset to achieve a high attacking performance. Thus, the malicious effect might also affect the predictions for other benign samples, resulting in losing attacking stealth. We adopt Eq. (\ref{eq:per}) to simultaneously achieve targeted attacks and maintain corruption accuracy.  Here, we use the \textbf{corruption accuracy degradation} on benign samples to measure the stealthiness. 


\begin{figure}[t]
  \centering
 \vspace{-4mm}
  \begin{tabular}{cccc}
    \includegraphics[width=0.21\textwidth]{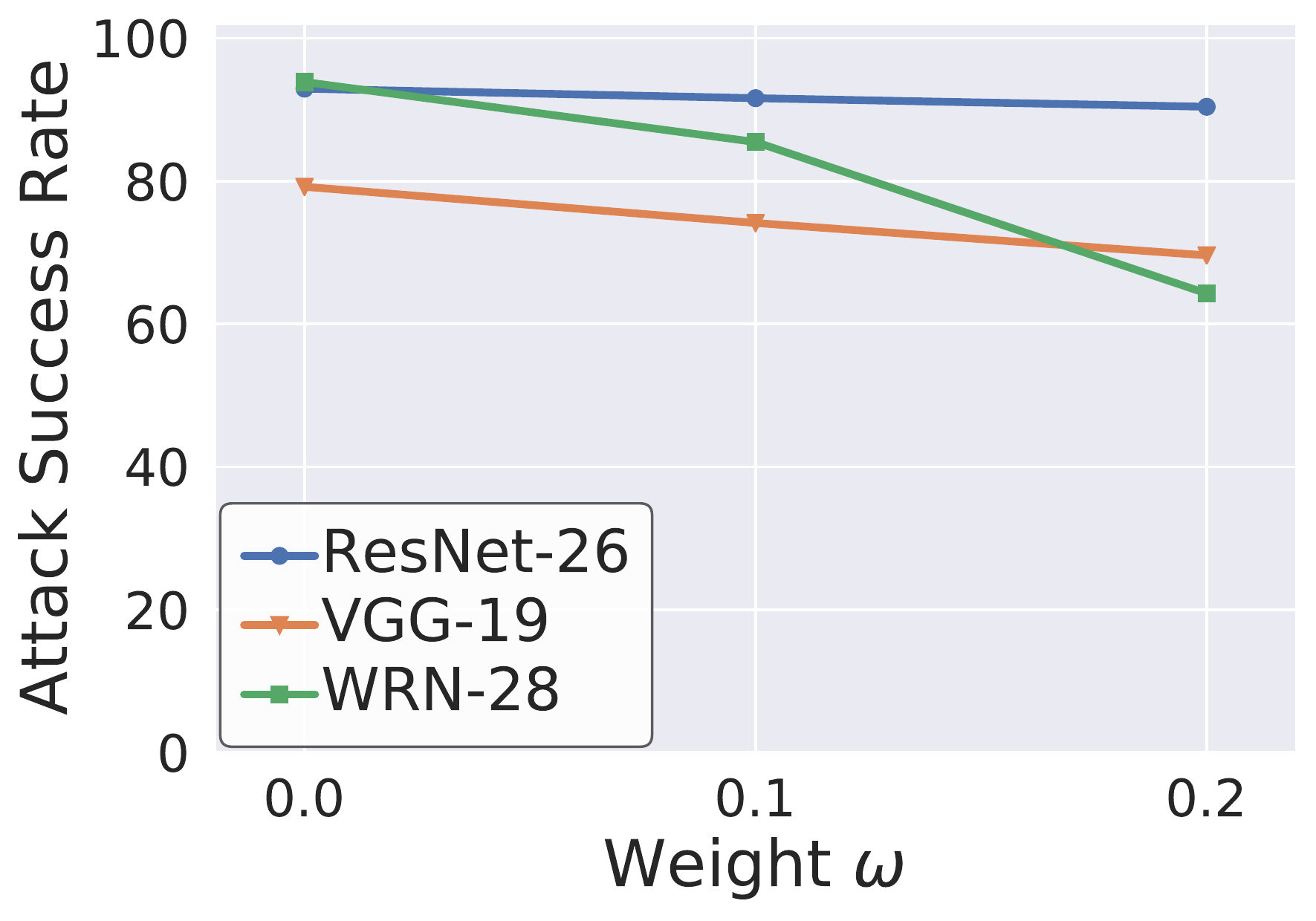} & 
   \includegraphics[width=0.21\textwidth]{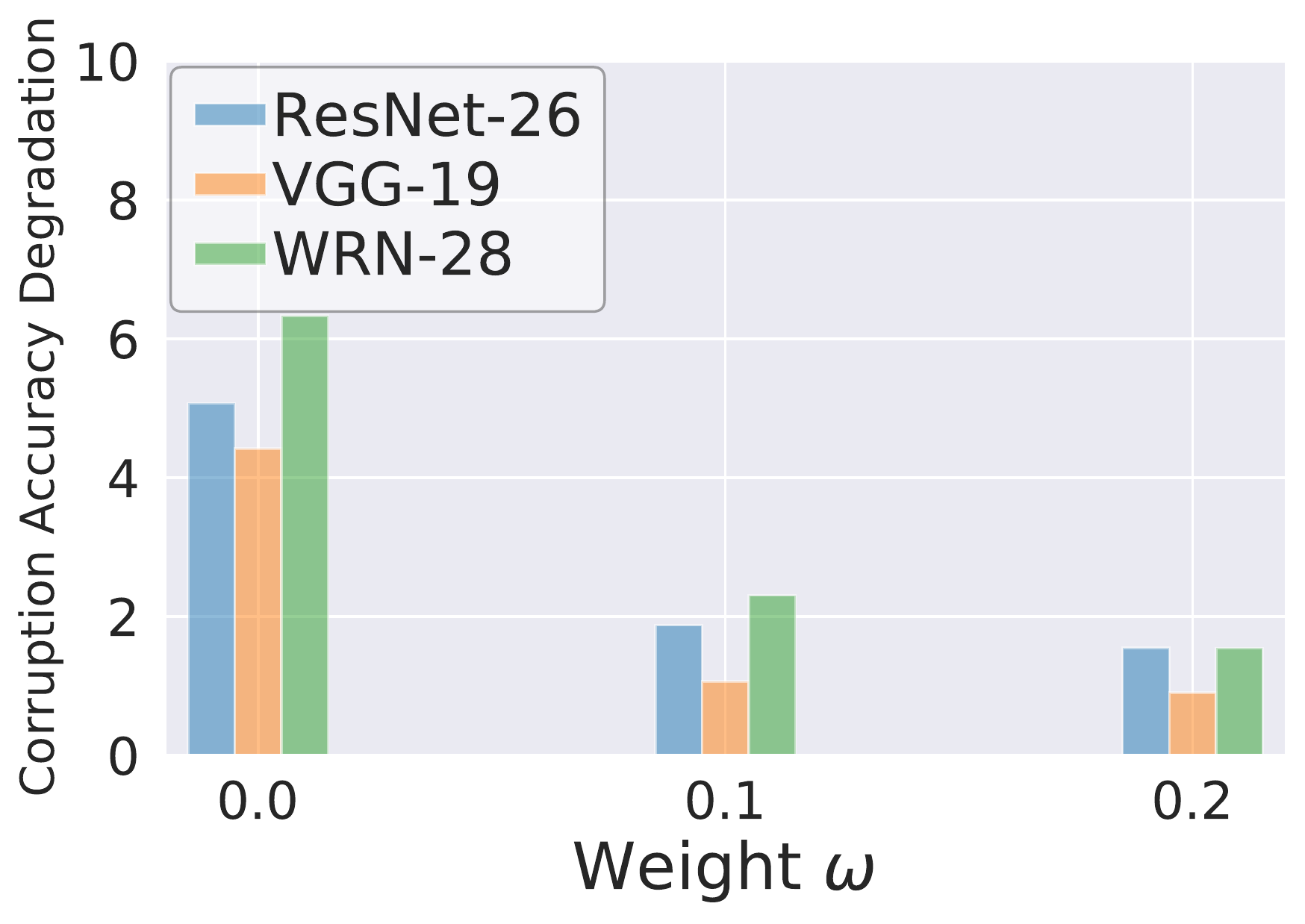} \\
  \end{tabular}
  \vspace{-4mm}
    \caption{ Adjusting the weight $\omega$ achieves a high attack success rate (line) and a high benign corruption accuracy (bar). [Benchmark: CIFAR-10-C, $N_m$=40, TTA method: \textbf{TeBN}] }
    \label{Fig:per}
    \vspace{-2mm}
  \end{figure}


\vspace{-1mm} \noindent\textbf{When $\omega = 0.1$, stealthy DIA can both achieve a high attack success rate and maintain corruption accuracy (Figure \ref{Fig:per}).} We select CIFAR-10-C with 40 malicious data and \textbf{TeBN} method, where $\omega = \{0, 0.1, 0.2\}$. 
We observe that if $\omega = 0.1$, the corruption accuracy degradation drops to $\sim$2\%. 
At the same time, the ASR remains more than 75\%. Appendix \ref{append:per} has more results on CIFAR-100-C.

\vspace{-2mm}
\subsection{ Additional Constraints on Malicious Images}
\label{sec:addcon}


\begin{figure}[t]
  \centering
    \begin{small}
  \begin{tabular}{cccc} 
  \includegraphics[width=0.09\textwidth]{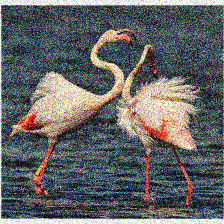} & 
    \includegraphics[width=0.09\textwidth]{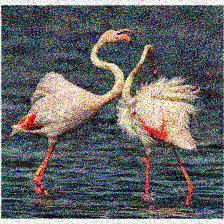} & 
  \includegraphics[width=0.09\textwidth]{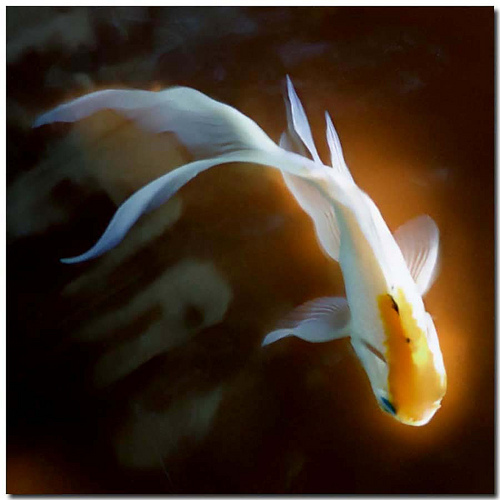} & 
    \includegraphics[width=0.09\textwidth]{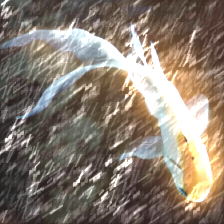} \\
     (a)
    & (b)
    & (c) 
    & (d)
  \\ 
  \end{tabular}
  \end{small}
  \vspace{-2mm}
  \caption{(a) Test data from ImageNet-C with Gaussian noise. (b) $L_\infty$ constrained DIA on test data ($\epsilon$=$8/255$). (c) A clean data from the ImageNet validation set. (d) Snow attack on the clean data.}
  \vspace{-2mm}
  \label{Tab:IN_Constrained}
  \end{figure}

We further consider the stealth of malicious samples to avoid suspicion. The model may reject test samples that are anomalous and refuse to adapt based on them. 

\vspace{-3mm}
\subsubsection{ $\ell_p$ Bound }
\vspace{-2mm}

\begin{figure}[t]
\centering
\vspace{-2mm}
\begin{tabular}{cc}
  \includegraphics[width=0.21\textwidth]{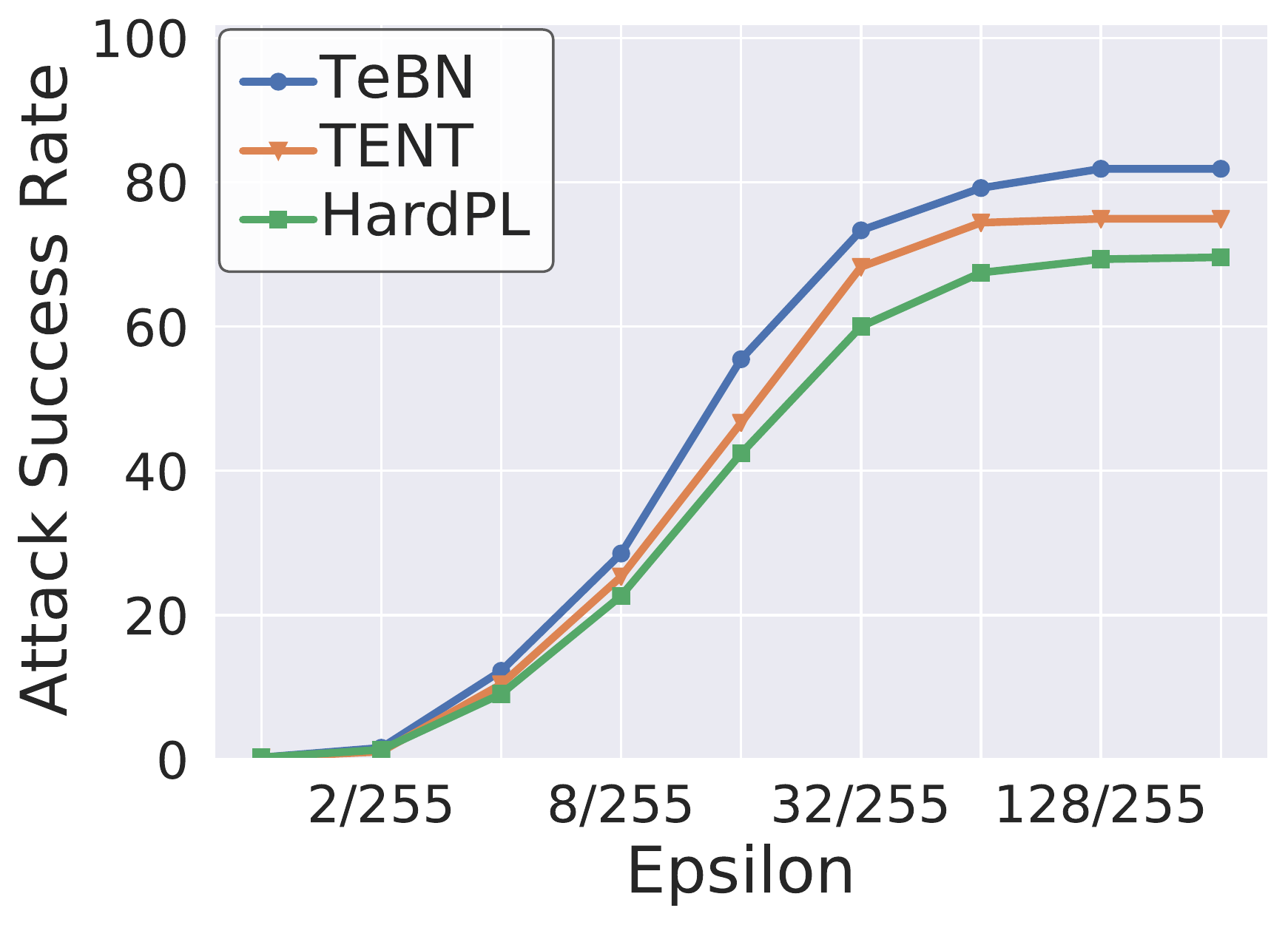} &
 \includegraphics[width=0.21\textwidth]{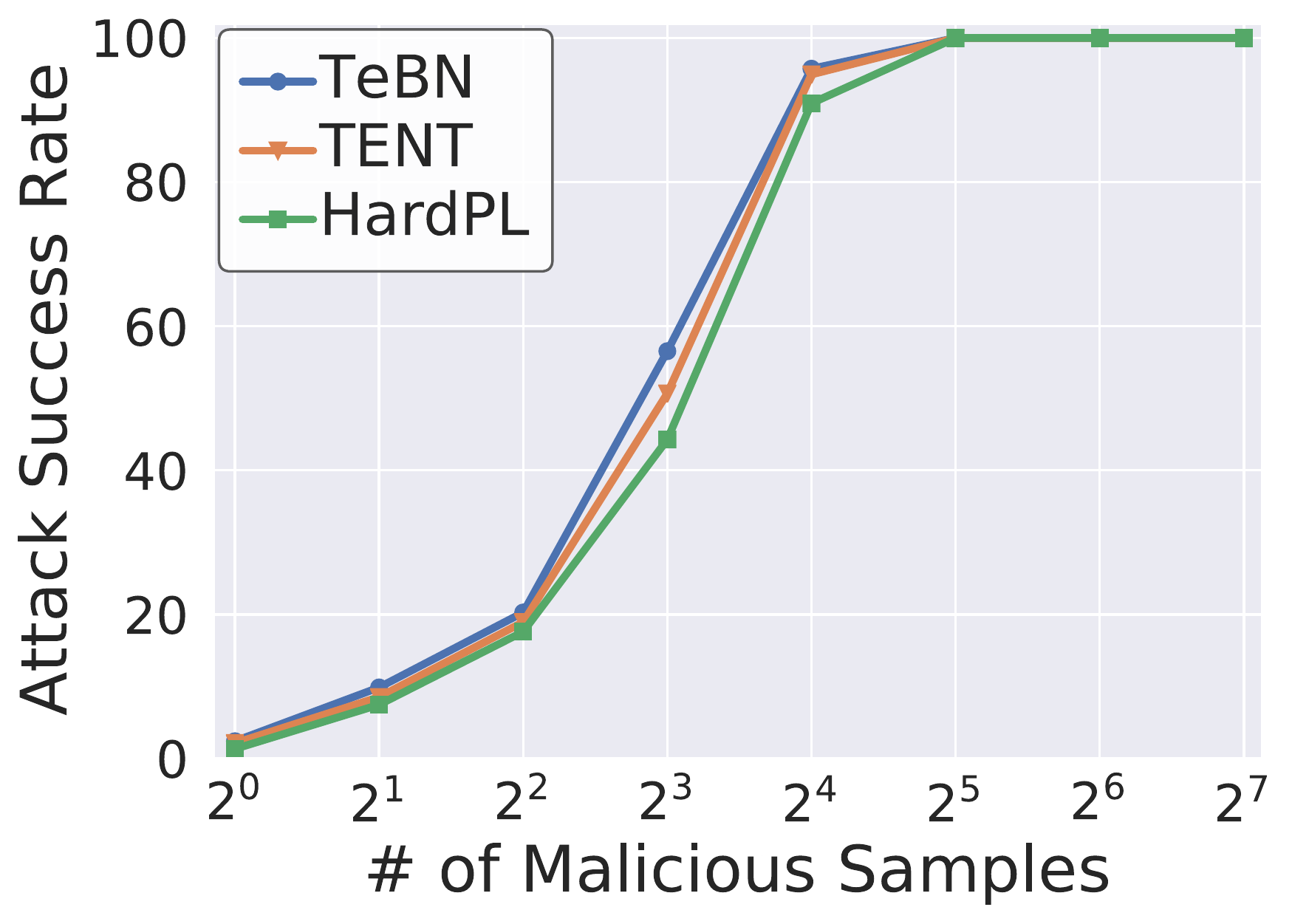}\\
\end{tabular}
\vspace{-5mm}
	\caption{Illustration of the attack success rate across various $\epsilon$ of $L_\infty$ constraints [$N_m = 5$]  (left) and the different number of malicious samples (right) [$\epsilon = 8/255$]. [Targeted Attack] }
	\label{Fig:INC_lp}
 \vspace{-3mm}
\end{figure}

Like much other literature in the adversarial machine learning community, we adopt the $\ell_p$ constraints as imperceptible metrics to perturb the malicious samples. Specifically, we conduct targeted attack experiments on the ImageNet-C by varying the $\ell_\infty$ bound of 5 malicious data samples.  Figure \ref{Fig:INC_lp} reports the attacking effectiveness trends in terms of various $\ell_\infty$ constraints, \textbf{where DIA with $\epsilon$ = $32/255$ reach similar performance with unconstrained attacks}. Furthermore, we specifically select the $\epsilon$ = $8/255$ to run extra experiments on the number of malicious samples, as the resulting images are almost imperceptible to the original images (showed in Figure \ref{Tab:IN_Constrained}(b) and Figure \ref{Tab:IN_EPS}). \textbf{As a result, DIA achieves near-100\% attack success rate, with 32 ($\epsilon$ = $8/255$) malicious samples for ImageNet-C.}


\vspace{-1mm}
\subsubsection{Simulated Corruptions}
\vspace{-2mm}

\begin{table}[t]
    \centering
    \caption{Average attack success rate of adversarially optimized snow corruptions and fog corruptions. [$N_m$=20; Targeted Attack]}
    \resizebox{0.48\textwidth}{!}{%
    \begin{tabular}{ccccc}
    \toprule
    \textbf{Dataset} & \textbf{ Corruption/Attack} & \textbf{TeBN(\%)} & \textbf{TENT(\%)} & \textbf{Hard PL(\%)}  \\ \midrule
    \multirow{2}{*}{\textbf{\begin{tabular}[c]{@{}c@{}}ImageNet-C\\ (ResNet50)\end{tabular}}}  
    & Snow/Snow Attack &   64.00 & 68.00 &   68.00  \\
    & Fog/Fog Attack &   84.00 & 76.00 &   72.00 \\
    \bottomrule       
    \vspace{-3mm}
    \end{tabular}%
    }
    \label{tab:fog_snow}
\end{table}

Since our out-of-distribution benchmark is composed of common corruptions, another idea is to leverage such intrinsic properties and generate ``imperceptible'' adversarial samples adaptively. For example,  we can apply the adversarially optimized snow distortions to the clean images and insert them into the test data in snow distribution.
Then, these injected malicious images (shown in Figure \ref{Tab:IN_Constrained}(d)) are hard to be distinguished from benign corrupted data. 
For implementation, we again compute the gradient of the loss function in Eq. (\ref{eq:mainproblem}) and adopt the same approach as \citet{Kang2019TestingRA}.

\vspace{-1mm} \noindent\textbf{Adversarially optimized simulated corruption is another effective and input-stealthy DIA vector (Table \ref{tab:fog_snow}).} We apply the Snow attack to the ImageNet-C with snow corruptions, similar to Fog.  By inserting 20 malicious samples, the attack success rate reaches at least 60\% and 72\% for Snow and Fog, respectively. Our findings show attackers can leverage test distribution knowledge to develop a better threat model. More details are presented in Appendix \ref{append:simulated}. 




\section{Mitigating Distribution Invading Attacks}
\label{sec:df}

Next, we turn our attention to developing mitigation methods. Our goal is two folds: maintaining the benefits of TTA methods and mitigating Distribution Invading Attacks. Note that we present additional results, including all CIFAR-C experiments, in Appendix~\ref{append:df}.

\vspace{-2mm}
\subsection{Leveraging Robust Model to Mitigate DIA}
\vspace{-1mm}

Our first idea is to replace the source model with a robust model (i.e., adversarially trained models \cite{Madry2018TowardsDL}).
Our intuition is that creating adversarial examples during training causes shifts in the batch norm statistics. (Adversarially) training with such samples will robustify the model to be resistant to BN perturbations. 
We evaluate the targeted DIA against robust ResNet-50 trained by \citet{Salman2020DoAR} and \citet{robustness}, which are the best ResNet-50 models from RobustBench \cite{croce2020robustbench}. Since our single-level attacks only exploit the vulnerabilities of re-estimating the BN statistics, evaluating defenses on TTA with updating parameters could give a false sense of robustness. Therefore, most countermeasure experiments mainly focus on the \textbf{TeBN} method. 

\begin{figure}[t]
\centering
\vspace{-2mm}
\begin{tabular}{cc}
  \includegraphics[width=0.22\textwidth]{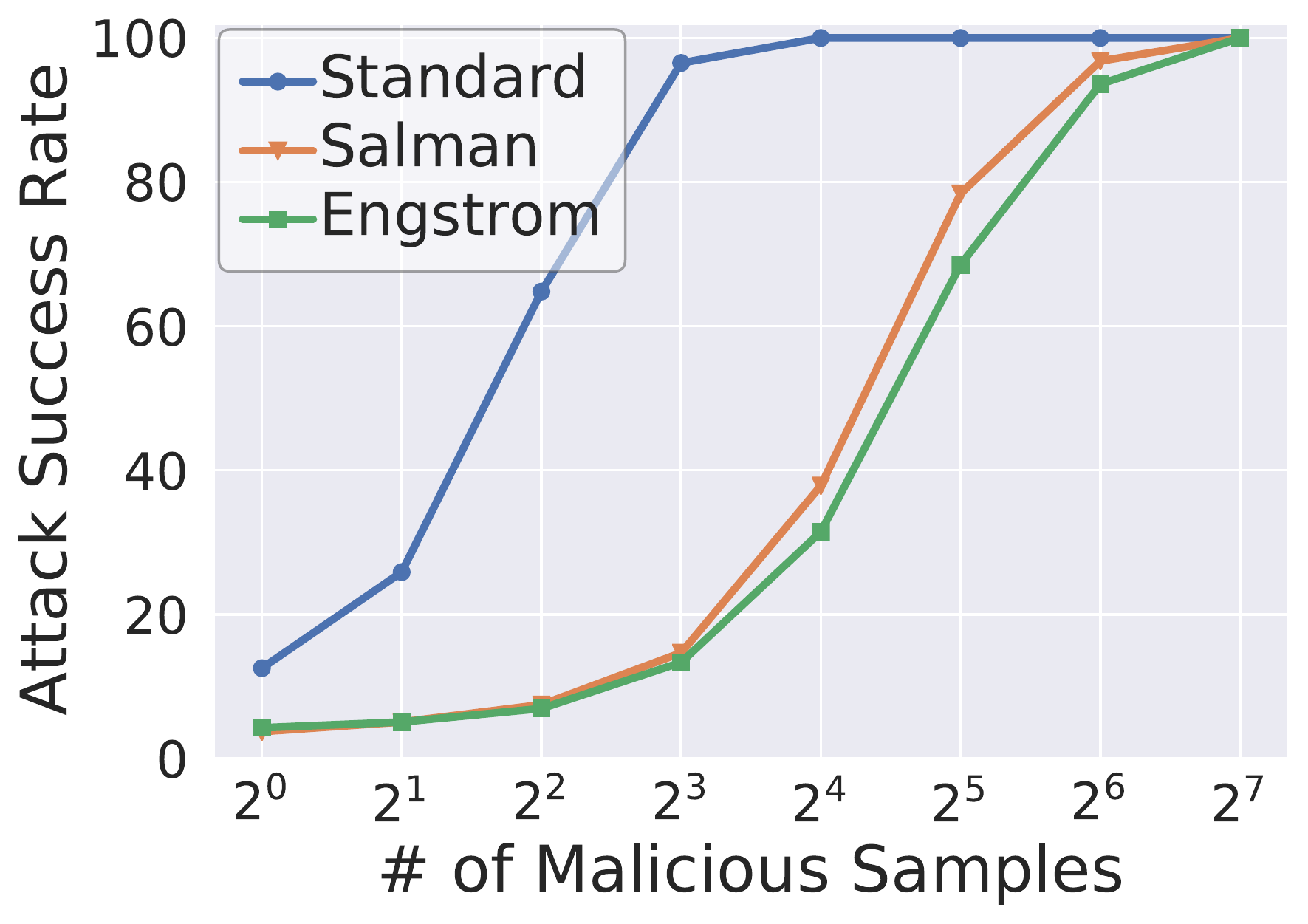}
  \includegraphics[width=0.22\textwidth]{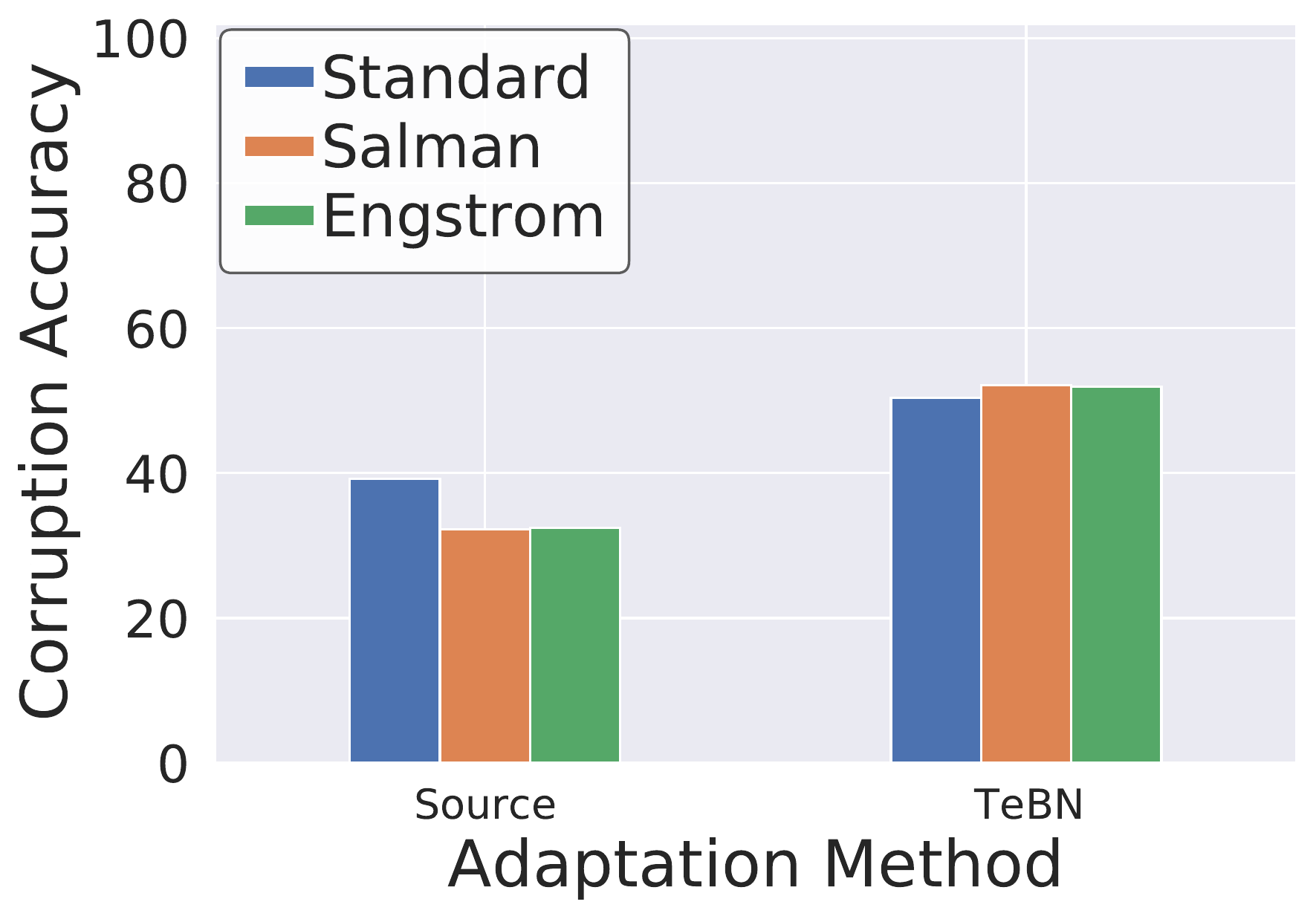}
\end{tabular}
\vspace{-5mm}
\caption{(Left) Attack success rate of DIA against robust models, including \citet{Salman2020DoAR} and \citet{robustness}, across numbers of malicious samples. As a reference, 128 is 64\% of the whole batch containing 200 samples. (Right) Corruption accuracy of robust models.  [TTA method: \textbf{TeBN}] }
\vspace{-4mm}
\label{Fig:INC_robust}
\end{figure}



\vspace{-1mm} \noindent\textbf{Adversarially trained models boost robustness against DIA and maintain the corruption accuracy (Figure \ref{Fig:INC_robust}).} We report the ASR curve of two robust models and observe that they significantly mitigate the vulnerabilities from the test batch. For example, with 8 malicious samples, robust models degrade ASR by $\sim$80\% for ImageNet-C. At the same time, the \textbf{TeBN} method significantly improves the corruption accuracy of robust models, reaching even higher results. 
However, DIA still achieves more than 70\% success rate with 32 malicious samples (16\% of the whole batch). 

\subsection{Robust Estimate of Batch Normalization Statistics}
\label{sec:dfBN}
We then seek to mitigate the vulnerabilities by robustifying the re-estimation of Batch Norm statistics.

\vspace{-1mm} \noindent\textbf{Smoothing via training-time BN statistics.} Since training-time BN cannot be perturbed by DIA, we can combine the training-time and test-time BN statistics. This approach is also mentioned in \cite{Schneider2020ImprovingRA, You2021TesttimeBS}; however, their motivation is to stabilize the BN estimation when batch size is small (e.g., $<$32) and improve the corruption accuracy. It can be formulated as $ \ 
    \bar{\mu}= \tau \mu_s+ (1-\tau) \mu_t, \quad \bar{\sigma}^2=\tau \sigma_s^2+(1-\tau) \sigma_t^2, \ 
$ where ($\bar{\mu}, \bar{\sigma}^2$) stand for final BN statistics, ($\mu_s, \sigma_s^2$) are training-time BN, and  ($\mu_t, \sigma_t^2$) are test-time BN. We view the training-time BN statistics as a robust prior for final estimation and adopt smoothing factor $\tau$ to balance the weight.

\begin{figure}[t]
  \centering
  \begin{tabular}{cc}
    \includegraphics[width=0.21\textwidth]{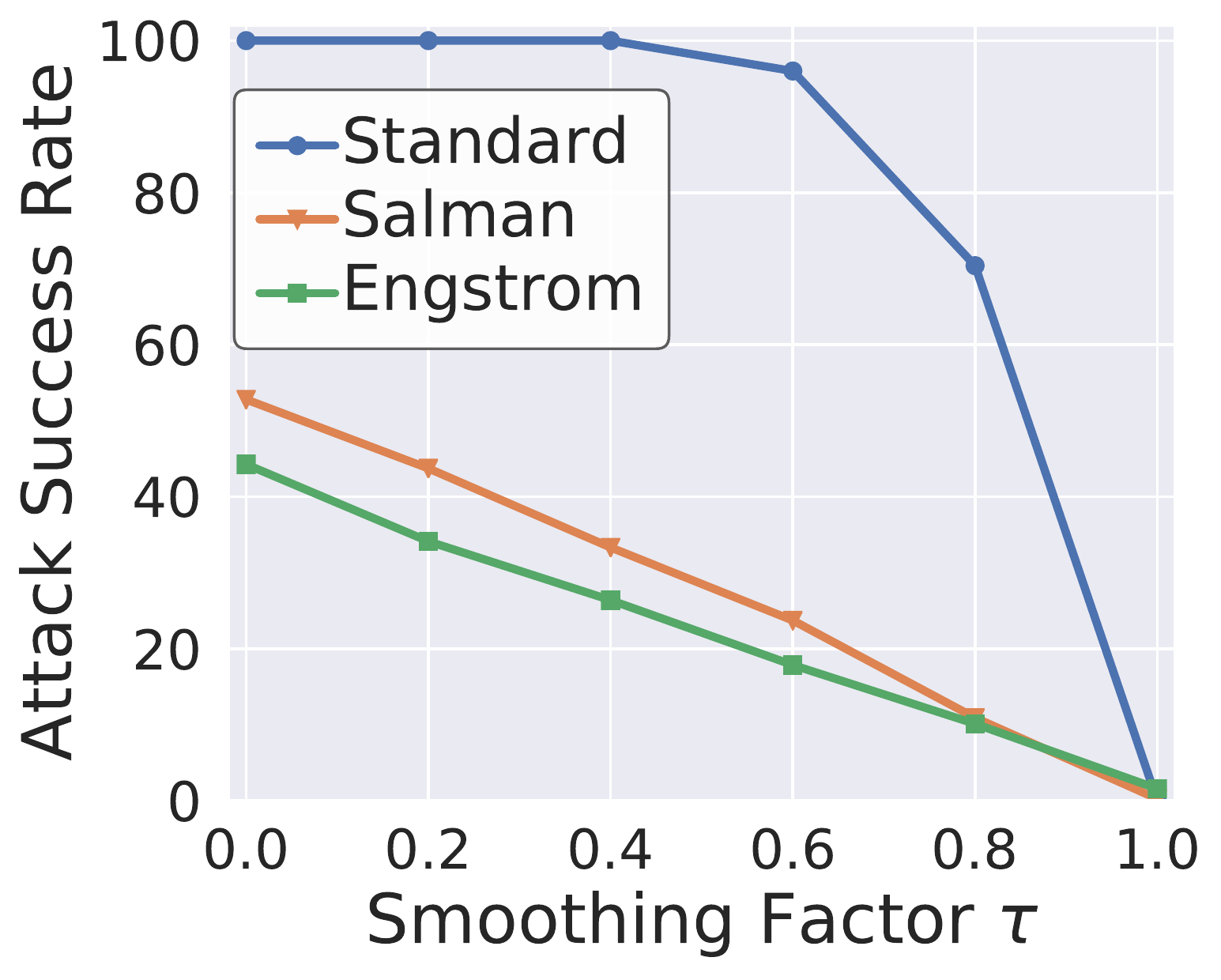} & 
   \includegraphics[width=0.21\textwidth]{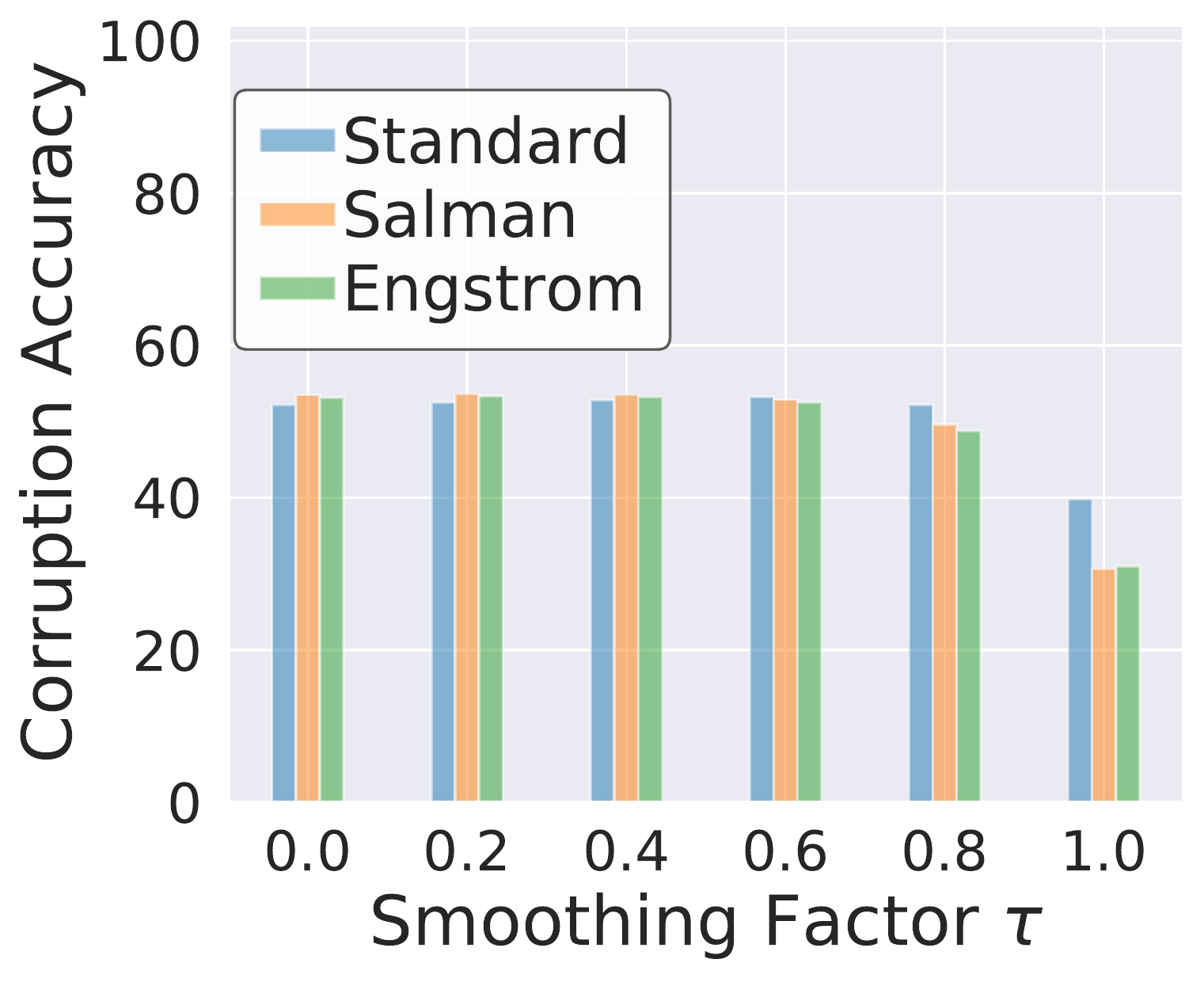} \\
  \end{tabular}
  \vspace{-4mm}
  \caption{Controlling $\tau$ = 0.6 can degrade the attack success rate (line) while maintaining high corruption accuracy (bar). [$N_m$=20]}
\label{Fig:IN_Bayes}
\vspace{-3mm}
  \end{figure}



\vspace{-1mm} \noindent \textbf{Leveraging Training-time Batch Norm statistics mitigates the vulnerabilities for both standard and adversarial models (Figure \ref{Fig:IN_Bayes}).} 
We specifically select $N_m$ = 20 and set $\tau = \{0.0, 0.2, 0.4, 0.6, 0.8, 1.0\}$, where $\tau = 0.0$ ignores the test-time BN and $\tau = 1.0$ ignores training-time BN. It appears that improving $\tau$ generally results in both ASR and corruption accuracy drops on ImageNet-C. However, the degradation in corruption accuracy happens only when $\tau > 0.6$. Therefore, setting $\tau = 0.6$ is a suitable choice, which can mitigate the ASR to $\sim$20\% for 20 malicious samples with robust models.

\vspace{-1mm} \noindent \textbf{Adptively Selecting Layer-wise BN statistics.} We also explore and understand the DIA by visualizing each layer BN in Appendix~\ref{sec:understandBN}. 
Given the discovery where BN statistics shift on the latter layers when applying DIA, we can strategically select training or test time BN statistics for different layers. Hence, we take advantage of training-time BN for the last few layers to constrain the malicious effects. 

\begin{figure}[t]
  \centering
  \begin{tabular}{cc}
    \includegraphics[width=0.21\textwidth]{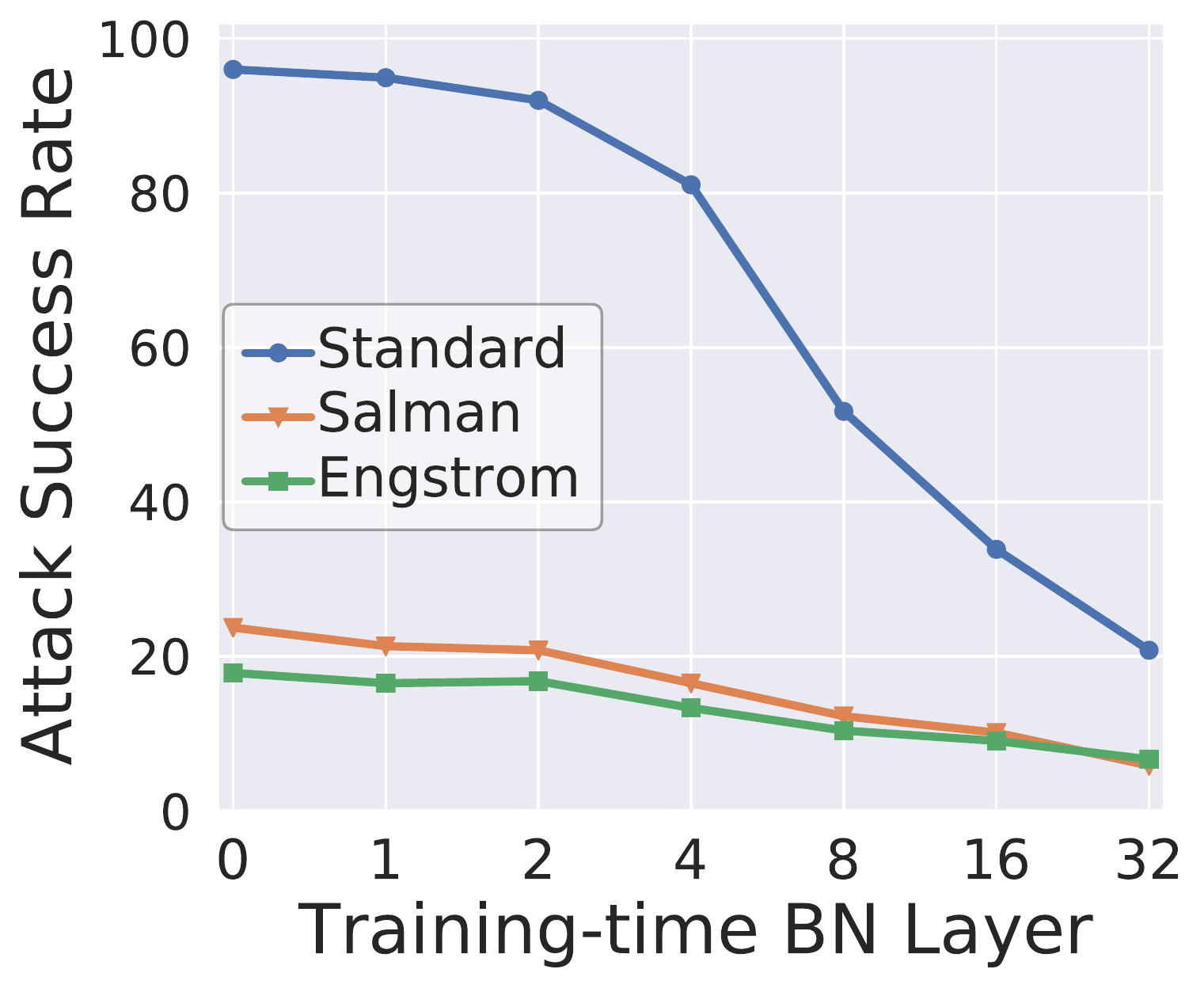} & 
   \includegraphics[width=0.21\textwidth]{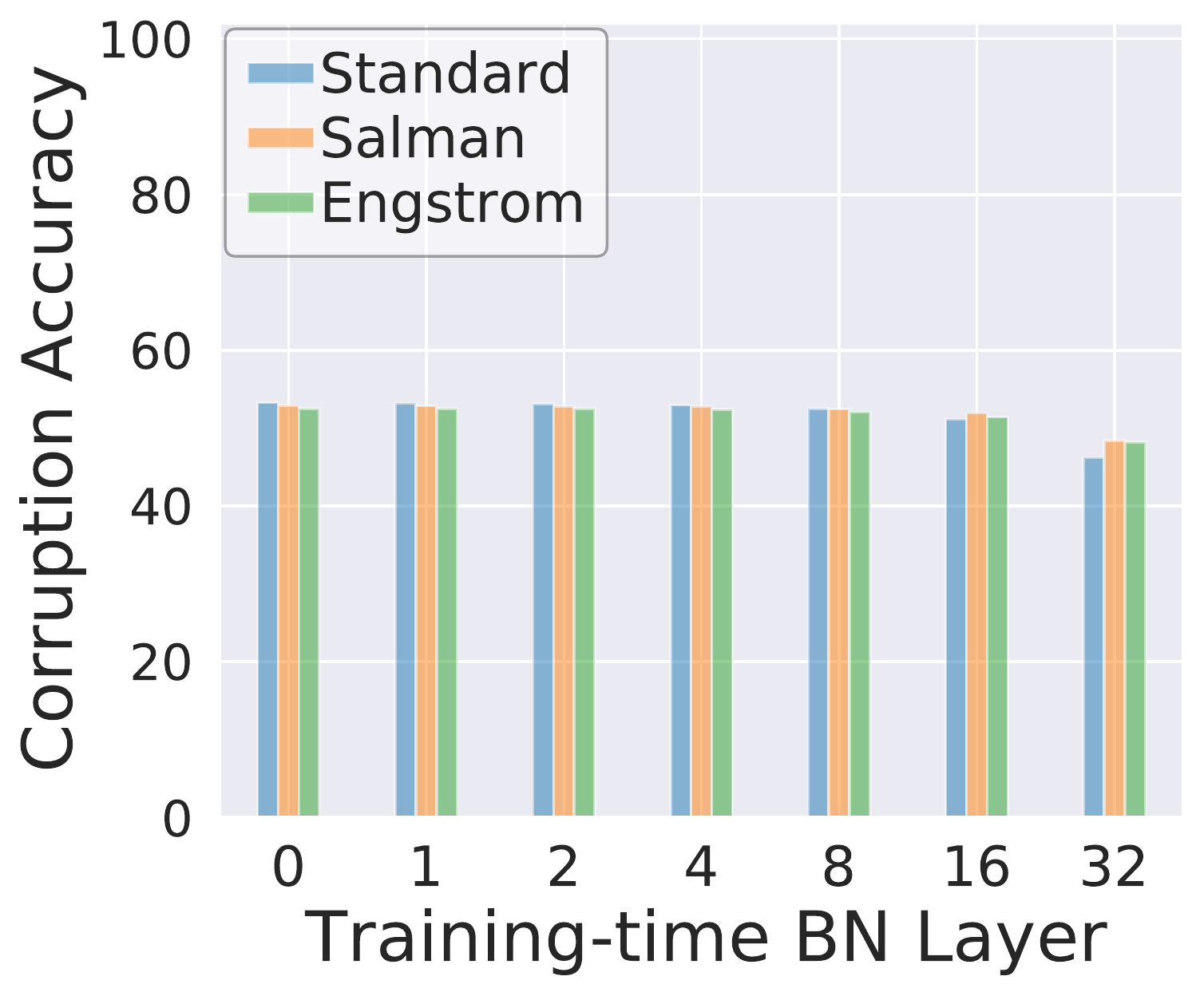} \\
  \end{tabular}
  \vspace{-4mm}
  \caption{Balancing layer-wise BN can degrade the ASR (line) while maintaining high corruption accuracy (bar). [$N_m$=20, $\tau$=0.6]}
  \vspace{-5mm}
  \label{Fig:IN_Bayes_layer}
  \end{figure}



\vspace{-1mm} \noindent\textbf{Adaptive Layer-wise BN further constrains the malicious effect from DIA (Figure \ref{Fig:IN_Bayes_layer}).} Given $\tau$ = 0.6 is a suitable choice for whole BN layers, we further leverage full training-time BN ($\tau$ = 1.0) for the last $N_{tr}$ = \{0, 1, 2, 4, 8, 16\} BN layers. It appears that increasing training-time BN layers (from $N_{tr}$ = 0 to $N_{tr}$ = 16) results in tiny corruption accuracy drops ($\sim$2\%) and generally helps the robustness against DIA. For example, if we set $N_{tr}$ = 8, ASR drops $\sim$40\% for the standard model with $N_m$ = 20.

In conclusion, applying both approaches, our best results achieve a negligible corruption accuracy degradation and mitigate ASR by $\sim$50\% for the standard model and  $\sim$40\% for the robust model on ImageNet-C. However, our mitigations have not fully resolved the issue, where increasing the number of malicious data samples may still allow successful DIA attacks with a high probability. We encourage future researchers to consider the potential adversarial risks when developing TTA techniques.

\vspace{-3mm}
\section{Discussion and Future Work}
\vspace{-1mm}

While our proposed attacks are promising in terms of effectiveness, several limitations exist.  
Many recent works in test-time adaptation have been proposed, which leverage different adaptation techniques (e.g., \cite{niu2022efficient}). 
Most of them still suffer from the two vulnerabilities we identified in Section \ref{sec:diamethod} and can be attacked by DIA directly (see Appendix \ref{append:addtta}). However, an adaptive attack design should be considered if the TTA techniques change significantly (e.g., methods neither using test-time BN nor self-learning). Furthermore, methods under the general transductive learning settings (i.e., predictions affected by test batch) also involve similar risks, which we encourage the community to explore in future studies. 

Another direction is relaxing the adversary's knowledge assumption. In our white-box threat model, we assume the adversary has the full knowledge of the test batch, including targeted and benign samples. However, obtaining this information in real-world scenarios might be difficult (e.g., some users keep the uploaded images secret \cite{deng2021separation}). One potential method to relax this assumption is to utilize outsourced information (e.g., online images) to replace the benign images and optimize the expected loss like EOT \cite{Athalye2017SynthesizingRA}. 
Furthermore, when model architectures and parameters are not exposed to adversaries (black-box threat model), launching DIA attacks could become more challenging. 
Future works should consider methods using  model ensemble \cite{Geiping2020WitchesBI} or diverse inputs \cite{Xie2018ImprovingTO} to boost the transferability of DIA's malicious data. 

\vspace{-3mm}
\section{Conclusion}
\vspace{-1mm}

In this work, we investigate  adversarial risks of test-time adaptation (TTA). First, we present a novel framework called Distribution Invading Attack, which can be used to achieve new malicious goals. Significantly, we prove that manipulating a small set of test samples can affect the predictions of other benign inputs if adopting TTA, which is not studied at all in  previous literature. We then explore mitigation strategies, such as utilizing an adversarially-trained model as a source model and robustly estimating BN statistics. Overall, our findings uncover the risks of TTA and inspire future works build robust and effective TTA techniques.

\section{Acknowledgements}

We are grateful to Ahmed Khaled, Chong Xiang, Ashwinee Panda, Tinghao Xie, William Yang, Beining Han, Liang Tong, and Prof. Olga Russakovsky for providing generous help and insightful feedback on our project. This work
was supported in part by the National Science Foundation
under grant CNS-2131859, the ARL’s Army Artificial Intelligence Innovation Institute (A2I2), Schmidt DataX award, Princeton E-ffiliates Award, and  Princeton Gordon Y. S. Wu Fellowship.

\bibliography{main_ref}
\bibliographystyle{icml2023}

\newpage
\appendix
\onecolumn

\section{Omitted Details in Background and Related Work}
\label{sec:ttabg}

In this appendix, we cover additional details on batch normalization (Appendix \ref{append:bnbg}) and self-learning (Appendix \ref{append:slbg}), along with TTA algorithms in Appendix \ref{append:ttabg}.
Then, we discuss conventional machine learning vulnerabilities including adversarial examples and data poisoning in Appendix \ref{sec:mlv}. 
Furthermore, we include a comprehensive review of the existing literature in Appendix \ref{append:related}.

\subsection{Batch Normalization}
\label{append:bnbg}

In batch normalization, we calculate the BN statistics for each BN layer $l$ by computing the mean $\mu$ and variance $\sigma^2$ of the pre-activations $\mathbf{z}_{l}$: $\mu \leftarrow \mathbb{E}[\mathbf{z}_{l}]$, $\sigma^2 \leftarrow \mathbb{E}[(\mathbf{z}_{l} - \mu)^2]$. The expectation $\mathbb{E}$ is computed channel-wise on the input.
We then normalize the pre-activations by subtracting the mean and dividing by the standard deviation: $\bar{\mathbf{z}}_l = (\mathbf{z}_l - \mu) / \sigma$. 
Finally, we scale and shift the standardized pre-activations $\bar{\mathbf{z}}_{l}$ to $\mathbf{z}'_{l} = \gamma_l^T \bar{\mathbf{z}}_{l} + \beta_l$ by learnable affine parameters $\{\gamma_l,\beta_l\}$.
In our experiments, the parameters updated during the TTA procedures are $\theta_{\mathcal{A}} = \{\gamma_l,\beta_l\}_{l = 1}^{L}$, where $L$ is the number of BN layers. The statistics of the source data are replaced with those of the test batch.

\subsection{Objectives of Self-Learning}
\label{append:slbg}

We provide detailed formulations of the TTA loss functions used in self-learning (SL) methods.
Let $h(\cdot;\theta)$ be the hypothesis function parameterized by $\theta$, which we consider to be the logit output (w.l.o.g.).
The probability that sample $\bx$ belongs to class $j$ is denoted as $p(j|\bx)=\sigma(h(\bx;\theta))$, where $\sigma(\cdot)$ is the softmax function. 
Here, we use $\theta$ to denote $\preparam$ for simplicity.

\smallskip \noindent \textbf{TENT~\cite{Wang2021TentFT}.} This method minimizes the entropy of model predictions.

\begin{equation} 
    \label{eq:entropy_min}
    \begin{aligned}
        &\mathcal{L}_{\text{TTA}}(\batchdata) := - \frac{1}{N_{Bt}}\sum_{i=1}^{N_{Bt}} \sum_j p(j|\batchsample )\log p(j|\batchsample  )
    \end{aligned}
\end{equation}

\smallskip \noindent \textbf{Hard PL~\cite{Galstyan2007EmpiricalCO, lee2013pseudo}.} The most likely class predicted by the pre-adapted model is computed as the pseudo label for the unlabeled test data.
\begin{equation} 
    \label{eq:hardPL}
    \begin{aligned}
        &\mathcal{L}_{\text{TTA}}(\batchdata, \bypls(\theta)) := -  \frac{1}{N_{Bt}}\sum_{i=1}^{N_{Bt}}\log p(\ypl (\theta)|\batchsample  ),\\
        &\text{where  }\ypl(\theta) = \underset{j}{\mathrm{argmax}} p(j|\batchsample ), \forall \batchsample  \in \batchdata
    \end{aligned}
\end{equation}
\smallskip \noindent \textbf{Soft PL~\cite{Galstyan2007EmpiricalCO, lee2013pseudo}.} Instead of using the predicted class, the softmax function is applied directly to the prediction to generate a pseudo label.

\begin{equation} 
    \label{eq:softPL}
    \begin{aligned}
        &\mathcal{L}_{\text{TTA}}(\batchdata, \bypls(\theta)) := -  \frac{1}{N_{Bt}}\sum_{i=1}^{N_{Bt}} \sum_j \ypl(\theta)\log p(j|\batchsample ),\\
        &\text{where  }\ypl(\theta) =  p(j|\batchsample), \forall \batchsample  \in \batchdata\\
    \end{aligned}
\end{equation}

\smallskip \noindent \textbf{Robust PL~\cite{rusak2022if}.} It has been shown that the cross-entropy loss is sensitive to label noise. To mitigate the side effect to the training stability and hyperparameter sensitivity, \textbf{Robust PL} replaces the cross-entropy (CE) loss of the \textbf{Hard PL} with a \textit{Generalized Cross Entropy (GCE)}.

\begin{equation} 
    \label{eq:RPL}
    \begin{aligned}
        &\mathcal{L}_{\text{TTA}}(\batchdata, \bypls(\theta)) := - \frac{1}{N_{Bt}}\sum_{i=1}^{N_{Bt}} q^{-1}(1- p(\ypl (\theta)|\batchsample  )^q),\\
        &\text{where  }\ypl (\theta) = \underset{j}{\mathrm{argmax}} p(j|\bx), \forall \batchsample \in \batchdata
    \end{aligned}
\end{equation}

where $q \in (0,1]$ adjusts the shape of the loss function. When $q \rightarrow 1$, the GCE loss approaches the MAE loss, whereas when $q \rightarrow 0$, it reduces to the CE losses.

\smallskip \noindent \textbf{Conjugate PL~\cite{Goyal2022TestTimeAV}.}
 We consider the source loss function, which can be expressed as $\mathcal{L}(h(\bx;\theta), y) = g(h(\bx;\theta))-y^Th(\bx;\theta)$, where $g$ is a function and $y$ is a one-hot encoded class label. The TTA loss for self-learning with conjugate pseudo-labels can be written as follows:
\begin{equation} 
    \label{eq:CPL}
    \begin{aligned}
        &\mathcal{L}_{\text{TTA}}(\batchdata, \bypls(\theta)) := - \frac{1}{N_{Bt}}\sum_{i=1}^{N_{Bt}}\mathcal{L}(h(\batchsample ;\theta)/T, \ypl (\theta)),\\  
        &\text{where  }\ypl (\theta) = \nabla g(h(\batchsample )/T;\theta), \forall \batchsample  \in \batchdata
    \end{aligned}
\end{equation}

The temperature $T$ is used to scale the predictor.


\subsection{TTA algorithms in details}
\label{append:ttabg}

In this subsection, we present the detailed algorithm of test-time adaptation  (Algorithm~\ref{alg:tta}). In our setting, the model is adapted online when a batch of test data comes and then makes the prediction immediately.
\begin{algorithm}[H]
   \caption{Test-Time Adaptation}\label{alg:tta}
    \begin{algorithmic}[1]
    \STATE \textbf{Input:} 
    Source model parameters $\srcparam$, 
     number of steps $N$, 
     TTA update rate: $\eta$
   \STATE \textbf{Initialization:} $\theta = \srcparam$
   \FOR{step $=1, 2, \dots,  N$}
    \STATE Obtain a new test batch $\mathbf{X}^t_{B}$ from the test domain.
    \STATE $\theta_{\B} \leftarrow \{\mu(\batch), \sigma^2(\batch)\}$ 
    \STATE $\theta_{\A} \leftarrow  \theta_{\A} - \eta \cdot \partial \mL_{\text{TTA}}(\batch)/ \partial \theta_{\A}$ 
   \STATE $\theta \leftarrow \adatheta \cup \bntheta  \cup \fixtheta$ 
   \STATE Make the prediction $\hat{\mathbf{y}}^t_B =  f(\mathbf{X}^t_{B};\theta)$
   \ENDFOR
\end{algorithmic}
\end{algorithm}

\subsection{Conventional Machine Learning Vulnerabilities}
\label{sec:mlv}

The following subsection provides additional background on ML vulnerabilities and their relevance to DIA. 
We then emphasize the distinctiveness of the proposed attack, which differentiates it from other known adversarial examples and data poisoning attacks.

\smallskip \noindent \textbf{Adversarial Examples.}
The vast majority of work studying vulnerabilities of deep neural networks concentrates on finding the imperceptible \textit{adversarial examples}. 
Those examples have been successfully used to fool the model at test time~\cite{Goodfellow2015ExplainingAH, Carlini2017TowardsET, Vorobeychik18book}. 
Commonly, generating an adversarial example $\hatx = \mathbf{x}^{t} +\hat{\pert}$ can be formulated as follows: 
\begin{equation}
  \hat{\pert} = \underset{\pert}{\argmax } \ \mL(f(\mathbf{x}^{t} +\pert ; \srcparam), y^{t}) \quad s.t. 
  \norm{\pert}_p \leq \eps, 
\end{equation}

\noindent where $\mL$ is the loss function (e.g., cross-entropy loss), $\mathbf{x}^{t}, y^{t}$ denote one test data, $\norm{\pert}_p \leq \eps$ is the $\ell_p$ constraint of the perturbation $\pert$. 
The objective is to optimize $\pert$ to maximize the prediction loss $\mL$. 
For $\ell_\infty$ constraint, the problem can be solved by Projected Gradient Descent \cite{Madry2018TowardsDL} as

\begin{equation}
\label{equ:pgd}
\pert \leftarrow \Pi_{\eps} (\pert + \alpha \ \mathrm{sign} (\nabla_{\pert} L(f(\mathbf{x}^{t} +\pert ; \srcparam), y^{t}))).
\end{equation}

\noindent Here, $\Pi_{\eps}$ denotes projecting the updated perturbation back to the constraint, and $\alpha$ is the attack step size. By iteratively updating through Eq. (\ref{equ:pgd}), the attacker will likely cause the model to mispredict. 
Our single-level variant of the DIA attack employs an algorithm similar to the projected gradient descent.

One characteristic of \textit{adversarial examples} in ML pipeline is that the perturbations have to be made on the targeted data. 
Therefore, as long as the user keeps the test data securely (not distorted), the machine learning model can always make benign predictions.
Our attack, targeted at TTA, differs from adversarial examples, where our malicious samples can be inserted into a test batch and attack the benign and unperturbed data. 
Therefore, there exists an increasing security risk if deploying TTA.

\smallskip \noindent \textbf{Data Poisoning Attacks.} 
In terms of attacking benign samples without directly modifying the data, another pernicious attack variant, \textit{data poisoning}, can accomplish this goal.
Specifically, an adversary injects poisoned samples into a training set and aims to cause the trained model to mispredict the benign sample. 
There are two main objectives for a poisoning attack: targeted poisoning and indiscriminate. 
For \textit{targeted poisoning attacks} \cite{Koh2017UnderstandingBP, Carlini2021PoisoningAB, Carlini2021PoisoningTU, Geiping2020WitchesBI}, the attacker's objective is to attack one particular sample with a pre-selected targeted label, which can be written as follows: 

\begin{equation}
\label{equ:tgtpoi}
\min _{\pert} \mathcal{L}\left(f\left(\xtgt, \theta^*(\pert)\right), \ytgt\right) \quad s.t. \  \theta^*(\pert) = \underset{\theta}{\arg \min } \frac{1}{N_s} \sum_{i=1}^{N_s} \mathcal{L}\left(f\left(\mathbf{x}^{s}_i +\pert_i , \theta\right), y^{s}_i\right)
\end{equation}

where $\xtgt$ is the targeted samples and $\ytgt$ is the correpsonding incorrect targeted labels. $\mathbf{x}^{s}_i$ is the training data and $y^{s}_i$ is the corresponding ground truth. We use the $\pert_i$ to denote the perturbation on training data $\mathbf{x}^{s}_i$, and since there are $N_m$ poisoning samples, some $\pert_i$ stays zero to represent clean samples.

For \textit{indiscriminate attacks} \cite{Nelson2008ExploitingML, biggio2012poisoning}, the adversary seeks to reduce the accuracy of all test data, which can be formulated as:  

\begin{equation}
\label{equ:tgtind}
\min _{\pert} -\frac{1}{N_t} \sum_{j=1}^{N_t}\mathcal{L}\left(f\left(\mathbf{x}^{t}_j, \theta^*(\pert)\right), y^{t}_j\right) \quad s.t. \  \theta^*(\pert) \in \underset{\theta}{\arg \min } \frac{1}{N_s} \sum_{i=1}^{N_s} \mathcal{L}\left(f\left(\mathbf{x}^{t}_i +\pert_i , \theta\right), y^{t}_i\right)
\end{equation}

where $\mathbf{x}^{t}_j$ is the test data and $y^{t}_j$ is the ground truth of it. 
These objectives also guide the design of our adversary's goals.

However, the key characteristic of data poisoning for the machine learning model is the assumption of access to training data for adversaries. 
Therefore, if the model developer obtains the training data from a reliable source and keeps the database secure, then there is no chance for any poisoning attacks to be effective.  
Our attack for TTA differs from data poisoning, in which DIA only requires test data access. This makes our attack easier to access, as data in the wild environment (test data) are less likely to be monitored.

\subsection{Extended Related Work}
\label{append:related}
We present more related works of our paper in this subsection.

\smallskip\noindent\textbf{Unsupervised Domain Adaptation.}
Unsupervised domain adaptation (UDA) deals with the problem of adapting a model trained on a source domain to perform well on a target domain, using both labeled source data and unlabeled target data.
One common approach~\cite{ganin2016domain, long2018conditional} is to use a neural network with shared weights for both the source and target domains and introduce an additional loss term to encourage the network to learn domain-invariant features.
 Other approaches include explicitly~\cite{saito2018maximum, damodaran2018deepjdot} or implicitly~\cite{li2016revisiting, wang2019transferable} aligning the distributions of the source and target domains. One category related to our settings is source-free domain adaptation \cite{liang20a, Kundu2020UniversalSD, Li2020ModelAU}, where they assume a source model and the entire test set. For instance, SHOT \cite{liang20a} uses information maximization and pseudo-labels to align the source and target domain during the inference stage.

\noindent\textbf{Test-time Adaptation.}
TTA obtains the model trained on the source domain and performs adaptation on the target domain. Some methods \cite{Sun2020TestTimeTW, Liu2021TTTWD, Gandelsman2022TestTimeTW} modify the training objective by adding a self-supervised proxy task to facilitate test-time training. However, in many cases, access to the training process is unavailable. Therefore some methods for test-time adaptation only revise the Batch Norm (BN) statistics \cite{Singh2019EvalNormEB, Nado2020EvaluatingPB, Schneider2020ImprovingRA, You2021TesttimeBS, Hu2021MixNormTA}.
Later, TENT~\cite{Wang2021TentFT}, BACS~\cite{zhou2021bayesian}, and MEMO~\cite{Zhang2021MEMOTT} are proposed to improve the performance by minimizing entropy at test time.
Other approaches~\cite{Galstyan2007EmpiricalCO, lee2013pseudo, rusak2022if, Goyal2022TestTimeAV, Wang2022TowardsUG} use pseudo-labels generated by the source (or other) models to self-train an adapted model. 
As an active research area, recent efforts have been made to further improve TTA methods in various aspects. For example,
\citet{niu2022efficient} seeks to solve the forgetting problem of TTA. \citet{wang2022continual, gong2022note, huang2022extrapolative} propose methods to address the continual domain shifts. \citet{Iwasawa2021TestTimeCA} improves the pseudo-labels by using the pseudo-prototype representations. \citet{Zhang2021DomainPL, Kojima2022RobustifyingVT, Gao2022VisualPT} leverage recent vision transformer model architecture to improve TTA.

Most TTA methods assume a batch of test samples is available, while some papers consider the test samples coming individually \cite{Zhang2021MEMOTT, gao2022back, bartler22a, Mirza2021TheNM, Dbler2022RobustMT}. For instance, MEMO \cite{Zhang2021AdaptiveRM} leverages various augmentations on single test input and then adapts the model parameters. \citet{gao2022back} propose using diffusion to convert the out-of-distribution samples back to the source domain. Such a ``single-sample'' adaption paradigm makes an independent prediction on each data, avoiding the risk (i.e., DIA) of TTA. However, batch-wise test samples provide more information about the distribution, usually achieving better performance.

\noindent \textbf{Adversarial Examples and Defenses.}
A host of works have explored the imperceptible adversarial examples~\cite{Goodfellow2015ExplainingAH, Carlini2017TowardsET, Vorobeychik18book}  which fool the model at test time and raise security issues. In response, adversarial training~\cite{madry2017towards} (training with adversarial samples) has been proposed as an effective technique for defending against adversarial examples. Later, there existed a host of enhanced methods on robustness~\cite{Wu2020AdversarialWP, Sehwag2022RobustLM, Dai2022ParameterizingAF, Gowal2021ImprovingRU}, time efficiency~\cite{shafahi2019free, wong2019fast}, and utility-robustness tradeoff~\cite{Pang2022RobustnessAA, Wang2022RemovingBN}. 


\noindent\textbf{Adversarial Defense via Transductive Learning.} The security community has also explored the use of transductive learning to enhance the robustness of ML 
 \cite{goldwasser2020beyond, pmlr-v119-wu20f, Wang2021FightingGW}.  However, the latter work by \citet{chen2021towards, Croce2022EvaluatingTA} demonstrated these defenses through transductive learning have only a small improvement in robustness. Our paper focuses on another perspective, where transductive learning induces new vulnerabilities in benign and unperturbed data.

\noindent\textbf{Data Poisoning Attacks and Defenses.} 
Data poisoning attacks refer to injecting poisoned samples into a training set and causing the model to predict incorrectly. 
It has been applied to different machine learning algorithms, including the Bayesian-based method \cite{Nelson2008ExploitingML},  support vector machines \cite{biggio2012poisoning}, as well as neural networks \cite{Koh2017UnderstandingBP}. 
Later, \citet{Carlini2021PoisoningTU} tries to poison the unlabeled training set of semi-supervised learning and \citet{Carlini2021PoisoningAB} target at contrastive learning with an extremely limited number of poisoning samples. 
Recently, researchers have utilized poisoning attacks to achieve other goals (e.g., enhancing membership inference attacks \cite{Tramr2022TruthSP} and breaking machine unlearning \cite{Marchant2021HardTF, Di2022HiddenPM}).
For defense, \citet{geiping2021doesn} has shown that adversarial training can also effectively defend against data poisoning attacks.

\noindent\textbf{Adversarial Risk in Domain Adaptation.}
Only a few papers discuss adversarial risks in the context of domain adaptation settings. 
One example is \cite{NEURIPS2021_90cc440b}, which proposes several methods for adding the poisoned data to the source domain data exploiting both the source and target domain information.
Another example is \cite{10.1145/3447548.3467214}, which introduces \textit{I2Attack}, an indirect, invisible poisoning attack that only manipulates the source examples but can cause the domain adaptation algorithm to make incorrect predictions on the targeted test examples. However, both approaches assume that the attacker has access to the source data and cannot be applied to the source-free (TTA) setting.

\newpage
\section{Theorical Analysis of Distribution Invading Attacks}

\newcommand{\w}{\Vec{w}}
\newcommand{\mus}{\Vec{\mu}_s}
\newcommand{\sigmas}{\Vec{\sigma}_s}
\newcommand{\mut}{\Vec{\mu}_t}
\newcommand{\sigmat}{\Vec{\sigma}_t}
\newcommand{\del}{\partial}

In this appendix, we provide technical details of computing the DIA gradient described in Section \ref{sec:diamethod}. Then, we analyze why smoothing via training-time BN can mitigate our attacks. 
For simplicity, we consider the case for a single layer, and the gradient computation can be generalized to the case of a multi-layer through backpropagation. 

\subsection{Understanding the Vulnerabilities of Re-estimating BN Statistics}
\label{append:UBN}
Recall the setting where $\xtgt \in \R^d$, $\batch = (x_{i, j})_{i=1 \ldots n, j=1 \ldots d} \in \R^{n \times d}$, $\w \in \R^d$ and $b \in \R$. 
The output of a linear layer with batch normalization\footnote{For the purpose of this analysis, we will set the scale parameter to 1 and the shift parameter to 0. This assumption does not limit the generalizability of our results.} on a single dimension can be written as 

\begin{align}
    f(\xtgt) 
    = \frac{\xtgt - \mu(\batch) } { \sqrt{ \sigma^2(\batch)}  } \Vec{w} + b
\end{align}

where $\mu$ and $\sigma^2: \R^{n \times d} \rightarrow \R^d$ denote the coordinate-wise mean and variance. In addition, the square root $\sqrt{\cdot}$ is also applied coordinate-wise. Suppose we can perturb $x_{i}$ in $\batch$, which corresponds to a malicious data point in $\mal$. 
In order to find the optimal perturbation direction that causes the largest deviation in the prediction of $f(\xtgt)$, we compute the following derivative:
\begin{align*}
    \frac{\del f(\xtgt)}{\del x_{i, j^*}}  &= \frac{ (-\frac{1}{n}) (  \sigma(\batch)  )_j - 
    [ (\xtgt)_j -  \mu(\batch)_j ]
    \frac{ \del (\sigma(\batch))_j }{ \del x_{i, j} }  }{ (  \sigma(\batch)  )_j^2 } w_j \\
    & = 
    \frac{
    (-\frac{1}{n})\left\{
     (\sigma(\batch))_j 
    - \left[
    (\xtgt)_j -  \mu(\batch)_j
    \right] 
    (\sigma(\batch))_j^{-1}
    ( x_{i, j^*} - \mu(\batch)_j)
    \right\} 
    }{ (\sigma(\batch)  )_j^2  }
    w_j\\
    \text{where} \ \ &\ (\mu(\batch))_j = \frac{1}{n} \sum_{i=1}^n x_{i, j} \\
    &\ (\sigma(\batch))_j = \sqrt{ \frac{1}{n} \sum_{i=1}^n x_{i, j}^2 - \left( \frac{1}{n} \sum_{i=1}^n x_{i, j} \right)^2 } 
\end{align*}

Then, we can leverage the above formula to search for the optimal malicious data. 

\subsection{Analysis of Smoothing via Training-time BN Statistics}
We robustly estimate the final BN statistics by $
    \bar{\mu}= \tau \mus+ (1-\tau) \mut,  \bar{\sigma}^2=\tau \sigmas^2+(1-\tau) \sigmat^2  
$, where ($\mus, \sigmas^2$) are training-time BN and ($\mut = \mu(\batch), \sigmat^2 = \sigma^2(\batch)$) are test-time BN (shown in Section \ref{sec:dfBN}).
Therefore, the output of a linear layer with smoothed batch normalization on a single dimension can be re-written as 

\begin{align}
    f(\xtgt) 
    = \frac{\xtgt - ((1-\tau) \mu(\batch) + \tau \mus)} { \sqrt{(1-\tau) \sigma^2(\batch) + \tau \sigmas^2} } \Vec{w} + b
\end{align}

The new gradient can be computed by  

\begin{align*}
    \frac{\del f(\xtgt)}{\del x_{i, j^*}} 
    &= \frac{ (-\frac{1-\tau}{n}) \Tilde{\sigma}_j
    - 
    [ (\xtgt)_j - (1-\tau) \mu(\batch)_j - \tau (\mus)_j ]
    \frac{1}{2} (\Tilde{\sigma}_j)^{-1}
    (1-\tau)
    \frac{ \del (\sigma(\batch))_j }{ \del x_{i, j} }  }{ (1-\tau) (\sigma(\batch))_j^2+ \tau (\sigmas^2 )_j } w_j \\
    &= \frac{
    (-\frac{1-\tau}{n})\left\{
     (\sigma(\batch))^2_j 
    - \left[
    (\xtgt)_j - (1-\tau) (\mu(\batch))_j - \tau (\mus)_j
    \right] 
    (\Tilde{\sigma}_j)^{-1}
    ( x_{i, j^*} - (\mu(\batch))_j )
    \right\} 
    }{   (1-\tau) (\sigma(\batch))_j^2+ \tau (\sigmas^2 )_j}
    w_j\\
    \text{where} \ \ &\ (\mu(\batch))_j = \frac{1}{n} \sum_{i=1}^n x_{i, j}, \\
&\ (\sigma(\batch))_j = \sqrt{ \frac{1}{n} \sum_{i=1}^n x_{i, j}^2 - ( \frac{1}{n} \sum_{i=1}^n x_{i, j} )^2 } \\ 
&\ \Tilde{\sigma}_j := \sqrt{  (1-\tau) (\sigma(\batch))_j^2+ \tau (\sigmas^2 )_j }. 
\end{align*}

Here, whenever $\mu(\batch) \approx \mus$ , and  $\sigma^2(\batch) \approx \sigmas^2$, the norm of graident $||\frac{\del f(\xtgt)}{\del x_{i, j^*}} ||$ will decrease as $\tau$ increases. Specifically, $||\frac{\del f(\xtgt)}{\del x_{i, j^*}} || = 0$ when $\tau = 1$, which means the BN statistics is only based on the training-time BN and do not involve any adversarial risks.








\newpage
\section{Preliminary Results}

\subsection{Sanity Check of Utilizing Training-time BN Statistics for TTA}
\label{sec:trbn}
In this experiment, we evaluate the performance of TTA using training-time BN statistics on CIFAR-10-C, CIFAR-100-C, and ImageNet-C with the ResNet architecture \cite{He2016DeepRL}.
Note that we did not include the result of the \textbf{TeBN} method, as it is identical to the \textbf{Source} method (directly using the source model without any TTA method). 
For other TTA methods, only affine transformation parameters (i.e.,  scale $\gamma_{l}$ and shift $\beta_{l}$) within the BN layer are updated. All other hyperparameters stay the same with experiments in \citet{Goyal2022TestTimeAV}, including  batch size 200 and  TTA learning rate $\eta$ = $0.001$. 

\begin{figure}[H]
\begin{center}
\begin{tabular}{ccc}
\subfigure{  \includegraphics[width=0.28\textwidth]{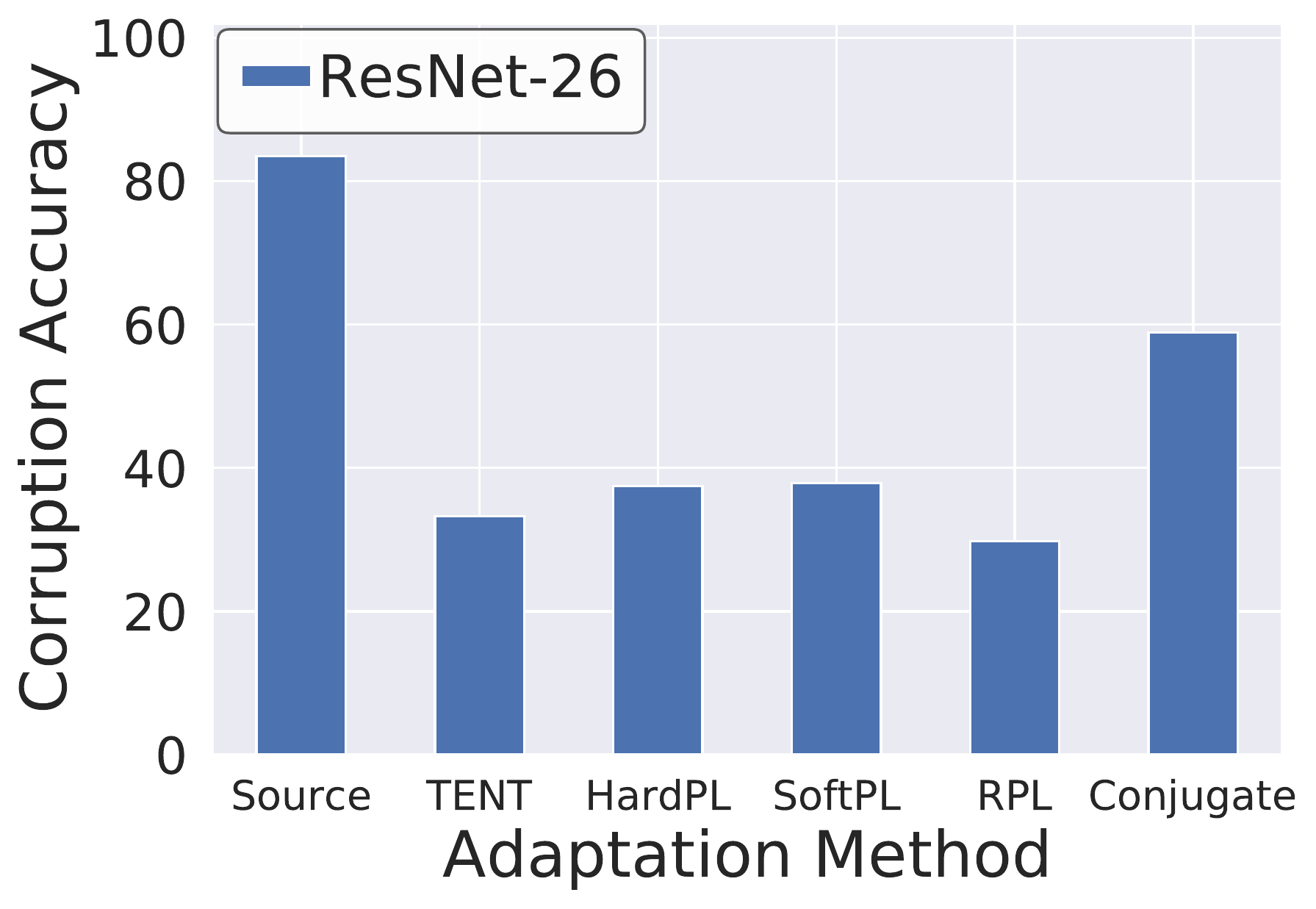}} &
 \subfigure{ \includegraphics[width=0.28\textwidth]{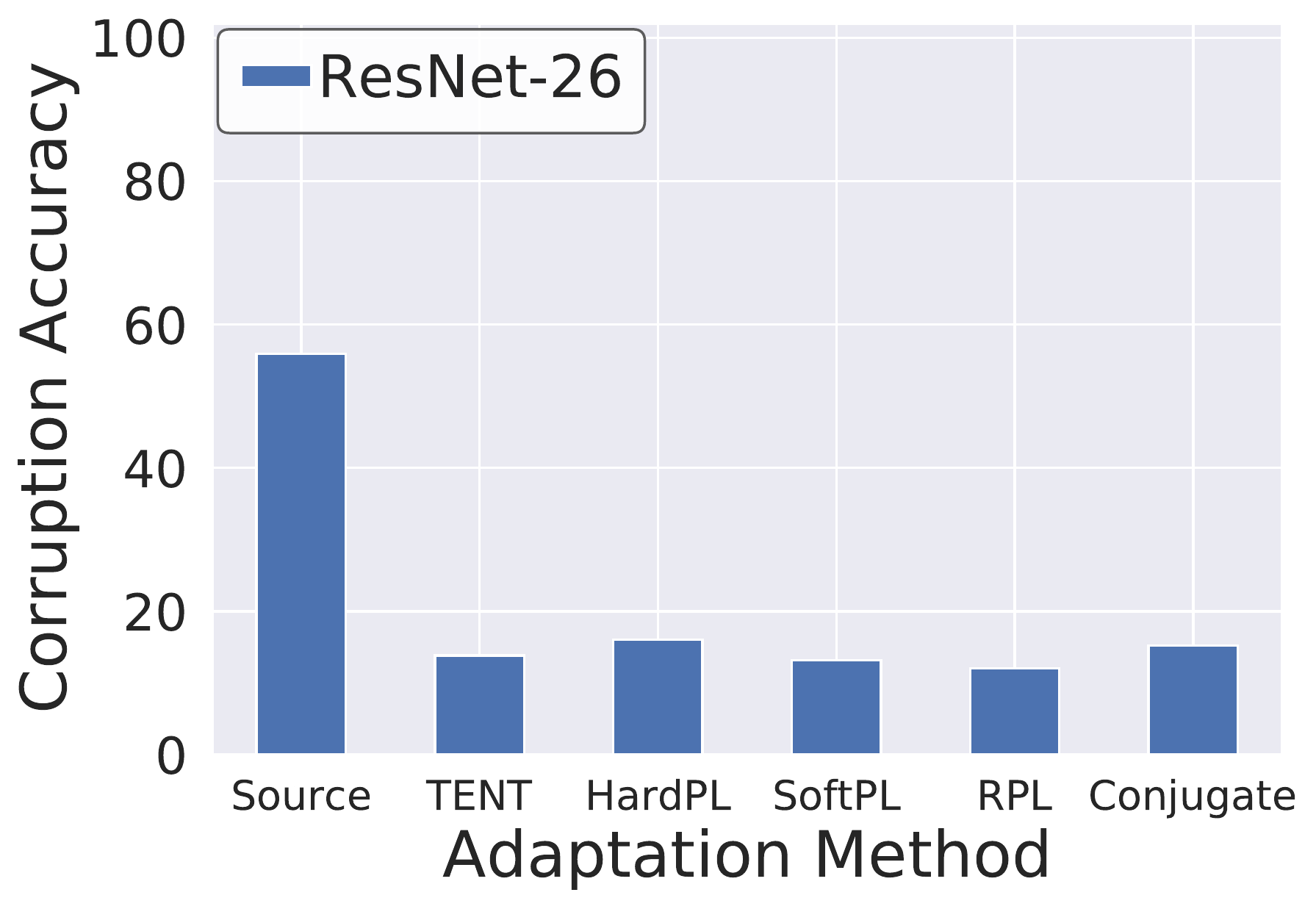}} &  
 \subfigure{ \includegraphics[width=0.28\textwidth]{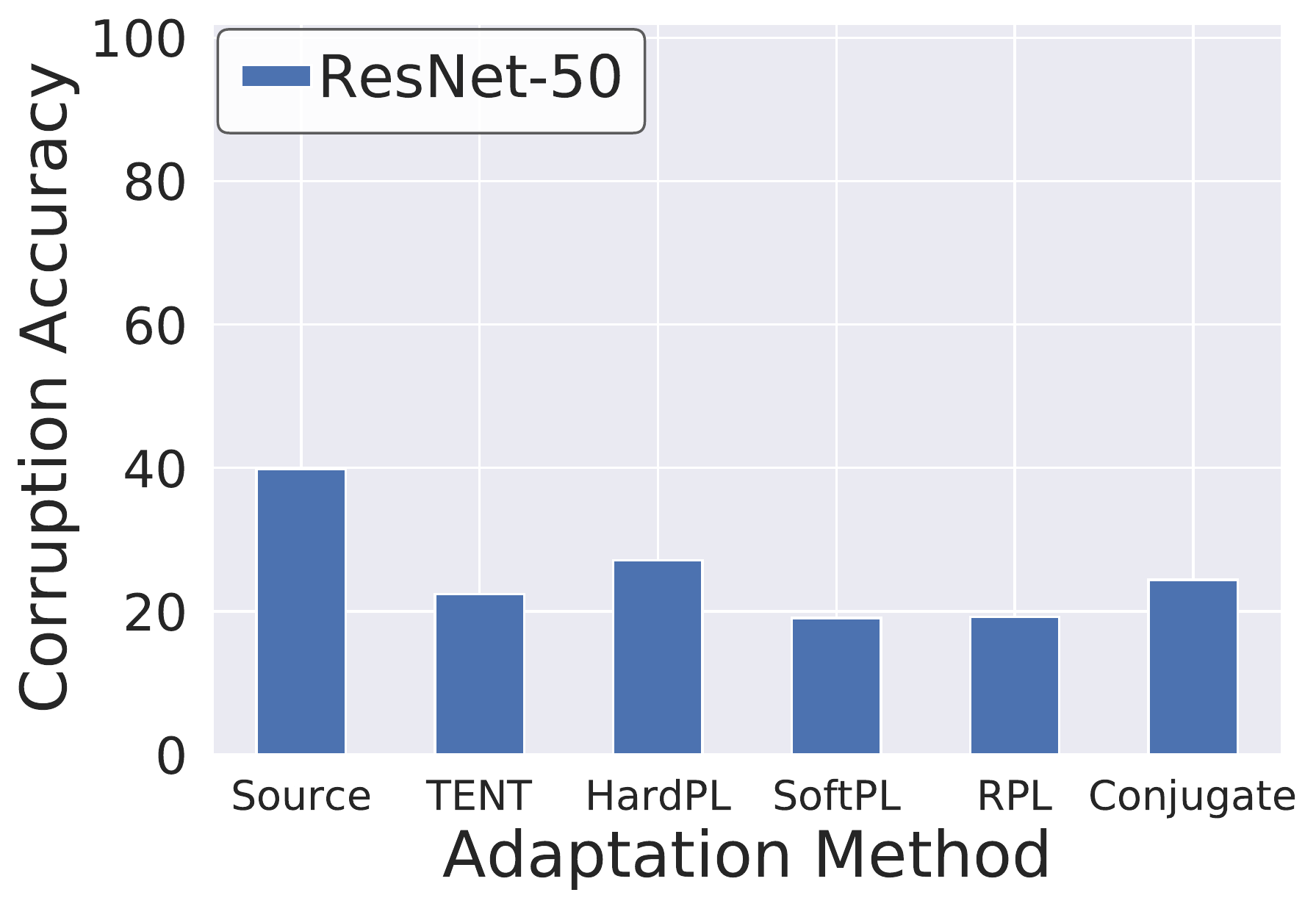}}\\
 (a) CIFAR-10-C & (b) CIFAR-100-C & (c) ImageNet-C \\
\end{tabular}
	\caption{Test-time adaptations exhibit an obvious degradation on corruption datasets when using training-time BN statistics. [Severity Level: 3] }
	\label{Fig:corr_acc_trbn}
\end{center}
\end{figure}

\smallskip\noindent\textbf{Adopting training-time BN completely ruins performance gain for TTA (Figure \ref{Fig:corr_acc_trbn}).} 
We observe that all TTA methods exhibit a significant degradation (from $\sim$15\% to $\sim$50\%) across benchmarks. 
It can be inferred that the utilization of test-time batch normalization statistics is paramount to all TTA methods implemented.
\subsection{Preliminary Results of Bilevel Optimization}
\label{sec:bilevelres}

We study the effectiveness of utilizing bilevel optimization to find the malicious data $\hatmal$. 
\begin{table}[H]
\centering
\caption{Attack success rate of Distribution Invading Attack with and without bilevel optimization. (We omit \textbf{TeBN} as it is identical for both methods) }
\resizebox{0.98\textwidth}{!}{%
\begin{tabular}{cccccccc}
\toprule
\textbf{Dataset} & \textbf{ $N_m$} & \textbf{Bilevel}  & \textbf{TENT(\%)} & \textbf{Hard PL(\%)} & \textbf{Soft PL(\%)} & \textbf{Robust PL(\%)} & \textbf{Conjugate PL(\%)}  \\ \midrule
\multirow{6}{*}{\textbf{\begin{tabular}[c]{@{}c@{}}CIFAR-10-C\\ (ResNet26)\end{tabular}}}  
& 10 (5\%)  & \xmark       & 23.20 &   25.33 &   23.20 & 24.80 &      23.60  \\
& 10 (5\%)  & $\checkmark$ & 22.93 &   23.73 &   22.80 & 24.00 &      23.73 \\
& 20 (10\%) & \xmark       & 45.73 &   48.13 &   47.47 & 49.47 &      45.73 \\
& 20 (10\%) & $\checkmark$ & 44.40 &   47.33 &   46.40 & 47.60 &      45.20 \\
& 40 (10\%) & \xmark       & 83.87 &   84.27 &   82.93 & 86.93 &      85.47 \\ 
& 40 (10\%) & $\checkmark$ & 82.80 &   84.53 &   82.40 & 84.53 &      84.67 \\
\hline
\multirow{6}{*}{\textbf{\begin{tabular}[c]{@{}c@{}}CIFAR-100-C\\ (ResNet26)\end{tabular}}} 
& 10 (5\%)  & \xmark       & 26.40 &   31.20 &   27.60 & 32.13 &      26.13 \\
& 10 (5\%)  & $\checkmark$ & 26.93 &   31.87 &   28.80 & 32.93 &      26.40 \\
& 20 (10\%) & \xmark       & 72.80 &   87.33 &   78.53 & 82.93 &      71.60  \\
& 20 (10\%) & $\checkmark$ & 72.27 &   85.87 &   78.13 & 85.47 &      72.13 \\
& 40 (10\%) & \xmark        & 100.00 &   99.87 &  100.00 & 99.87 &     100.00\\ 
& 40 (10\%) & $\checkmark$  & 100.00 &   99.73 &  100.00 & 100.00 &     100.00 \\
\hline
\multirow{6}{*}{\textbf{\begin{tabular}[c]{@{}c@{}}ImageNet-C\\ (ResNet50)\end{tabular}}}  

& 5  (2.5\%)  &  \xmark        &  75.73 &   69.87 &   62.67 & 66.40 &      57.87   \\ 
& 5  (2.5\%)  &  $\checkmark$  & 74.93 &   70.13 &   62.13 & 67.47 &      59.43 \\
& 10 (5\%)  & \xmark        & 98.67 &   96.53 &   94.13 & 96.00 &      92.80 \\
& 10 (5\%)  & $\checkmark$  & 98.40 &   97.07 &   94.40 & 96.27 &      93.43 \\
& 20 (10\%)  & \xmark        & 100.00 &  100.00 &  100.00 & 100.00 &     100.00 \\ 
& 20 (10\%)  & $\checkmark$  & 100.00 &  100.00 &  100.00 & 99.73 &     100.00 \\     
\bottomrule
\end{tabular}%
}
\label{tab:bilevel}
\end{table}

\smallskip\noindent\textbf{Our proposed single-level solution yields comparable effectiveness to the Bi-level Optimization approach (Table \ref{tab:bilevel}).}
We evaluate the DIA method with and without the bilevel optimization (inner loop) step, revealing that, in most cases, the difference between them is generally less than 1\%.
In addition, we observe that using bilevel optimization results in a substantial increase (by a factor of $\sim$10) in computational time.

\subsection{Effectiveness of Test-time Adaptations on Corruption Datasets}
\label{sec:ttap}

We demonstrate the effectiveness of applying TTA methods to boost corruption accuracy. We use \textbf{Source} to denote the performance of the source model without TTA.  

\begin{figure}[H]
  \begin{center}
  \begin{tabular}{ccc}
  \subfigure{  \includegraphics[width=0.28\textwidth]{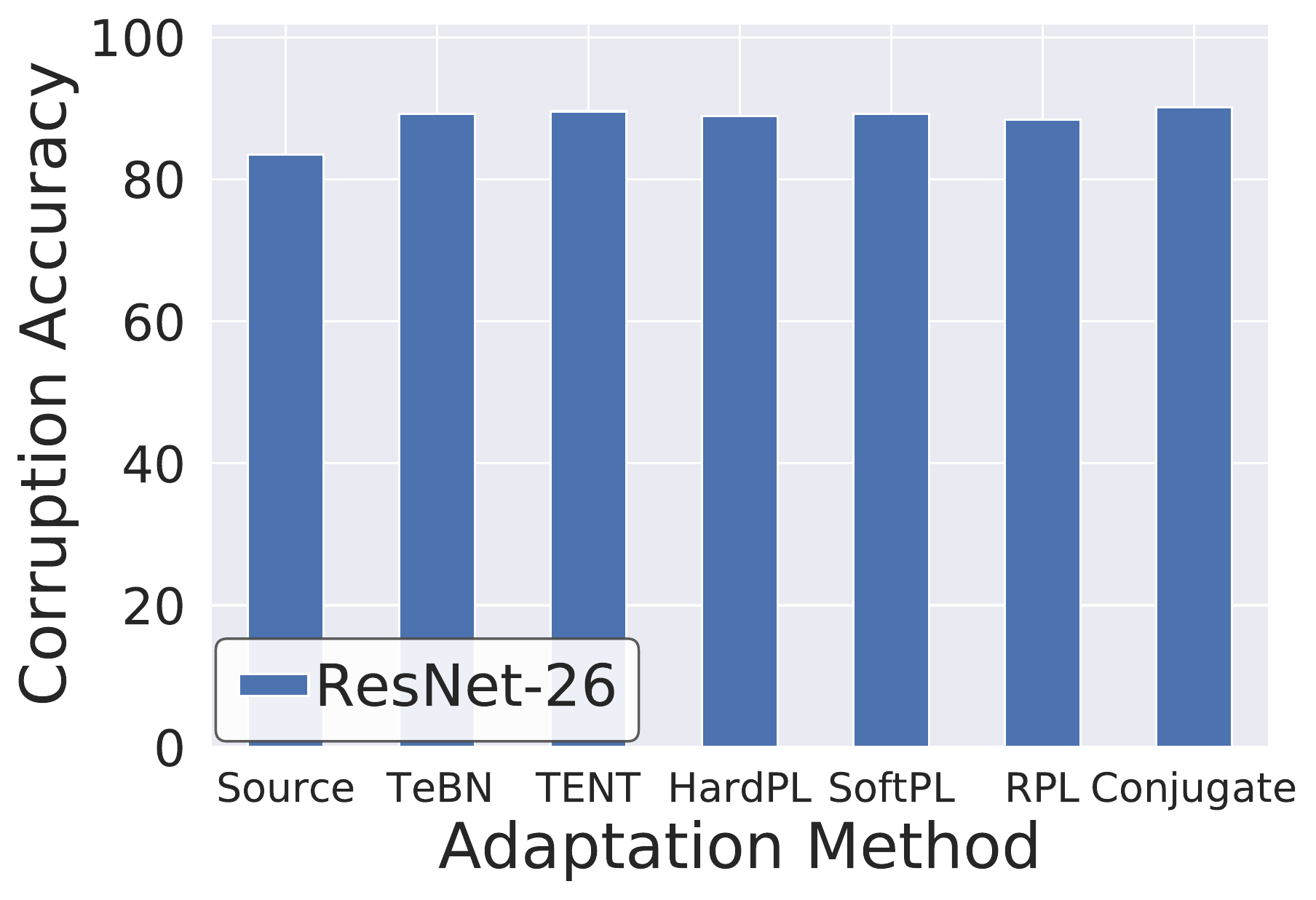}} &
   \subfigure{ \includegraphics[width=0.28\textwidth]{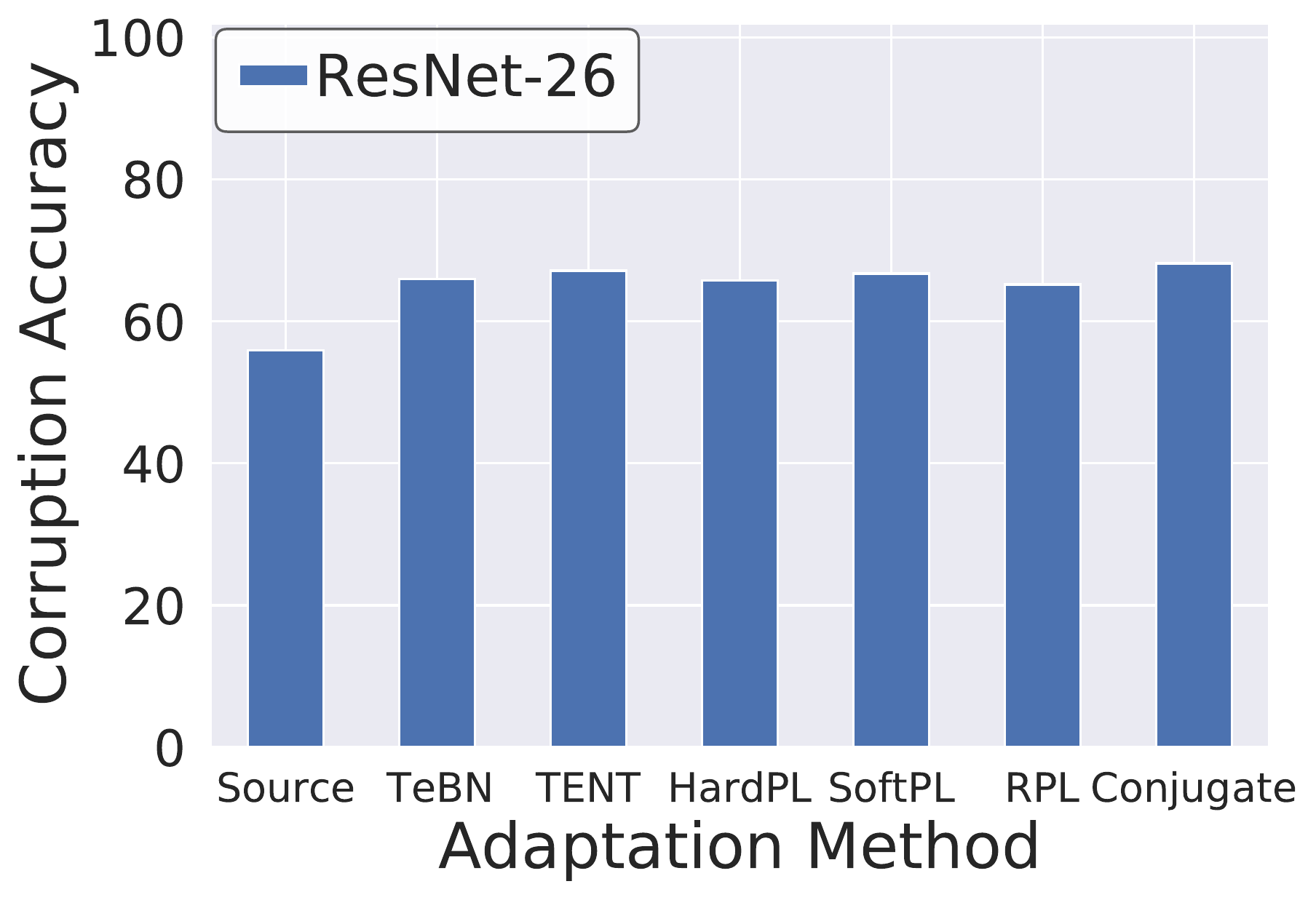}} &  
   \subfigure{ \includegraphics[width=0.28\textwidth]{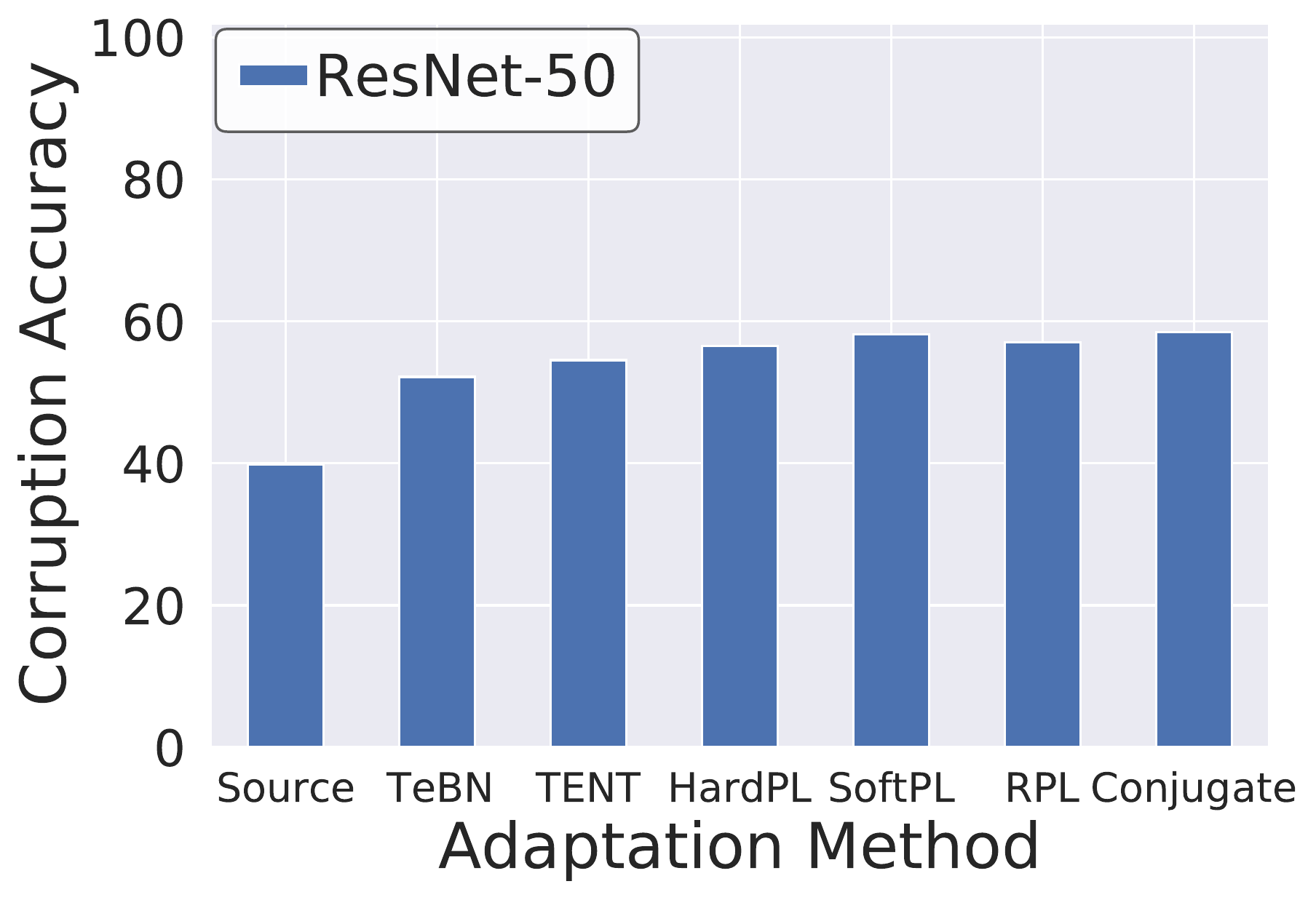}}\\
   (a) CIFAR-10-C & (b) CIFAR-100-C & (c) ImageNet-C \\
  \end{tabular}
    \caption{ Test-time adaptations consistently improve the accuracy on corruption datasets with severity level 3.  }
    \label{Fig:corr_acc}
  \end{center}
  \end{figure}

\smallskip\noindent\textbf{All TTA methods significantly boost the accuracy on distribution shift benchmarks (Figure \ref{Fig:corr_acc}).}
We observe the absolute performance gains of TTA on corrupted inputs are \textbf{$>$5\%} for CIFAR-10-C, \textbf{$>$10\%} for CIFAR-100-C, and \textbf{$>$12\%} for ImageNet-C.
These promising results incentivize 
the use of TTA methods
in various applications when test data undergoes a shift in distribution.

\newpage
\section{Additional Experiment Details and Results of Distribution Invading Attacks}
\label{sec:adddia}

In this appendix, we show some supplementary experimental details and results in Section \ref{sec:dia}. We begin by outlining our experimental setup in Appendix~\ref{sec:setup}. This is followed by the experiments that study the effectiveness of DIA by varying the number of malicious samples in Appendix~\ref{sec:diasample} and using various models and data augmentations in Appendix~\ref{append:arch}. Then, we delve into different types of attack objectives and constraints, including (1) indiscriminate attack (Appendix~\ref{append:IA}), (2) stealthy targeted attack (Appendix~\ref{append:per}), and (3) distribution invading attack via simulated corruptions (Appendix~\ref{append:simulated}). Besides, we examine various TTA design choices, such as larger learning rates (Appendix~\ref{append:large_lr}) and increased optimization steps (Appendix~\ref{append:opt_step}). We also consider the DIA under various corruption choices by adjusting the severities of corruption in Appendix~\ref{append:simulated} and present the detailed results of all types of corruptions in Appendix~\ref{append:cor}. Other new baseline methods are applied in Appendix~\ref{append:addtta}.

\subsection{Addtional details of Experiment setup}
\label{sec:setup}

\textbf{Dataset.}  Our attacks are evaluated on CIFAR-10 to CIFAR-10-C, CIFAR-100 to CIFAR-100-C, and ImageNet to ImageNet-C \citep{Hendrycks2019BenchmarkingNN}, which contain 15 types of corruptions on test data. 
We select all corruption types and set the severity of the corruption as 3 for most experiments.\footnote{the severities of the corruption range from 1 to 5, and 3 can be considered as medium corruption degree.} 
Therefore, our CIFAR-10-C and CIFAR-100-C evaluation sets contain 10,000 $\times$ 15 = 150,000 images with $32\times32$ resolution from 10 and 100 classes, respectively. 
For ImageNet-C, there are 5,000 $\times$ 15 = 75,000 high-resolution ($224\times224$) images from 1000 labels to evaluate our attacks.  

\smallskip\noindent\textbf{Source Model Details.} We follow the common practice on distribution shift benchmarks \cite{Hendrycks2019BenchmarkingNN}, and train our models on the CIFAR-10, CIFAR-100, and ImageNet. For training models on the CIFAR dataset, we use the SGD optimizer with a 0.1 learning rate, 0.9 momentum, and 0.0005 weight decay. We train the model with a batch size of 256 for 200 epochs. We also adjust the learning rate using a cosine annealing schedule \cite{Loshchilov2016SGDRSG}. 
This shares the same configurations with \citet{Goyal2022TestTimeAV}.
As we mentioned previously, the architectures include ResNet with 26 layers \cite{He2016DeepRL}, VGG 19 with layers  \cite{Simonyan2015VeryDC}, Wide ResNet with 28 layers \cite{Zagoruyko2016WideRN}. 
For the ImageNet benchmark, we directly utilize the models downloaded from RobustBench~\cite{croce2020robustbench} (\href{https://robustbench.github.io/}{https://robustbench.github.io/}), including standard trained ResNet-50 \cite{He2016DeepRL}, AugMix \cite{Hendrycks2020AugMixAS}, DeepAugment \cite{Hendrycks2021TheMF}, robust models from \citet{Salman2020DoAR}, and \citet{robustness}.  We want to emphasize that TTA does not necessarily need to train a model from scratch, and downloading from outsourcing is acceptable and sometimes desirable.   

\smallskip\noindent\textbf{Test-time Adaptation Methods.} Six test-time adaptation methods are selected, including \textbf{TeBN}~\cite{Schneider2020ImprovingRA}, \textbf{TENT}~\cite{Wang2021TentFT},  \textbf{Hard PL} ~\cite{lee2013pseudo,Galstyan2007EmpiricalCO},  \textbf{Soft PL} ~\cite{lee2013pseudo,Galstyan2007EmpiricalCO},
\textbf{Robust PL}~\cite{rusak2022if},
 and \textbf{Conjugate PL}~\cite{Goyal2022TestTimeAV}, where they all obtain obvious performance gains when data distribution shifts. 
We use the same default hyperparameters with \citet{Wang2021TentFT} and  \citet{Goyal2022TestTimeAV}, where the code is available at \href{https://github.com/DequanWang/tent}{github.com/DequanWang/tent} and \href{https://github.com/locuslab/tta_conjugate}{github.com/locuslab/tta-conjugate}.
Besides setting the batch size to \textbf{200}, we use  Adam for the TTA optimizer, $\eta=0.001$ for the TTA learning rate, and 1 for temperature. TTA is done in 1 step for each test batch. 

\smallskip\noindent\textbf{Attack Setting.} Each test batch is an individual attacking trial, which contains 200 samples. 
Therefore, there are 150,000/200 = 750 trials for CIFAR-C and 75,000/200 = 375 trials for ImageNet-C. 
All of our experimental results are averaged across all trials. 
For targeted attacks, we use \textbf{attack success rate (ASR)} as the evaluation metrics for attack effectiveness, which means the ratio of DIA can flip the targeted sample to the pre-select label. 
For indiscriminate attacks, we use corruption \textbf{corruption error rate} (i.e., the error rate on the benign corrupted data) to measure the effectiveness. For stealthy targeted attacks, we again use \textbf{attack success rate (ASR)} as the attacking effectiveness metrics. 
In addition, since we want to maintain the corruption accuracy, we leverage the \textbf{corruption accuracy degradation}  (i.e., the accuracy drop of benign corrupted data compared to ``no attack'') as the metric. 

For other hyperparameters, we set the attacking steps $N=500$ and attacking optimization rate $\alpha = 1/255$. In addition, the attack effectiveness can be further improved if applying more iterations or multiple random initializations like \citet{Madry2018TowardsDL}. We leave more enhancing DIA techniques as future directions. 

\newpage
\subsection{More Experiments of DIA across Number of Malicious Samples}
\label{sec:diasample}

\begin{figure}[H]
\centering
\begin{tabular}{ccc}
  \includegraphics[width=0.28\textwidth]{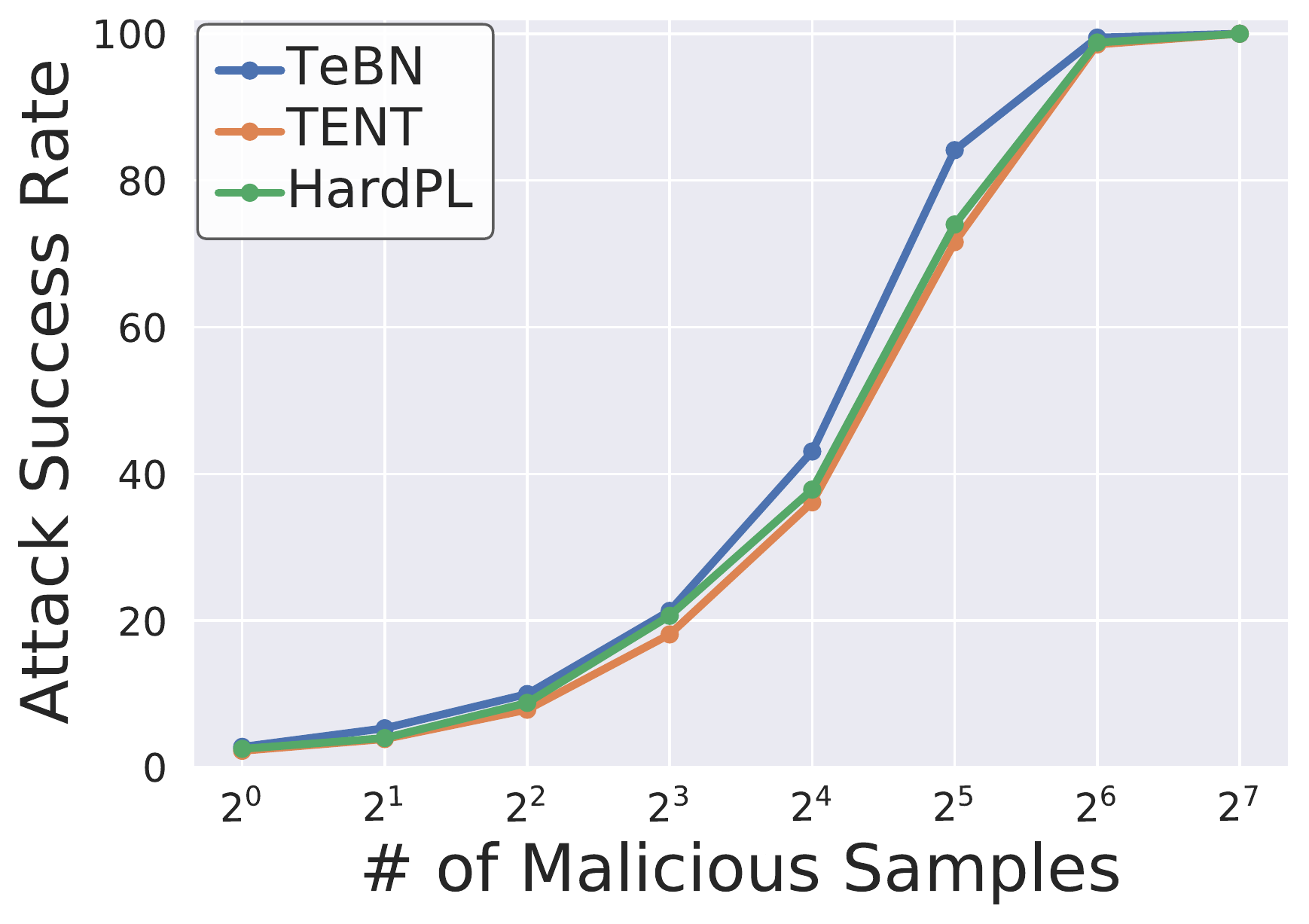} &
 \includegraphics[width=0.28\textwidth]{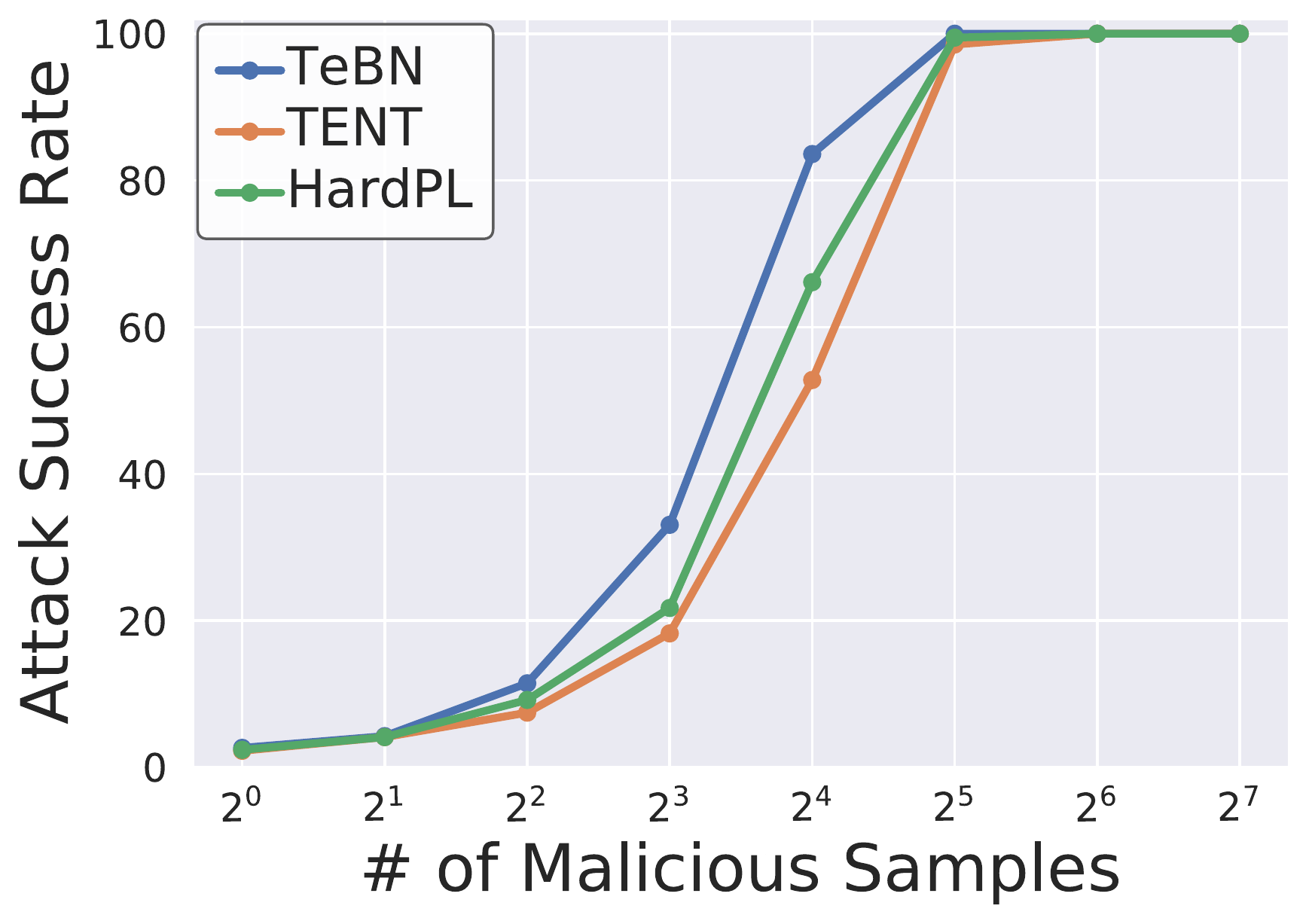} &  \includegraphics[width=0.28\textwidth]{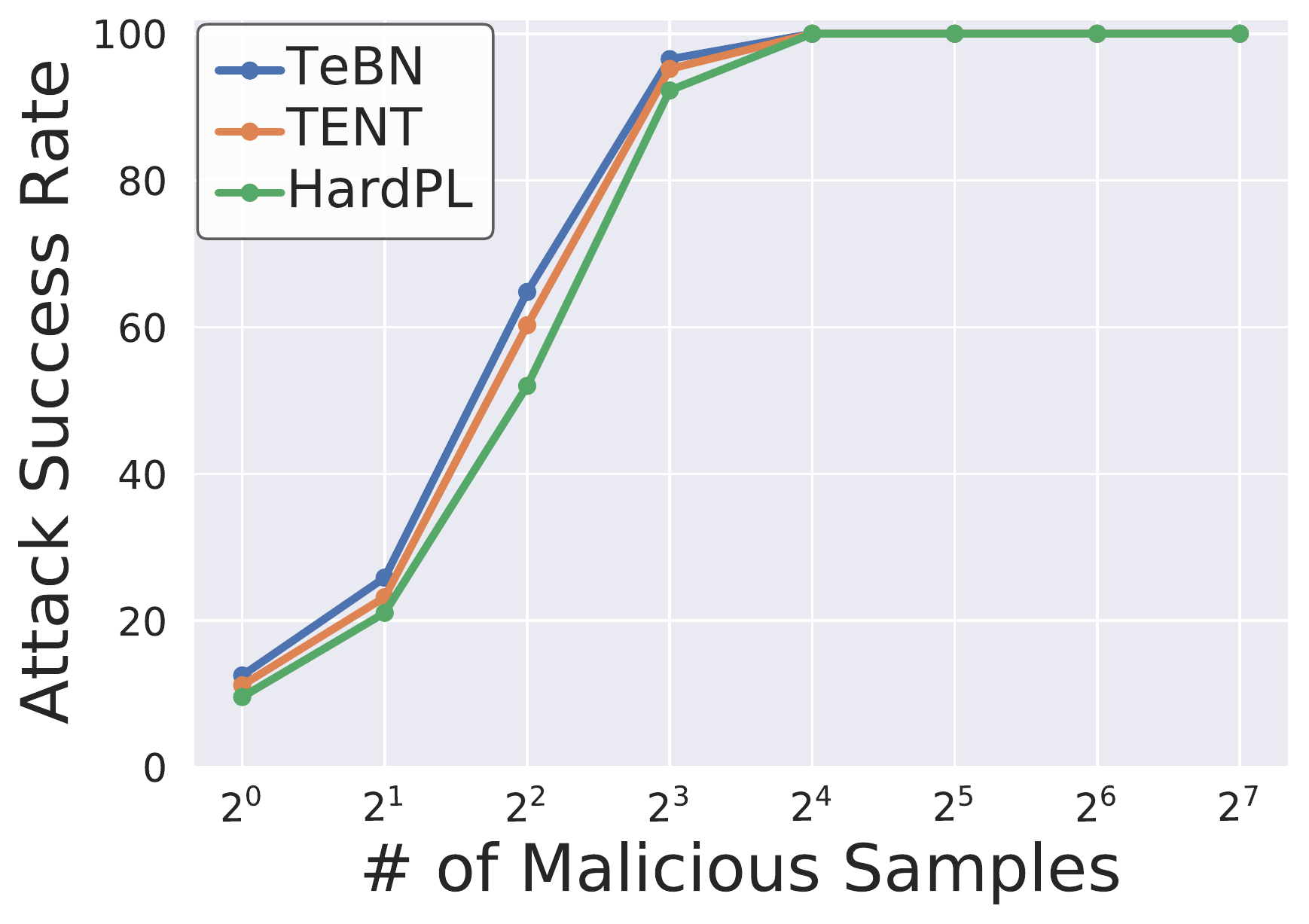}\\
(a) CIFAR-10-C & (b) CIFAR-100-C & (c) ImageNet-C \\
\end{tabular}
	\caption{ Success rate of our proposed attack across numbers of malicious samples from 1 to 128 (0.5\% to 64\%). [TTA: \textbf{TeBN}, \textbf{TENT}, and \textbf{Hard PL}] (Extended version of Figure \ref{Fig:mal_smaple}) }
	\label{Fig:mal_smaple_full}
\end{figure}

Figure~\ref{Fig:mal_smaple_full} depicts the effectiveness of the DIA success rate in relation to the number of malicious samples.
We selected 
\textbf{TeBN}, \textbf{TENT}, and \textbf{Hard PL} as demonstrated TTA methods and observed them perform similarly. Our conclusion ``DIA obtains near-100\% ASR, using 64 malicious samples for  CIFAR-10-C, 32 for CIFAR-100-C, and 16 for ImageNet-C'' still holds. 

\subsection{Effectiveness of DIA across Various Model Architectures and Data Augmentations}
\label{append:arch}

Then we conduct more comprehensive experiments on DIA across model architectures, data augmentations, and the number of malicious samples.
In Table \ref{tab:arch_all} (full version of Table \ref{tab:arch}), we observe that the attack success rate of the VGG \cite{Simonyan2015VeryDC} architecture is less affected by the number of malicious samples compared to the Wide ResNet \cite{Simonyan2015VeryDC} architecture. 
For example, the ASR of VGG improves from $\sim$42\% to $\sim$79\% while Wide ResNet improves from $\sim$16\% to $\sim$94\% if $N_m$ increases from 10 to 40 for the \textbf{TeBN} method on CIFAR-10-C. 
Therefore, our previous hypothesis should be modified to ``\textbf{with a sufficient number of malicious data}, more BN layers expose more vulnerabilities.''  
For ImageNet-C, we are still observing the mild mitigation effect from strong data augmentations.

\begin{table}[t]
      \centering
      \caption{ Effectiveness of distribution invading attack across various model architectures, data augmentation, and the number of malicious data. (Full version of Table \ref{tab:arch})}
      \resizebox{0.98\textwidth}{!}{%
      \begin{tabular}{@{}lcccccccc@{}}
      \toprule
      \textbf{Dataset} & \textbf{$N_m$} & \textbf{Architectures} & \textbf{TeBN(\%)} & \textbf{TENT(\%)} & \textbf{Hard PL(\%)} & \textbf{Soft PL(\%)} & \textbf{Robust PL(\%)} & \textbf{Conjugate PL(\%)}  \\ \midrule
      \multirow{9}{*}{\textbf{CIFAR-10-C}} & \multirow{3}{*}{\textbf{10 (5\%)}} 
            & ResNet-26        & 25.87 & 23.20 &   25.33 &   23.20 & 24.80 &      23.60 \\
      &     & VGG-19       & 41.73 & 30.27 &   30.13 &   28.00 & 33.07 &      29.07 \\
      &     & WRN-28     & 16.27 & 14.40 &   14.67 &   14.53 & 15.73 &      13.87 \\
      \cmidrule(l){2-9}
       & \multirow{3}{*}{\textbf{20 (10\%)}} 
            & ResNet-26         & 55.47 & 45.73 &   48.13 &   47.47 & 49.47 &      45.73 \\
      &     & VGG-19        & 60.13 & 44.80 &   46.67 &   44.13 & 46.67 &      44.67 \\
      &     & WRN-28      & 46.00 & 41.47 &   43.33 &   41.60 & 44.00 &      40.53 \\
      \cmidrule(l){2-9}
       & \multirow{3}{*}{\textbf{40 (20\%)}} 
            & ResNet-26         & 92.80 & 83.87 &   84.27 &   82.93 & 86.93 &      85.47 \\
      &     & VGG-19        & 79.07 & 65.33 &   67.47 &   64.13 & 67.47 &      64.13 \\
      &     & WRN-28      & 93.73 & 89.60 &   90.80 &   90.13 & 91.20 &      89.33 \\ \midrule
      \multirow{9}{*}{\textbf{CIFAR-100-C}} & \multirow{3}{*}{\textbf{10 (5\%)}} 
            & ResNet-26        & 46.80 & 26.40 &   31.20 &   27.60 & 32.13 &      26.13 \\
      &     & VGG-19       & 42.13 & 32.00 &   41.33 &   33.60 & 33.87 &      37.33 \\
      &     & WRN-28     & 29.60 & 14.67 &   16.53 &   15.47 & 18.13 &      14.53 \\\cmidrule(l){2-9}
       & \multirow{3}{*}{\textbf{20 (10\%)}} 
            & ResNet-26        & 93.73 & 72.80 &   87.33 &   78.53 & 82.93 &      71.60 \\
      &     & VGG-19       & 71.60 & 57.33 &   71.60 &   61.87 & 63.60 &      66.13 \\
      &     & WRN-28     & 64.80 & 38.67 &   44.13 &   40.93 & 46.40 &      39.87 \\\cmidrule(l){2-9}
       & \multirow{3}{*}{\textbf{40 (20\%)}} 
            & ResNet-26        & 100.00 & 100.00 &   99.87 &  100.00 & 99.87 &     100.00 \\
      &     & VGG-19       &  88.00 &  82.93 &   86.53 &   85.73 & 87.33 &      85.47 \\
      &     & WRN-28     &  97.47 &  83.60 &   87.07 &   86.67 & 90.13 &      79.73 \\ 
                                            \toprule
      \textbf{Dataset} & \textbf{$N_m$} & \textbf{Augmentations} & \textbf{TeBN(\%)} & \textbf{TENT(\%)} & \textbf{Hard PL(\%)} & \textbf{Soft PL(\%)} & \textbf{Robust PL(\%)} & \textbf{Conjugate PL(\%)}  \\ \midrule
      \multirow{9}{*}{\textbf{ImageNet-C}}  & \multirow{3}{*}{\textbf{5 (2.5\%)}}
      &  Standard      & 80.80 & 75.73 &   69.87 &   62.67 & 66.40 &      57.87 \\
      & & AugMix       & 72.53 & 65.60 &   59.47 &   53.60 & 56.27 &      49.07 \\
      & & DeepAugment  & 67.20 & 63.73 &   58.67 &   53.87 & 57.33 &      53.33 \\\cmidrule(l){2-9}                       
       & \multirow{3}{*}{\textbf{10 (5\%)}}
      &  Standard        & 99.47 & 98.67 &   96.53 &   94.13 & 96.00 &      92.80 \\
      & & AugMix         & 98.40 & 96.00 &   93.60 &   88.27 & 92.00 &      87.20 \\
      & & DeepAugment    & 96.00 & 94.67 &   91.20 &   87.47 & 89.60 &      86.67 \\\cmidrule(l){2-9}          
       & \multirow{3}{*}{\textbf{20 (10\%)}}
      &  Standard              & 100.00 & 100.00 &  100.00 &  100.00 & 100.00 &     100.00 \\
      & & AugMix        & 100.00 & 100.00 &  100.00 &  100.00 & 100.00 &     100.00 \\
      & & DeepAugment      & 100.00 & 100.00 &  100.00 &  100.00 & 100.00 &     100.00 \\                             
      \bottomrule
      \end{tabular}%
      }
      \label{tab:arch_all}
      \end{table}

\subsection{Effectiveness of Indiscriminate Attack}
\label{append:IA}

We further present detailed results of indiscriminate attacks with more TTA methods and options for the number of malicious samples.
Since the benign \textbf{corruption error rates} are different for different numbers of malicious data, we include the corruption error rate improvement in the bracket with \textcolor{red}{red} color.  

\begin{table}[H]
\centering
\caption{Average corruption error rate of TTA when deploying
indiscriminate attack. The \textcolor{red}{red} number inside the bracket is the corruption error rate improvement. (Extended version of Table \ref{tab:all})}
\resizebox{0.98\textwidth}{!}{%
\begin{tabular}{cccccccc}
\toprule
\textbf{Dataset} & \textbf{ $N_m$} & \textbf{TeBN(\%)} & \textbf{TENT(\%)} & \textbf{Hard PL(\%)} & \textbf{Soft PL(\%)} & \textbf{Robust PL(\%)} & \textbf{Conjugate PL(\%)}  \\ \midrule
\multirow{3}{*}{\textbf{\begin{tabular}[c]{@{}c@{}}CIFAR-10-C\\ (ResNet26)\end{tabular}}}  
& 10 (5\%)  &  15.66 \rr{(+4.96)} & 15.16 \rr{(+4.74)}  &   15.86 \rr{(+4.84)}&  15.56 \rr{(+4.86)}& 16.60 \rr{(+5.04)}&  14.32 \rr{(+4.51)} \\
& 20 (10\%) &  20.09 \rr{(+9.37)} & 19.43 \rr{(+9.00)}  &   20.14 \rr{(+9.10)}&  19.77 \rr{(+9.06)}& 21.04 \rr{(+9.46)}&  18.51 \rr{(+8.70)}\\
& 40 (20\%) &  28.02 \rr{(+17.29)}& 27.01 \rr{(+16.54)} &   27.91 \rr{(+16.86)}& 27.47 \rr{(+16.69)}&28.81 \rr{(+17.20)}& 26.09 \rr{(+16.26)}\\   
 \hline
\multirow{3}{*}{\textbf{\begin{tabular}[c]{@{}c@{}}CIFAR-100-C\\ (ResNet26)\end{tabular}}} 
& 10 (5\%)  &    43.84 \rr{(+7.22)}& 42.90  \rr{(+7.32)}&  44.65 \rr{(+7.92)}&    43.45 \rr{(+7.56)}&  45.34 \rr{(+8.02)}&  42.01 \rr{(+7.53)}\\
& 20 (10\%) &    51.99 \rr{(+13.44)}& 50.33 \rr{(+12.60)}& 52.00 \rr{(+13.33)}&   50.80 \rr{(+12.82)}& 52.52 \rr{(+13.21)}& 49.39 \rr{(+12.71)}\\
& 40 (20\%) &    58.41 \rr{(+23.89)}& 54.44 \rr{(+21.13)}& 56.59 \rr{(+21.93)}&   54.93 \rr{(+21.22)}& 56.97 \rr{(+21.72)}& 53.33 \rr{(+21.06)}\\
\hline
\multirow{3}{*}{\textbf{\begin{tabular}[c]{@{}c@{}}ImageNet-C\\ (ResNet50)\end{tabular}}}  
& 5 (2.5\%) & 57.42 \rr{(+9.45)} & 54.53 \rr{(+8.90)} &   52.32 \rr{(+8.70)} &   50.53 \rr{(+8.50)} & 51.52 \rr{(+8.43)}&   50.18 \rr{(+8.52)}\\
& 10 (5\%)  & 67.18 \rr{(+19.21)}& 62.39 \rr{(+16.76)}&   58.70 \rr{(+15.09)}&   55.90 \rr{(+13.88)}& 57.08 \rr{(+13.97)}&  55.60 \rr{(+13.94)}\\
& 20 (10\%) & 79.03 \rr{(+31.24)}& 75.01 \rr{(+29.57)}&   70.26 \rr{(+26.84)}&   65.67 \rr{(+23.81)}& 67.53 \rr{(+24.60)}&  65.30 \rr{(+23.80)}\\
\bottomrule                   
\end{tabular}%
}
\label{tab:all_all}
\end{table}

\noindent \textbf{The indiscriminate DIA causes a large error rate increase on benign samples (Table \ref{tab:all_all}).}  We observe that the ratio between the fraction of malicious data and error rate improvement is about 0.9$\times$ for CIFAR-10-C and 1.2$\times$ for CIFAR-100-C. For example, the error rate increases $\sim$12\% with 20 (10\%) malicious data for CIFAR-100-C. Significantly, 20 (10\%)  malicious data cause the error rate improves $\sim$25\% for ImageNet-C.
In addition, compared to other TTA methods, \textbf{TeBN} is still the most vulnerable method against DIA.  

\subsection{Effectiveness of Stealthy Targeted Attack}
\label{append:per}

\begin{figure}[H]
\centering
\begin{tabular}{cccc}
  \includegraphics[width=0.23\textwidth]{Figures/cifar_per/cifar10_per_norm_asr.pdf} & 
 \includegraphics[width=0.23\textwidth]{Figures/cifar_per/cifar10_per_norm_acc.pdf} & 
  \includegraphics[width=0.23\textwidth]{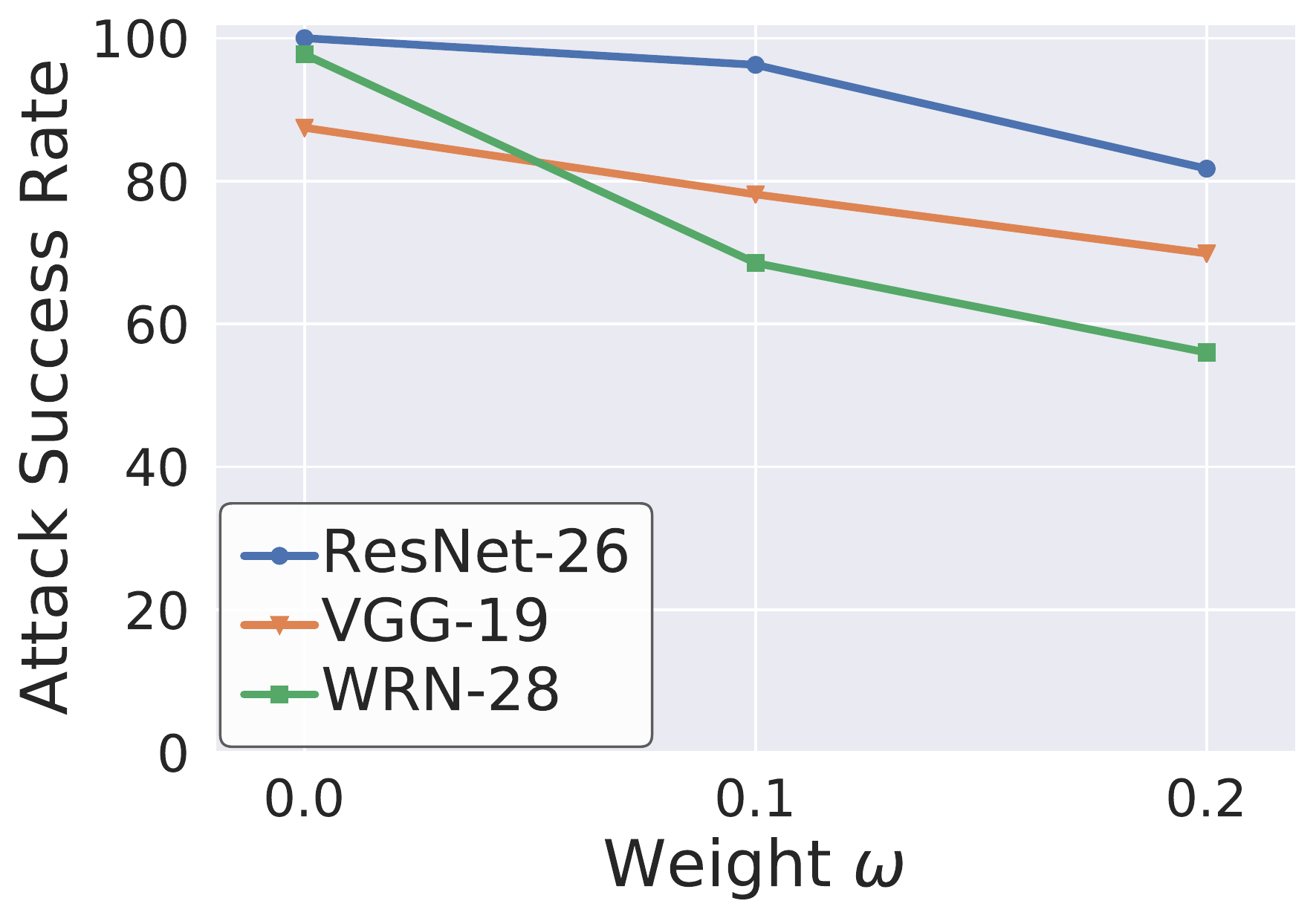} & 
 \includegraphics[width=0.23\textwidth]{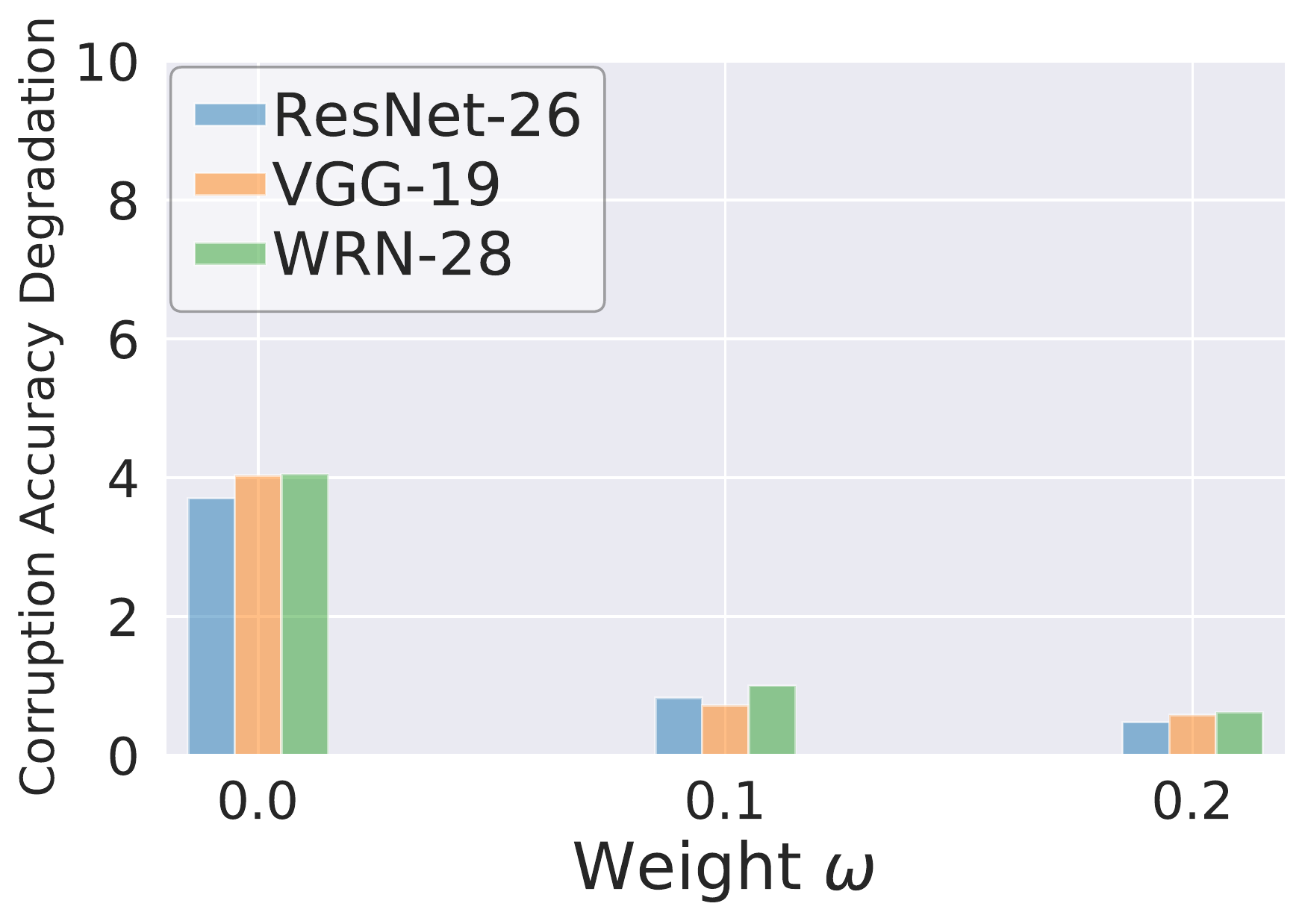} \\
 \multicolumn{2}{c}{(a) CIFAR-10-C}  & \multicolumn{2}{c}{(b) CIFAR-100-C} \\
\end{tabular}
\caption{ Adjusting the weight $\omega$ achieves a high attack success rate (line) and a minimal benign corruption accuracy degradation (bar). [$N_m$=40, TTA method: Norm] (Extended version of Figure \ref{Fig:per}) }
	\label{Fig:perall}
\end{figure}

Next, we present more results of the stealthy targeted attack on the CIFAR-C dataset in Figure \ref{Fig:perall}. We use 40 malicious data and \textbf{TeBN} method, where $\omega = \{0, 0.1, 0.2\}$. Specifically, we select three architectures, including ResNet-26, VGG-19, and WRN-28. Since different architectures result in various corruption accuracy, we use the \textbf{corruption accuracy degradation} to measure the stealthiness, where the best attack should have 0\% degradation.
We observe that we can obtain $<$ 2\%  degradation by sacrificing the attack success rates about 10\% when $\omega = 0.1$. 
In addition, we find the attack success rates of Wide ResNet-28 drop much more than other model architectures. 

\subsection{Distribution Invading Attack via Simulated Corruptions}
\label{append:simulated}

This subsection presents the details of generating simulated corruption and then demonstrates its effectiveness. 

\begin{figure}[H]
\centering
\includegraphics[width=0.75\textwidth]{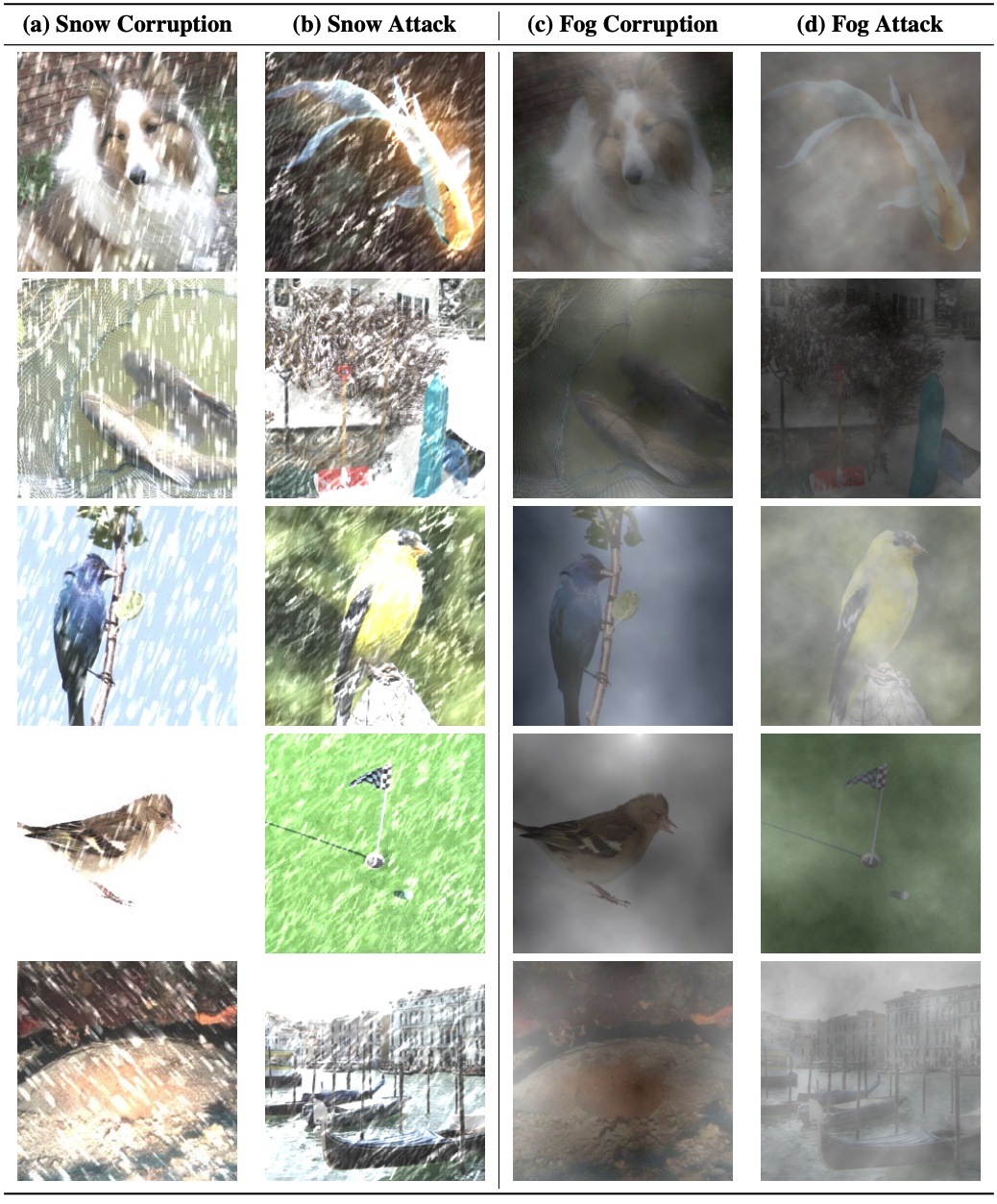}
\caption{Demonstrations of the corrupted snow and fog dataset from ImageNet-C with severity level 3 and adversarially corrupted images (DIA). We observe that manually distinguishing malicious and benign samples is hard. }
\label{fig:IN_CORR}
\end{figure}

\smallskip\noindent\textbf{Constructing Simulated Corruptions.} We directly utilize the methods of \citet{Kang2019TestingRA} to construct adversarially snow and fog effects, which are available at \href{https://github.com/ddkang/advex-uar}{github.com/ddkang/advex-uar}. Specifically, we use many tiny occluded image regions representing snowflakes at randomly chosen locations. Then, the intensity and directions of those ``snow pieces'' are adversarially optimized. Furthermore, the adversarial fog is generated by adversarially optimizing the diamond-square algorithm~\cite{Fournier1982ComputerRO}, a technique commonly used to create a random, stochastic fog effect. We present some snow and fog attack examples in Figure \ref{fig:IN_CORR}.

\smallskip\noindent\textbf{Simulated corruptions can be used as an input-stealthy DIA approach (Table \ref{tab:fog_snow_full}).} For evaluation, we use snow to attack clean images and insert them into the validation set of snow corruption. 
Similarly,  we use a fog attack for the fog corruption dataset. 
We observe that either adversarial snow or fog can effectively attack all TTA approaches with at least 60\% of attack success rate.

\begin{table}[H]
\centering
\caption{Average attack success rate of adversarially optimized snow corruptions and fog corruptions. [$N_m$=20; Target Attack] (Extended version of Table \ref{tab:fog_snow}) }
\resizebox{0.96\textwidth}{!}{%
\begin{tabular}{cccccccc}
\toprule
\textbf{Dataset} & \textbf{ Corruption} & \textbf{TeBN(\%)} & \textbf{TENT(\%)} & \textbf{Hard PL(\%)} & \textbf{Soft PL(\%)} & \textbf{Robust PL(\%)} & \textbf{Conjugate PL(\%)} \\ \midrule
\multirow{2}{*}{\textbf{\begin{tabular}[c]{@{}c@{}}ImageNet-C\\ (ResNet-50)\end{tabular}}} 
& Snow &   64.00 & 68.00 &   68.00 &   60.00 & 68.00 &     60.00 \\
& Fog &   84.00 & 76.00 &   72.00 &   76.00 & 76.00 & 76.00 \\
\bottomrule                   
\end{tabular}%
}\label{tab:fog_snow_full}
\end{table}

\subsection{Effectiveness of DIA across Severities of Corruption}
\label{append:severities}

\begin{figure}[H]
  \centering
  \begin{tabular}{cc}
    \includegraphics[width=0.35\textwidth]{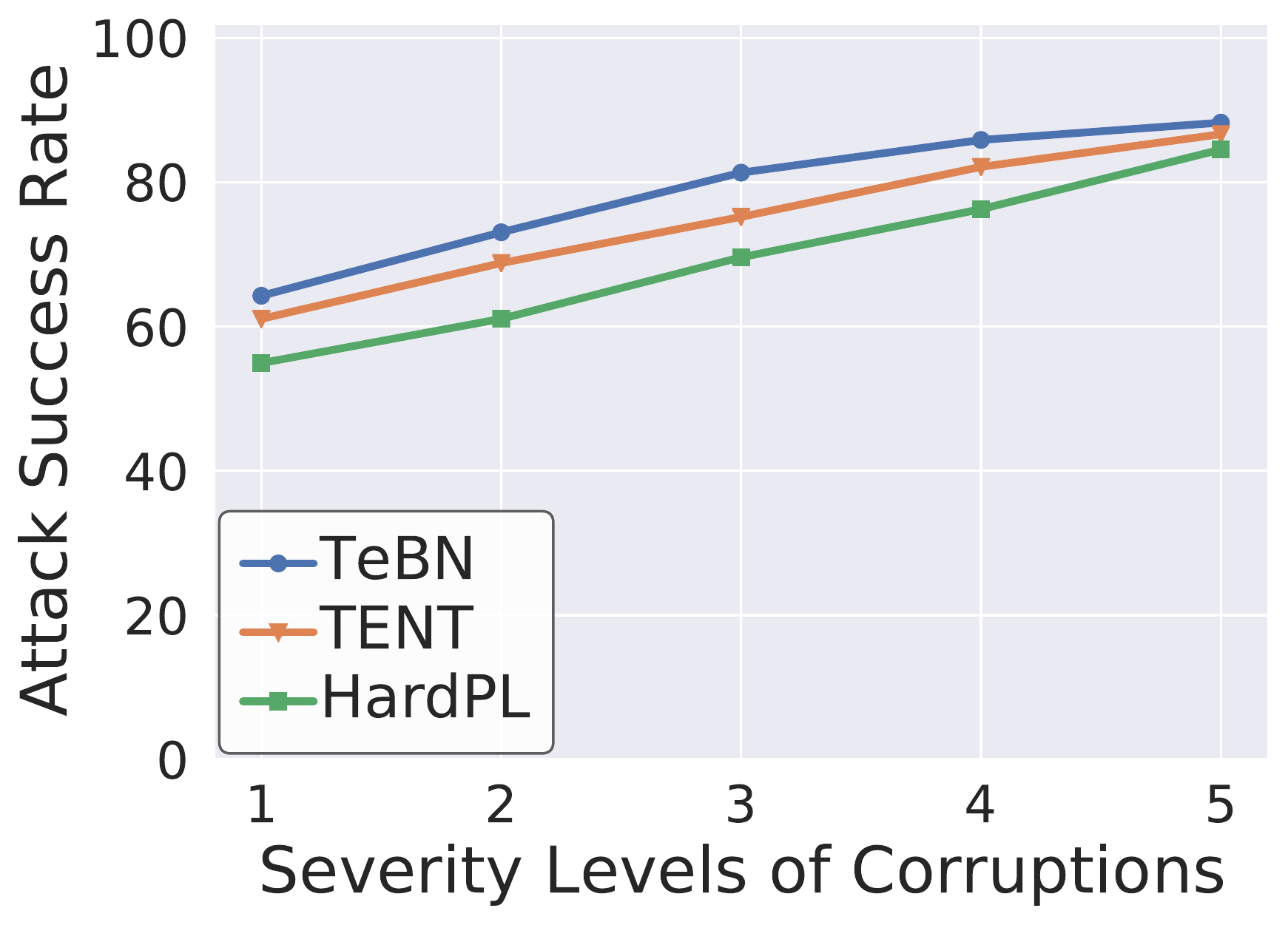} & 
   \includegraphics[width=0.35\textwidth]{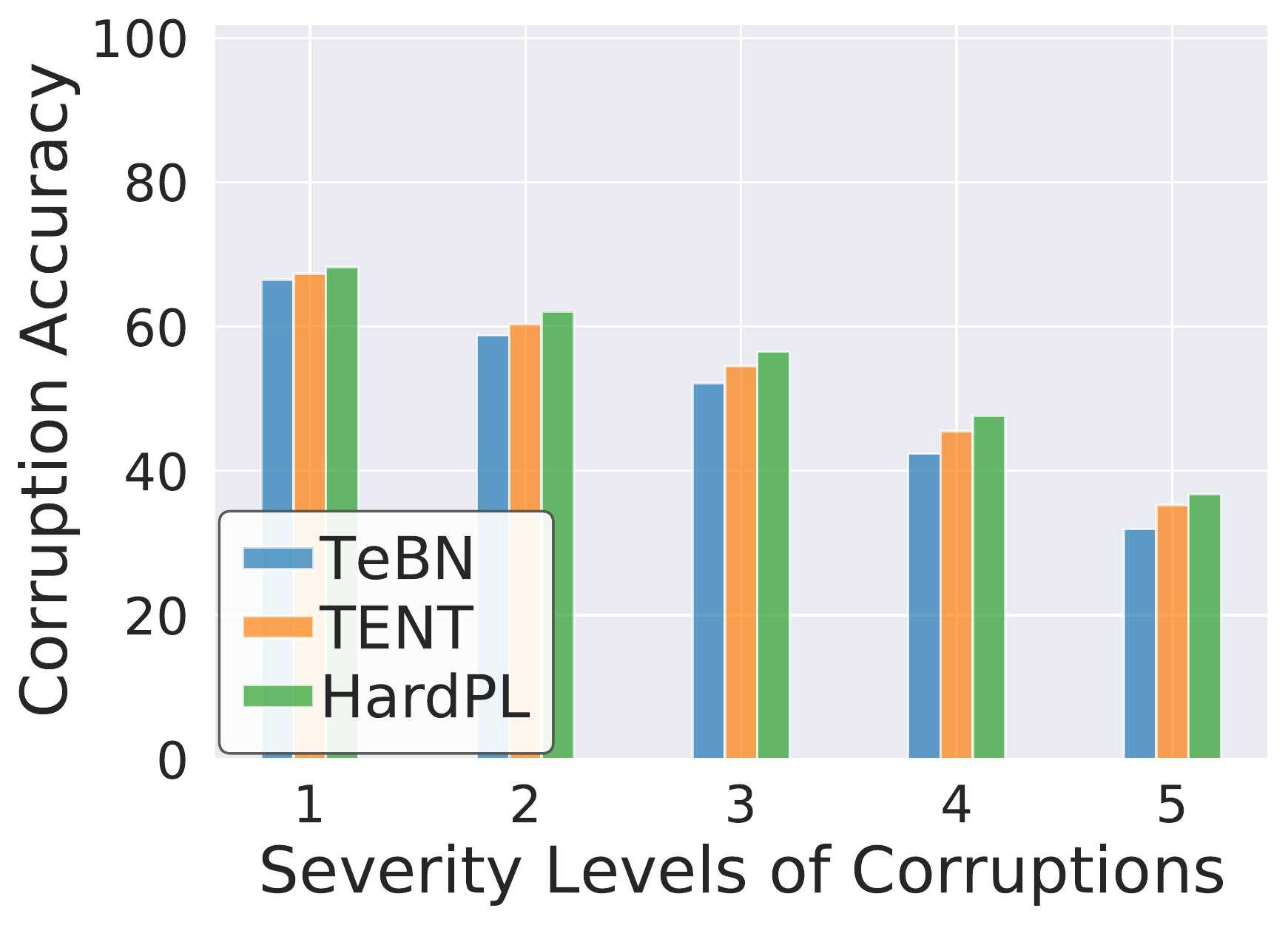} \\
    (a) Attack Success Rate & (b) Corruption Accuracy\\
  \end{tabular}
      \caption{Illustration of how severity levels of corruption affect the corruption accuracy and attack success rate of our attack on the ImageNet-C dataset. (Line plot: attack success rate; bar plot: corruption accuracy) [$N_m$=5; Target Attack] }
      \label{Fig:INC_augall}
  \end{figure}

  \begin{figure}[H]
  \centering
  \begin{tabular}{cccc}
    \includegraphics[width=0.23\textwidth]{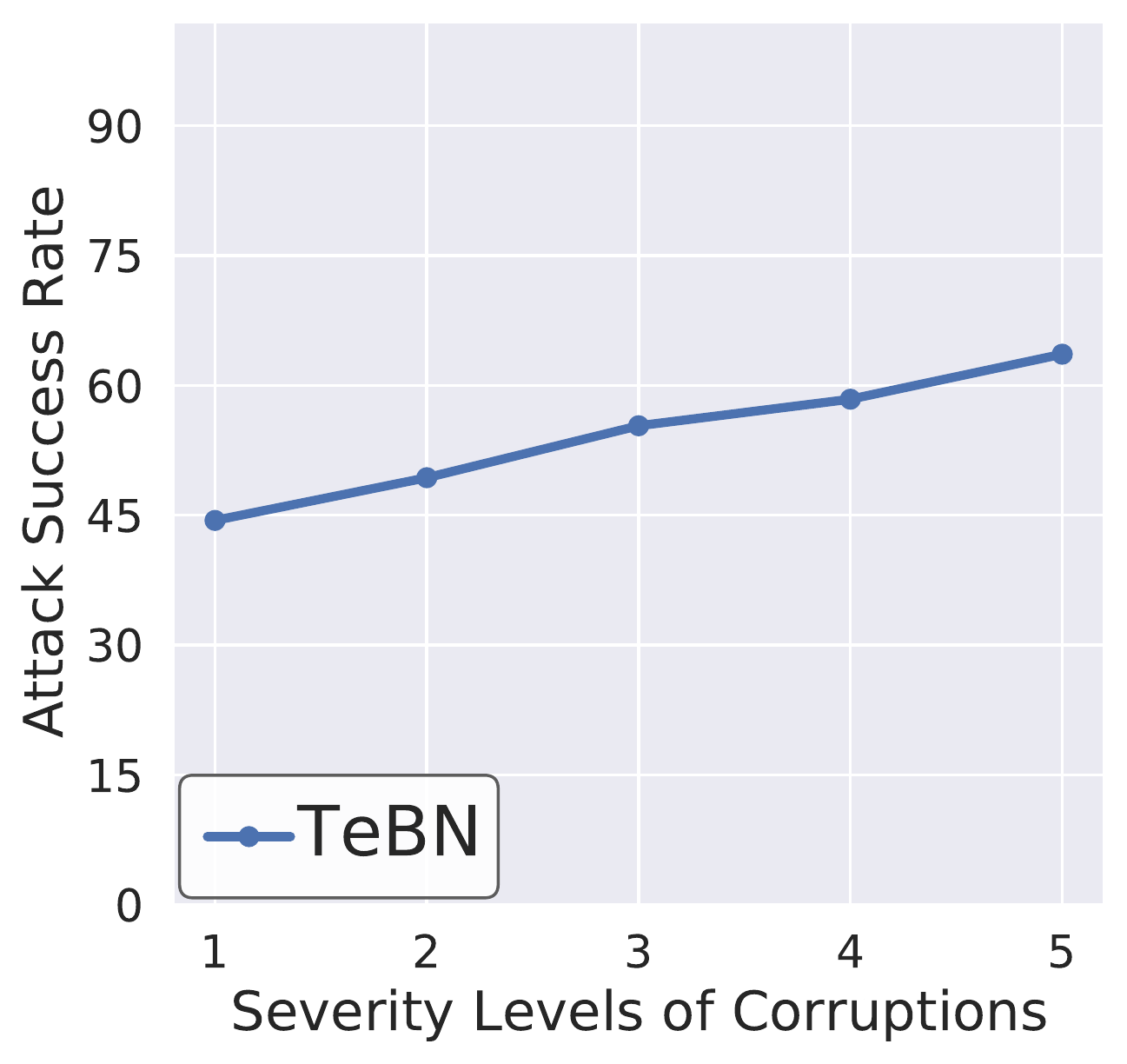} & 
   \includegraphics[width=0.23\textwidth]{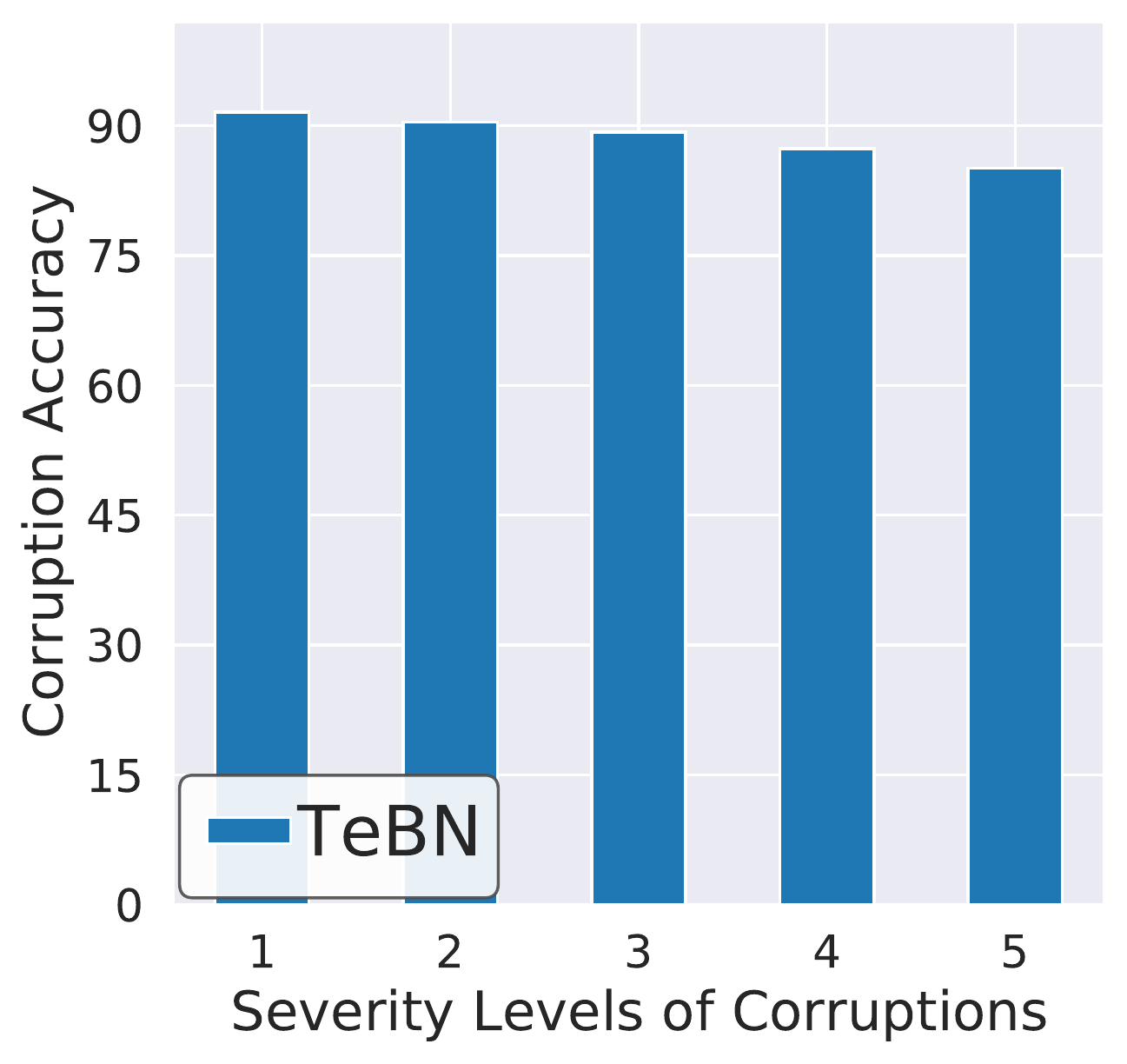}&
    \includegraphics[width=0.23\textwidth]{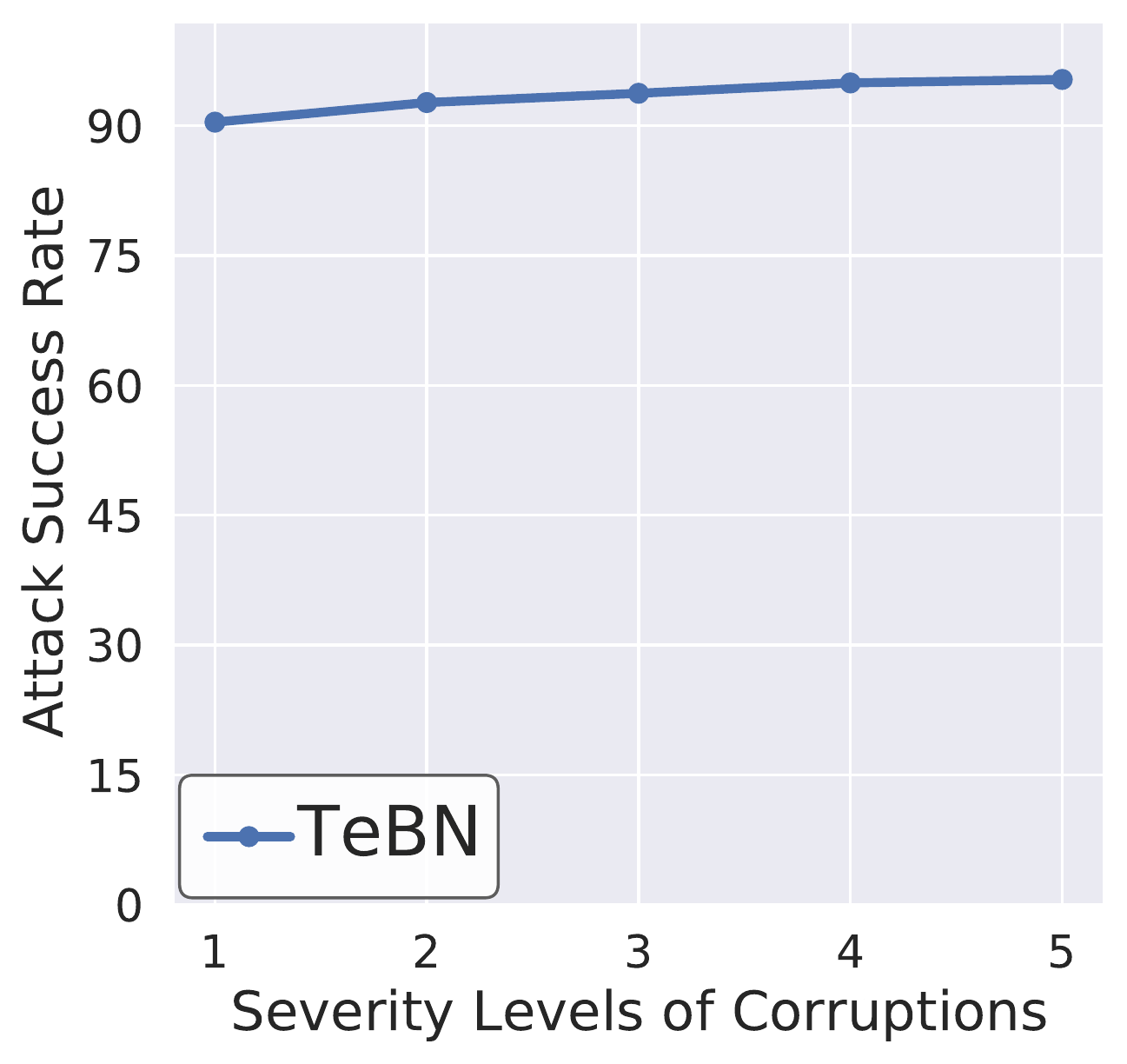} & 
   \includegraphics[width=0.23\textwidth]{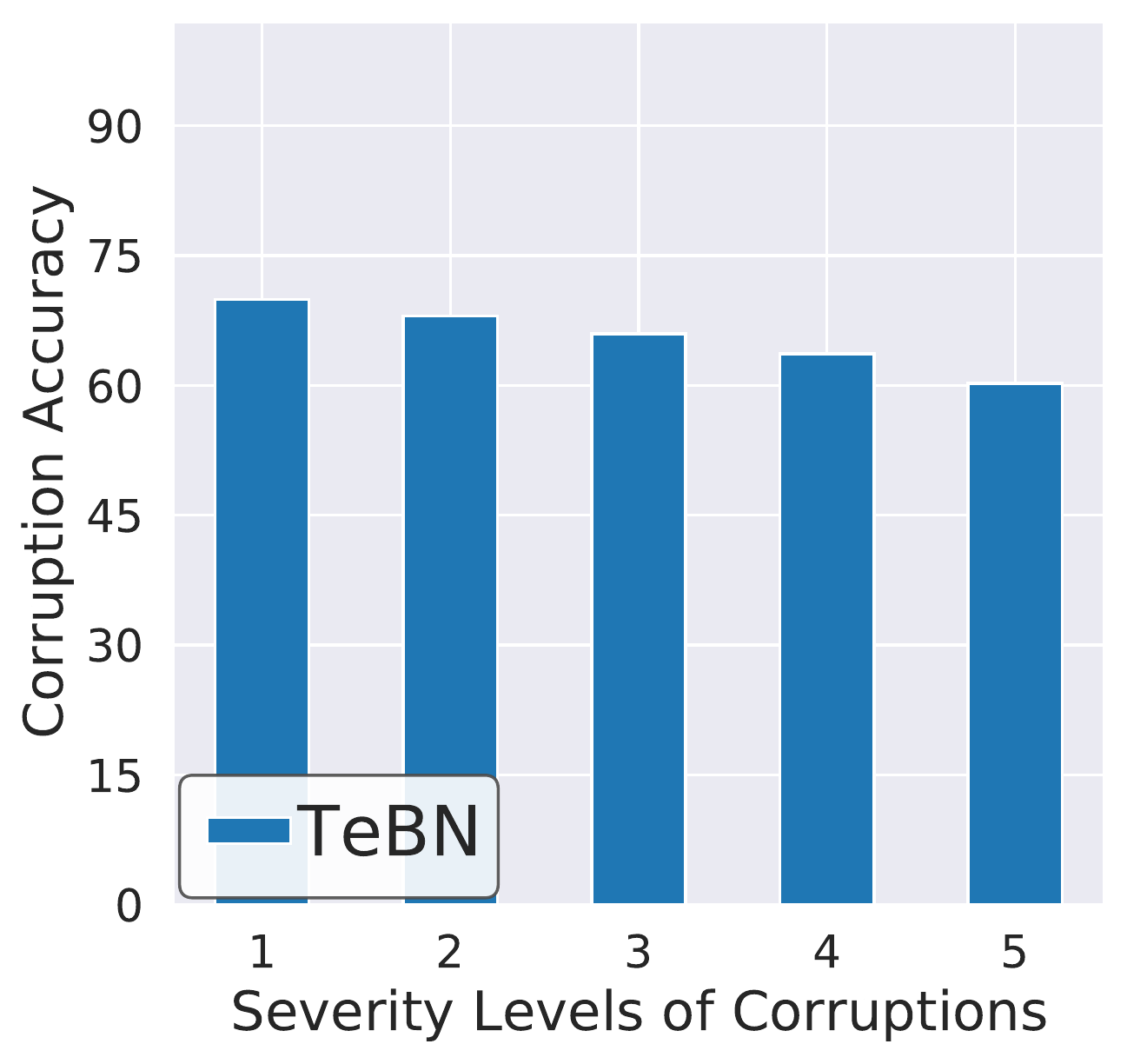} \\
     \multicolumn{2}{c}{(a) CIFAR-10-C  }  & \multicolumn{2}{c}{(b) CIFAR-100-C  }   \\
  \end{tabular}
      \caption{Illustration of how severity levels of corruption affect the corruption accuracy and attack success rate of our attack on the CIFAR-C dataset. (Line plot: attack success rate; bar plot: corruption accuracy) [$N_m$=20; Target Attack] }
      \label{Fig:CIFARC_augall}
  \end{figure}

  \smallskip\noindent\textbf{Larger corruption severity tends to be more vulnerable against  Distribution Invading Attack (Figure \ref{Fig:INC_augall} and Figure \ref{Fig:CIFARC_augall}).} Our previous evaluation only concentrates on the level 3 severity (medium) of corruption; now, we analyze the impact of corruption severity. First, the accuracy significantly drops when corruptions are more severe, e.g., the average degradation from level 1 to level 5 is nearly 30\% on ImageNet-C. Therefore, as expected, the ASR of the Distribution Invading Attack gets higher as the models (updated by TTA methods) tend not to be confident in their predictions. Similar behaviors are observed for the CIFAR-C dataset. 
  
\subsection{Effectiveness of DIA with Larger TTA Learning Rate}
\label{append:large_lr}

We conduct ablation studies (shown in Table \ref{tab:ttalr}) on the TTA learning rate $\eta$ (set as 0.001 for previous experiments). 
Concretely, we increase $\eta$ to 0.005 and evaluate DIA. Our single-level optimization method exhibits some performance degradations when the TTA learning rate rises but still reaches near-100\% ASR with 20 (10\%) malicious data. Furthermore, we also observe the corruption accuracy with $\eta = 0.005$ drop $\sim$4\% on average. 
\begin{table}[H]
\centering
\caption{Effectiveness of distribution invading attack with larger TTA learning rate ($\eta$) on ImageNet-C dataset.}
\resizebox{0.95\textwidth}{!}{%
\begin{tabular}{@{}ccccccccc@{}}
\toprule
 \textbf{$N_m$ }& \textbf{$\eta$} & \textbf{TeBN(\%)} & \textbf{TENT(\%)} & \textbf{Hard PL(\%)} & \textbf{Soft PL(\%)} & \textbf{Robust PL(\%)} & \textbf{Conjugate PL(\%)}  \\
 \midrule 
 \multirow{2}{*}{ \textbf{5 (2.5\%)}  } 
&  0.001      & 80.80 & 75.73 &   69.87 &   62.67 & 66.40 &      57.87 \\
&  0.005     & 80.80 &  52.00 &   46.93 &   35.73 &  45.33 &      32.00 \\
\midrule 
\multirow{2}{*}{  \textbf{10 (5\%)}  } 
&   0.001   & 99.47   & 98.67 &   96.53 &   94.13 & 96.00 &      92.80 \\
&   0.005    &  99.47 &  87.73 &   83.73 &   74.67 &  82.13 &      71.47 \\
 \midrule 
 \multirow{2}{*}{  \textbf{20 (5\%)}   } 
&    0.001   &  100.00 & 100.00 &  100.00 &  100.00 & 100.00 &     100.00 \\      
&    0.005   &  100.00 & 100.00 &   99.73 &   98.13 & 100.00 &      98.67 \\           
\bottomrule
\end{tabular}%
}
\label{tab:ttalr}
\end{table}



\subsection{Effectiveness of DIA with More TTA Optimization Steps}
\label{append:opt_step}
In this subsection, we conduct ablation studies (shown in Table \ref{tab:ttastep}) where the TTA optimization step is no longer 1 step but 5 steps.
We observe a degradation (0\% to 25\%) in DIA performance if the TTA optimizes the unsupervised loss with 5 steps. However, with 20 malicious data, the ASR is still near-100\%.

\begin{table}[H]
\centering
\caption{Effectiveness of distribution invading attack with more TTA optimization steps on ImageNet-C dataset.}
\resizebox{0.95\textwidth}{!}{%
\begin{tabular}{@{}ccccccccc@{}}
\toprule
 \textbf{$N_m$ }& \textbf{Steps} & \textbf{TeBN(\%)} & \textbf{TENT(\%)} & \textbf{Hard PL(\%)} & \textbf{Soft PL(\%)} & \textbf{Robust PL(\%)} & \textbf{Conjugate PL(\%)}  \\
 \midrule 
 \multirow{2}{*}{ \textbf{5 (2.5\%)}  } 
&  1      & 80.80 & 75.73 &   69.87 &   62.67 & 66.40 &      57.87 \\
&  5     &  80.80 &  58.13 &   53.07 &   39.73 &  49.87 &      36.00  \\
\midrule 
\multirow{2}{*}{  \textbf{10 (5\%)}  } 
&   1   & 99.47   & 98.67 &   96.53 &   94.13 & 96.00 &      92.80 \\
&   5    &  99.47 &  93.60 &   88.53 &   79.73 &  87.20 &      77.87 \\
 \midrule 
 \multirow{2}{*}{  \textbf{20 (5\%)}   } 
&    1   &  100.00 & 100.00 &  100.00 &  100.00 & 100.00 &     100.00 \\      
&    5   & 100.00 & 100.00 &  100.00 &   99.73 & 100.00 &      99.47 \\           
\bottomrule
\end{tabular}%
}
\label{tab:ttastep}
\end{table}



\subsection{Effectiveness of DIA with Additional Baseline Methods}
\label{append:addtta}

In this subsection, we present our DIA can also attack advanced methods of \textbf{TENT}, like \textbf{ETA} \cite{niu2022efficient}. In general, \textbf{ETA} introduces a technique to boost the efficiency of \textbf{TENT} by actively selecting reliable samples for updating the base model. However, they still suffer from the vulnerabilities of re-estimating BN statistics. Therefore, we can directly apply DIA, and our results demonstrate that \textbf{ETA is even more vulnerable than TENT  for the CIFAR-C dataset (Table \ref{tab:addtable}).} Furthermore, 20 malicious data can still achieve near-100\% ASR on the ImageNet-C benchmark. 

\begin{table}[H]
\centering
\caption{Attack success rate of Distribution Invading Attack (targeted attacks) across benchmarks and TTA methods.  
}
\resizebox{0.6\textwidth}{!}{%
\begin{tabular}{clccc}
\toprule
\textbf{Dataset} & \textbf{ $N_m$} & \textbf{TeBN(\%)} & \textbf{TENT(\%)} & \textbf{ETA(\%)} \\ \midrule
\multirow{3}{*}{\textbf{\begin{tabular}[c]{@{}c@{}}CIFAR-10-C\\ (ResNet26)\end{tabular}}}  
& 10 (5\%) & 25.87 & 23.20   & 24.53      \\
& 20 (10\%) & 55.47 & 45.73  & 53.07     \\
& 40 (20\%)& 92.80 & 83.87   & 89.87     \\ \hline
\multirow{3}{*}{\textbf{\begin{tabular}[c]{@{}c@{}}CIFAR-100-C\\ (ResNet26)\end{tabular}}} 
& 10 (5\%) & 46.80 & 26.40       &  33.60       \\             
& 20 (10\%) & 93.73 & 72.80     &  86.67       \\             
& 40 (20\%) & 100.00 & 100.00  & 100.00     \\ \hline
\multirow{3}{*}{\textbf{\begin{tabular}[c]{@{}c@{}}ImageNet-C\\ (ResNet50)\end{tabular}}}  
& 5 (2.5\%) &  80.80 & 75.73    & 41.33    \\ 
& 10 (5\%) & 99.47   & 98.67    & 81.60    \\
& 20 (10\%) & 100.00 & 100.00   & 99.47   \\ \bottomrule                                              
\end{tabular}%
}
\label{tab:addtable}
\end{table}

\newpage
\subsection{Effectiveness of DIA across Various Corruption Types}
\label{append:cor}

We then report the result of DIA for all 15 corruption types on Table \ref{tab:all_corr}. 
We select $N_m = 20$ for CIFAR-10-C and CIFAR-100-C and $N_m = 5$ for ImageNet-C.
We observe that the predictions under Brightness corruption are the hardest to attack. This is the expected result because the benign accuracy of Brightness is the highest \cite{Schneider2020ImprovingRA}. 

\begin{table}[H]
\centering
\caption{Effectiveness of Distribution Invading Attack across various corruption types. (CIFAR-C: $N_m$=20; ImageNet-C: $N_m$=5)}
\label{tab:all_corr}
\resizebox{0.98\textwidth}{!}{%
\begin{tabular}{cccccccc}
\toprule
\textbf{Dataset} & \textbf{ Corruption Type } & \textbf{TeBN(\%)} & \textbf{TENT(\%)} & \textbf{Hard PL(\%)} & \textbf{Soft PL(\%)} & \textbf{Robust PL(\%)} & \textbf{Conjugate PL(\%)}  \\ \midrule
\multirow{15}{*}{\textbf{\begin{tabular}[c]{@{}c@{}}CIFAR-10-C\\ (ResNet-26)\\($N_m$=20 (10\%))\end{tabular}}}  
& Gaussian Noise     &  68.0 &  48.0 &    52.0 &    50.0 & 56.0 &       50.0 \\
& Shot Noise         &  70.0 &  54.0 &    56.0 &    62.0 & 62.0 &       54.0 \\
& Impulse Noise      &  76.0 &  48.0 &    58.0 &    54.0 & 62.0 &       50.0 \\
& Defocus Blur       &  44.0 &  38.0 &    38.0 &    40.0 & 40.0 &       40.0 \\
& Glass Blur         &  50.0 &  44.0 &    52.0 &    44.0 & 48.0 &       42.0 \\
& Motion Blur        &  54.0 &  48.0 &    46.0 &    50.0 & 52.0 &       46.0 \\
& Zoom Blur          &  44.0 &  46.0 &    44.0 &    46.0 & 44.0 &       44.0 \\
& Snow               &  58.0 &  44.0 &    42.0 &    44.0 & 50.0 &       44.0 \\
& Frost              &  48.0 &  46.0 &    48.0 &    50.0 & 48.0 &       48.0 \\
& Fog                &  60.0 &  40.0 &    46.0 &    44.0 & 48.0 &       40.0 \\
& Brightness         &  38.0 &  38.0 &    40.0 &    38.0 & 38.0 &       40.0 \\
& Contrast           &  62.0 &  48.0 &    48.0 &    48.0 & 50.0 &       50.0 \\
& Elastic Transform  &  56.0 &  56.0 &    58.0 &    50.0 & 54.0 &       52.0 \\
& Pixelate           &  44.0 &  46.0 &    46.0 &    44.0 & 40.0 &       44.0 \\
& JPEG Compression   &  60.0 &  42.0 &    48.0 &    48.0 & 50.0 &       42.0 \\
\midrule
& All  &  55.47 & 45.73 &   48.13 &   47.47 & 49.47 &      45.73 \\
\midrule
\midrule
\multirow{15}{*}{\textbf{\begin{tabular}[c]{@{}c@{}}CIFAR-100-C\\ (ResNet-26)\\($N_m$=20 (10\%))\end{tabular}}}      
& Gaussian Noise     & 100.0 &  80.0 &    92.0 &    86.0 & 94.0 &       74.0 \\
& Shot Noise         & 100.0 &  80.0 &    98.0 &    82.0 & 86.0 &       82.0 \\
& Impulse Noise      &  96.0 &  82.0 &    92.0 &    90.0 & 90.0 &       74.0 \\
& Defocus Blur       &  86.0 &  64.0 &    82.0 &    72.0 & 82.0 &       68.0 \\
& Glass Blur         &  96.0 &  78.0 &    80.0 &    78.0 & 78.0 &       66.0 \\
& Motion Blur        &  94.0 &  74.0 &    82.0 &    80.0 & 82.0 &       74.0 \\
& Zoom Blur          &  90.0 &  60.0 &    80.0 &    62.0 & 70.0 &       60.0 \\
& Snow               &  92.0 &  72.0 &    80.0 &    74.0 & 72.0 &       66.0 \\
& Frost              &  94.0 &  72.0 &    94.0 &    74.0 & 82.0 &       78.0 \\
& Fog                &  94.0 &  76.0 &    92.0 &    86.0 & 92.0 &       76.0 \\
& Brightness         &  86.0 &  64.0 &    84.0 &    72.0 & 72.0 &       68.0 \\
& Contrast           &  96.0 &  80.0 &    94.0 &    88.0 & 94.0 &       78.0 \\
& Elastic Transform  & 100.0 &  74.0 &    90.0 &    80.0 & 84.0 &       70.0 \\
& Pixelate           &  90.0 &  64.0 &    84.0 &    80.0 & 78.0 &       72.0 \\
& JPEG Compression   &  92.0 &  72.0 &    86.0 &    74.0 & 88.0 &       68.0 \\
\midrule
& All   & 93.73 & 72.80 &   87.33 &   78.53 & 82.93 &      71.60  \\
\midrule
\midrule
\multirow{15}{*}{\textbf{\begin{tabular}[c]{@{}c@{}}ImageNet-C\\ (ResNet-50)\\($N_m$=5 (2.5\%))\end{tabular}}}  
& Gaussian Noise      &  92.0 &  92.0 &    72.0 &    56.0 & 64.0 &       52.0 \\
& Shot Noise          &  88.0 &  76.0 &    68.0 &    56.0 & 60.0 &       48.0 \\
& Impulse Noise       &  92.0 &  84.0 &    80.0 &    68.0 & 80.0 &       60.0 \\
& Defocus Blur        & 100.0 &  96.0 &    88.0 &    84.0 & 84.0 &       80.0 \\
& Glass Blur          &  92.0 &  84.0 &    84.0 &    68.0 & 72.0 &       64.0 \\
& Motion Blur         &  88.0 &  88.0 &    84.0 &    68.0 & 76.0 &       68.0 \\
& Zoom Blur           &  84.0 &  76.0 &    76.0 &    72.0 & 76.0 &       68.0 \\
& Snow                &  72.0 &  64.0 &    56.0 &    56.0 & 56.0 &       56.0 \\
& Frost               &  88.0 &  72.0 &    72.0 &    64.0 & 68.0 &       52.0 \\
& Fog                 &  88.0 &  84.0 &    80.0 &    76.0 & 80.0 &       68.0 \\
& Brightness          &  48.0 &  48.0 &    44.0 &    40.0 & 40.0 &       40.0 \\
& Contrast            &  96.0 &  96.0 &    88.0 &    84.0 & 88.0 &       80.0 \\
& Elastic Transform   &  56.0 &  60.0 &    52.0 &    48.0 & 48.0 &       40.0 \\
& Pixelate            &  56.0 &  52.0 &    52.0 &    44.0 & 48.0 &       36.0 \\
& JPEG Compression    &  72.0 &  64.0 &    52.0 &    56.0 & 56.0 &       56.0 \\
\midrule
& All   & 80.80 & 75.73 &   69.87 &   62.67 & 66.40 &      57.87 \\
\bottomrule                   
\end{tabular}%
}
\end{table}

\newpage
\section{Additional Experiments on Robust Models and Mitigation Methods}
\label{append:df}

In this appendix, we show some additional findings and results of our mitigating methods as the complement of Section~\ref{sec:df}. 

The roadmap of this section is as follows.
We first investigate the role of robust models from two aspects.
In Appendix~\ref{append:adv_model}, we apply TTA to the adversarially trained models and surprisingly witness an improvement in clean accuracy.
We then evaluate the robust models' performance facing the corruption accuracy in Appendix~ \ref{sec:corr_robust} and analyze their effectiveness in resisting DIA attacks in Appendix~\ref{append:romodel_DIA}.
Furthermore, we extensively study the Robust Batch Normalization (BN) estimation. In Appendix~\ref{sec:understandBN}, we visualize the behaviors of Layer-wise BN. Then we provide the additional results of Robust BN Estimation on CIFAR-C in Appendix~\ref{append:RoBNCifarC}, ImageNet-C in Appendix~\ref{append:RoBNImnC}. Finally, we conduct a parameter search for Robust BN estimation in Appendix~\ref{append:param_search}.

\subsection{Additional Findings of Adversarial Models and TTA on Clean Data}
\label{append:adv_model}

\begin{table}[H]
\centering
\resizebox{0.98\textwidth}{!}{%
\begin{tabular}{cccccccccc}
\toprule
\textbf{Dataset} & \textbf{ Models}  &\textbf{Source(\%)} & \textbf{TeBN(\%)} & \textbf{TENT(\%)} & \textbf{Hard PL(\%)} & \textbf{Soft PL(\%)} & \textbf{Robust PL(\%)} & \textbf{Conjugate PL(\%)}  \\ \midrule
\multirow{4}{*}{\textbf{\begin{tabular}[c]{@{}c@{}}CIFAR-10 \end{tabular}}}  
& \cite{Dai2022ParameterizingAF}  WRN-28 &   87.02 &  87.00 &  \textbf{91.74} &   90.83 &   91.62 &  90.93 &      91.68 \\
& \cite{Wu2020AdversarialWP}  WRN-28     &   88.25 &  86.35 &  92.81 &   91.35 &   \textbf{93.07} &  93.02 &      91.59 \\
& \cite{Gowal2021ImprovingRU} WRN-28    &   87.49 &  87.42 &  88.39 &   88.22 &   \textbf{88.44} &  88.21 &      \textbf{88.44} \\
& \cite{Sehwag2022RobustLM}  RN-18      &   84.59 &  84.38 &  87.24 &   87.25 &   \textbf{87.63} &  87.05 &      86.73 \\
 \hline
\hline
\multirow{4}{*}{\textbf{\begin{tabular}[c]{@{}c@{}}CIFAR-100 \end{tabular}}} 
& \cite{Sehwag2022RobustLM}  WRN-34 &    65.93 &  67.37 &  73.96 &   70.63 &   \textbf{74.06} &  73.93 &      72.81 \\
& \cite{Wu2020AdversarialWP}  WRN-34     &    60.38 &  55.37 &  62.01 &   59.43 &   \textbf{62.19} &  61.89 &      61.87 \\
& \cite{Chen2022EfficientRT}  WRN-34    &    64.07 &  61.94 &  67.58 &   61.29 &   68.11 &  \textbf{69.09} &      65.02 \\
& \cite{Jia2022LASATAT}  WRN-34-20       &     67.31 &  64.17 &  69.72 &   68.60 &   \textbf{70.21} &  69.97 &      69.71 \\
\bottomrule
\end{tabular}%
}
\caption{ Clean accuracy can be improved by the test-time adaptation for adversarially trained models on CIFAR-10 and CIFAR-100. As a reference, the accuracy of the standard Wide ResNet with 28 layers (WRN-28) is 94.78\% for CIFAR-10 and 78.78\% for CIFAR-100. (RN-18 is ResNet-18, WRN-34 is Wide ResNet 34 )}
\label{tab:clean_cifar}
\end{table}

One of the notable weaknesses of adversarial training is the apparent performance degradation on clean test data \cite{Madry2018TowardsDL, Xie2019FeatureDF}, owing to the divergent distributions of adversarial and clean inputs for batch normalization (BN). Therefore, \cite{Xie2020AdversarialEI} proposed to use distinct BN layers to handle data from two distributions and \cite{Wang2022RemovingBN} leveraged normalizer-free networks (removing BN layers) to improve clean accuracy.               
As we discussed, TTA methods exhibit promising improvement in mitigating distribution shifts. Therefore, we consider if they can improve the accuracy of clean input when training data draws from the distribution of adversarial data. 

\smallskip\noindent\textbf{TTA methods achieve a non-trivial clean accuracy improvement for adversarially trained models on CIFAR-10 and CIFAR-100 (Table~\ref{tab:clean_cifar}).} 
We select a list of adversarial training methods~\cite{Dai2022ParameterizingAF, Wu2020AdversarialWP, Gowal2021ImprovingRU, Sehwag2022RobustLM, Chen2022EfficientRT, Jia2022LASATAT} as the source model and use TTA methods on clean inputs. 
All of them are downloaded from RobustBench~\cite{croce2020robustbench}.
As shown, \textbf{Soft PL} significantly improves the average accuracy $\sim$3.35\% for CIFAR-10 and $\sim$4.22\% for CIFAR-100.  
The most significant improvement even boosts 8.13\% compared to no TTA method. 
This result is interesting as another perspective to understand TTA better, and we hope future works can explore more on it. 
We also observe that simply replacing the BN statistics does not work as well as using adversarial models on ImageNet.

\subsection{Corruption Accuracy of Robust Models}
\label{sec:corr_robust}

We evaluate the corruption accuracy, where Wide ResNet with 28 layers trained by \citet{Gowal2021ImprovingRU} and \citet{Wu2020AdversarialWP} are chosen for CIFAR-10-C, and Wide ResNet with 28 layers developed by \citet{Pang2022RobustnessAA} and \citet{Rebuffi2021FixingDA} are selected for CIFAR-100-C. Note that all adversarial models are downloaded from RobustBench \cite{croce2020robustbench}. 

\begin{figure}[H]
\centering
\begin{tabular}{ccc}
\includegraphics[width=0.24\textwidth]{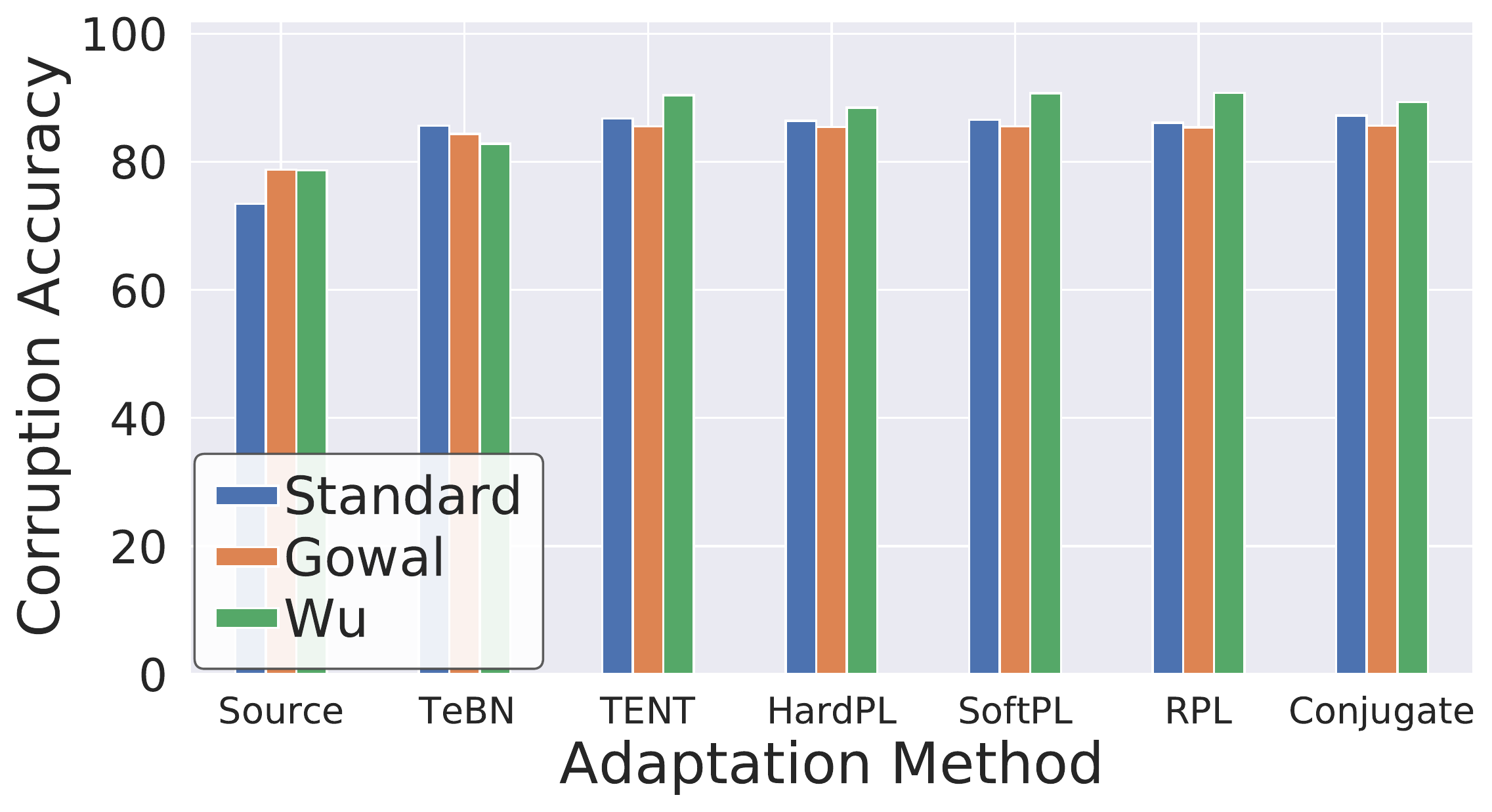} &
\includegraphics[width=0.24\textwidth]{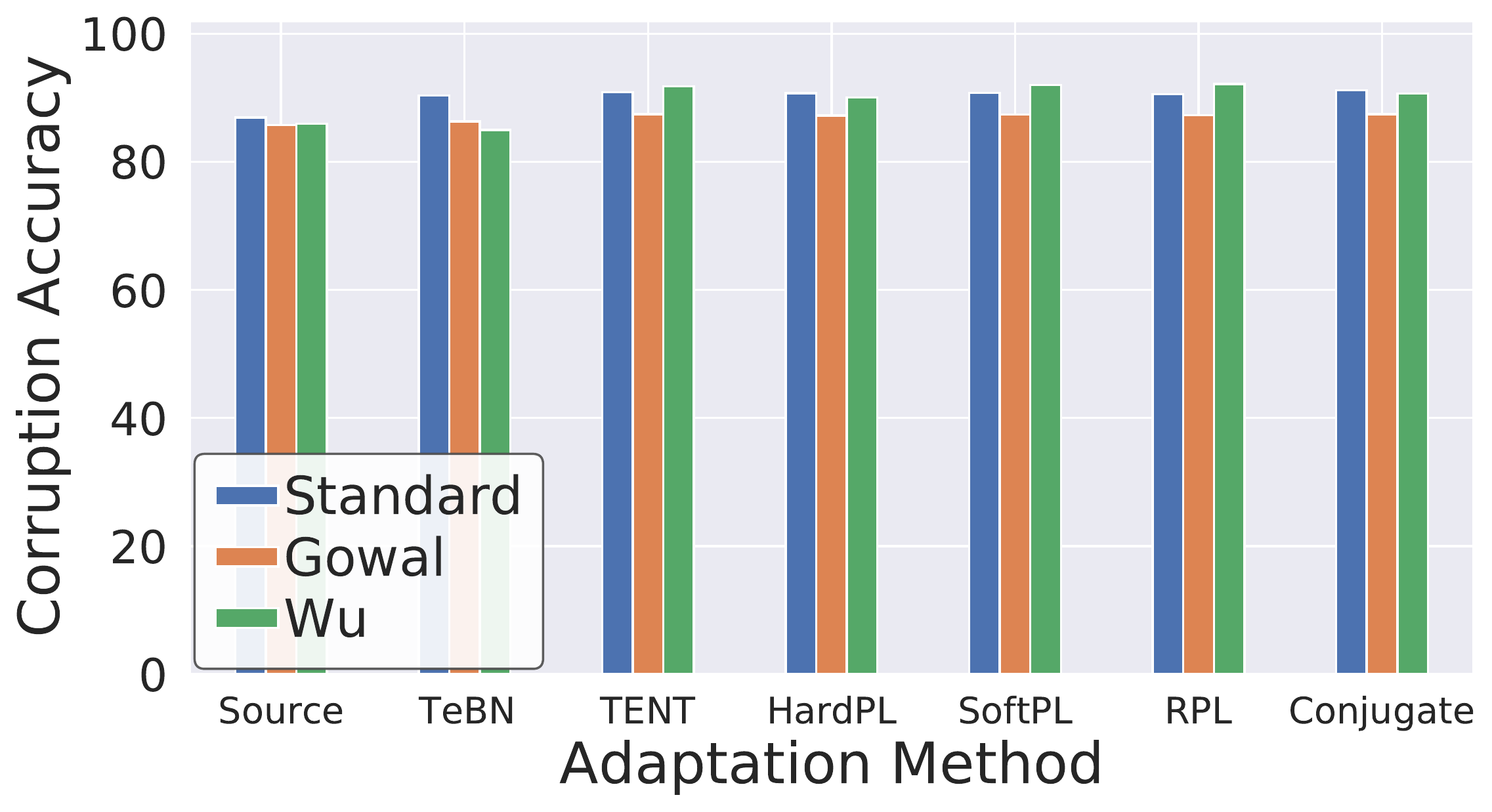} &
 \includegraphics[width=0.24\textwidth]{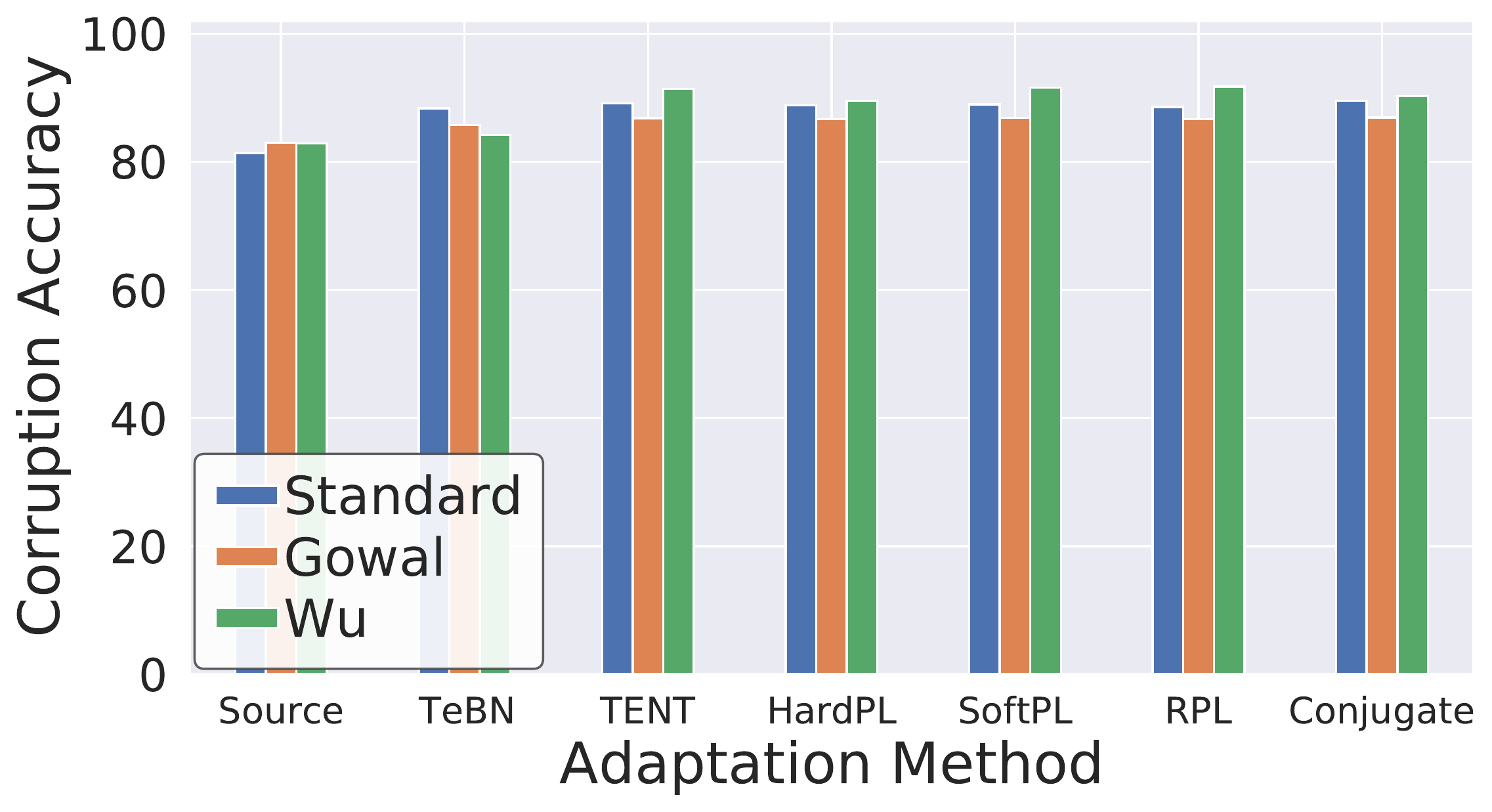} \\
(a) Average severity& (b) Severity 1 & (c) Severity 2 \\
     &   &   \\
\includegraphics[width=0.24\textwidth]{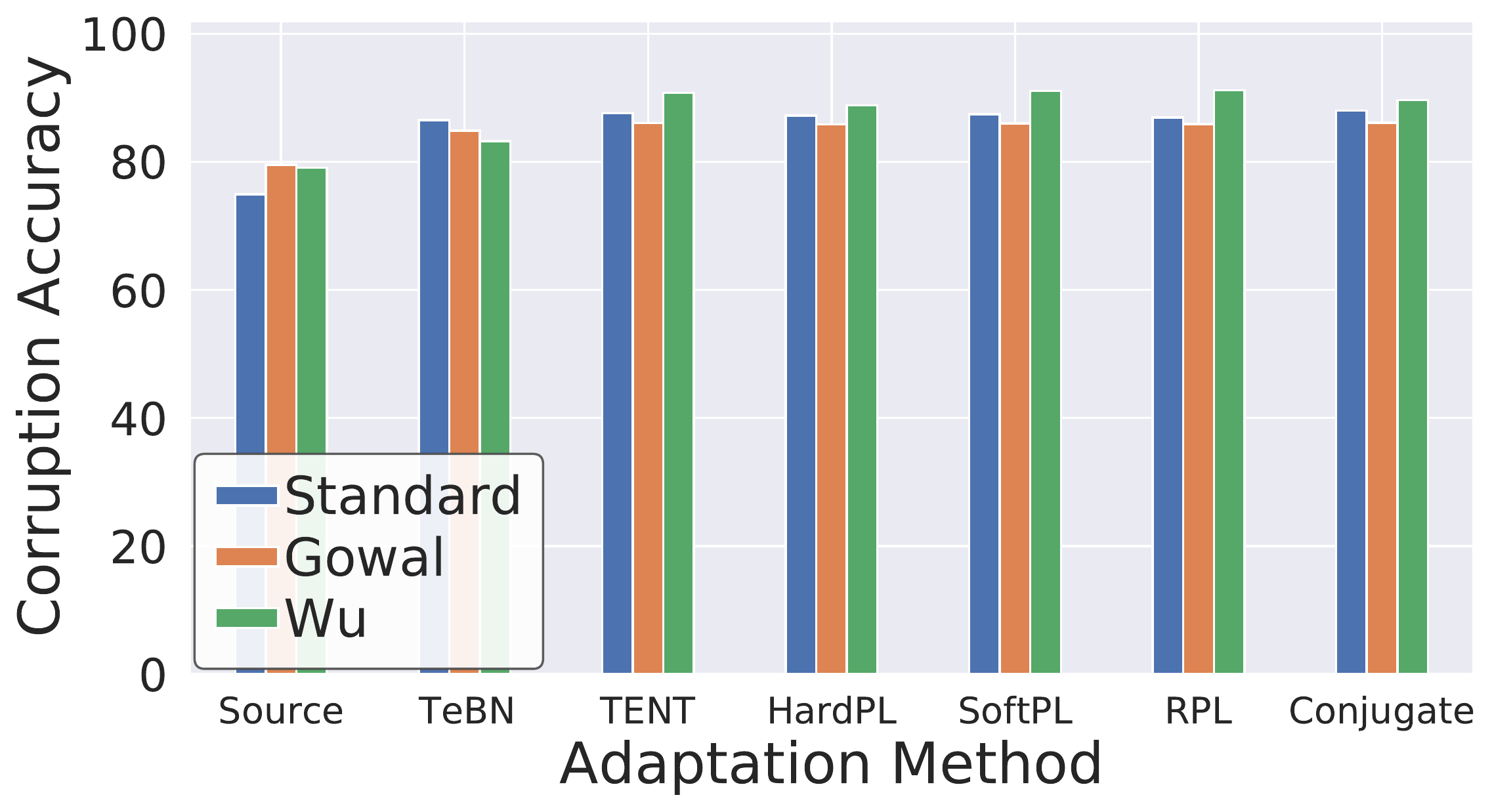}&
\includegraphics[width=0.24\textwidth]{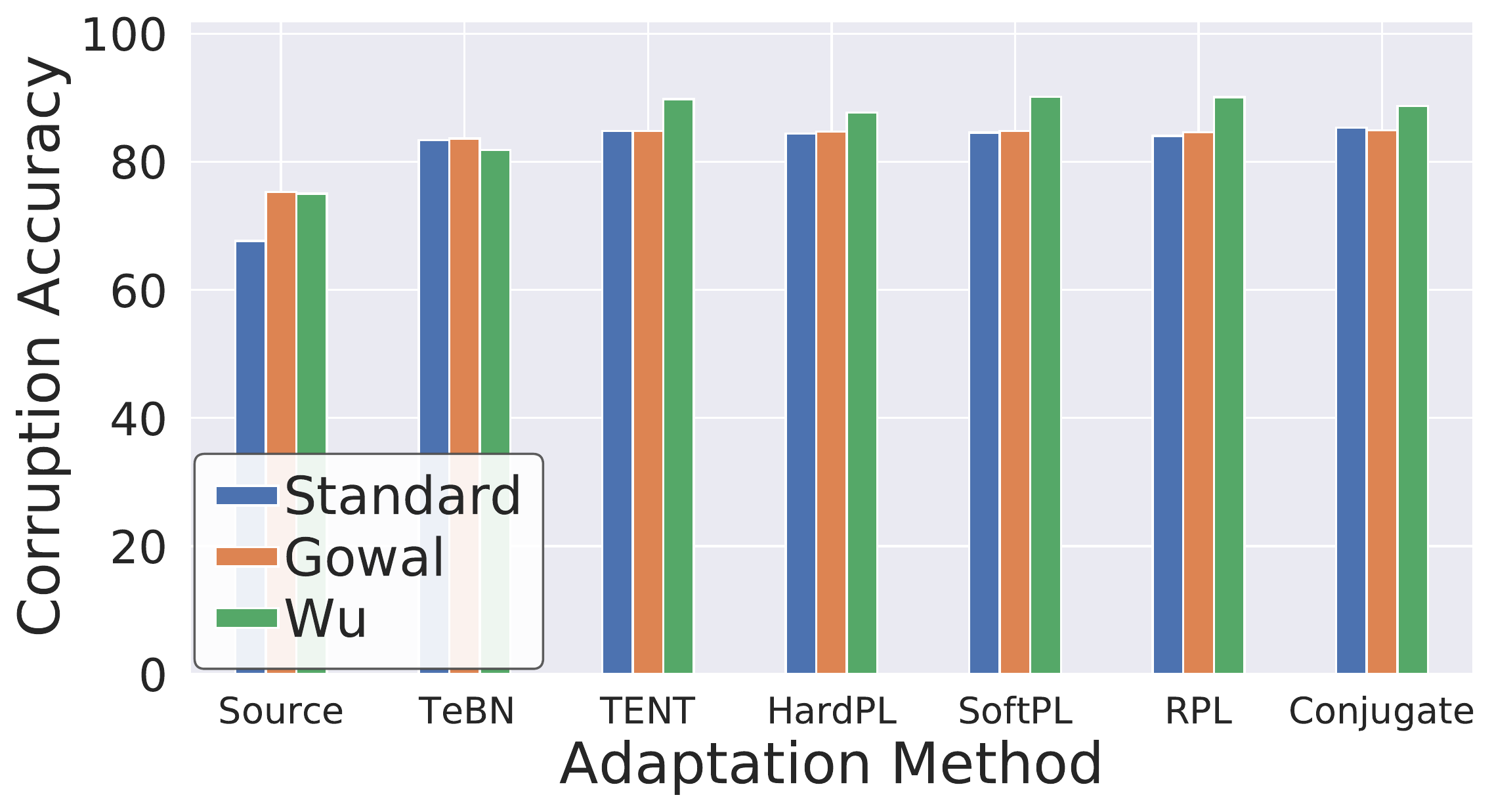} &
\includegraphics[width=0.24\textwidth]{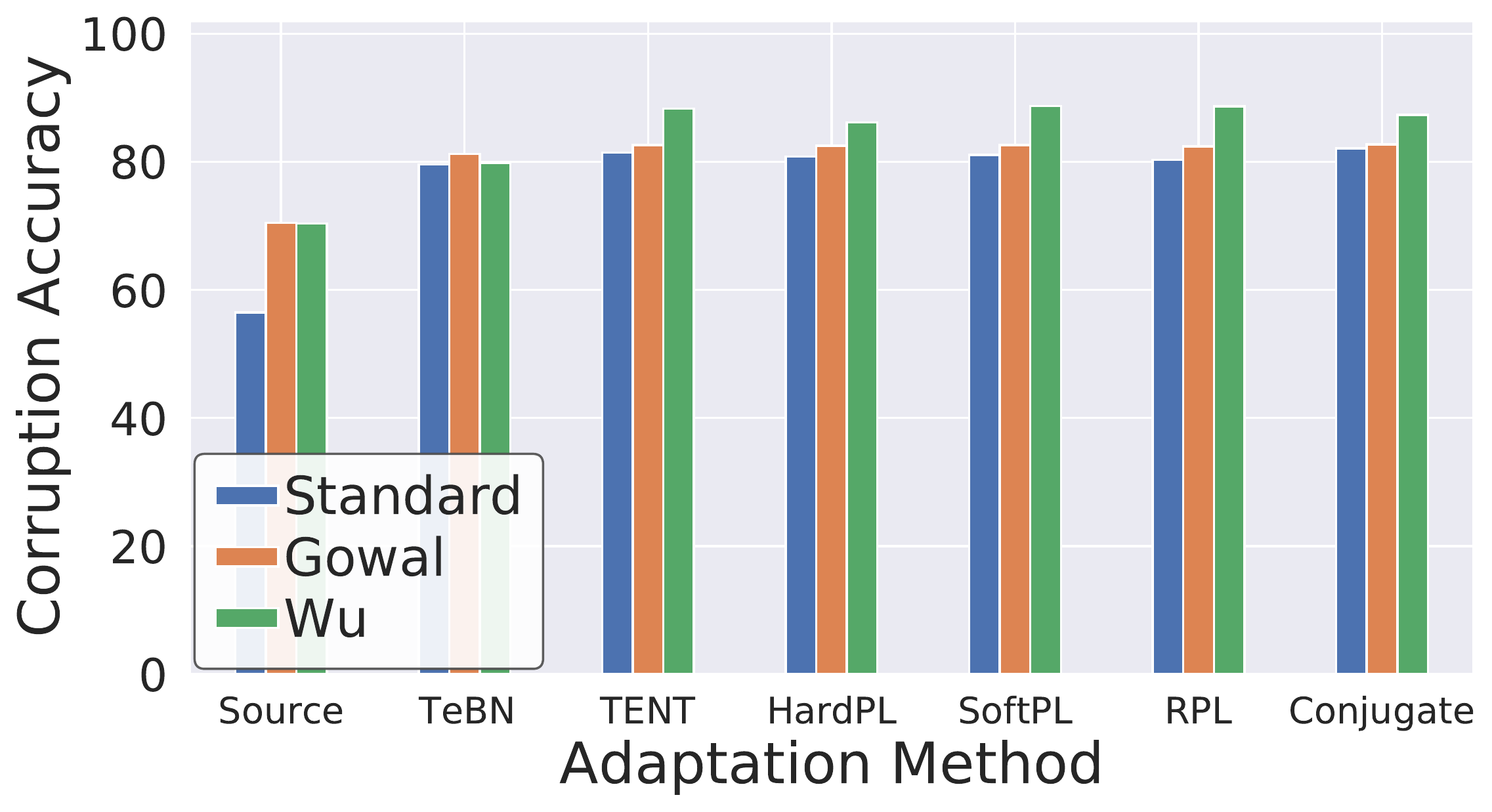}\\
 (d) Severity 3 & (e) Severity 4 &  (f) Severity 5  \\
\end{tabular}
	\caption{Corruption accuracy of the standard and robust models (\citet{Gowal2021ImprovingRU}  and  \citet{Wu2020AdversarialWP}) on CIFAR-10-C under different severity levels.}
	\label{Fig:C10C_robust_acc}
\end{figure}
\begin{figure}[H]
\centering
\begin{tabular}{ccc}
\includegraphics[width=0.24\textwidth]{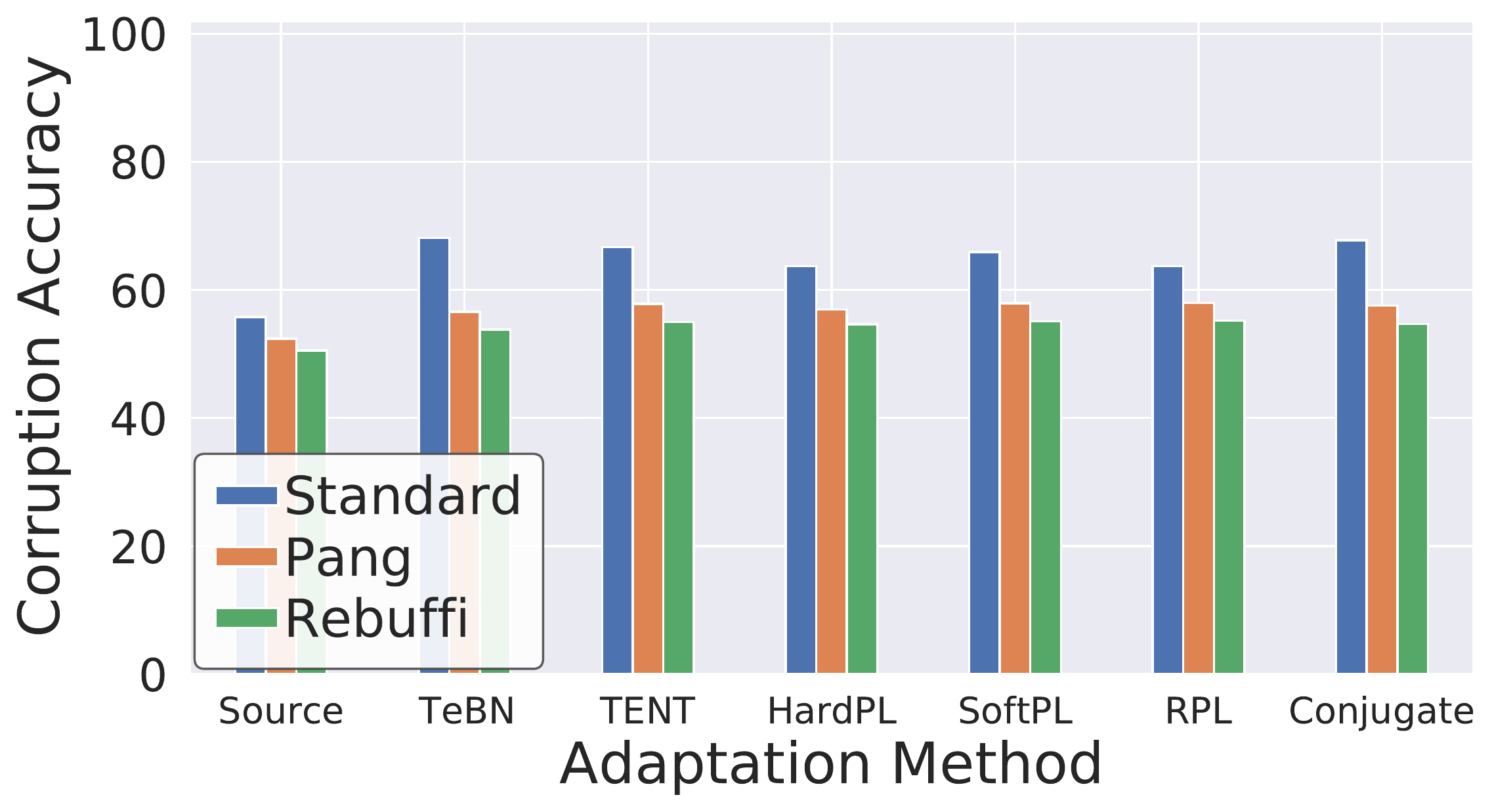} &
\includegraphics[width=0.24\textwidth]{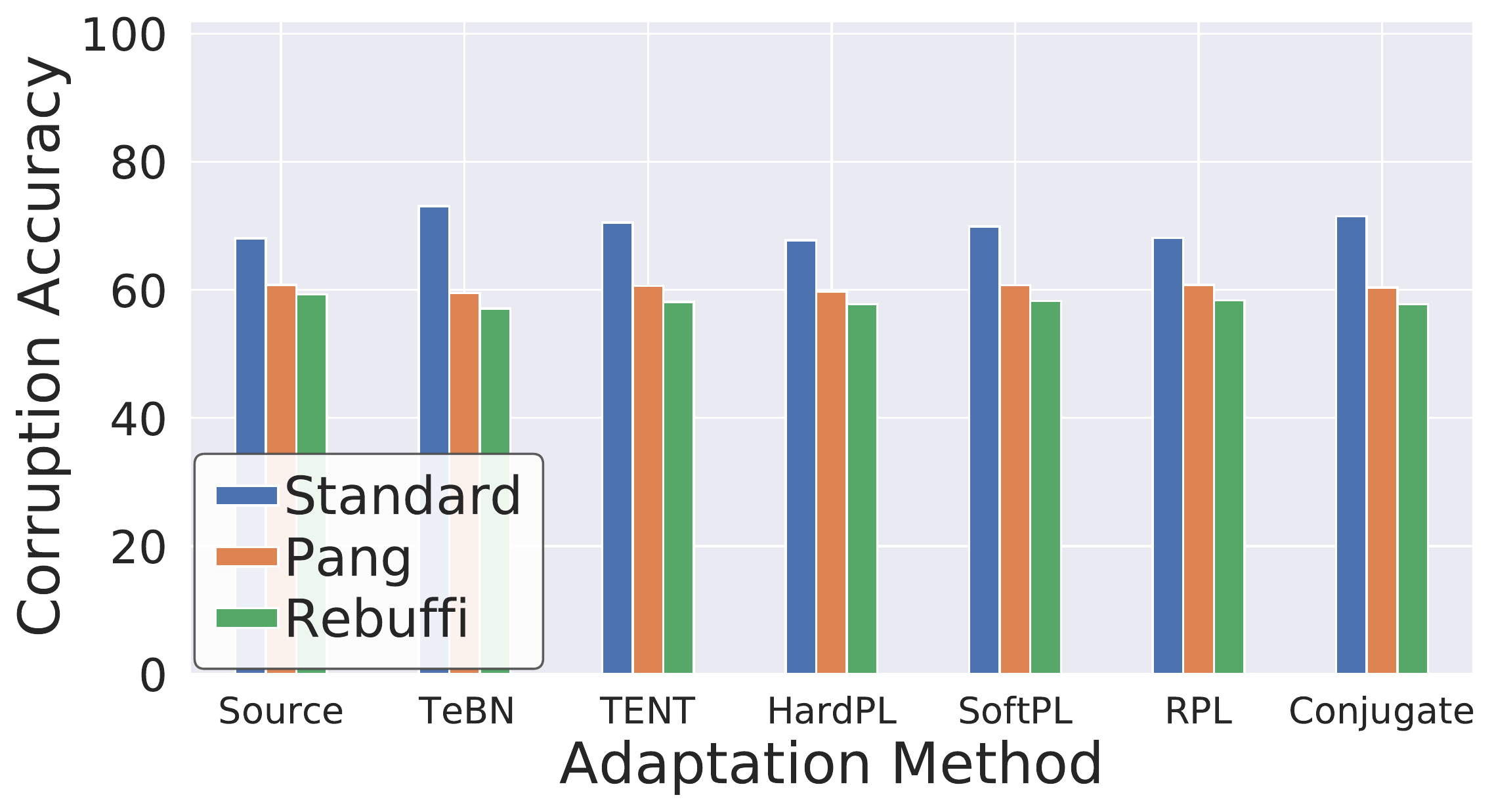} &
 \includegraphics[width=0.24\textwidth]{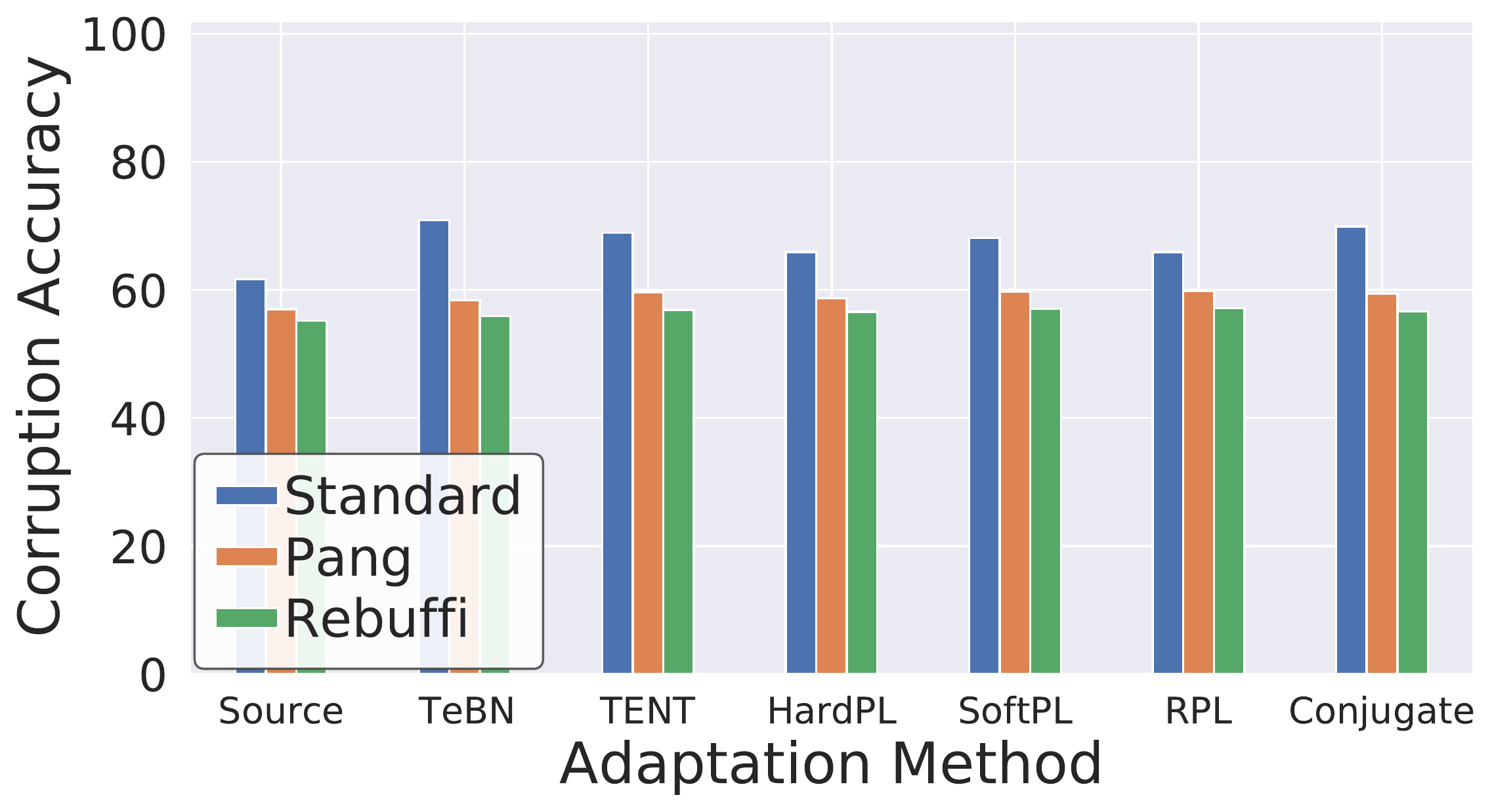} \\
(a) Average severity& (b) Severity 1 & (c) Severity 2 \\
     &   &   \\
\includegraphics[width=0.24\textwidth]{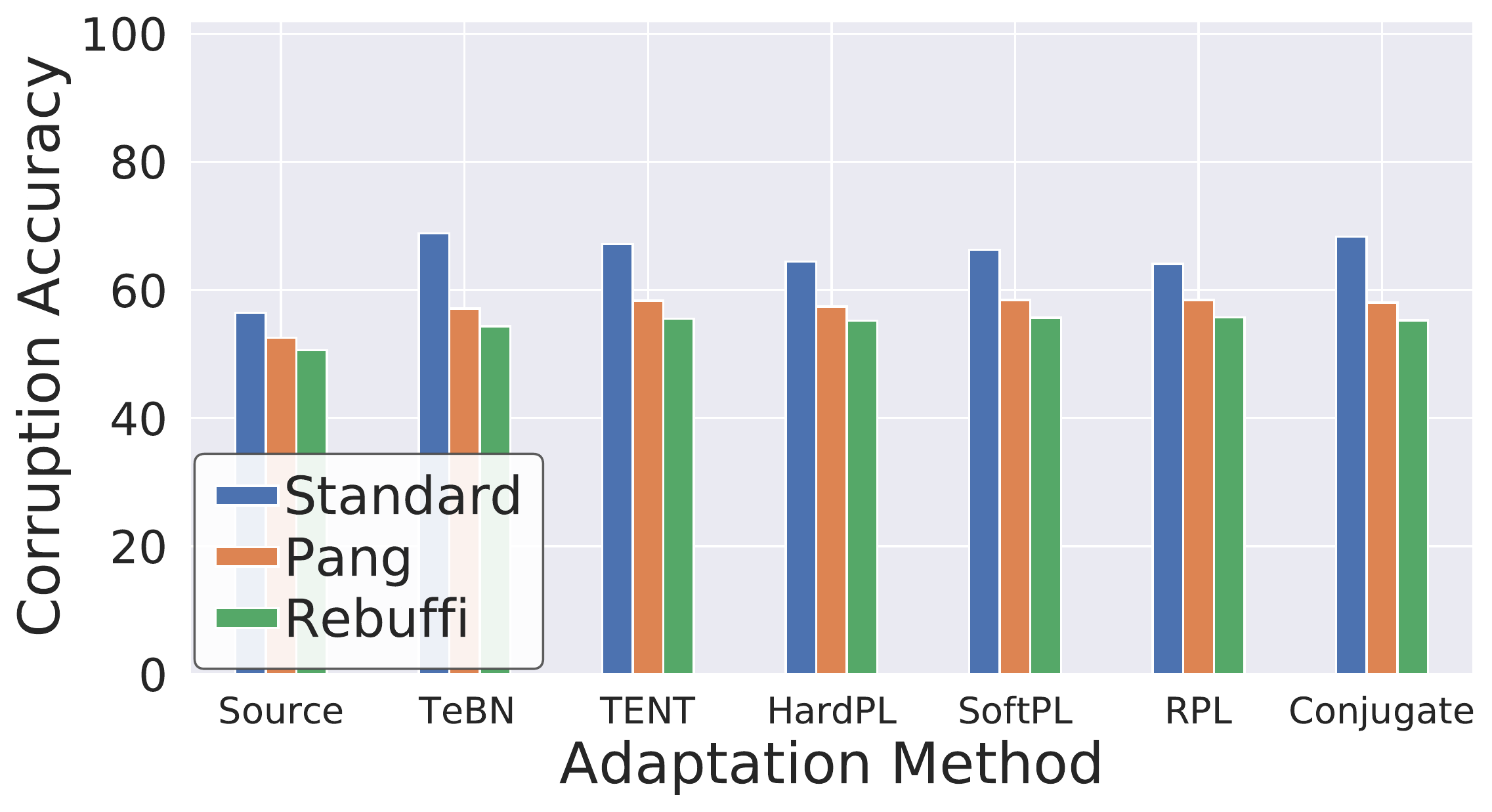}&
\includegraphics[width=0.24\textwidth]{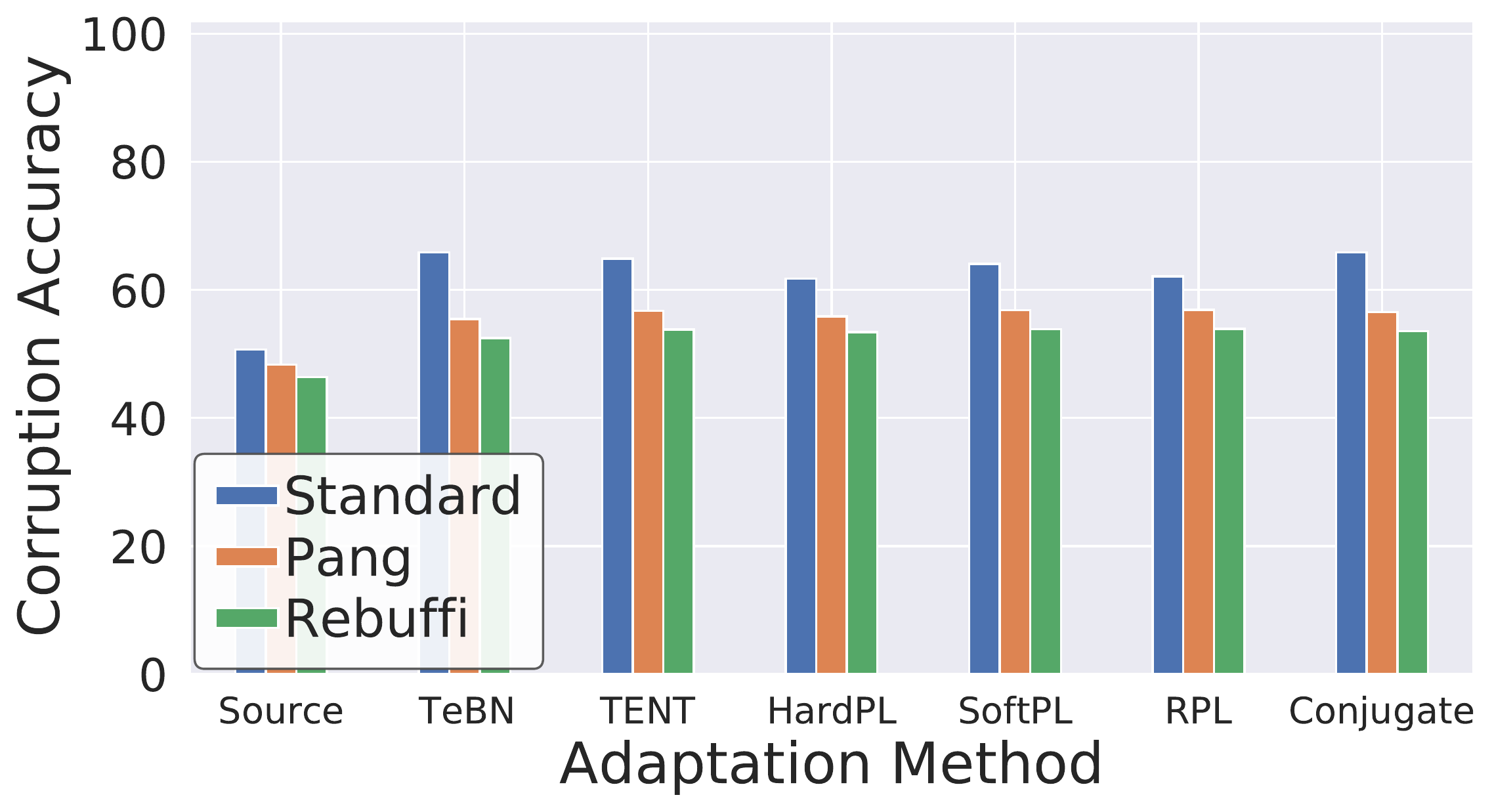} &
\includegraphics[width=0.24\textwidth]{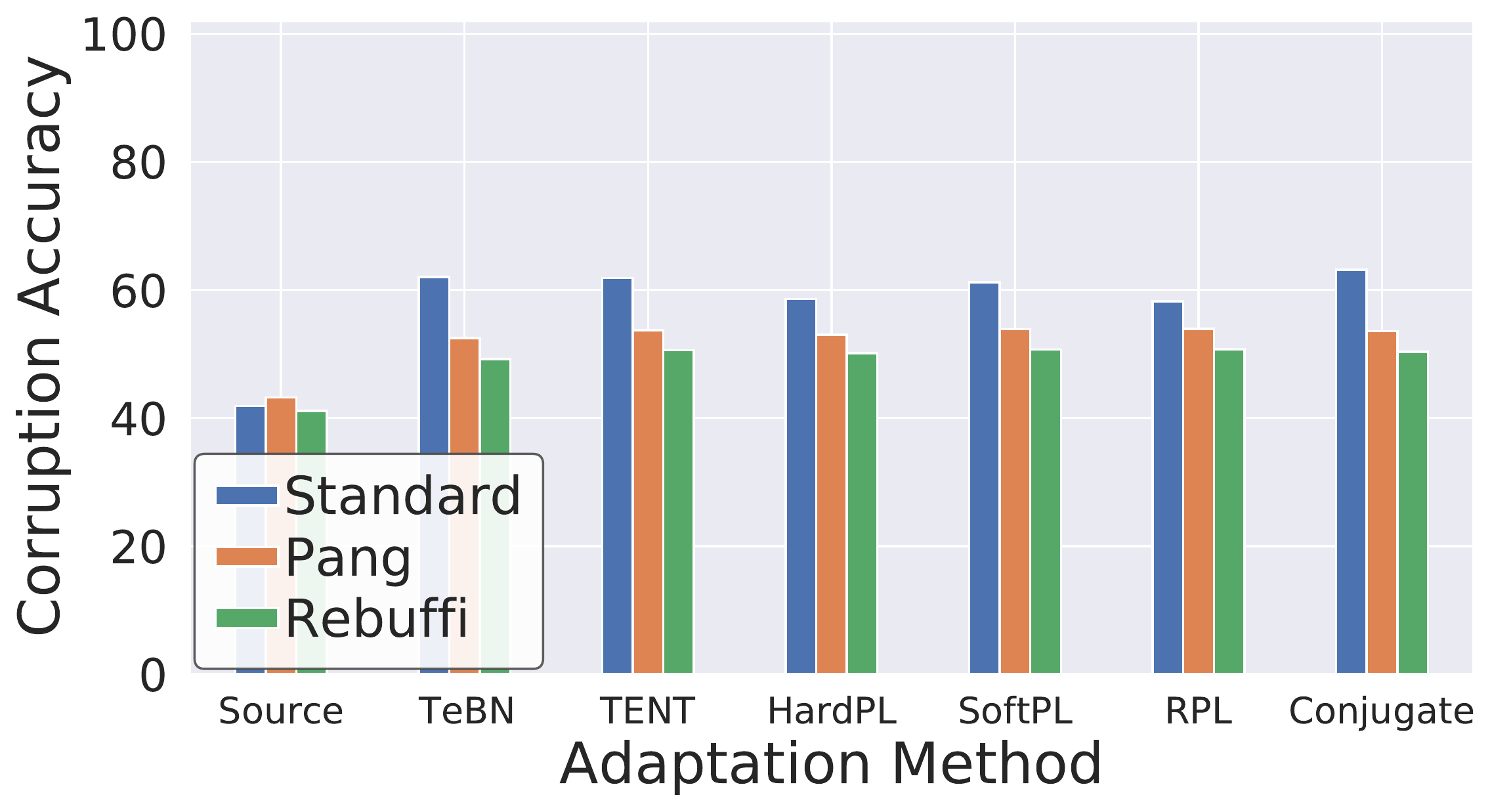}\\
 (d) Severity 3 & (e) Severity 4 &  (f) Severity 5  \\
\end{tabular}
	\caption{ Corruption accuracy of the standard and robust models (\citet{Pang2022RobustnessAA} and \citet{Rebuffi2021FixingDA}) on CIFAR-100-C  under different severity levels.}
	\label{Fig:C100C_robust_acc}
\end{figure}

\smallskip\noindent\textbf{Robust models with TTA achieve better performance than the standard model on CIFAR-10-C, but worse performance on CIFAR-100-C (Figure \ref{Fig:C10C_robust_acc} and Figure \ref{Fig:C100C_robust_acc}).} 
For CIFAR-10-C, we observe a performance boost from using robust models. For example, \citet{Wu2020AdversarialWP} achieves a $\sim$5\% improvement than the standard one. 
For CIFAR-100-C, robust models cannot perform similarly to the standard model. 
However, leveraging adversarial models and TTA can still exceed the standard model without TTA. 
Since they may use different training configurations (e.g., training batch size, data augmentations, inner optimization steps), we leave the question of ``training the best adversarial source model for TTA'' as future works.

\begin{figure}[H]
  \centering
  \begin{tabular}{ccc}
  \includegraphics[width=0.24\textwidth]{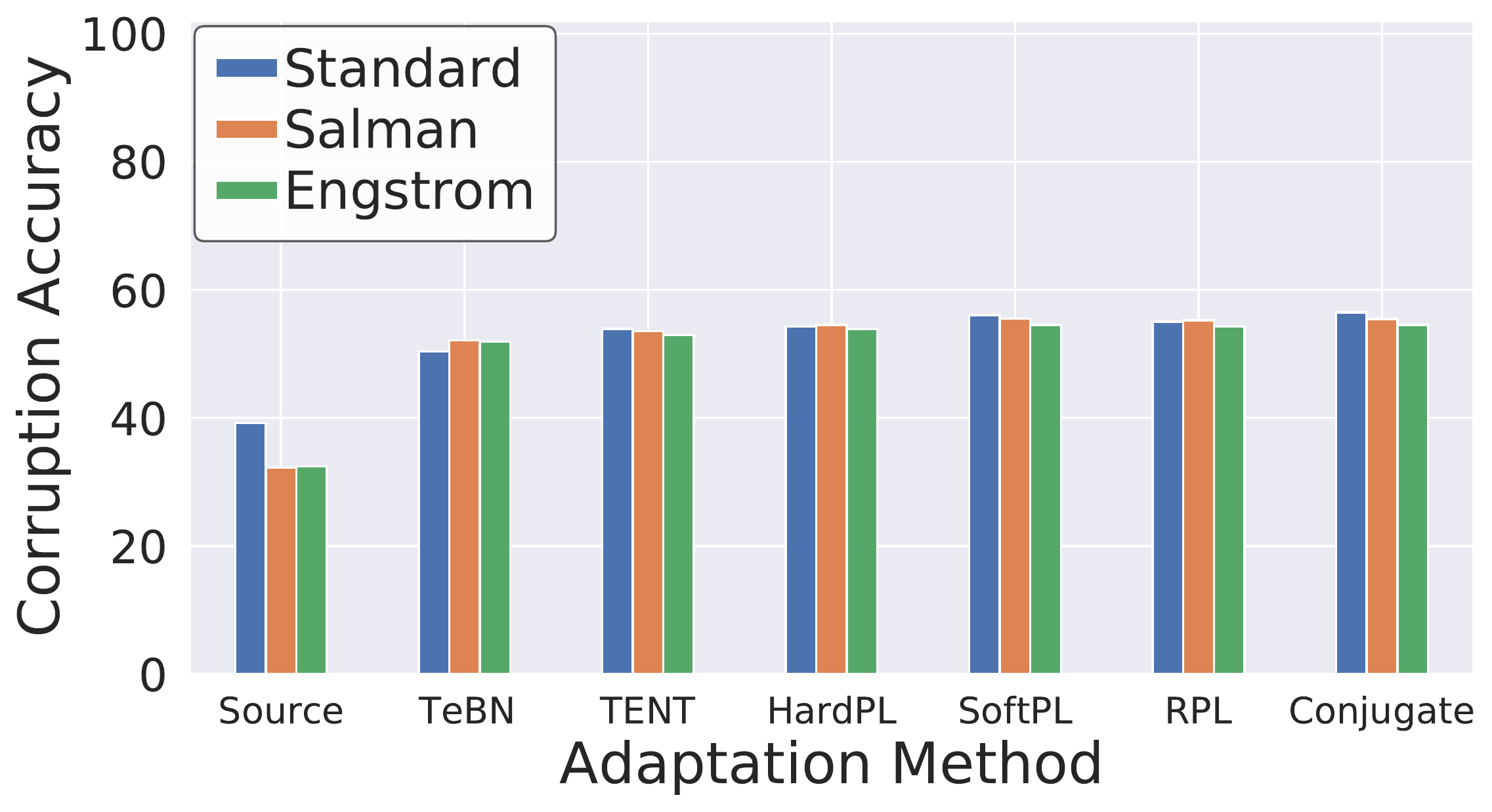} &
  \includegraphics[width=0.24\textwidth]{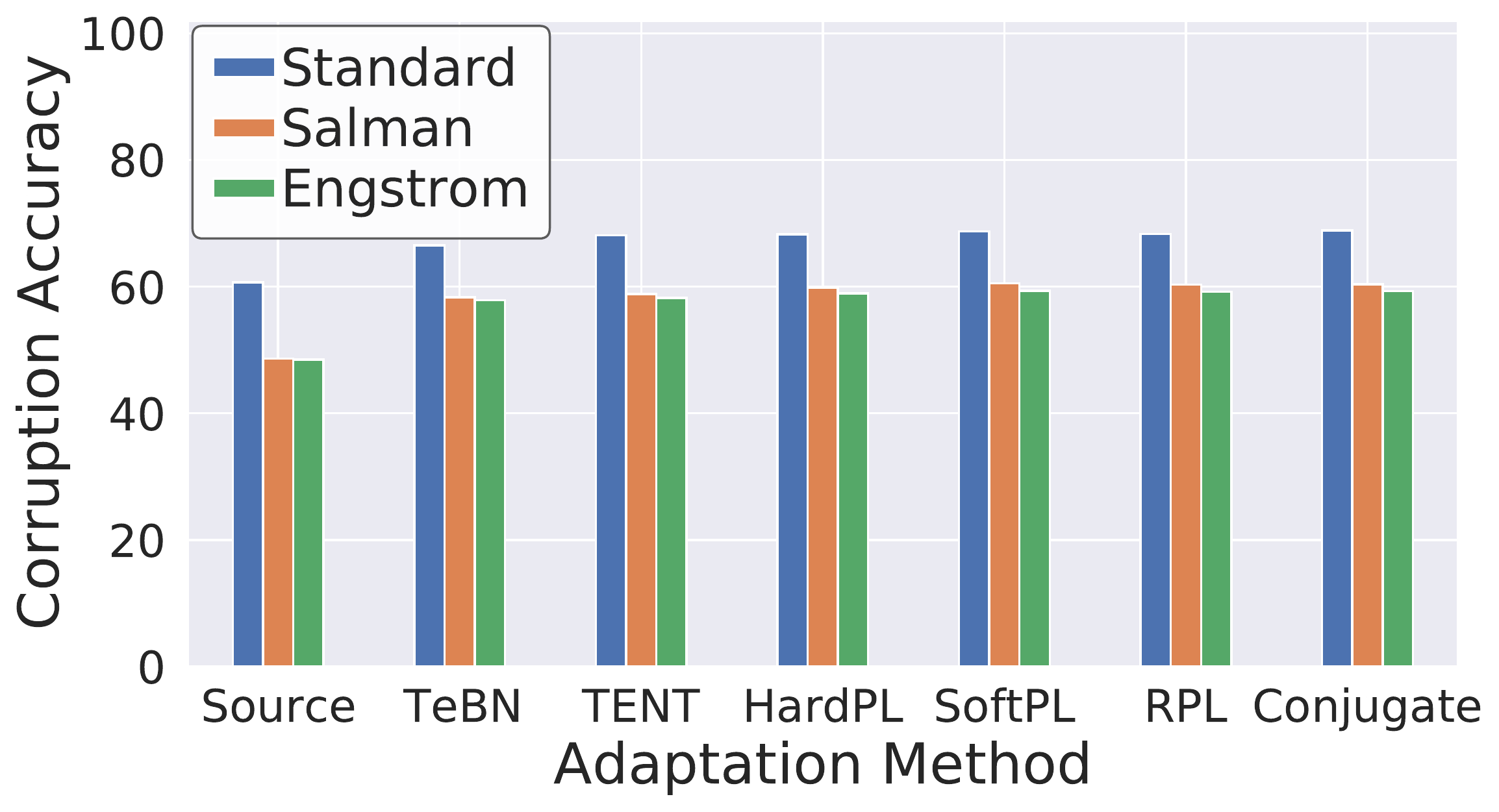} &
   \includegraphics[width=0.24\textwidth]{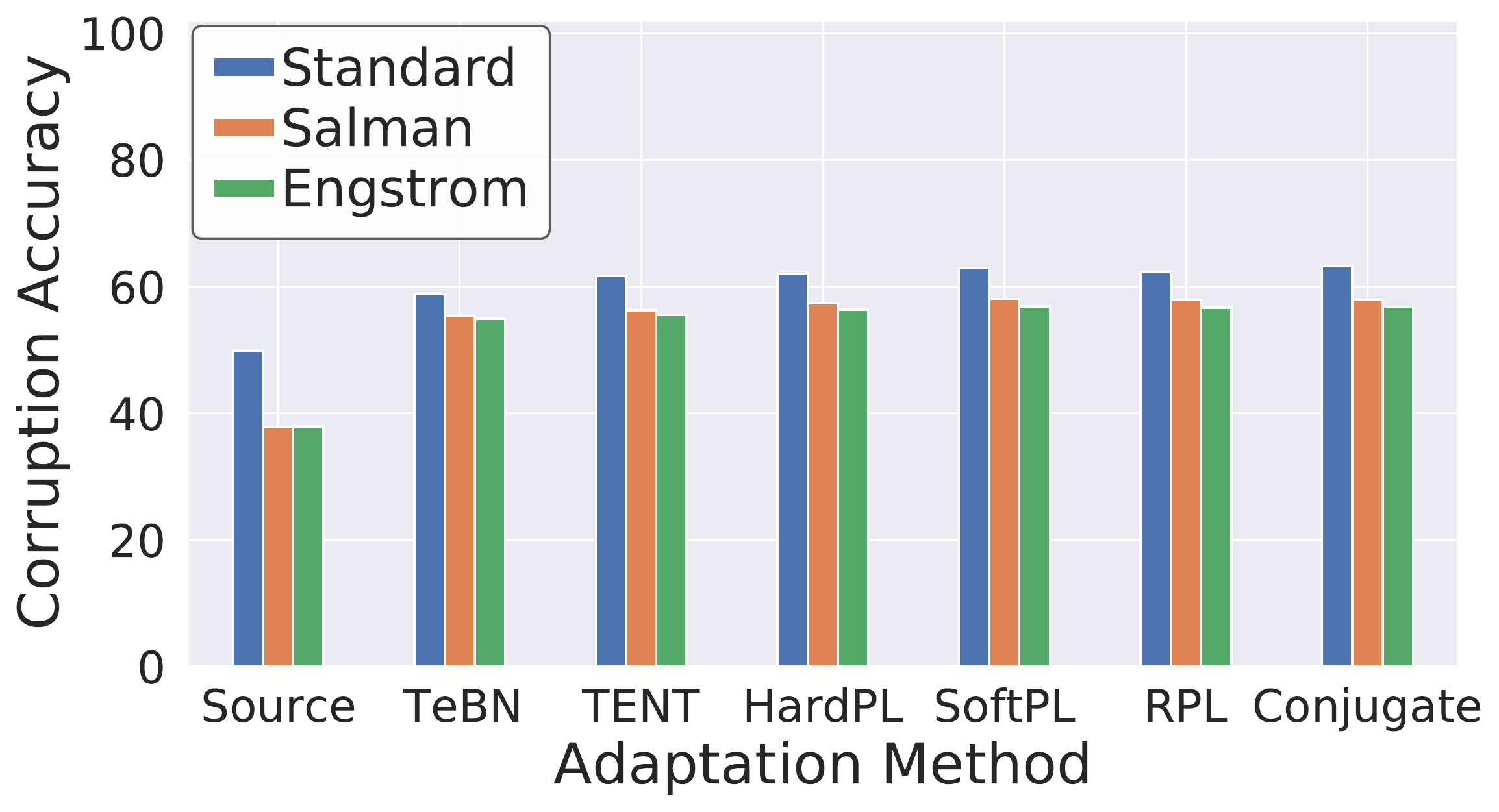} \\
  (a) Average severity& (b) Severity 1 & (c) Severity 2 \\
       &   &   \\
  \includegraphics[width=0.24\textwidth]{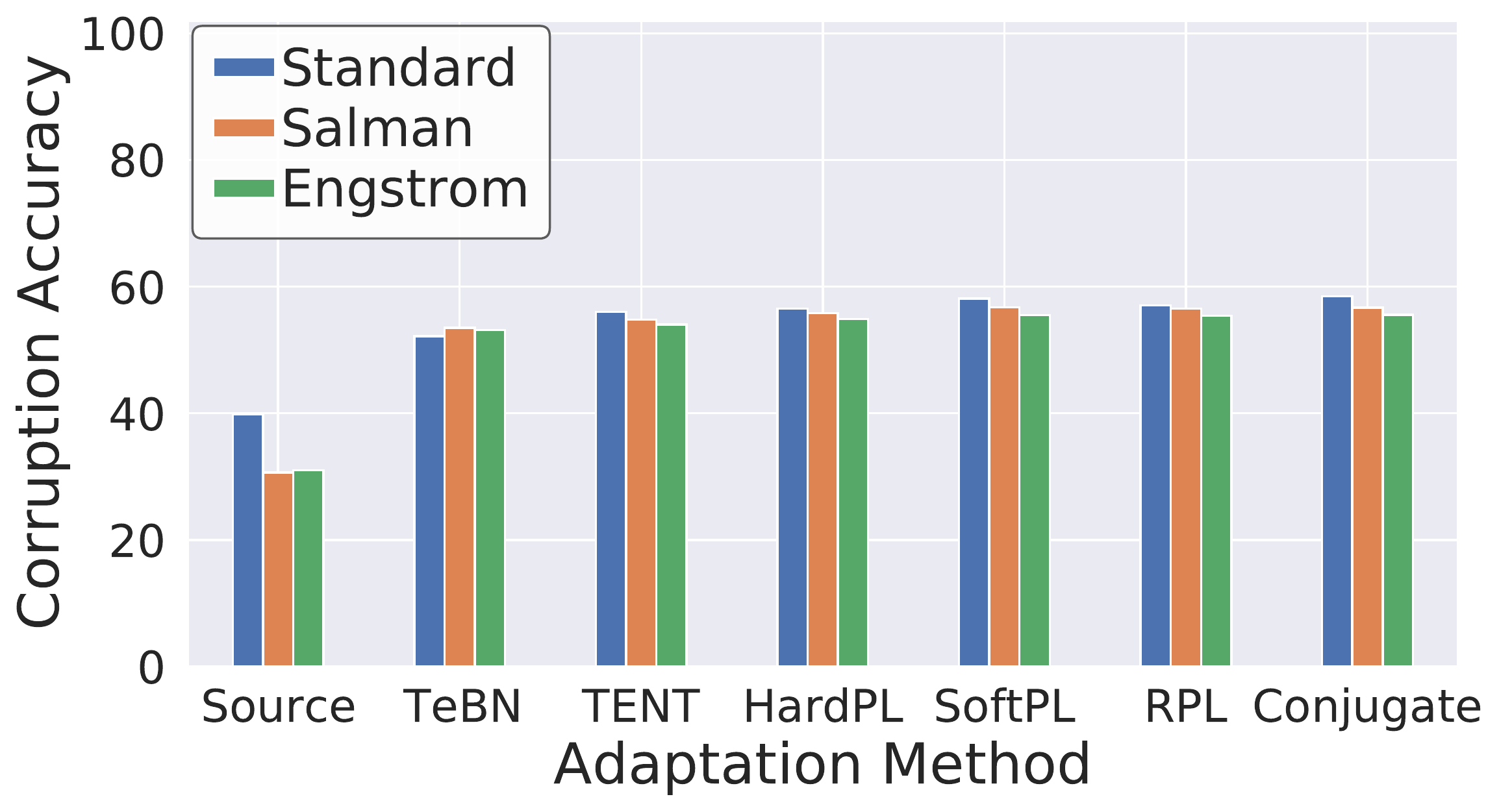}&
  \includegraphics[width=0.24\textwidth]{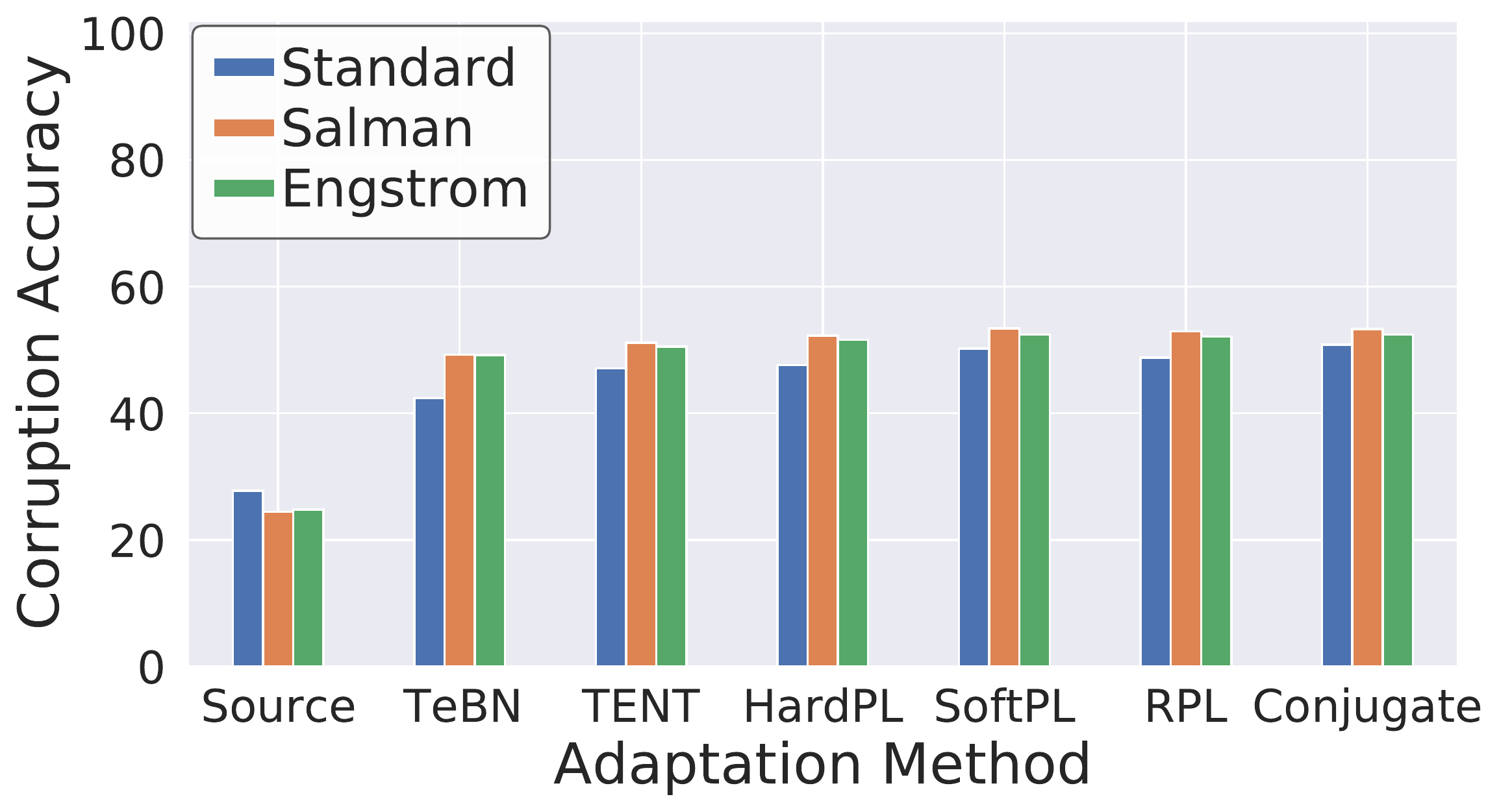} &
  \includegraphics[width=0.24\textwidth]{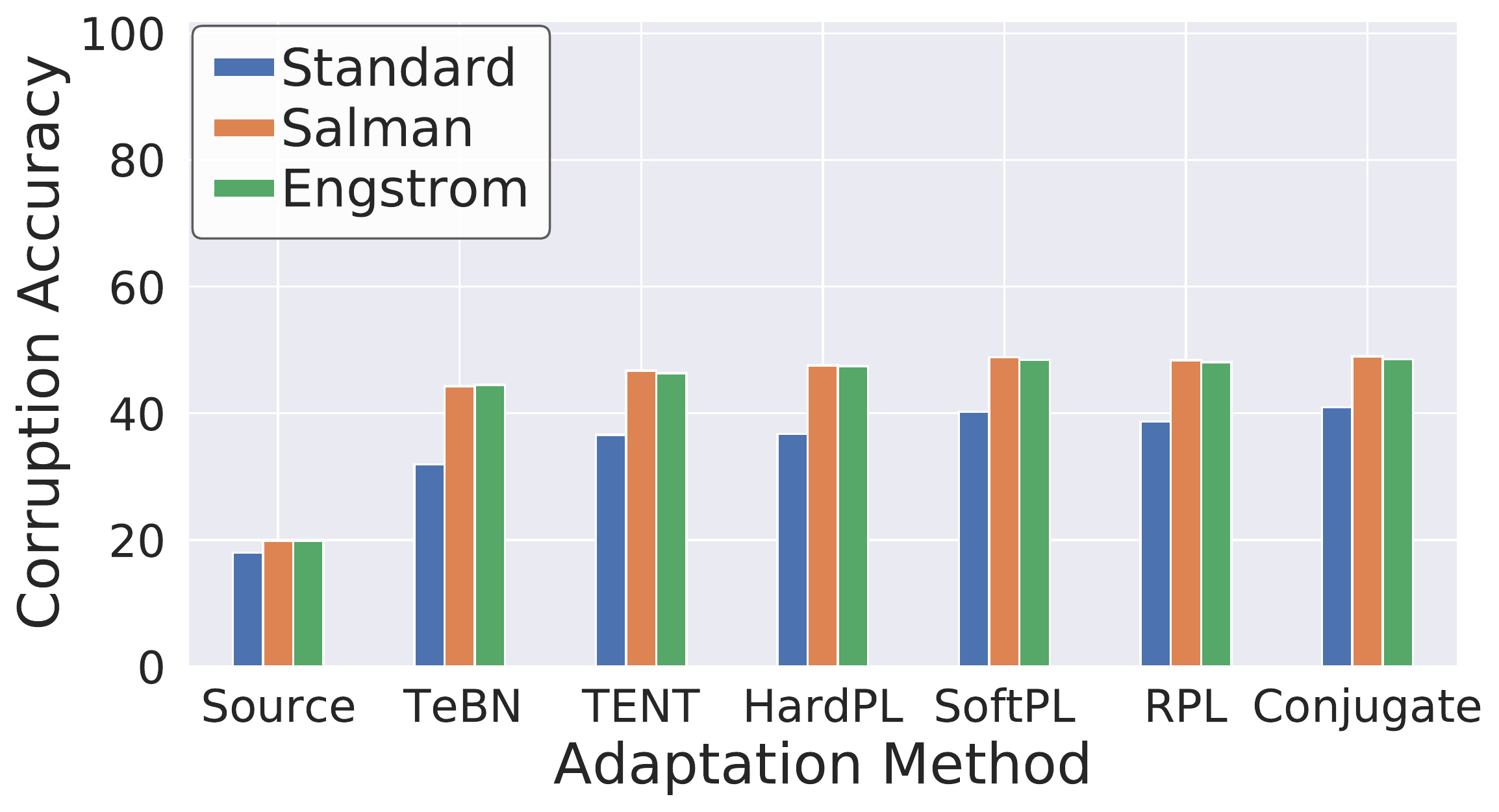}\\
   (d) Severity 3 & (e) Severity 4 &  (f) Severity 5  \\
  \end{tabular}
    \caption{ Corruption accuracy of the standard and robust models (\citet{Salman2020DoAR} and  \citet{robustness}) on ImageNet-C under different severity levels.}
    \label{Fig:INC_robust_acc}
  \end{figure}

\smallskip\noindent\textbf{Robust models achieve comparable performance with the standard model on ImageNet-C if TTA is deployed (Figure \ref{Fig:INC_robust_acc}).} We then evaluate the corruption accuracy of robust models on the ImageNet-C dataset.
Specifically, adversarially trained ResNet 50 models originated by \citet{Salman2020DoAR} and \citet{robustness} are selected. 
We observe that the average corruption accuracy of adversarial models can significantly benefit from test-time adaptation methods.
For example, the average corruption accuracy of robust models is $\sim$10\% less than the standard model but reaches similar performance when TTA is deployed. 
Furthermore, robust models generally gain higher accuracy on larger severity than the standard ones (e.g., $\sim$10\% improvement on severity 5 with TTA).

\subsection{Mitigating Distribution Invading Attacks by Robust Models on CIFAR-C Benchmarks}
\label{append:romodel_DIA}

\begin{figure}[H]
\centering
\begin{tabular}{cc}
  \includegraphics[width=0.35\textwidth]{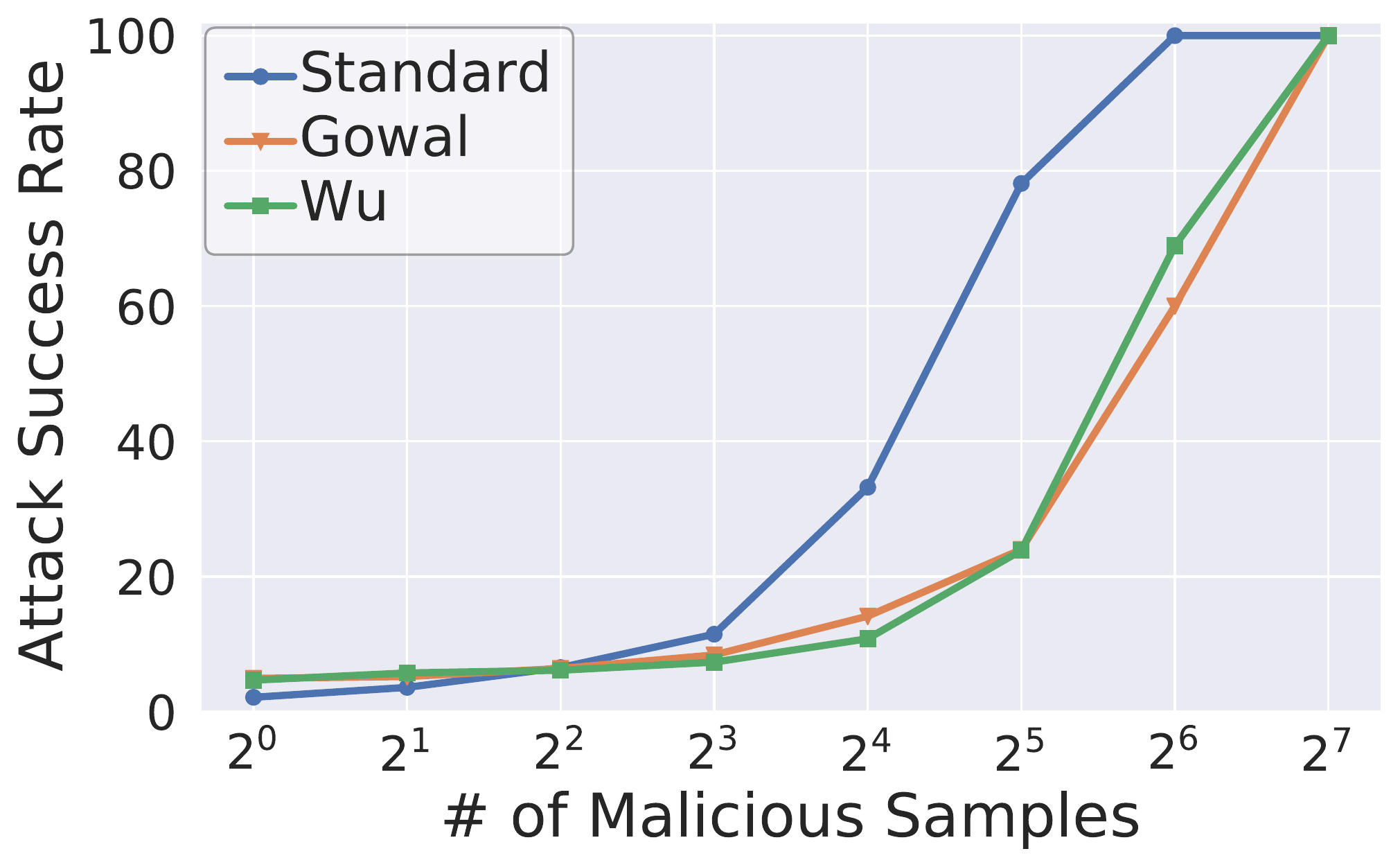} &
 \includegraphics[width=0.35\textwidth]{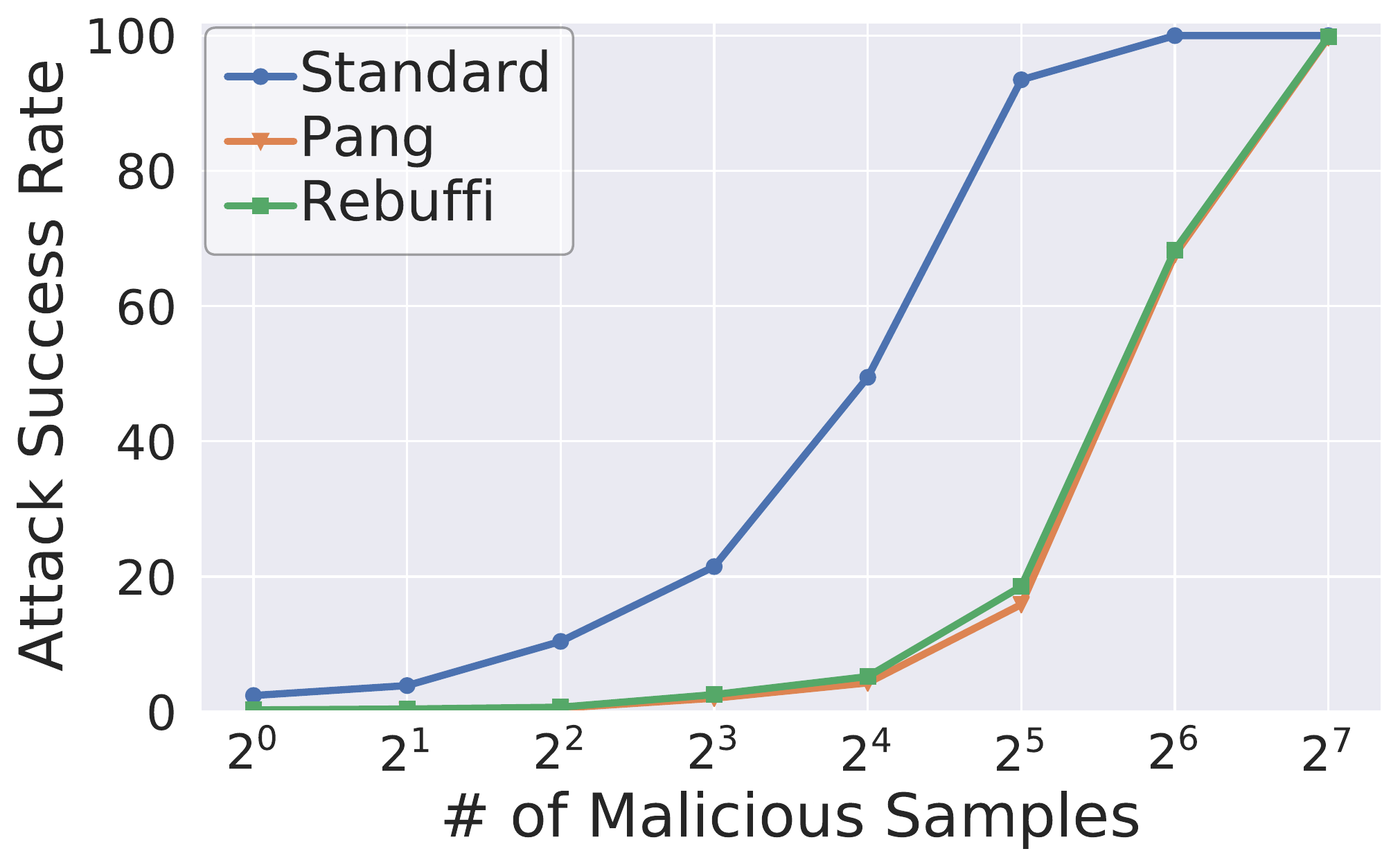} \\
 (a) CIFAR-10-C & (b) CIFAR-100-C   \\
\end{tabular}
	\caption{Attack success rate of our proposed attacks against robust models across 
 the numbers of malicious samples. As a reference, 128 is 64\% of the whole batch containing 200 samples. Robust models for CIFAR-10-C are \citet{Gowal2021ImprovingRU} and  \citet{Wu2020AdversarialWP}; robust models for CIFAR-100-C are \citet{Pang2022RobustnessAA} and \citet{Rebuffi2021FixingDA}. [TTA method: \textbf{TeBN}] }
	\label{Fig:CIFAR_robust}
\end{figure}

In Figure~\ref{Fig:CIFAR_robust}, we report the ASR of standard and two robust models across the number of malicious samples on the CIFAR-C dataset. 
\textbf{Adversarial models mitigate the vulnerability against DIA but cannot fully alleviate it.}
For example, with 32 malicious samples, robust models
degrade $\sim$60\% and  $\sim$70\% ASR for CIFAR-10-C and CIFAR-100-C, respectively. However, increasing malicious samples still breaks robust source models.

\subsection{Understanding the Layer-wise Batch Normalization behaviors}
\label{sec:understandBN}

\begin{figure}[htbp]
\centering
    \includegraphics[width=0.88\textwidth]{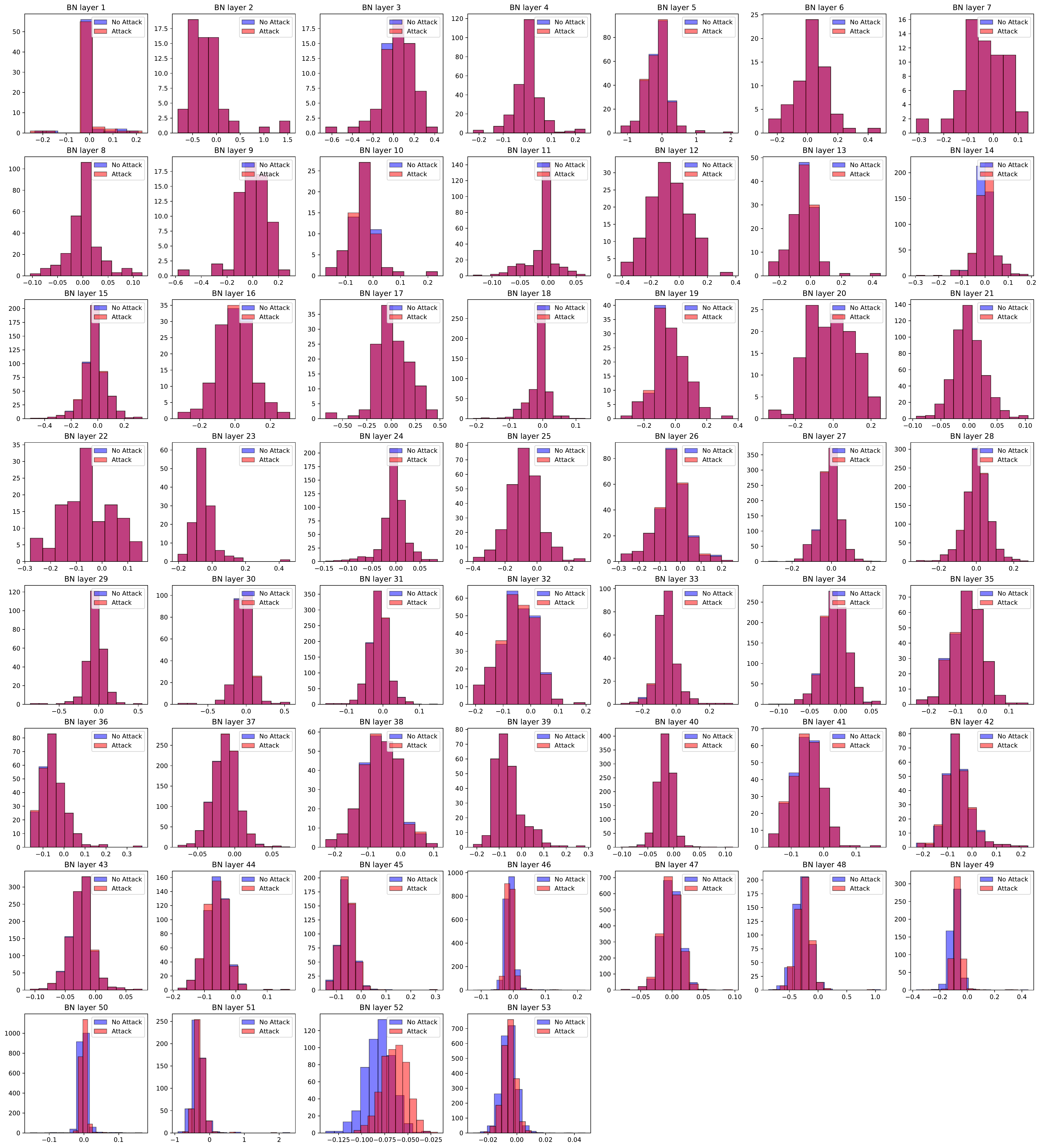}
    \caption{Full layer-by-layer 
 visualization of BN mean histogram with and without DIA attacks. The distribution differs on the last few layers.}
    \label{Fig:BN_mean_vis}
\end{figure}

\begin{figure}[htbp]
\centering
    \includegraphics[width=0.88\textwidth]{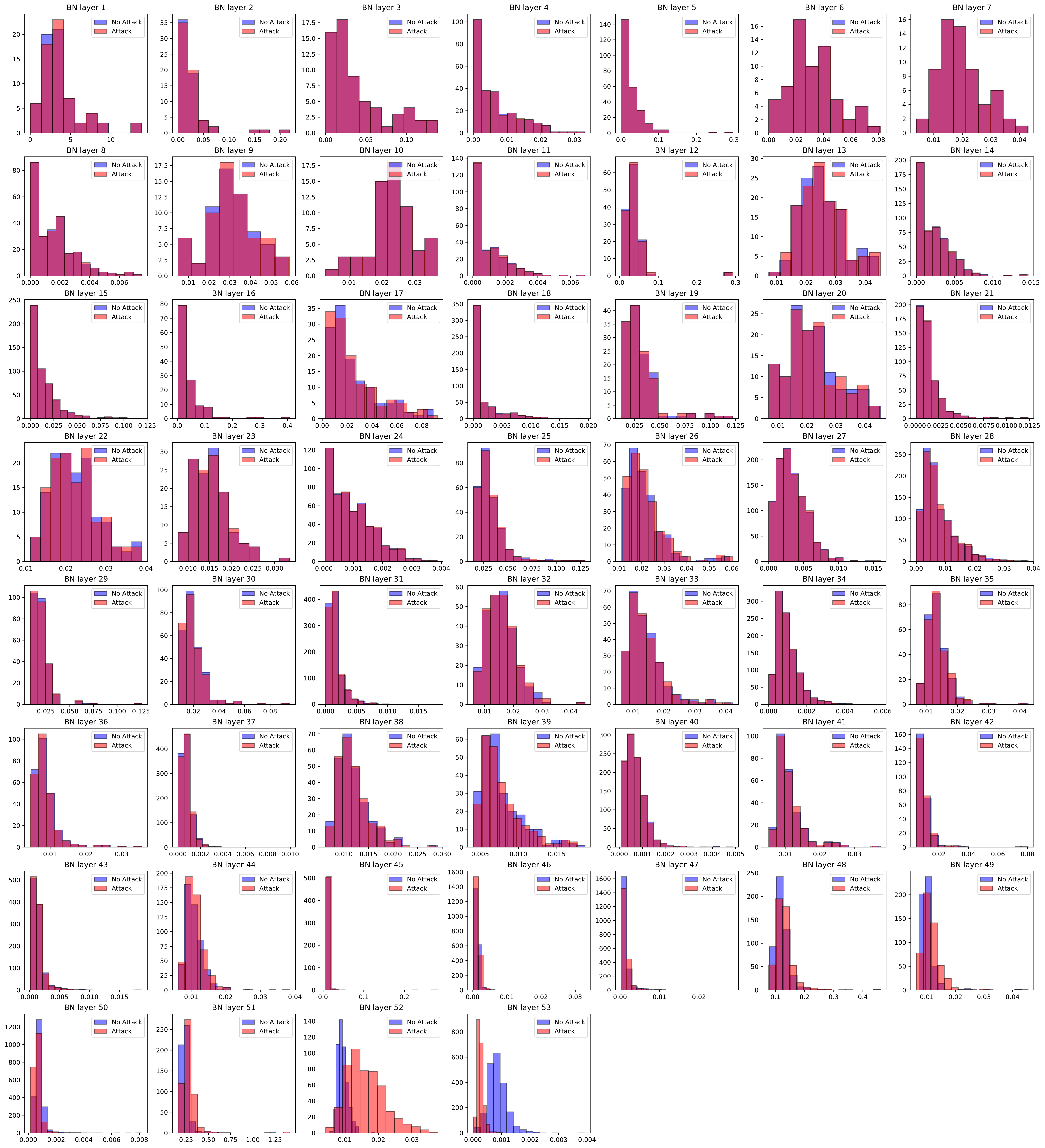}
    \caption{Full layer-by-layer visualization of BN variance histogram with and without DIA attacks. The distribution differs on the last few layers. }
    \label{Fig:BN_var_vis}
\end{figure}

To further design a more robust approach, we seek to understand what happens to BN statistics when deploying the Distribution Invading Attack. 
Therefore, we visualize the BN mean histogram and BN variance histogram of all 53 layers in Figure \ref{Fig:BN_mean_vis} and \ref{Fig:BN_var_vis}, with and without attacks. 
We use the \emph{Wasserstein distance} with $\ell_1$-norm to measure the discrepancy between two histograms. Specifically, given two probability measures $\mu_w, \mu_v$ over $\R$, the Wasserstein distance under Kantorovich formulation \citep{kantorovich1942translocation} is defined as
\begin{align}
\wasser(\mu_w, \mu_v) := \min_{\pi \in \Pi(\mu_w, \mu_v)} \int_{\R \times \R} |z - z'| d \pi(z, z')
\end{align}
where 
$
\Pi(\mu_w, \mu_v) 
:=\left\{\pi \in \mathcal{P}(\R \times \R) \mid \int_\R \pi(z, z') dz=\mu_w, \int_{\R} \pi(z, z') dz'=\mu_v\right\}
$
denotes a collection of couplings between two distributions $\mu_w$ and $\mu_v$. 
Here, we view each histogram as a discrete probability distribution in a natural sense. 
Note that the histograms for different layers may have different overall scales; hence, we normalize the $\ell_1$ distance $|z - z'|$ by the range of the histogram of benign samples for a fair comparison.



\begin{figure}[htbp]
\centering
\begin{tabular}{cccc}
  \includegraphics[width=0.22\textwidth]{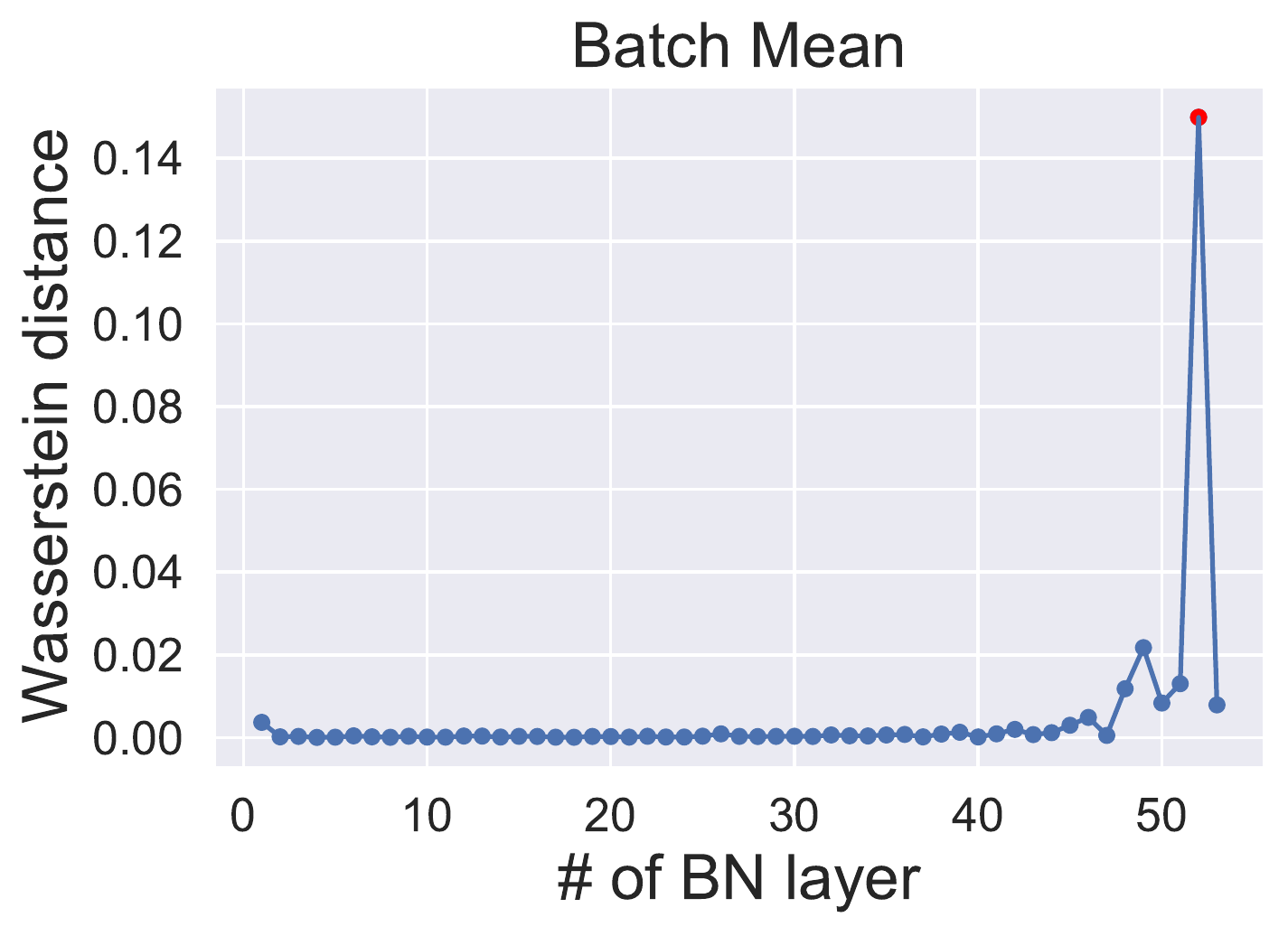} &
    \includegraphics[width=0.22\textwidth]{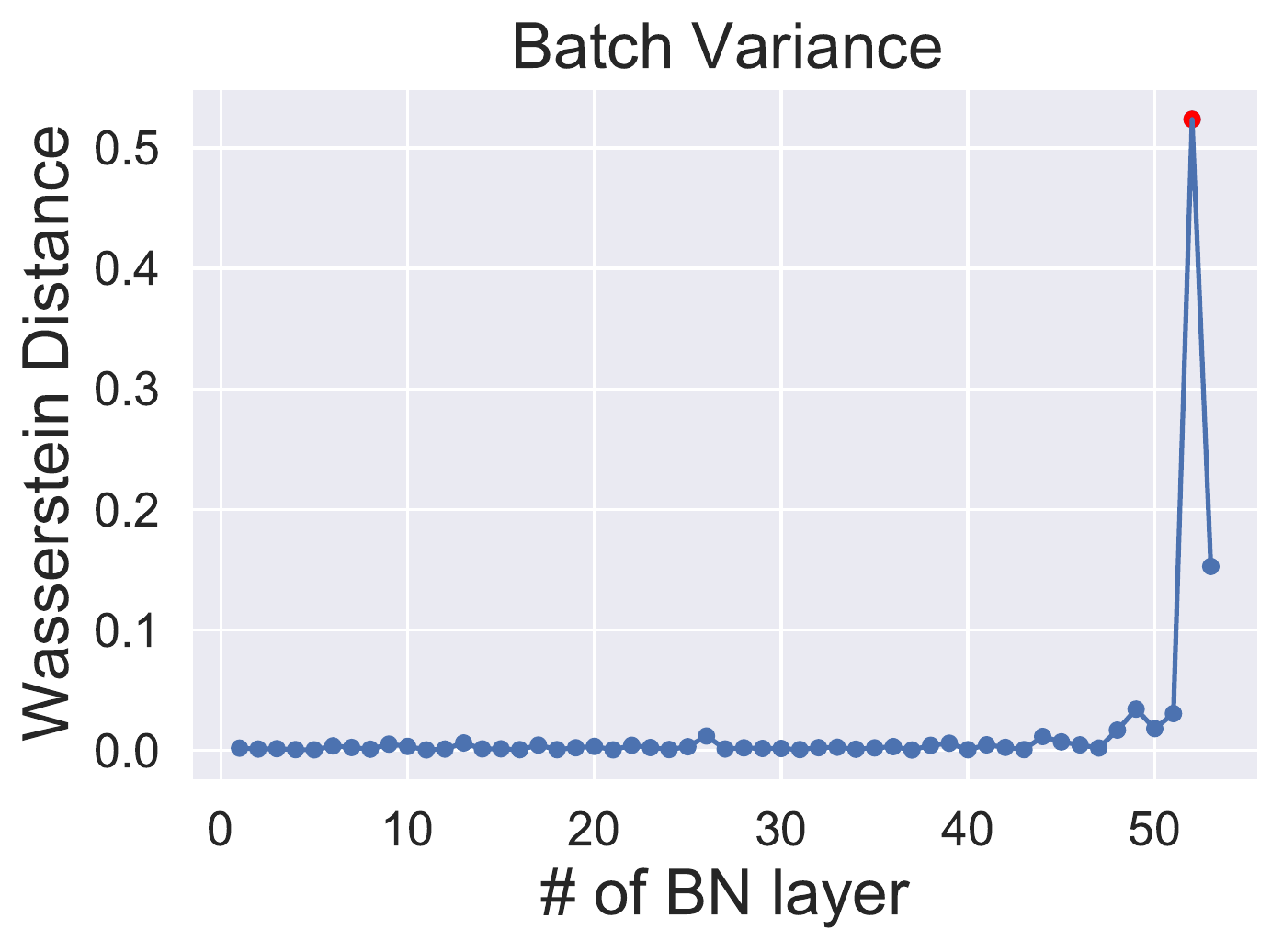} &
 \includegraphics[width=0.22\textwidth]{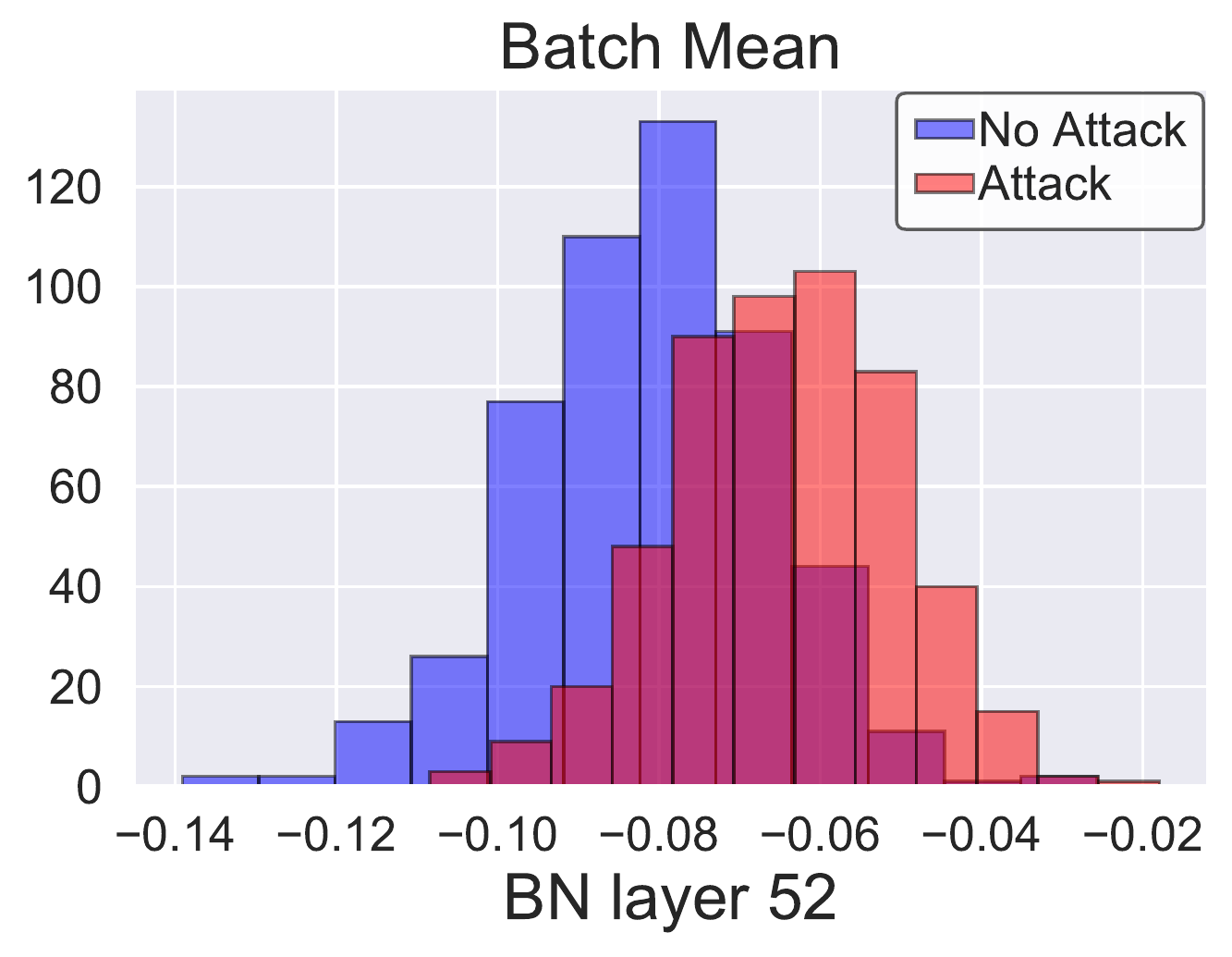}&
 \includegraphics[width=0.22\textwidth]{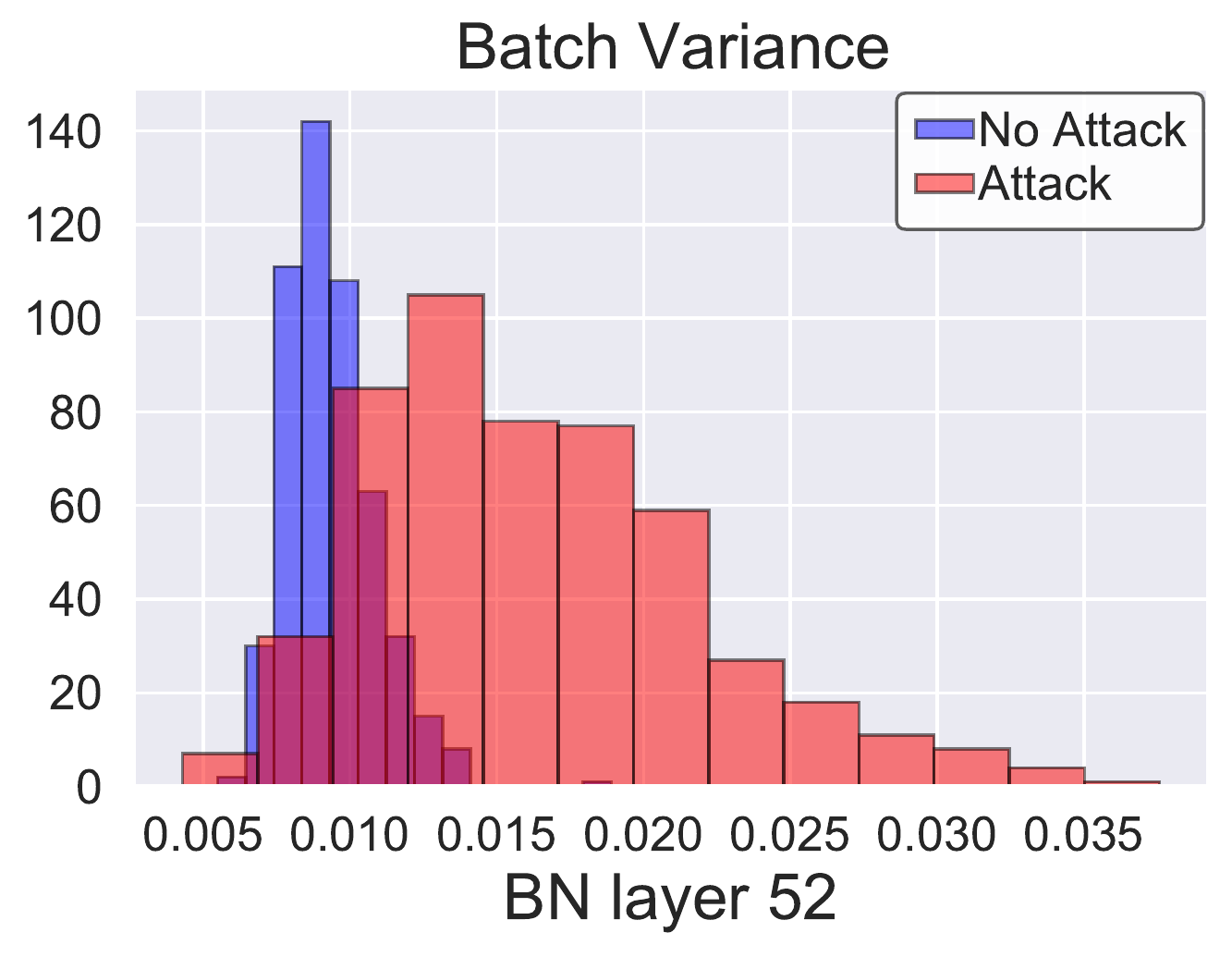}\\
 \multicolumn{2}{c}{(a) Wasserstein Distance}  & \multicolumn{2}{c}{(b) BN Statistics Histogram }  \\
\end{tabular}
\caption{ (a) Wasserstein Distance between malicious and benign BN Statistics gets unexpectedly high at BN Layer 52.  (b)  Histogram of batch mean and variance at BN Layer 52. (More figures are presented in Figure \ref{Fig:BN_mean_vis} and Figure \ref{Fig:BN_var_vis}). }
\label{Fig:IN_layer_BN}
\end{figure}

Figure \ref{Fig:IN_layer_BN}(a) presents the Wasserstein distance for each layer, and interestingly, we observe that $\wasser$ remains low on most layers but leaps on the last few layers, especially the BN layer 52. 
Then, we visualize the histograms of layer 52 in Figure \ref{Fig:IN_layer_BN}(b), which appears to be an obvious distribution shift.  

\subsection{Effectiveness of Robust BN Estimation on CIFAR-C}
\label{append:RoBNCifarC}

We report the effectiveness of our two robust estimation methods, smoothing via training-time BN statistics and adaptively selecting layer-wise BN statistics on CIFAR-C. Precisely, we use $N_m$ = 40, which is very challenging for defense (ASR reaches near-100\%). Again, we select $\tau = \{0.0, 0.2, 0.4, 0.6, 0.8, 1.0\}$ for balancing training-time and test-time BN. For adaptively selecting layer-wise BN statistics, we leverage full training-time BN ($\tau$ = 1.0) for the last $N_{tr}$ = \{0, 1, 2, 4, 8, 16\} BN layers. 

\begin{figure}[H]
\centering
\begin{tabular}{cccc}
   \includegraphics[width=0.23\textwidth]{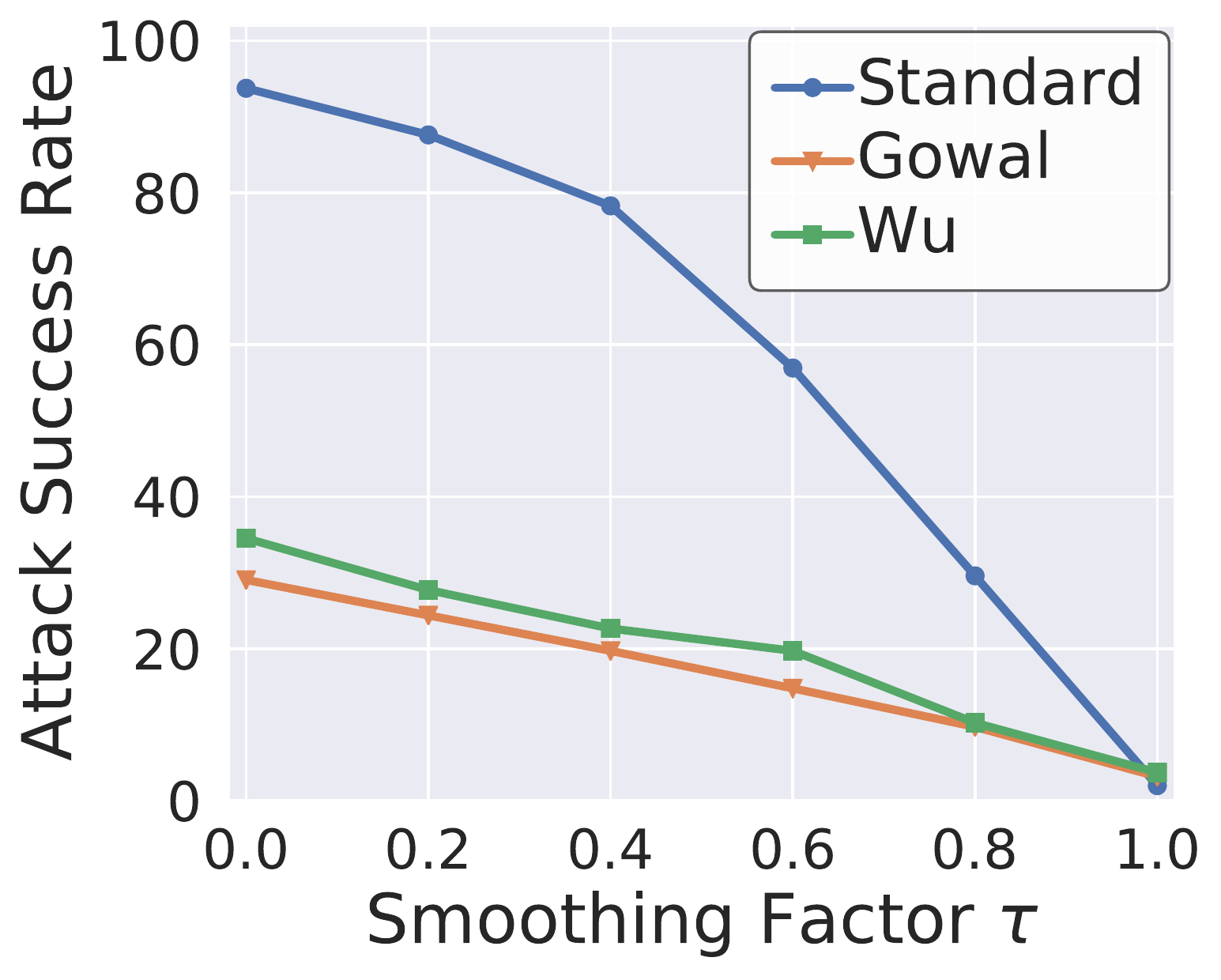} & 
 \includegraphics[width=0.23\textwidth]{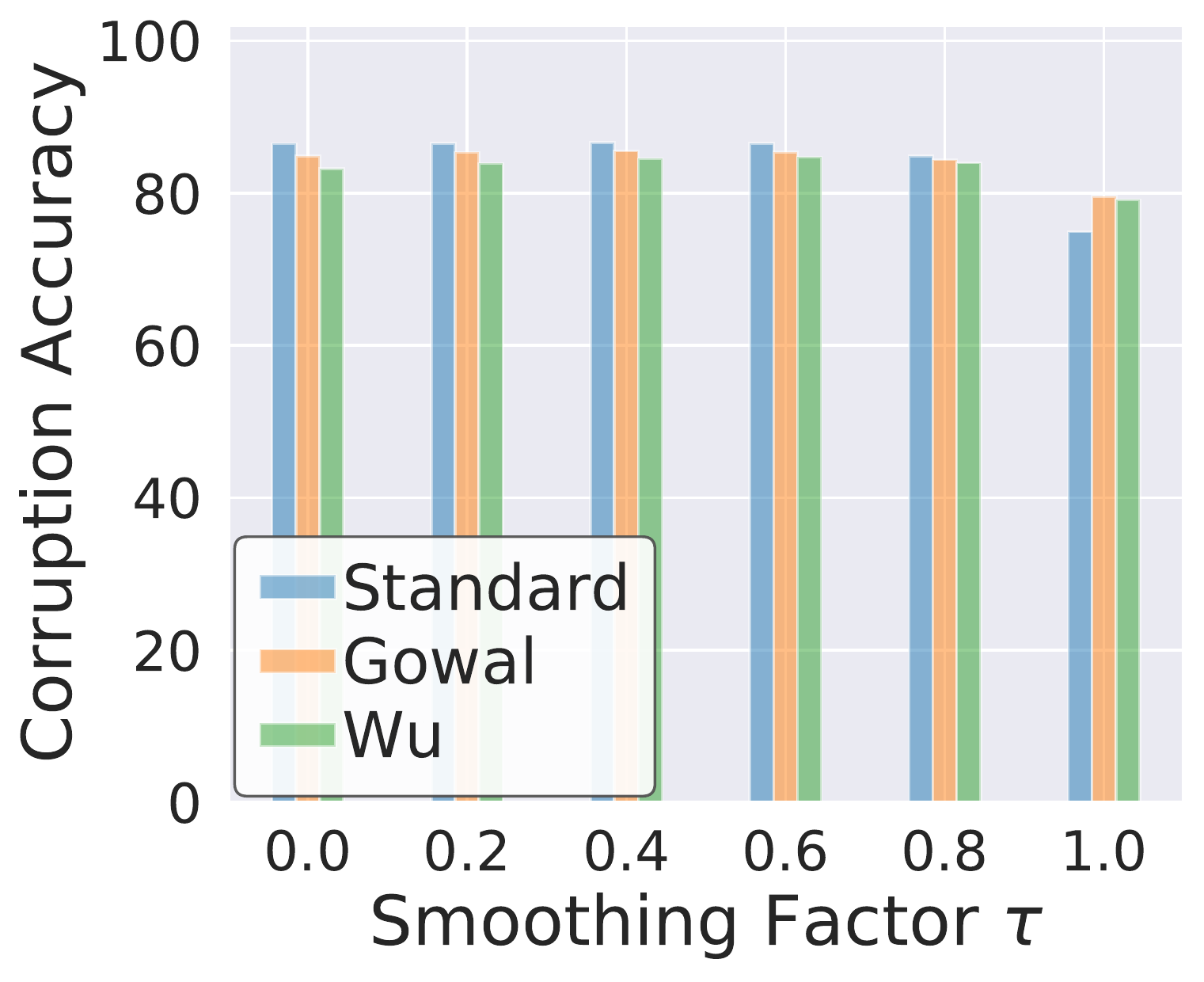} & 
  \includegraphics[width=0.23\textwidth]{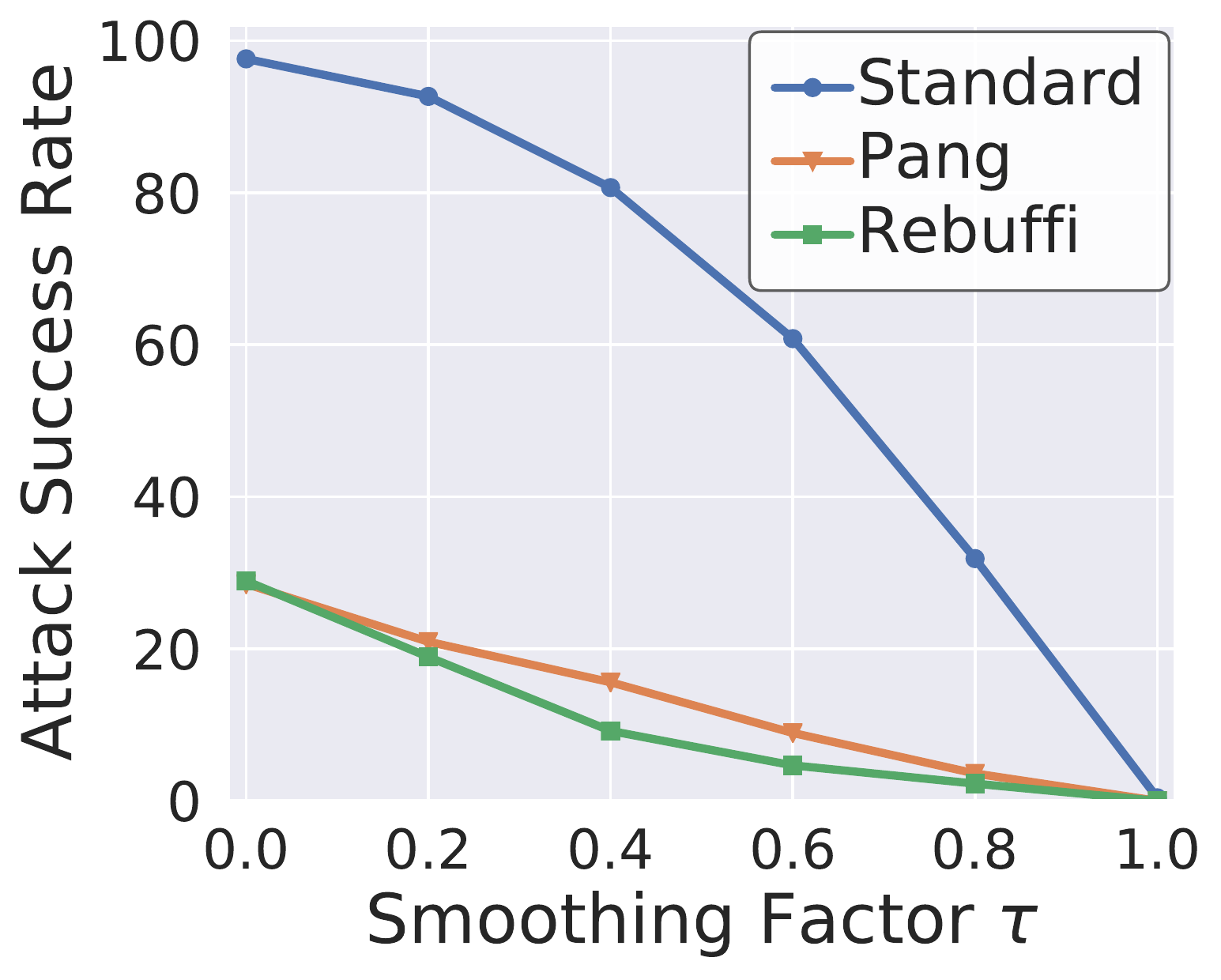} & 
 \includegraphics[width=0.23\textwidth]{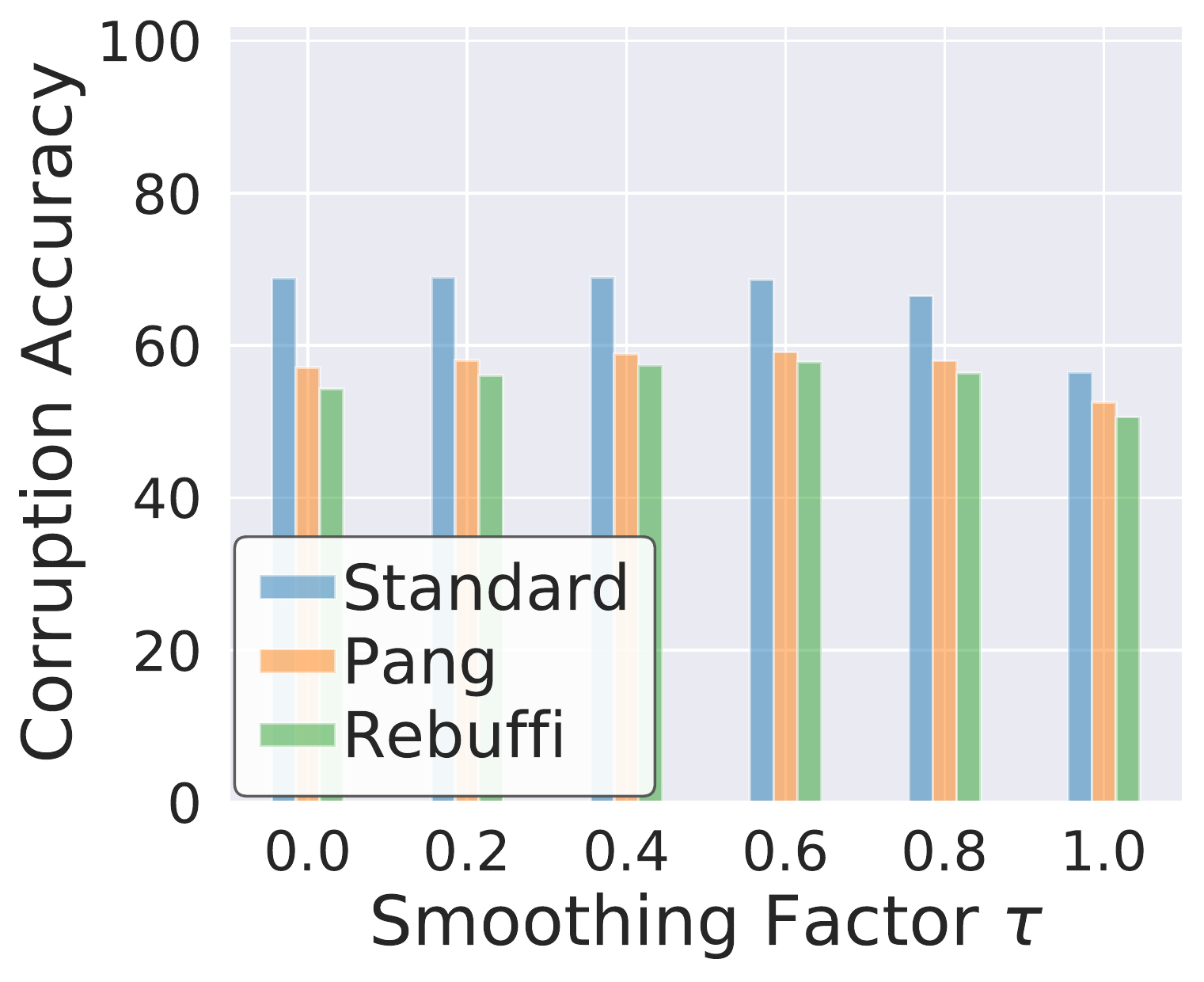} \\
 \multicolumn{2}{c}{(a) CIFAR-10-C  }  & \multicolumn{2}{c}{(b) CIFAR-100-C  } \\
\end{tabular}
\caption{Controlling $\tau$ = 0.6 can degrade the attack success rate (line) while maintaining high corruption accuracy (bar). [$N_m$=40]}
	\label{Fig:Cifar_Bayes}
\end{figure}

\begin{figure}[H]
\centering
\begin{tabular}{cccc}
  \includegraphics[width=0.23\textwidth]{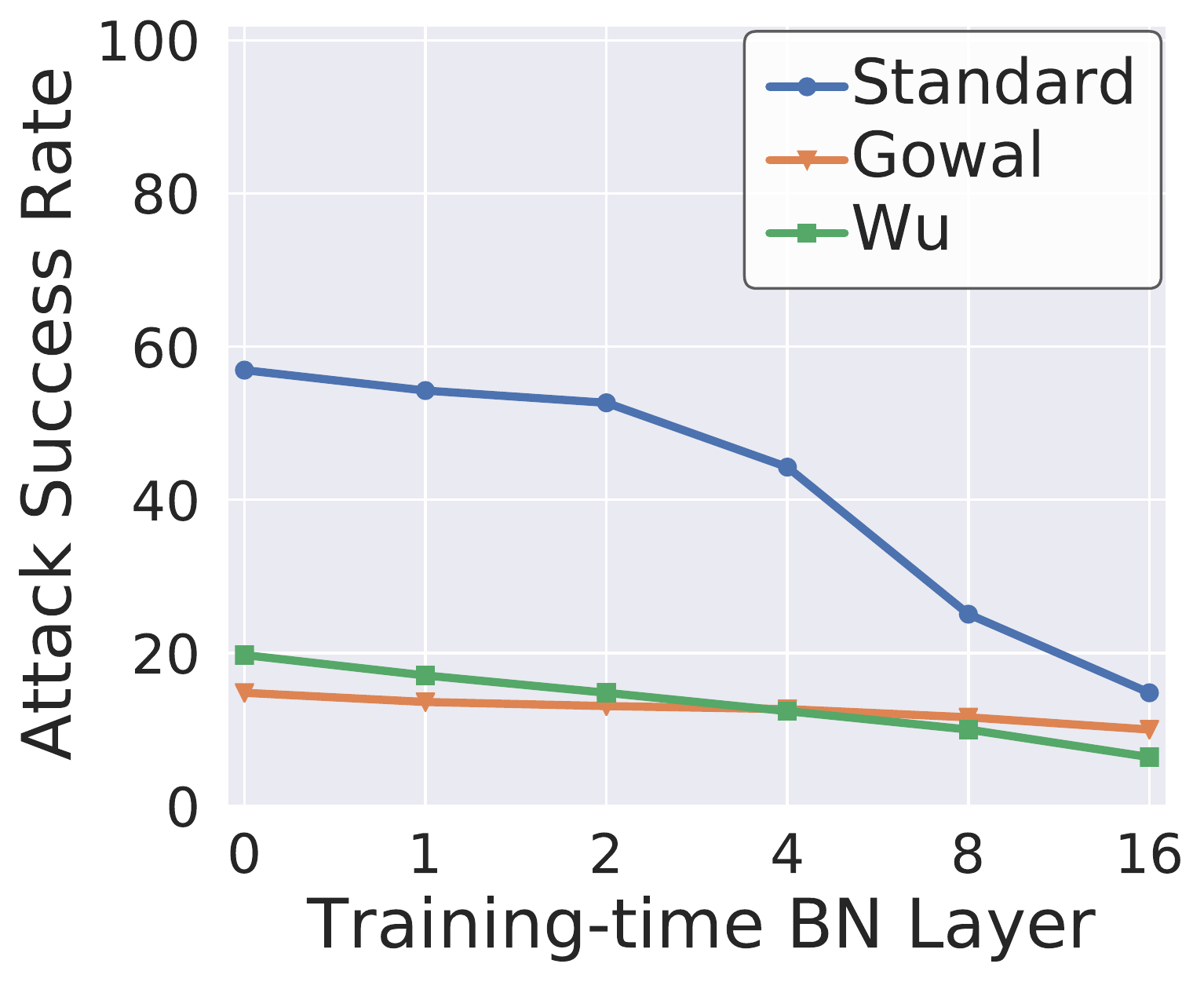} & 
 \includegraphics[width=0.23\textwidth]{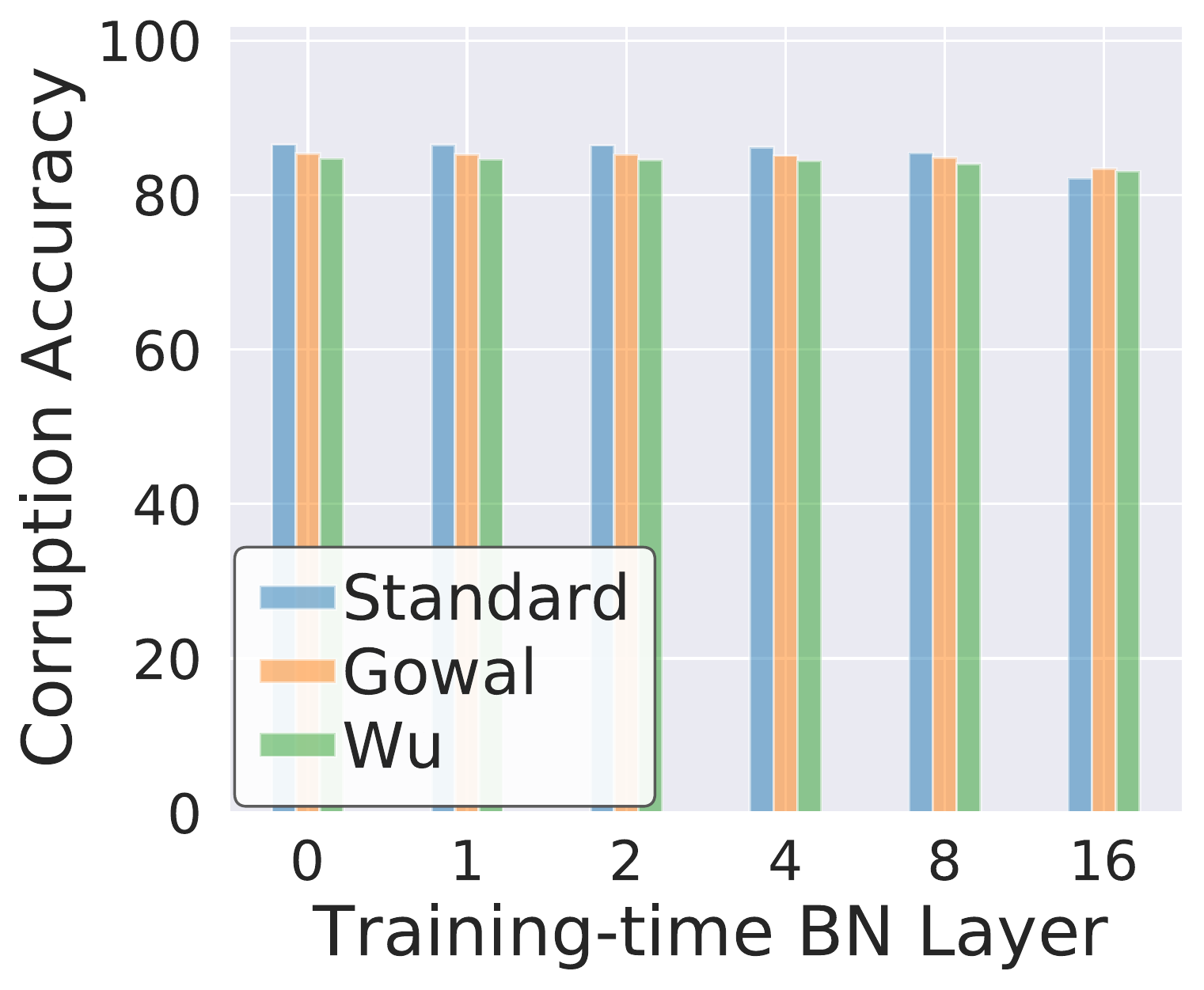} & 
  \includegraphics[width=0.23\textwidth]{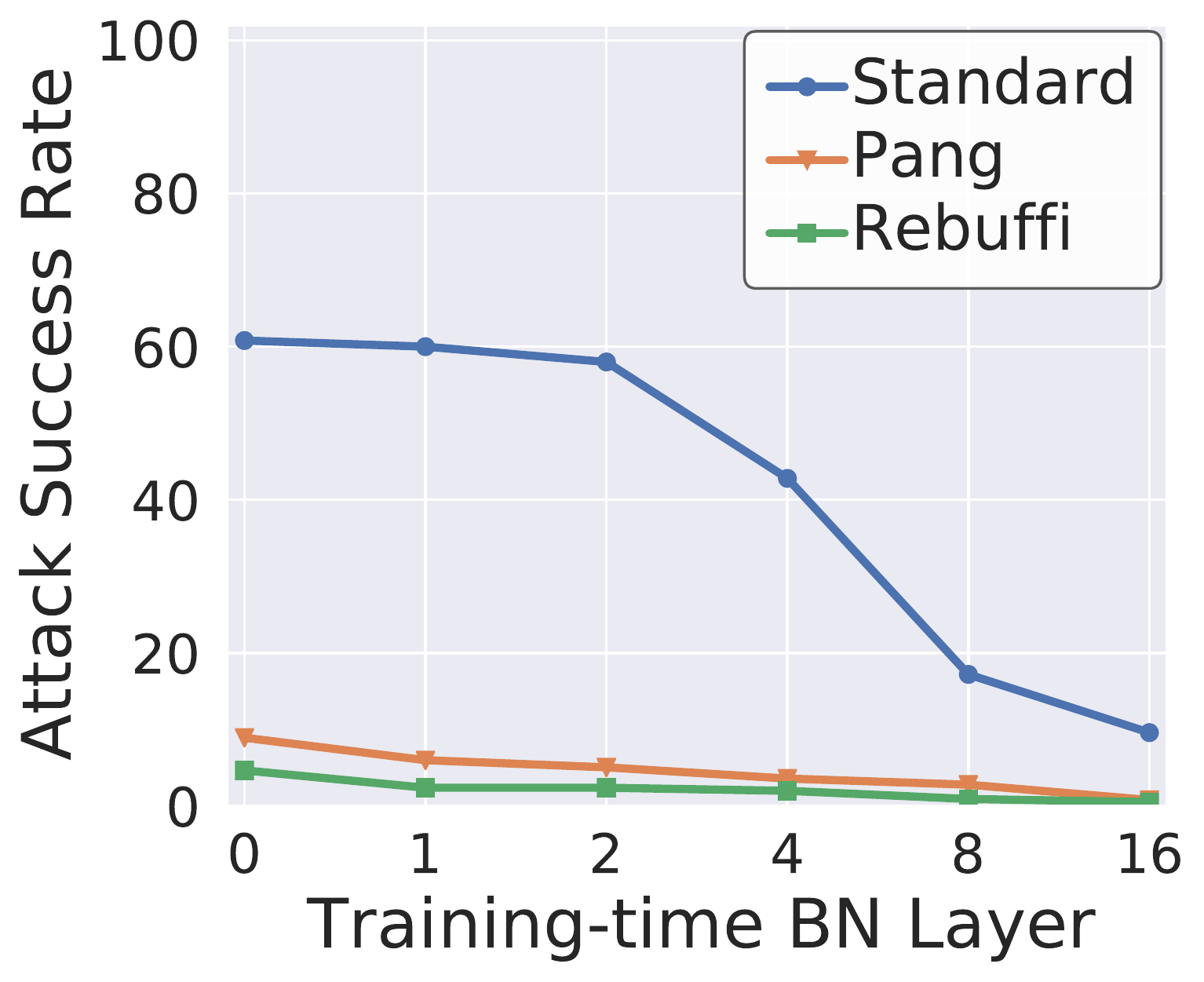} & 
 \includegraphics[width=0.23\textwidth]{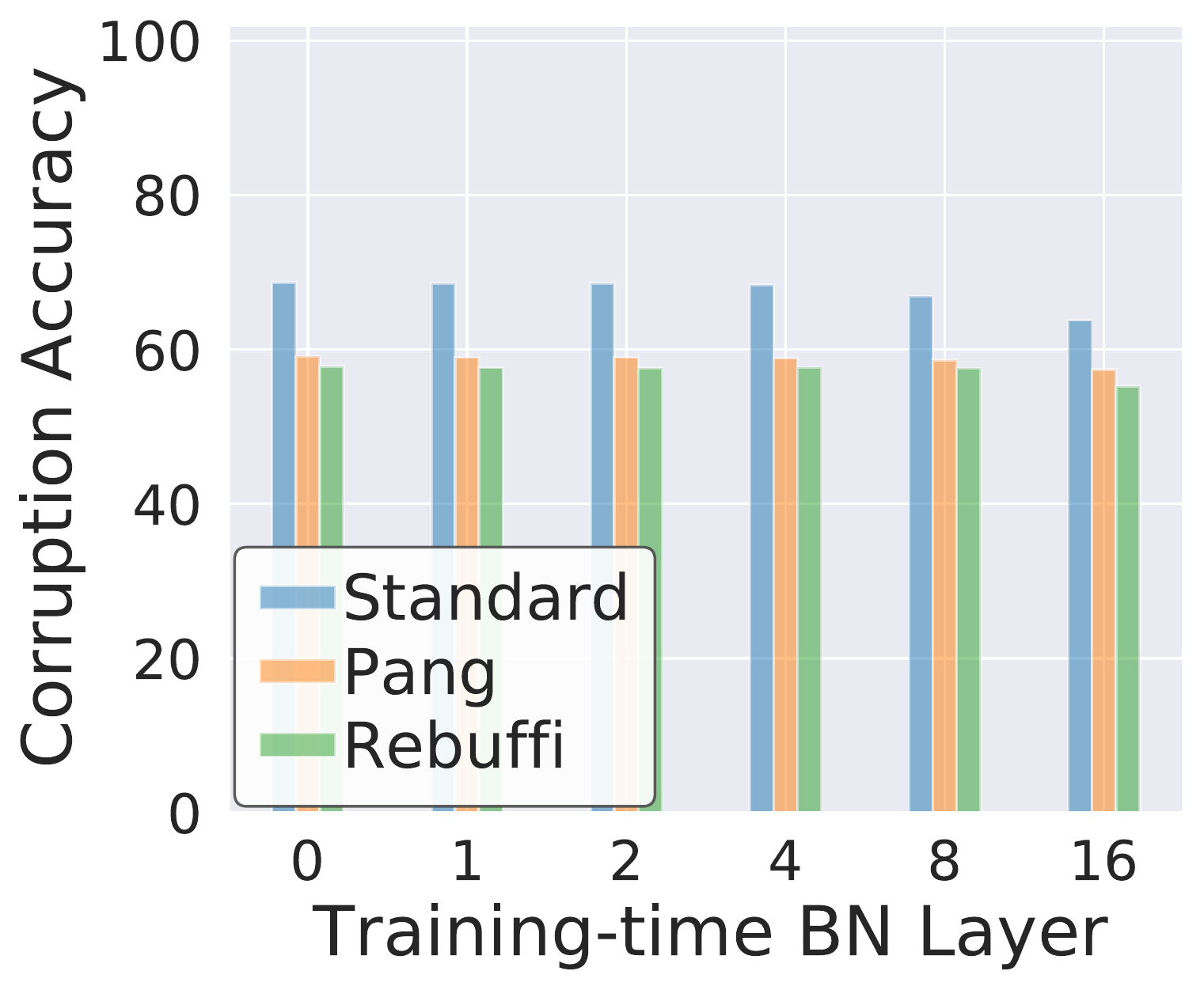} \\
 \multicolumn{2}{c}{(a) CIFAR-10-C  }  & \multicolumn{2}{c}{(b) CIFAR-100-C  } \\
\end{tabular}
\caption{Controlling layer-wise BN can degrade the attack success rate (line) while maintaining high corruption accuracy (bar). [$N_m$=40; $\tau$=0.6]}
	\label{Fig:Cifar_Bayes_layer}
\end{figure}

\smallskip\noindent\textbf{Robustly estimating the final BN statistics significantly enhance performance against DIA.}
Figure \ref{Fig:Cifar_Bayes} demonstrates the effectiveness of smoothing with training-time BN, which degrades the ASR $\sim$30\% for the standard model if we select $\tau$ = 0.6. Then, we stick with $\tau$ = 0.6 and apply layer-wise BN. As a result, the ASR further degrades by about 15\%  without sacrificing the corruption accuracy (shown in Figure \ref{Fig:Cifar_Bayes_layer}). 
In conclusion, applying two of our robust estimating methods can degrade $\sim$45\% for the standard model and $\sim$30\% for the adversarial model in total.


Our method involves a trade-off between TTA performance and robustness to DIA attacks. It does not inherently eliminate vulnerability when exploiting the test batch information. In addition, selecting the right hyper-parameters (i.e., $\tau$ and $N_{tr}$) without the out-of-distribution information is another challenging problem.

\subsection{Addtional Experiments of Robust BN Estimation on ImageNet-C}
\label{append:RoBNImnC}

\begin{figure}[H]
\centering
\begin{tabular}{cccc}
  \includegraphics[width=0.23\textwidth]{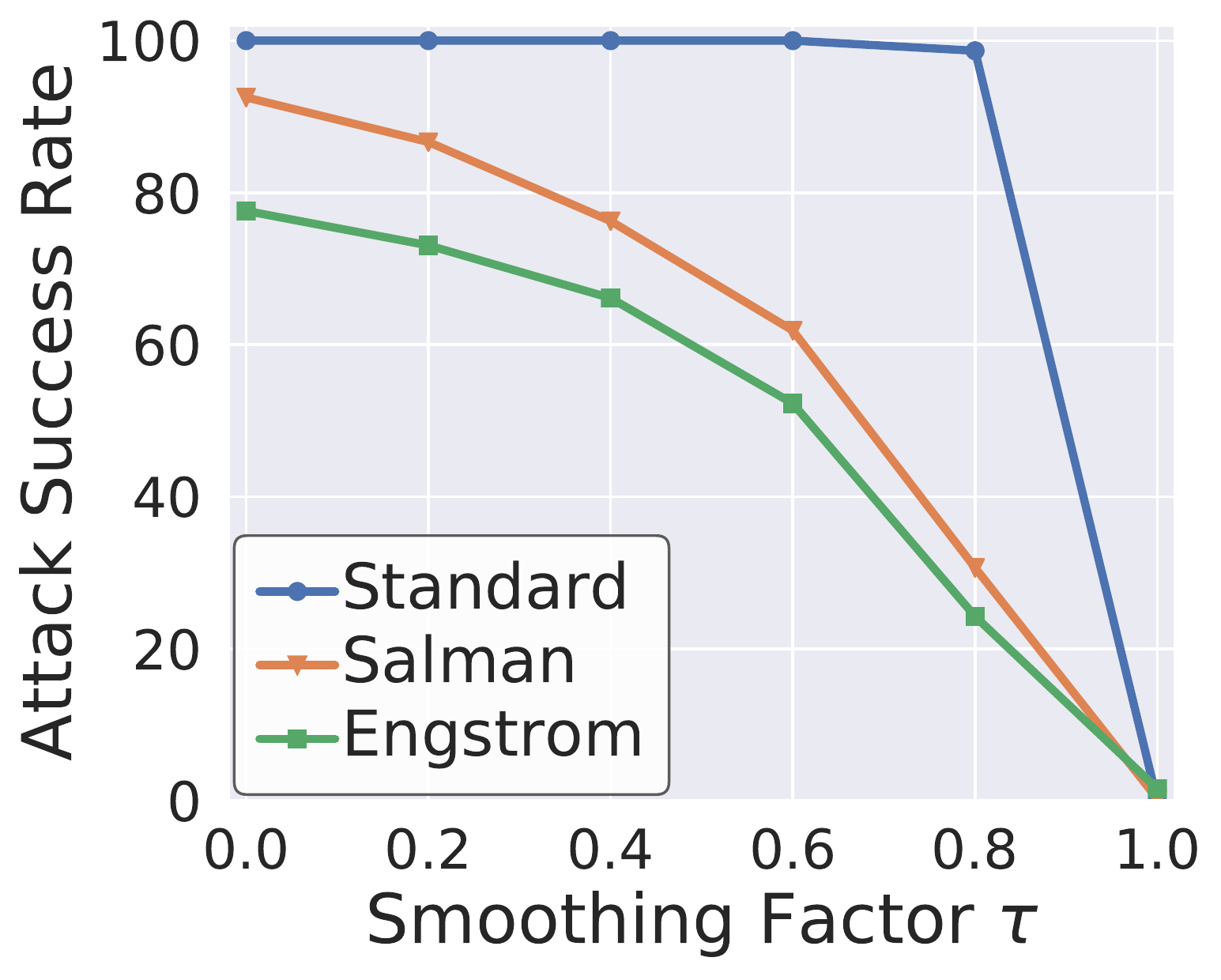} & 
 \includegraphics[width=0.23\textwidth]{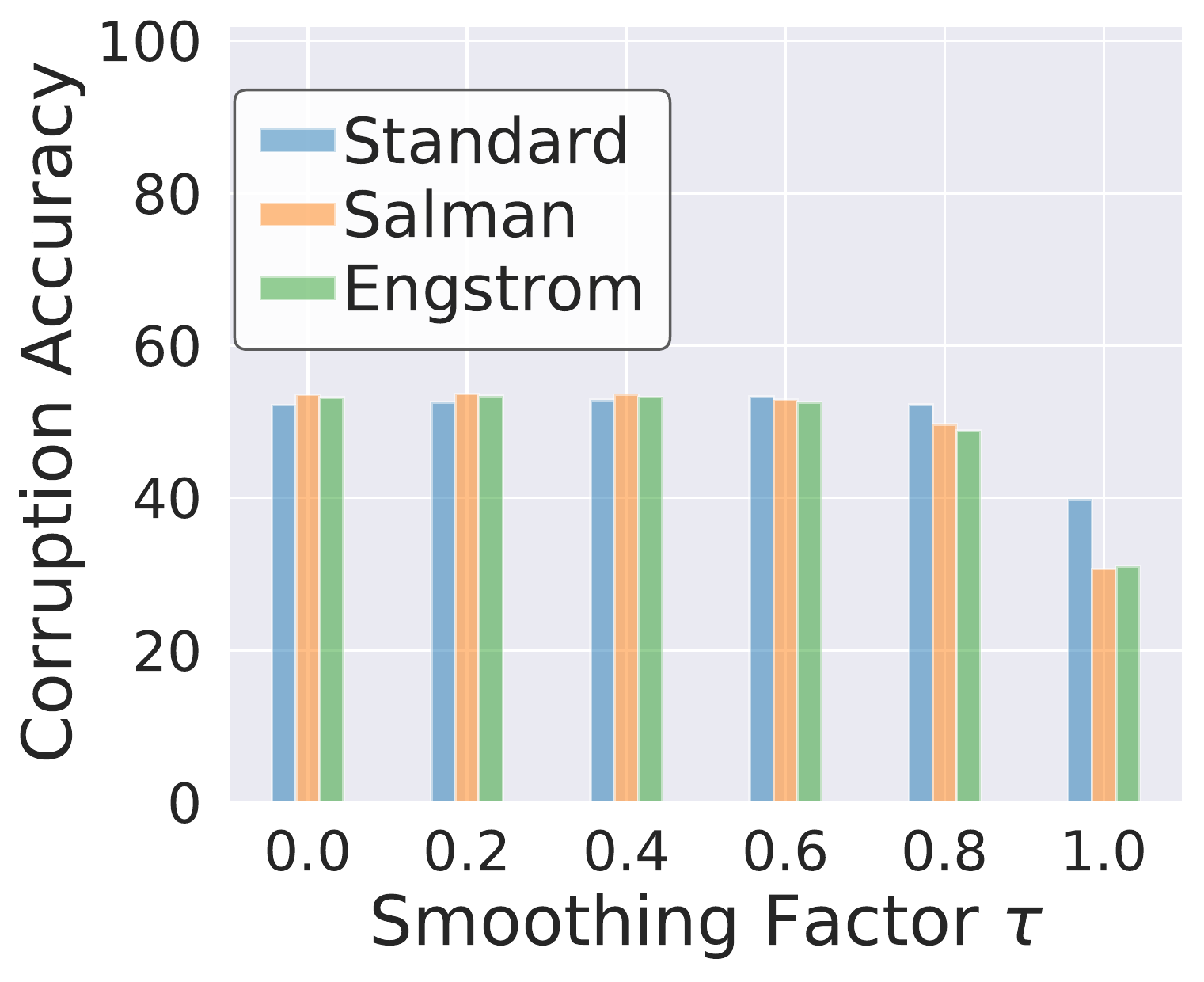} & 
  \includegraphics[width=0.23\textwidth]{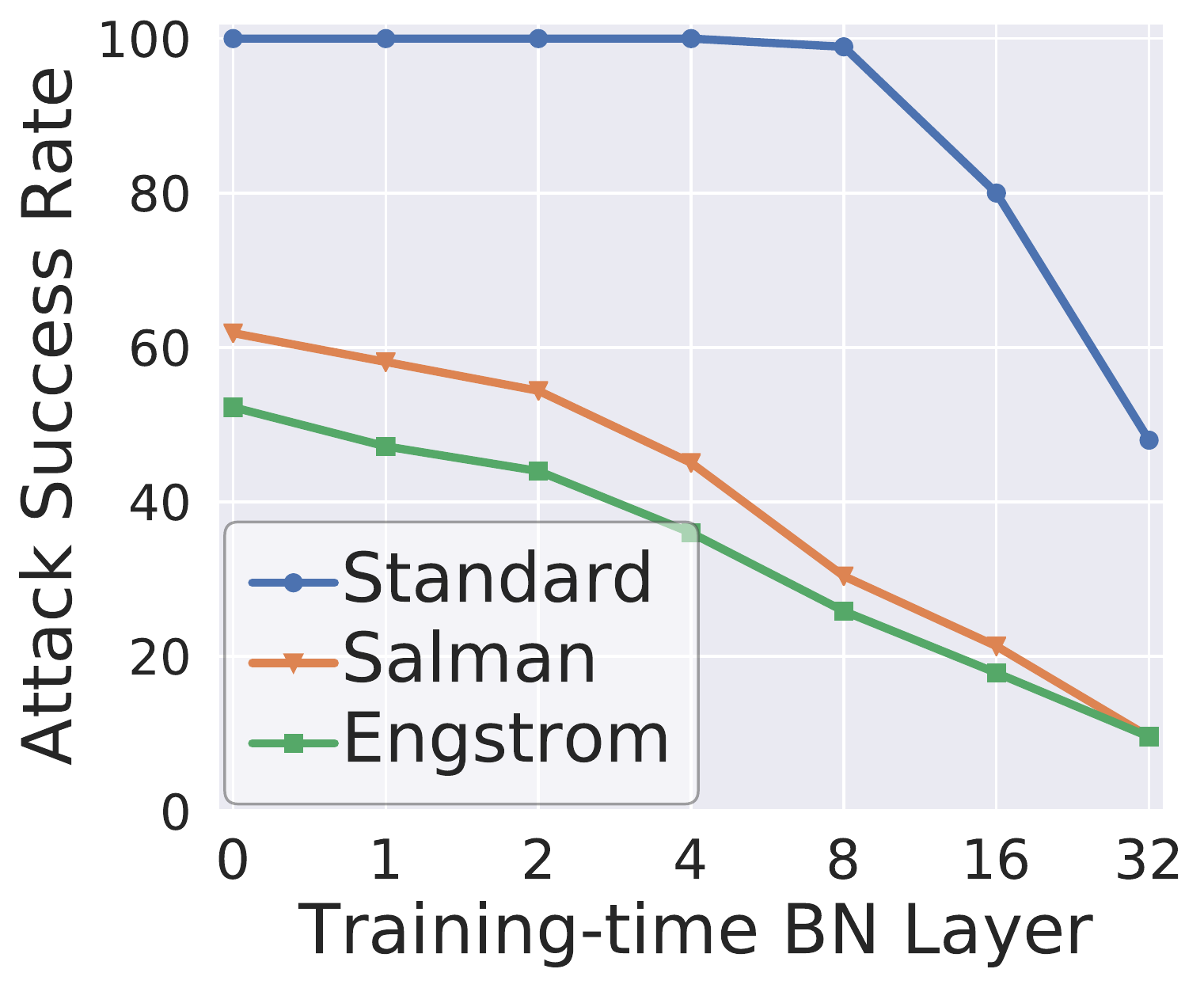} & 
 \includegraphics[width=0.23\textwidth]{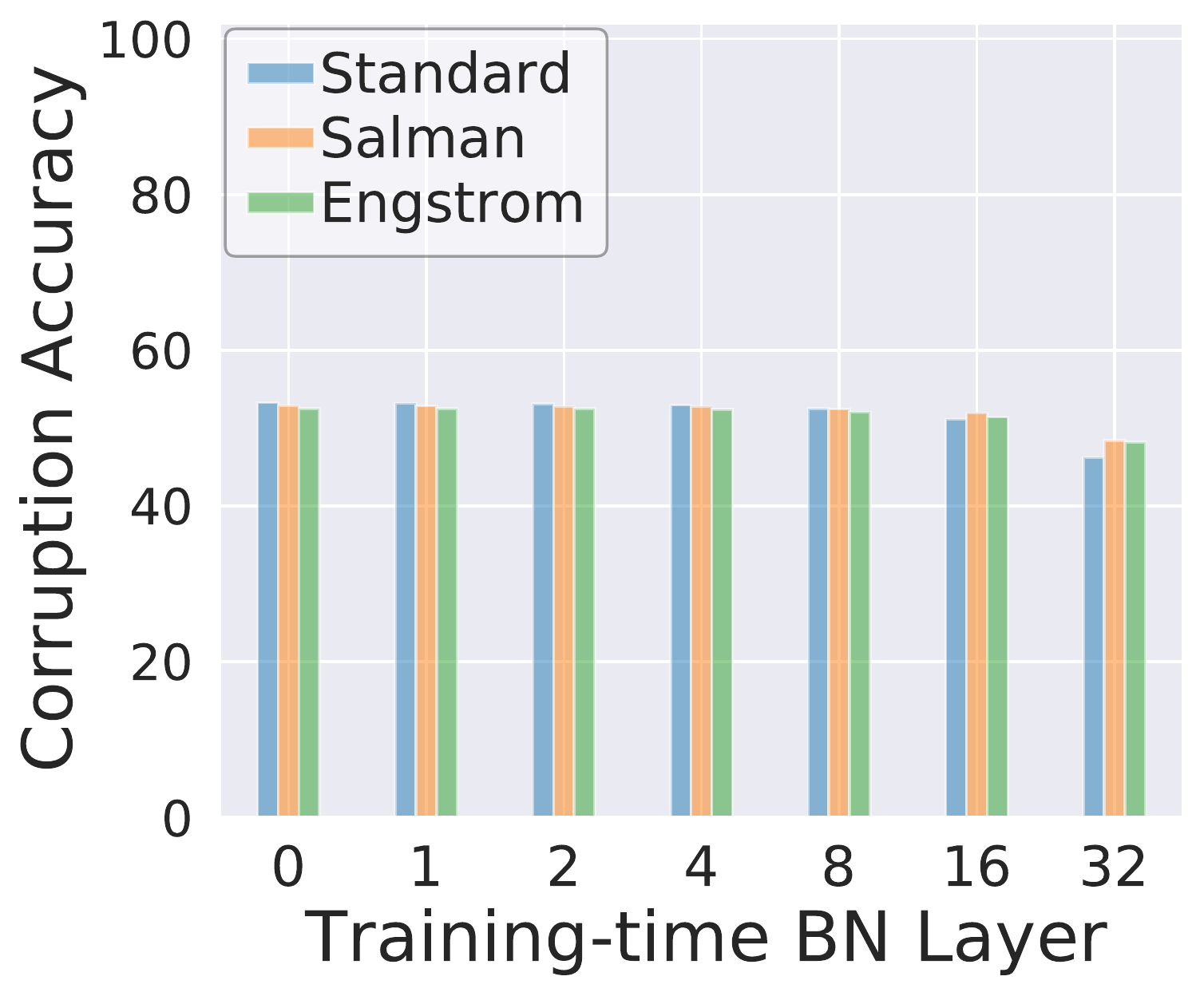} \\
 \multicolumn{2}{c}{(a) Smoothing via training-time BN statistics  }  & \multicolumn{2}{c}{(b)  Adaptively selecting layer-wise BN statistics. [$\tau$=0.6]  } \\
\end{tabular}
\caption{ Robustly estimating the final BN statistics can degrade the attack success rate (line) while maintaining high corruption accuracy (bar) on ImageNet-C. [$N_m$=40] This is the complementary evaluation of Figure~\ref{Fig:IN_Bayes} and Figure~\ref{Fig:IN_Bayes_layer} with more malicious samples. }
	\label{Fig:IN_rbn_def40}
\end{figure}

Figure~\ref{Fig:IN_rbn_def40}
shows the additional experiments of section~\ref{sec:dfBN} with a stronger attacking setting ($N_m$ = 40). We observe the conclusions keep the same where selecting an appropriate $\tau$ and $N_{tr}$ (i.e., $\tau$ = 0.6, $N_{tr}$ = 16) can decrease the ASR for $\sim$70\% and $\sim$20\% for the robust and standard model, respectively, and  maintain the corruption accuracy.

\subsection{More Evaluations of Corruption Accuracy for Robustly Estimating BN}
\label{append:param_search}

In this subsection, we present a more comprehensive parameters search for the smoothing factor $\tau$ and number of training-time BN  layers $N_{tr}$ to understand their effect on corruption accuracy.  
Specifically, we analyze standard and robust models on CIFAR-C and ImageNet-C benchmarks. 

\begin{figure}[H]
\centering
\begin{tabular}{ccc}
  \includegraphics[width=0.32\textwidth]{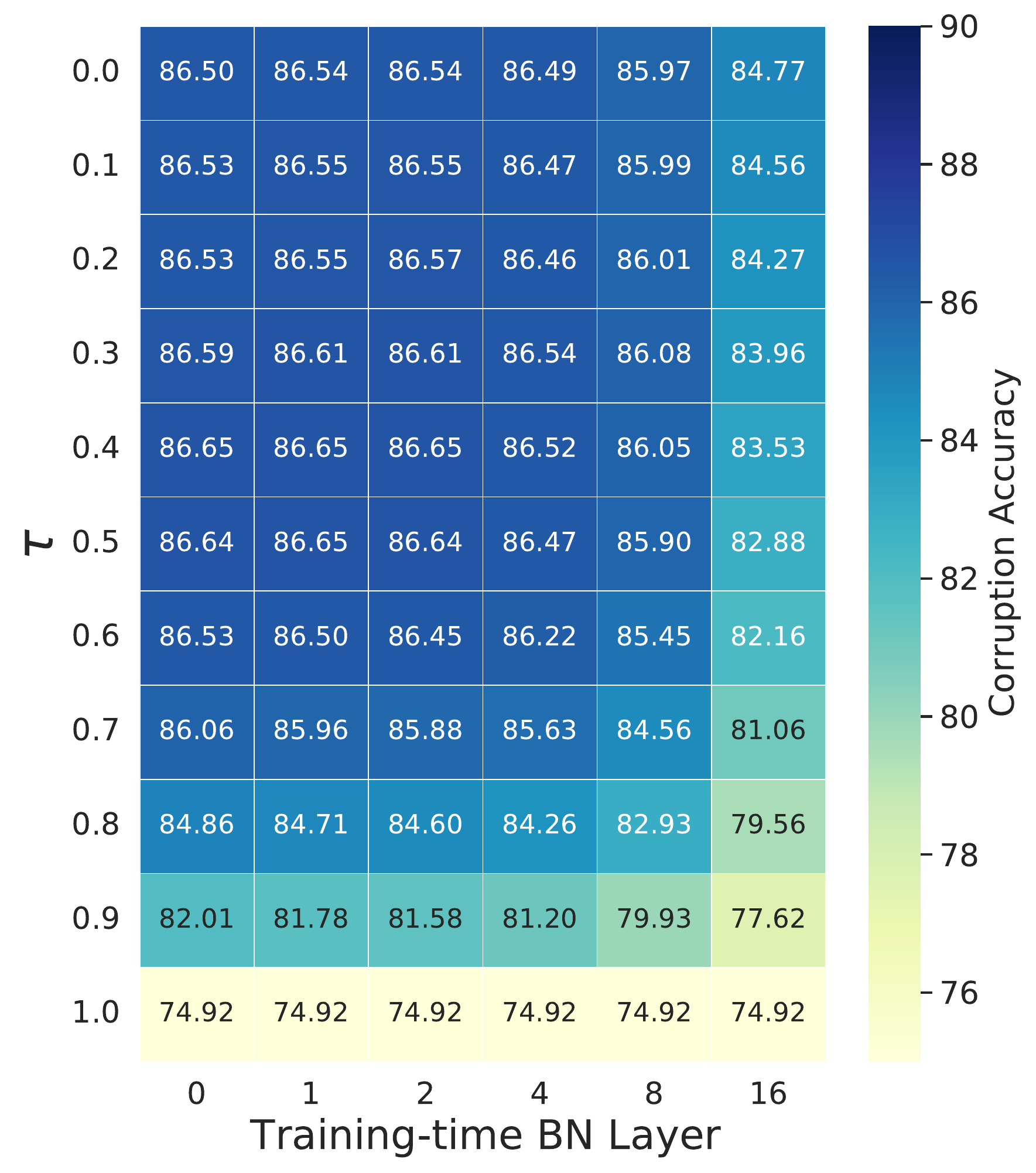} &
 \includegraphics[width=0.32\textwidth]{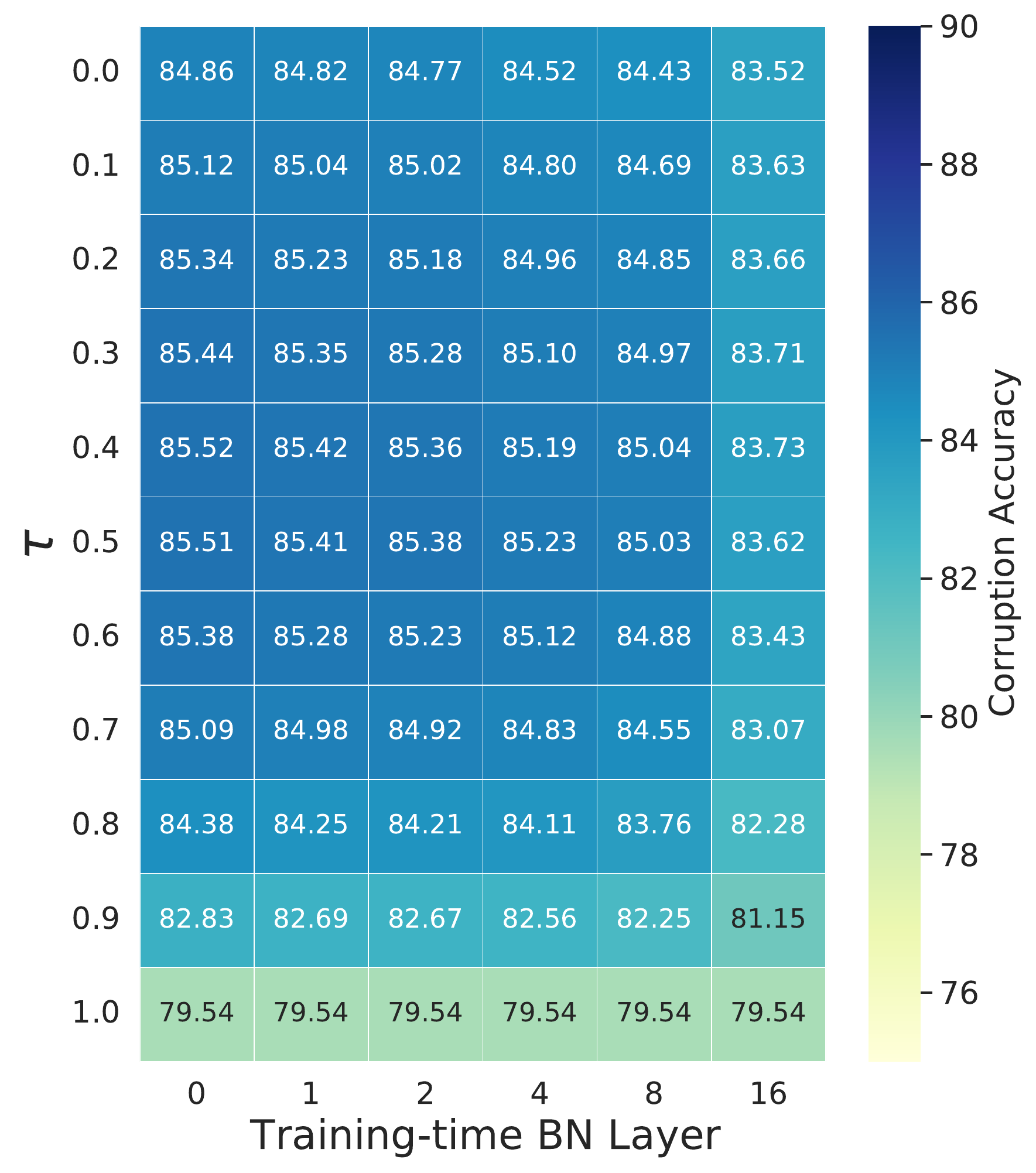} &
  \includegraphics[width=0.32\textwidth]{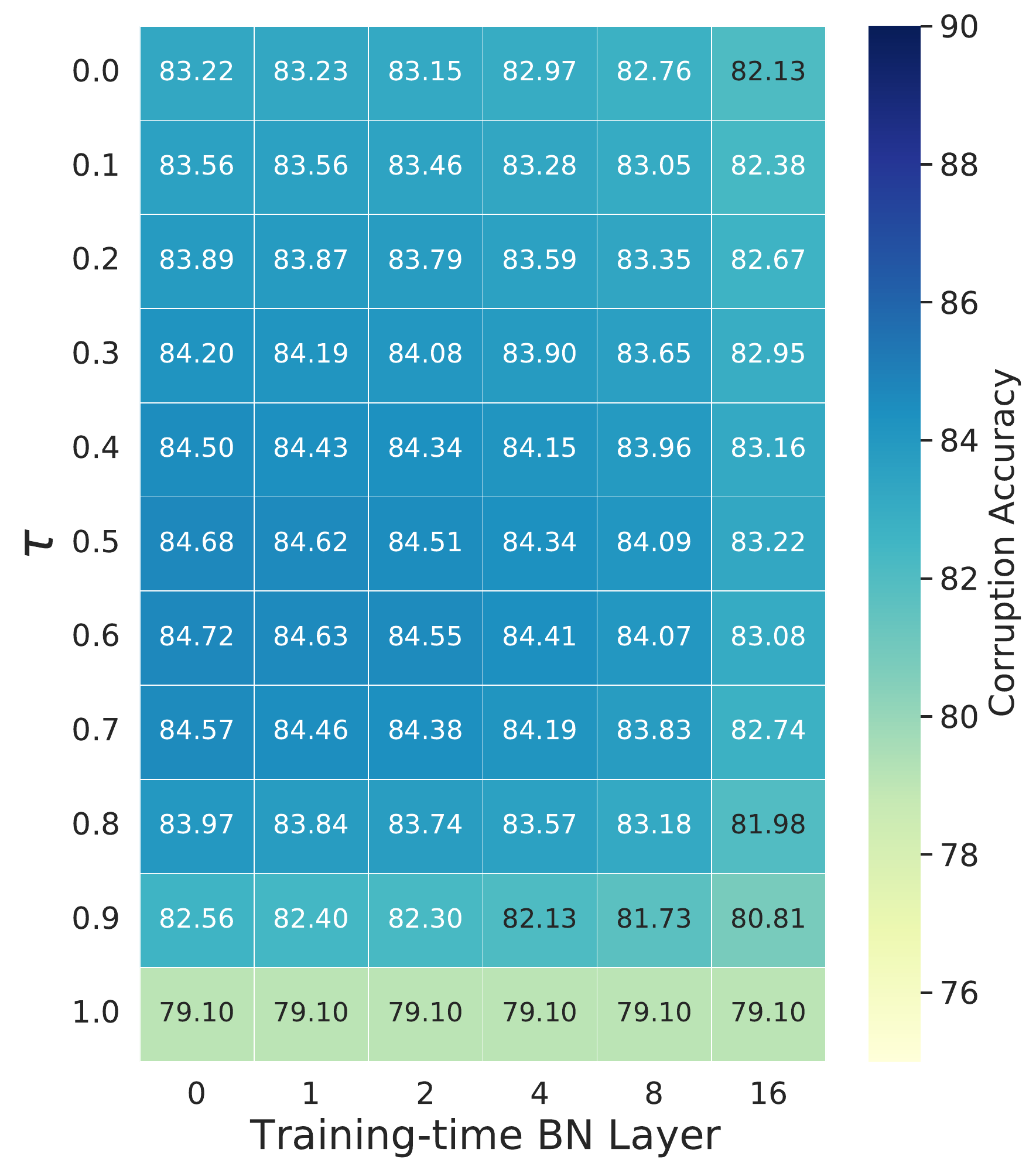}\\
  (a) Standard WRN-28 & (b) \citet{Gowal2021ImprovingRU} & (c) \citet{Wu2020AdversarialWP}\\
\end{tabular}
	\caption{Corruption accuracy of balancing training-time BN across $\tau$ and number of Training-time BN  layers on CIFAR-10-C. }
	\label{Fig:CIFAR10_BAYESBN}
\end{figure}

\begin{figure}[H]
\centering
\begin{tabular}{ccc}
  \includegraphics[width=0.32\textwidth]{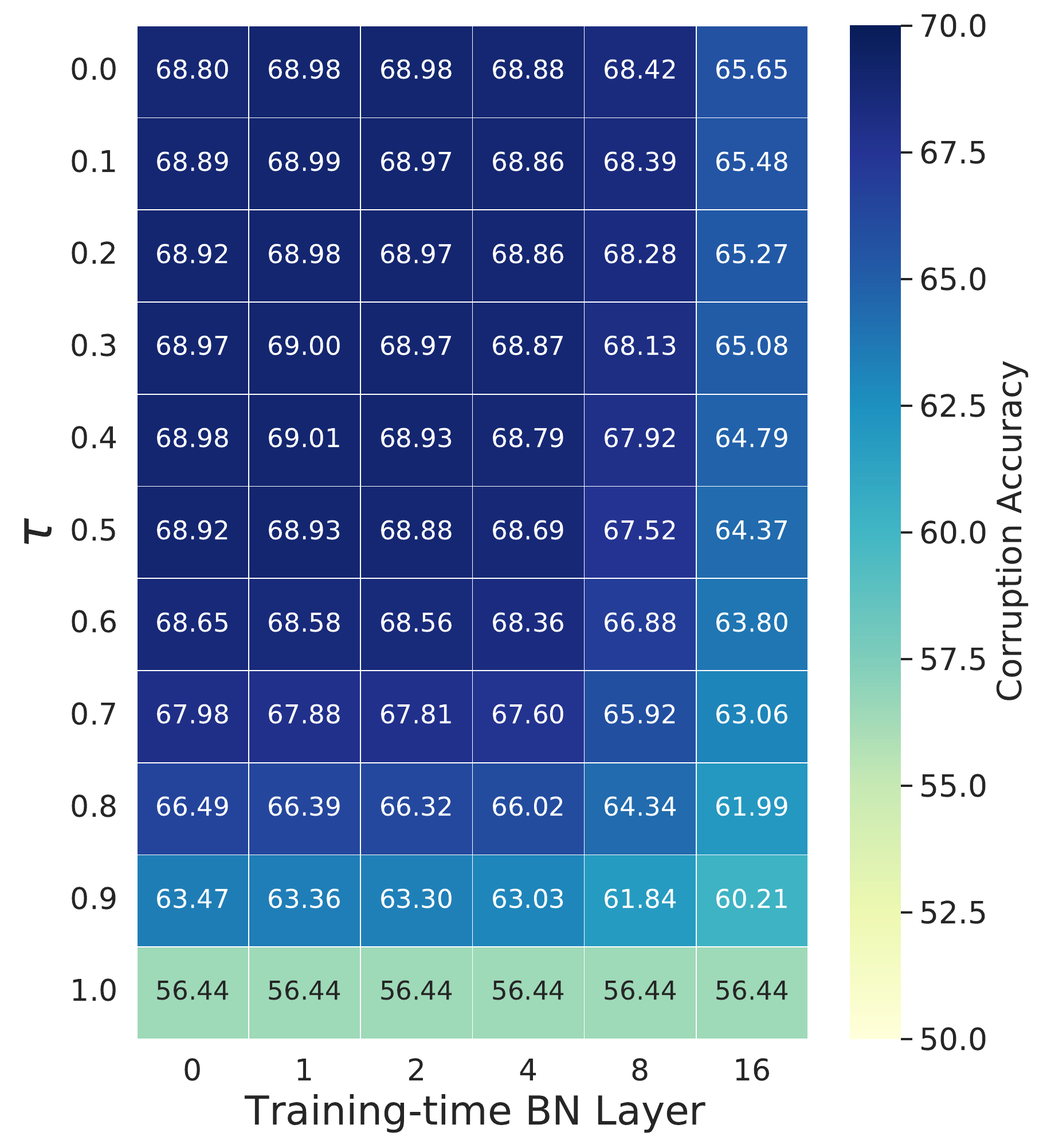} &
 \includegraphics[width=0.32\textwidth]{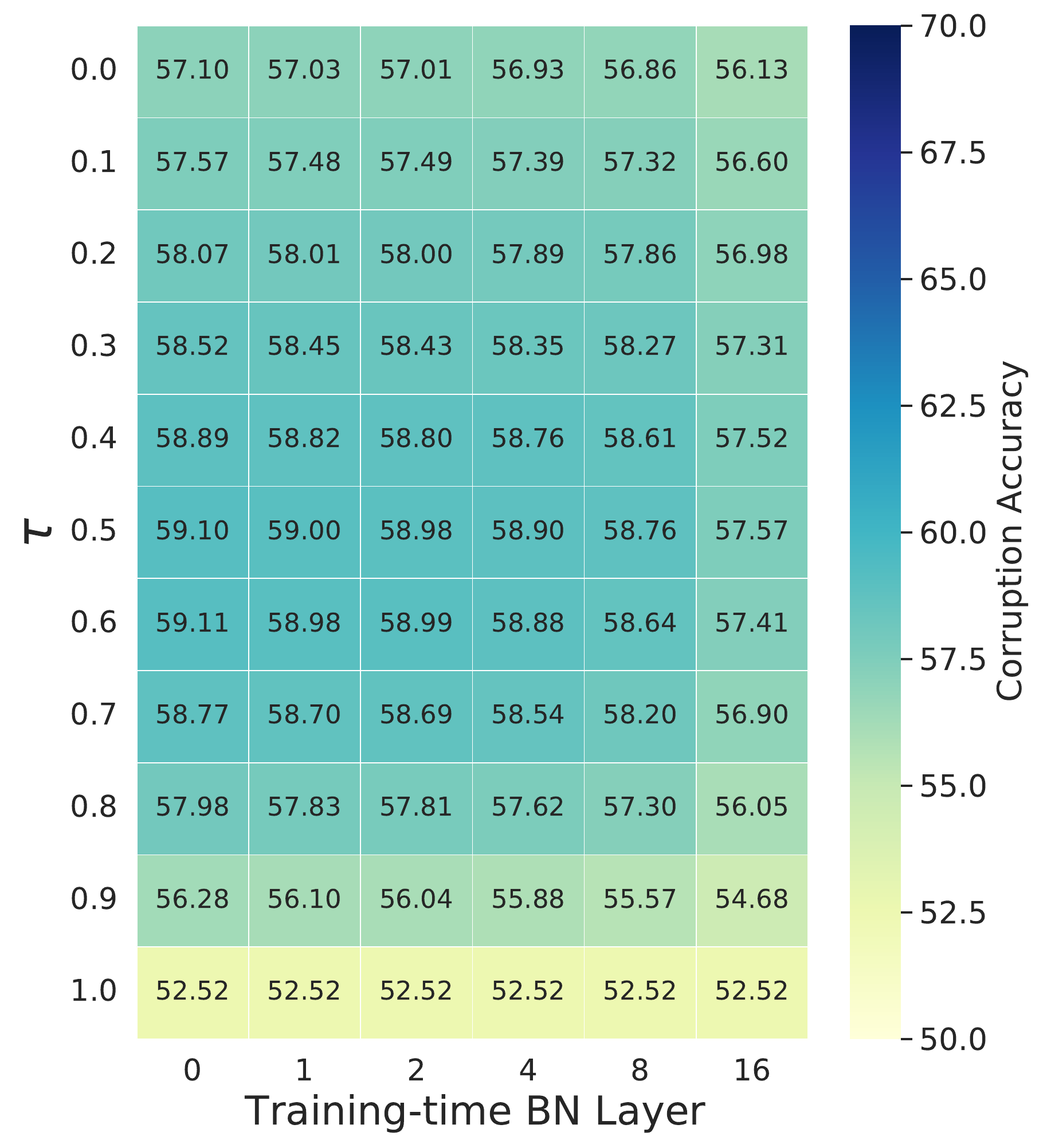} &
  \includegraphics[width=0.32\textwidth]{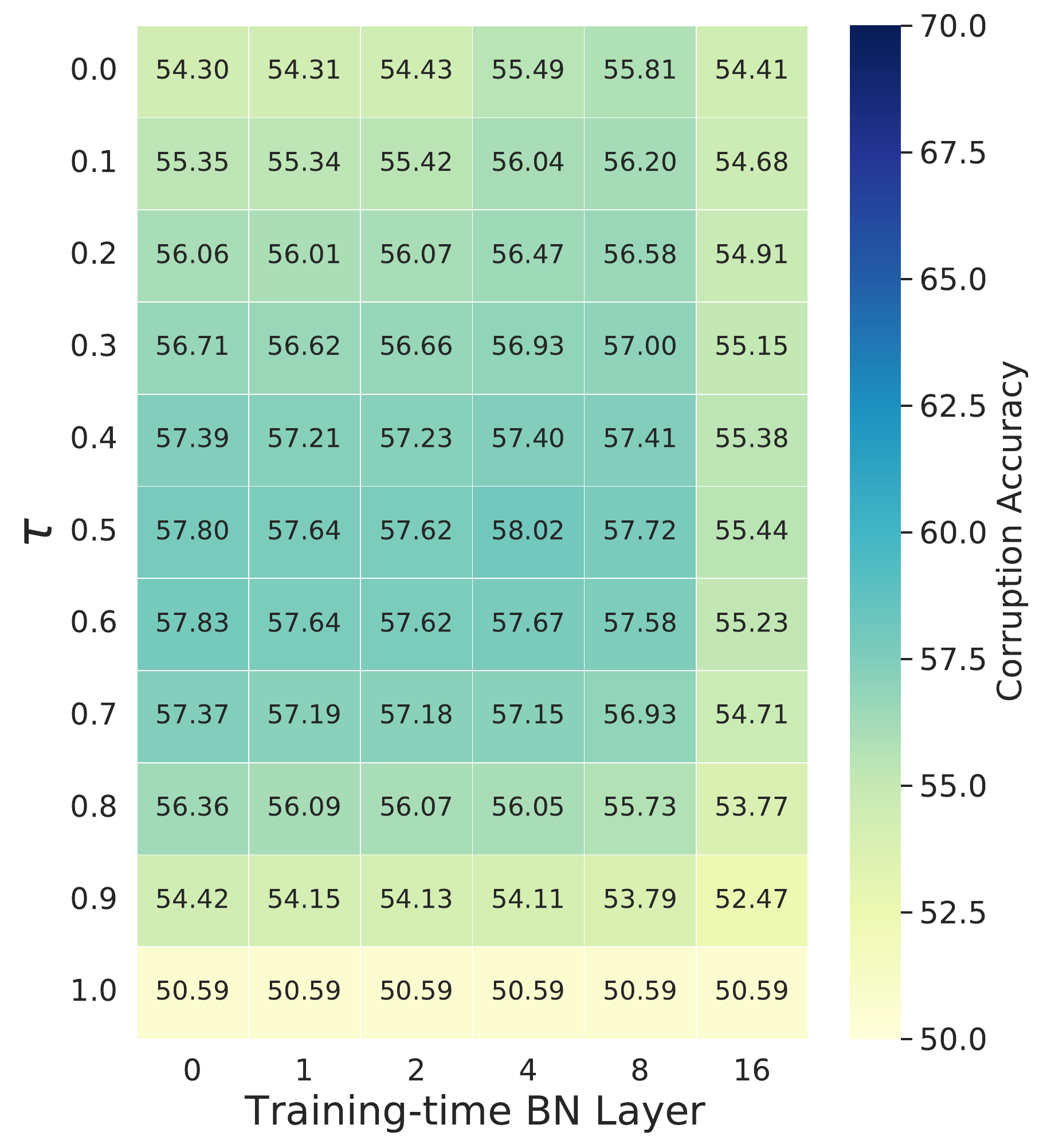}\\
  (a) Standard WRN-28 & (b) \citet{Pang2022RobustnessAA} & (c) \citet{Rebuffi2021FixingDA}\\
\end{tabular}
	\caption{Corruption accuracy of balancing training-time BN across $\tau$ and number of Training-time BN  layers on CIFAR-100-C. }
	\label{Fig:CIFAR100_BAYESBN}
\end{figure}

\begin{figure}[H]
\centering
\begin{tabular}{ccc}
  \includegraphics[width=0.32\textwidth]{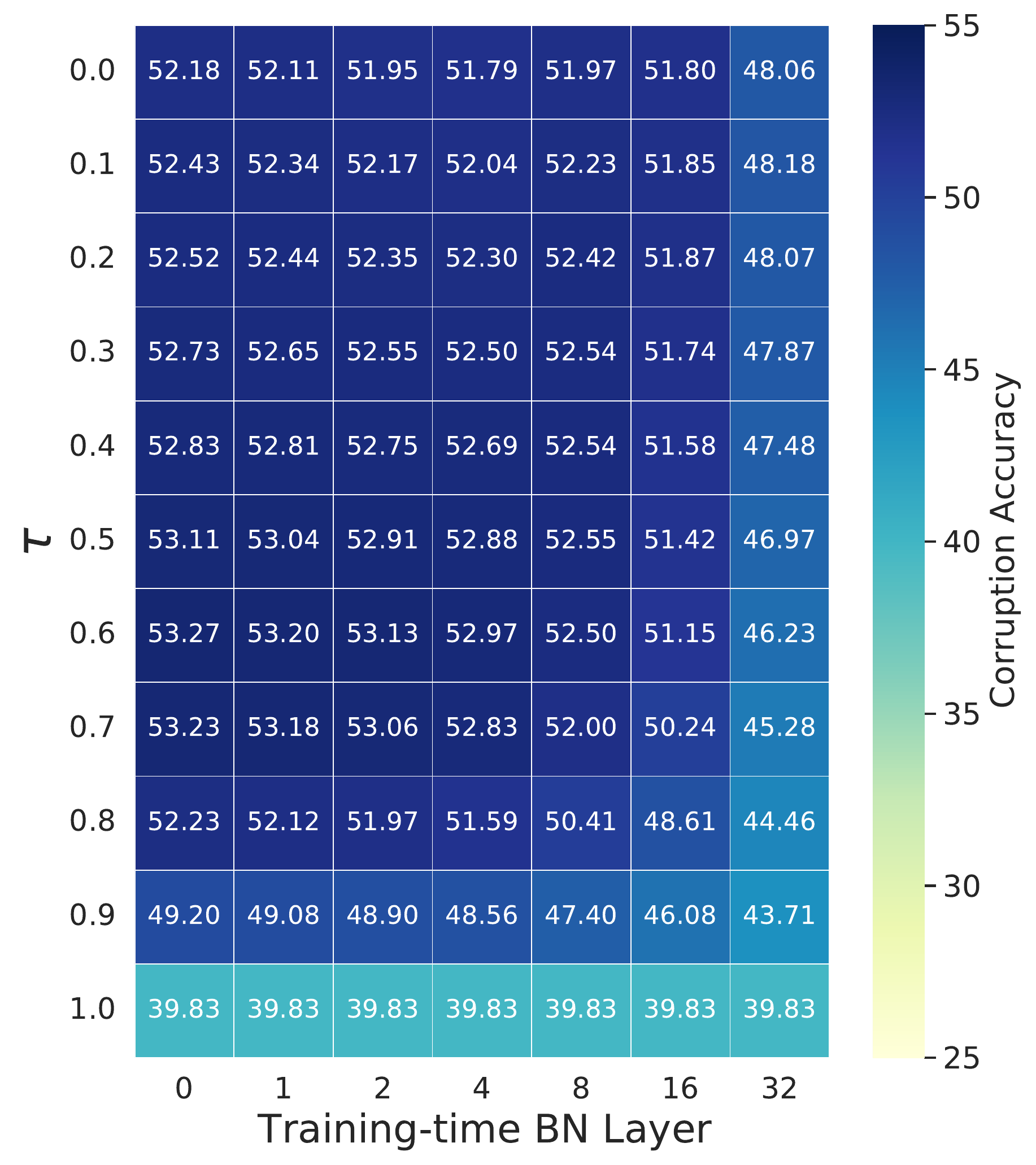} &
 \includegraphics[width=0.32\textwidth]{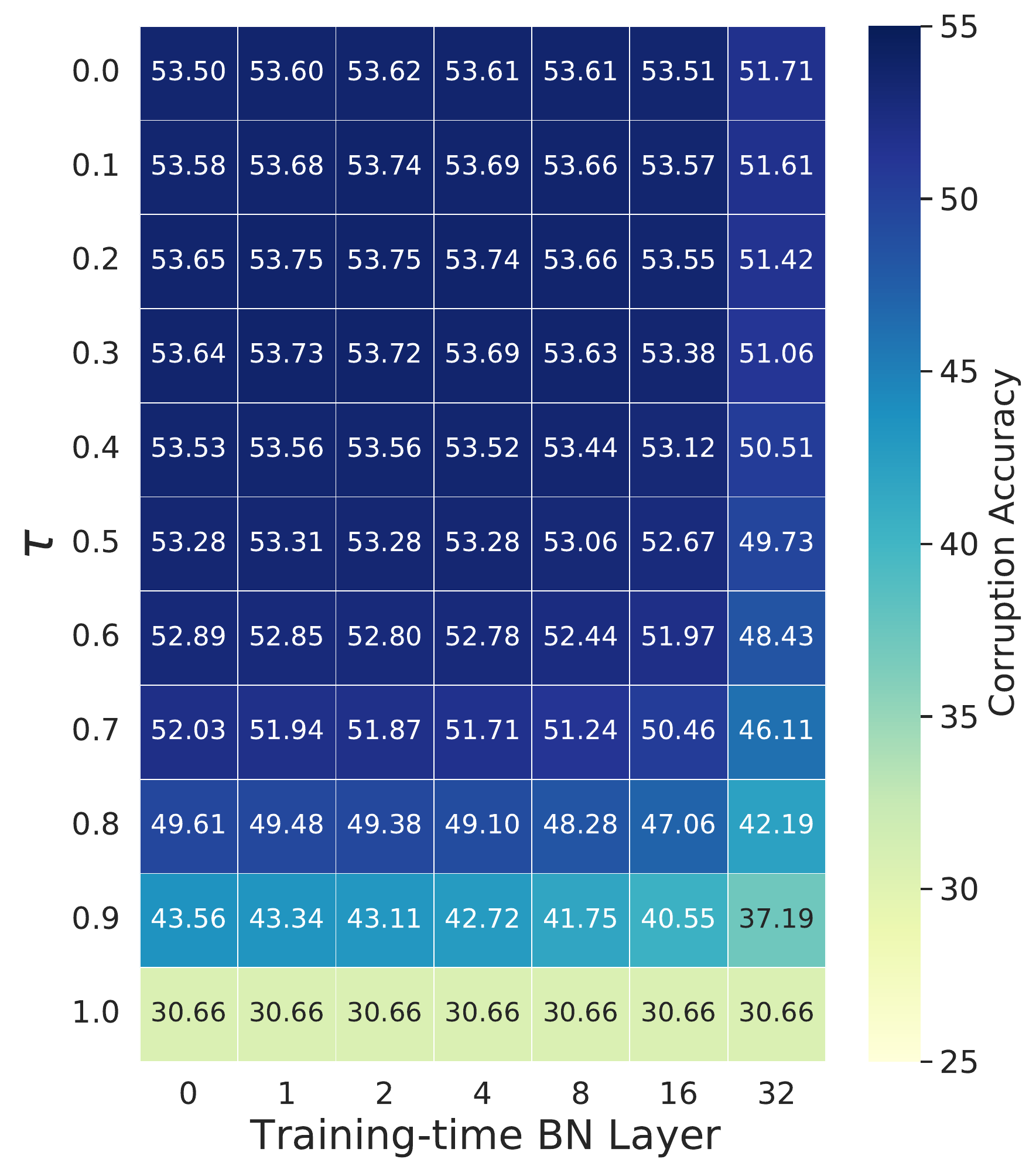} &
  \includegraphics[width=0.32\textwidth]{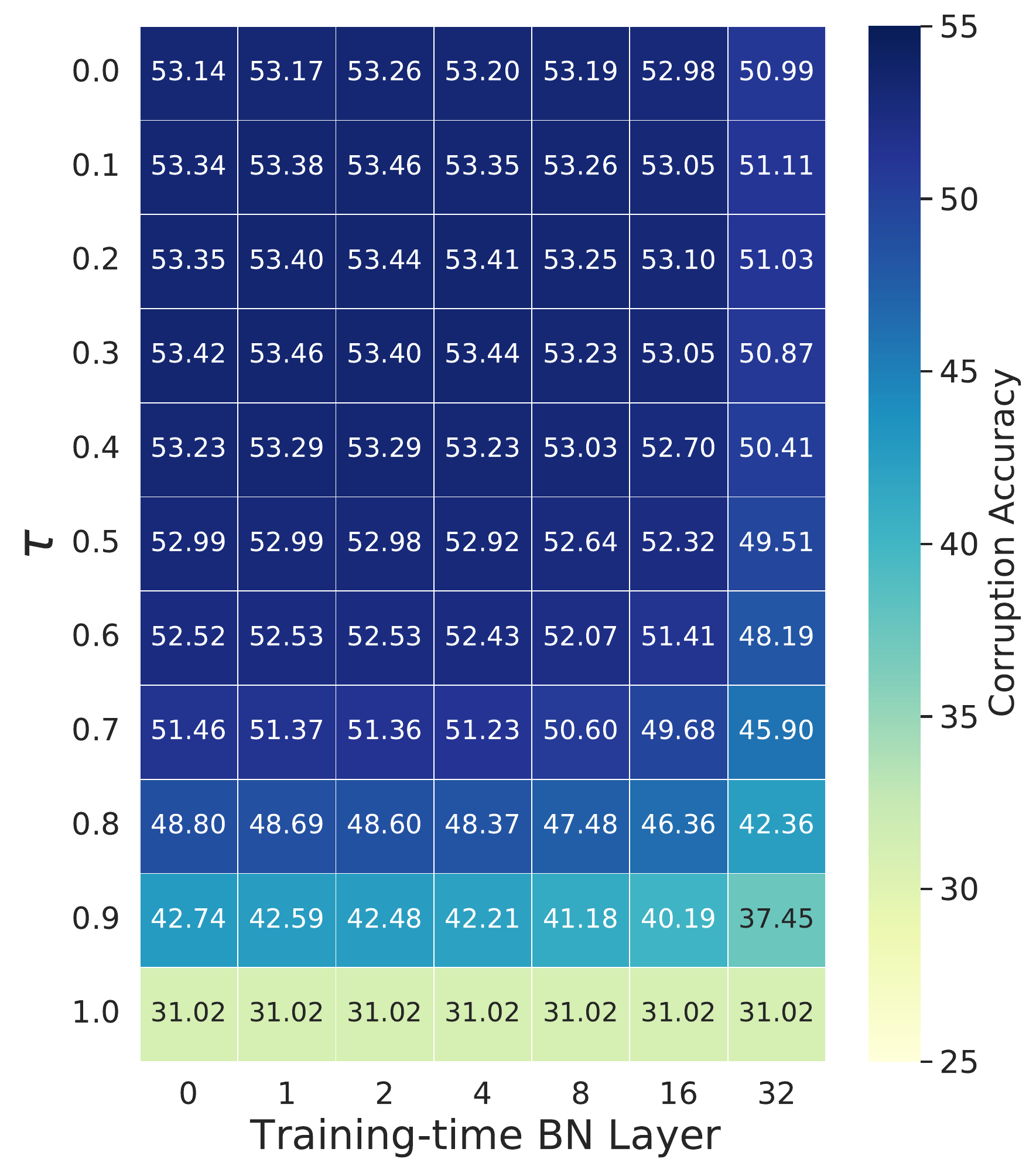}\\
  (a) Standard ResNet-50 & (b) \citet{Salman2020DoAR} & (c) \citet{robustness}\\
\end{tabular}
	\caption{Corruption accuracy of balancing training-time BN across $\tau$ and number of Training-time BN  layers on ImageNet-C. }
	\label{Fig:IN_BAYESBN}
\end{figure}

As shown in Figure~\ref{Fig:CIFAR10_BAYESBN}, Figure~\ref{Fig:CIFAR100_BAYESBN}, and Figure~\ref{Fig:IN_BAYESBN}, we observe that the corruption accuracy generally reaches the best when $\tau = 0.5$ and $N_{tr}=0$. The improvement compared to \textbf{TeBN} ($\tau = 0$ and $N_{tr}=0$) is limited ($\sim$2\%) except for the robust models on CIFAR-100-C. The corruption accuracy drops dramatically for $\tau > 0.7$. Furthermore, increasing $N_{tr}$ also decreases the accuracy of corrupted data.

\newpage
\section{Visiualization of Constrained Attacks}
\subsection{Examples of Corrupted Images and Corresponding $L_\infty$ Malicious Images }

\begin{figure}[H]
\centering
\resizebox{0.98\columnwidth}{!}{%
\begin{tabular}{c} 
\includegraphics[width=1.0\textwidth]{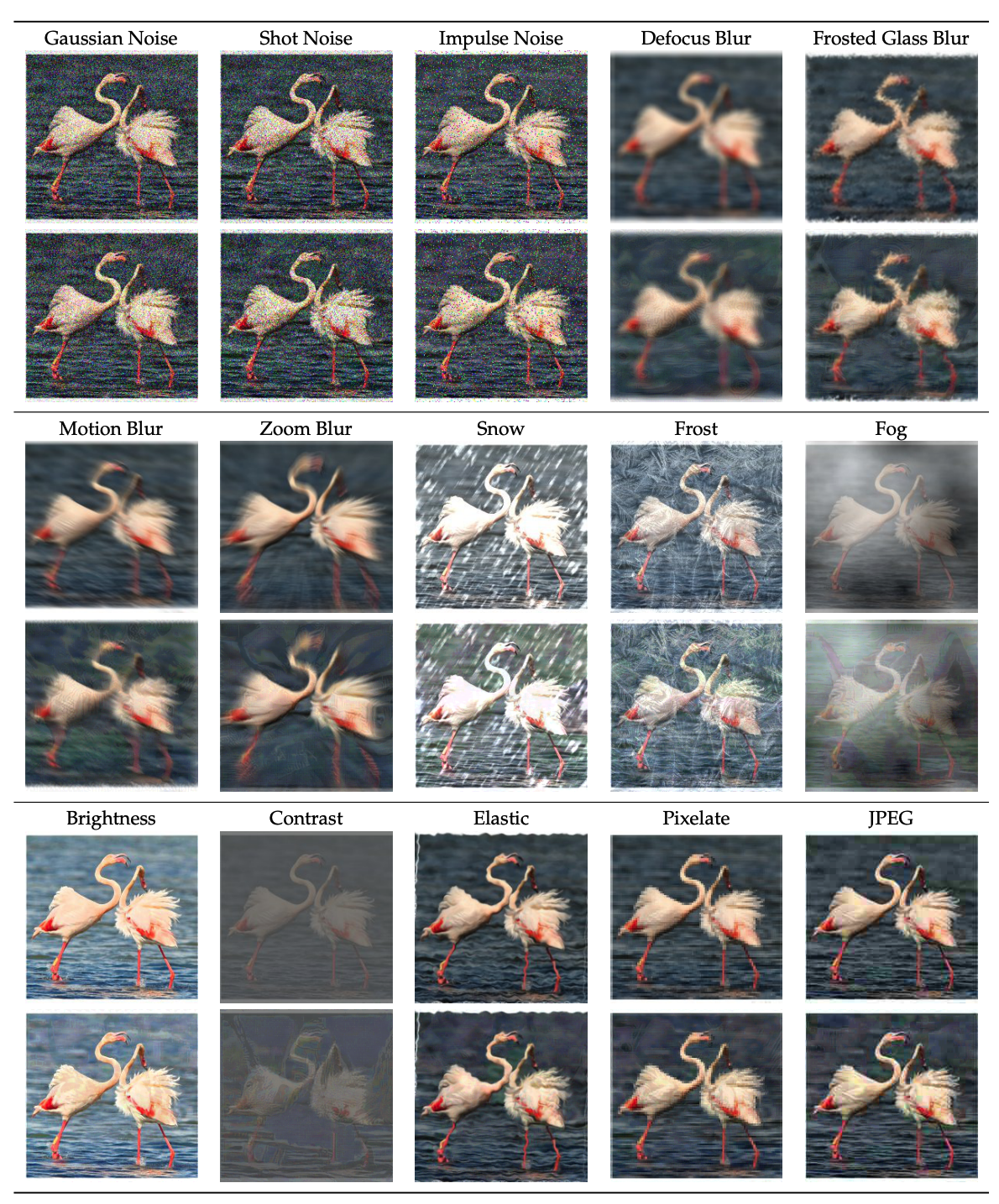}
\end{tabular}
}
\caption{Demonstrations of 15 types of corrupted benign images (upper) and malicious images with $\epsilon = 8/255$ (lower) from the ImageNet-C benchmark. Most  malicious images are visually imperceptible compared with their original ones. [Severity level: 3]}
\label{Tab:IN_EPS}
\end{figure}

\newpage
\subsection{Examples of Corrupted Images and Adversarial Corrupted Images }
\label{appendix:IN_CORR}

\newpage
\section{Visualization of Batch Normalization Statistics}

\end{document}